\documentclass[letterpaper]{article} 

\usepackage[T1]{fontenc}
\usepackage{aaai24}  
\usepackage{times}  
\usepackage{helvet}  
\usepackage{courier}  
\usepackage[hyphens]{url}  
\usepackage{graphicx} 
\usepackage{natbib}  
\usepackage{caption} 
\usepackage{algorithm}
\usepackage{algorithmic}
\usepackage{pifont}

\usepackage{microtype}
\usepackage{float}
\usepackage{array}
\usepackage{booktabs}
\usepackage{multirow}
\usepackage{subfig}
\usepackage{graphicx}
\usepackage{booktabs} 

\usepackage{amsmath}
\usepackage{amssymb}
\usepackage{mathtools}
\usepackage{amsthm}
\usepackage{pbsi}
\usepackage{booktabs}
\usepackage{colortbl}
\usepackage[capitalize,noabbrev]{cleveref}

\theoremstyle{plain}
\newtheorem{theorem}{Theorem}[section]

\theoremstyle{definition}
\newtheorem{definition}[theorem]{Definition}

\theoremstyle{remark}
\newtheorem{remark}[theorem]{Remark}

\usepackage[textsize=tiny]{todonotes}

\usepackage{newfloat}
\usepackage{listings}
\DeclareCaptionStyle{ruled}{labelfont=normalfont,labelsep=colon,strut=off} 
\floatstyle{ruled}
\newfloat{listing}{tb}{lst}{}
\floatname{listing}{Listing}
\title{Tensor Generalized Approximate Message Passing}
\author{
    Yinchuan Li, \textnormal{\textit{Member, IEEE}},
    Guangchen Lan,
    Xiaodong Wang, \textnormal{\textit{Fellow, IEEE}}
}
\affiliations{
    Columbia University\\

    New York, NY 10027 USA\\
    yinchuan.li.cn@gmail.com, \{gl2713, wangx\}@columbia.edu
}

\usepackage{bibentry}

\begin{document}

\maketitle
\allowdisplaybreaks[4]

\begin{abstract}
We propose a tensor generalized approximate message passing (TeG-AMP) algorithm for low-rank tensor inference, which can be used to solve tensor completion and decomposition problems. We derive TeG-AMP algorithm as an approximation of the sum-product belief propagation algorithm in high dimensions where the central limit theorem and Taylor series approximations are applicable. As TeG-AMP is developed based on a general TR decomposition model, it can be directly applied to many low-rank tensor types. Moreover, our TeG-AMP can be simplified based on the CP decomposition model and a tensor simplified AMP is proposed for low CP-rank tensor inference problems. Experimental results demonstrate that the proposed methods significantly improve recovery performances since it takes full advantage of tensor structures.
\end{abstract}

\section{Introduction}

\subsection{Motivations}

Multi-dimensional (MD) arrays, i.e., tensors, arise naturally in numerous applications~\cite{ghadermarzy2017near,gandy2011tensor}, including visual data reconstruction~\cite{liu2012tensor}, deep neural networks~\cite{tung2018deep,wang2018wide}, data mining~\cite{acar2005modeling}, seismic data processing~\cite{ely20135d} and signal processing~\cite{nion2010tensor,de2007parafac}. In many modern use cases of machine and deep learning (e.g., mobile phones~\cite{kim2015compression}, wearables, and IoT devices~\cite{lane2015early}), storage and computation resources are extremely limited. Hence, low-rank tensor approximation~\cite{wang2018wide} has thus drawn considerable attention in effective model compression, low generative error, and fast prediction speed~\cite{sokolic2017generalization}, and has naturally inspired more works with tensor decomposition and completion algorithms~\cite{liu2012tensor,li2019multi-d,li2019compressive}.


\subsection{Related Works}
Many tensor decomposition models have been proposed. In particular, the canonical polyadic (CP) decomposition~\cite{de2006link} attempts to approximate an observed tensor by a sum of rank one tensors. The Tucker decomposition~\cite{oseledets2008tucker} aims to approximate tensors by a core tensor and several factor matrices. The tensor train (TT) decomposition~\cite{oseledets2011tensor} attempts to approximate an observed tensor by a sequence of cores, and was improved in~\cite{zhao2016tensor} as the tensor ring (TR) decomposition. 
The most widely used algorithm to perform rank decomposition is alternating minimization (AltMin)~\cite{carroll1970analysis, wang2017efficient, lan2023improved}, which uses convex optimization techniques on different slices of the tensor. However, a major disadvantage of AltMin for tensor decomposition or completion is that it does not perform well in the presence of highly noisy measurements or missing a large number of measurements. Related variant algorithms~\cite{steinlechner2016riemannian,narita2012tensor,gandy2011tensor,kasai2016low,wein2019kikuchi,arous2020algorithmic} suffer from similar problems. As the above methods involve an {\em unfolding} process, i.e., transforming the tensor into a matrix, they are inherently {\em matrix-based} methods, while the tensor structure is not fully utilized.

The approximate message passing (AMP)~\cite{donoho2009message,kabashima2004bp,advani2013statistical,bayati2011dynamics} methods, based on the Gaussian approximations of loopy belief propagation, have attracted considerable attention in recovery and completion problems. \cite{rangan2011generalized} considered the estimation of an i.i.d. random vector observed through a linear transform, followed by a component-wise and probabilistic measurement channel and proposed the generalized approximate message passing (G-AMP), which provides a computationally efficient approximate implementation of sum-problem loopy belief propagation for such problems. AMP for belief propagation in graphical models and a rigorous analysis of AMP was presented in~\cite{bayati2011dynamics}. In~\cite{rangan2012iterative}, AMP was used for a rank-one matrix estimation problem, which has been extended to various low-rank matrix decomposition problems in~\cite{lesieur2015mmse,lesieur2017constrained,lesieur2017statistical,lesieur2015phase}. Moreover, bilinear generalized AMP (BiG-AMP) is introduced in~\cite{parker2014bilinear,parker2014bilinear2} and was applied to matrix completion.

Recently, low-rank tensor decomposition is studied in federated learning in~\cite{lan2023federated, wang2022tensor} and AMP in~\cite{lesieur2017statistical}, but their methods are limited to symmetric tensors and then necessarily requires cubic in shape. Generally, tensors that occur naturally in the wild are almost never cubic in shape, nor are they symmetric. A Bayesian AMP algorithm for arbitrarily shaped spiked-tensor decomposition is proposed in~\cite{kadmon2018statistical}, and a dynamic mean field theory is used to precisely characterize their performance. 
Unfortunately, this method is only based on spiked decomposition and cannot be applied to tensor completion problems. Tensor decomposition and completion algorithms based on AMP for various low-rank tensor models remain to be studied.

\subsection{Main Contributions}

In this paper, we propose an AMP-based algorithm for low TR-rank tensor inference and refer to it as {\em Tensor Generalized AMP} (TeG-AMP), which can be used to solve problems like high-dimensional tensor completion and decomposition.
Under the assumption of statistically independent tensor entries with known priors, we derive our TeG-AMP algorithm as an approximation of the sum-product belief propagation algorithm in the high-dimensional limit, where the central limit theorem reasonings and Taylor series approximations are applicable. Specifically, the main contributions of this paper are as follows:

1) To the best of authors' knowledge, our TeG-AMP is the first tensor inference algorithm that is well-designed to take advantage of tensor structures, which naturally makes our algorithm performance essentially better than conventional tensor completion/decomposition algorithms;

2) The loopy belief propagation of high-dimensional tensor is different from that of low-dimensional data (vector and matrix) in previous works~\cite{rangan2011generalized,bayati2011dynamics,lesieur2015mmse,lesieur2017constrained,lesieur2017statistical,lesieur2015phase,parker2014bilinear,parker2014bilinear2}. Our TeG-AMP contains adequate original derivations and analysis to close the loop of belief propagation;

3) As TeG-AMP is developed based on the general TR decomposition model, it can be applied to many low-rank tensor data types, and does not have excessive requirements on tensors, so it has high practical application value; In addition, our TeG-AMP can be simplified based on the CP decomposition model and a tensor simplified AMP (TeS-AMP) is proposed for low CP-rank tensor inference problems.



At last, we present the empirical analysis and compare TeG-AMP to state-of-the-art algorithms on tensor completion problems. Our numerical results demonstrate that TeG-AMP yields significantly higher reconstruction accuracy.




\section{Background and Problem Formulation}

\subsection{Tensor Ring Model}

A $d$-th order tensor ${\cal {U}} \in {\mathbb{R}}^{N_1\times N_2 \times \cdots \times N_d}$ can be decomposed into a sequence of latent tensors $\mathcal{Z}_{i} \in \mathbb{R}^{r_{i} \times N_{i} \times r_{i+1}},~i = 1,\cdots,d,$ as follows:
\begin{align} 
\label{TR-decomposition-1}
u_{\mathbf{x}} &={\rm Tr}\left\{\prod_{i=1}^{d} \mathcal{Z}_{i}(:,x_{i},:)\right\},
\end{align}
where $u_{\mathbf{x}}$ denotes the $(x_1,x_2,\cdots,x_d)$-th element of tensor ${\cal {U}}$, $\mathcal{Z}_{i}(:,x_{i},:) \in \mathbb{R}^{r_i \times r_{i+1}}$ denotes the $x_i$-th lateral slice matrix of the $i$-th latent tensor $\mathcal{Z}_{i}$, and ${\rm Tr}(\cdot)$ denotes the trace operator. 
The last latent tensor $\mathcal{Z}_{d}$ is of size ${r_{d} \times N_{d} \times r_{1}}$, 
which means $r_{d+1} = r_1$. The latent tensor $\mathcal{Z}_{i}$ is also called $i$-th core (or node), and the sizes of cores, denoted by a vector $\mathbf{r} = [r_1, r_2,\cdots, r_d]^T$, are called {\em TR-ranks}.

The matrix form of the TR-decomposition in \eqref{TR-decomposition-1} can be expanded as
\begin{align} 
u_{\mathbf{x}} &= \sum_{\ell_{1}, \ldots, \ell_{d}=1}^{r_{1}, \ldots, r_{d}} \prod_{i=1}^{d}  \mathcal{Z}_{i}(\ell_i,x_{i},\ell_{i+1}) \triangleq  \sum_{\mathbf{l}=1}^{\mathbf{r}} \prod_{i=1}^{d} \mathcal{Z}_{i}^{\ell_{i}, \ell_{i+1} }(x_{i}). \nonumber
\end{align}
Then, ${\cal {U}}$ can then be decomposed as
\begin{align}
{\mathcal U} = \sum_{\mathbf{l}=1}^{\mathbf{r}} \mathcal{Z}_{1}(\ell_1,:,\ell_{2}) \otimes \mathcal{Z}_{2}(\ell_2,:,\ell_{3}) \otimes \cdots \otimes \mathcal{Z}_{d}(\ell_d,:,\ell_{1}), \nonumber
\end{align}
where $\mathcal{Z}_{i}(\ell_i,:,\ell_{i+1}) \in \mathbb{R}^{N_{i}}$ and $\otimes$ denotes the tensor product (outer product). Hence tensor ${\cal {U}}$ can be decomposed as a sum of $\prod_{i=1}^{d} r_i$ rank-1 tensors each being the tensor product of $d$ vectors taken from each core.

\subsection{Problem Formulation}

We provide an algorithmic framework for the general problem of estimating tensor ring decomposition components from the original tensor $\mathcal{U}$ as follows
\begin{align} 
\label{V-U}
\mathcal{V} = \mathcal{U}_{\Omega} + \mathcal{W},
\end{align}
where $\mathcal{U}_{\Omega}$ is partially observed from $\mathcal{U}$, and $\mathcal{W}$ is the noise. $\mathcal{V}$ can be represented as a likelihood function of the decoupled form
\begin{align} 
\label{likelihood-decoupled}
p(\mathcal{V} | \mathcal{U})
=\prod_{\mathbf{x}} p\left(v_{\mathbf{x}} | u_{\mathbf{x}}\right).
\end{align}
Note that \eqref{likelihood-decoupled} subsumes tensor completion and decomposition as special cases~\cite{xie2017kronecker,tang2019social}.

We assume that a prior of the tensor $\mathcal{U}$ is composed of independent priors of each element of each core
\begin{align}
\label{pdf-decoupled}
p(\mathcal{U})
=&~ \prod_{i=1}^{d} p\left(\mathcal{Z}_{i}\right) = \prod_{i=1}^{d} \prod_{\ell_i=1}^{r_{i}} \prod_{x_i=1}^{N_{i}} \prod_{\ell_{i+1}=1}^{r_{i+1}} p\left(\mathcal{Z}_{i}(\ell_i,x_i,\ell_{i+1})\right).
\end{align}

The goal is to compute the minimum mean-squared error (MMSE) estimates of $\{\mathcal{Z}_{i}\}_{i=1}^{d}$ from the observation ${\mathcal V}$ based on \eqref{likelihood-decoupled} and \eqref{pdf-decoupled}. Although the exact computation of these quantities is difficult, it can be efficiently approximated by the loopy belief propagation approach. 
According to Bayes' rule, the problem is equivalent to computing the posterior
\begin{align} 
\label{posterior-1}
p(u_{\mathbf{x}} | v_{\mathbf{x}}) \propto ~ p(v_{\mathbf{x}} | u_{\mathbf{x}}) \prod_{i=1}^{d} p\left(\mathcal{Z}_{i}(:,x_i,:) \right).
\end{align}
The above posterior distribution can be represented by a factor graph as shown in Figure~\ref{figure:TR-AMP}, where  there are $\sum_{i=1}^{d}{N_i}$ variable nodes, each representing a random variable, which is matrix $\mathcal{Z}_{i}(:,x_i,:),~x_i=1,\cdots, N_i,~i=1,\cdots,d,$ and there are $\prod_{i=1}^{d}{N_i}$ factor nodes, each representing one term of $p(\mathcal{V} | \mathcal{U})$ in \eqref{likelihood-decoupled}. We next show how to obtain the marginal posteriors based on the sum-product algorithm (SPA).

\section{Tensor Generalized AMP}

\subsection{Loopy Belief Propagation}

For loopy factor graphs, the exact inference is in general NP-hard~\cite{cooper1990computational} and hence loopy belief propagation does not guarantee the correct posteriors. However, extensive applications (e.g., inference on Markov random fields~\cite{freeman2000learning}, LDPC decoding~\cite{mackay2003information}, turbo decoding~\cite{mceliece1998turbo}, multi-user detection~\cite{boutros2002iterative}, and compress sensing~\cite{donoho2009message}) show that loopy belief propagation (BP) has state-of-the-art performance.

In loopy belief propagation, beliefs about the random variables, in the form of pdfs or log-pdfs, are propagated between the nodes of the factor graph until they converge. The typical approach to compute these beliefs is known as SPA~\cite{pearl2014probabilistic,kschischang2001factor}. Specifically, in the SPA, there are two types of messages in the form of pdfs, the message $p_{\mathbf{x} \leftarrow x_{i}}(\mathcal{Z}_{i}(:,x_{i},:))$ from a factor node $p\left({ v_{\mathbf{x}} | u_{\mathbf{x}}}\right)$ to a variable node $\mathcal{Z}_{i}(:,x_{i},:)$, and the message $p_{\mathbf{x} \rightarrow x_{i}}(\mathcal{Z}_{i}(:,x_{i},:))$ from a variable node $\mathcal{Z}_{i}(:,x_{i},:)$ to a factor node $p\left({ v_{\mathbf{x}} | u_{\mathbf{x}}}\right)$. The updating rules are introduced in the next subsection.

\begin{remark}[\textbf{Differences}]\label{tensor-structure} Note that the factor graphs in conventional AMP algorithms are all ``\emph{one step}'' relationship. For example, for vector-AMP,  the variable node is the element in the vector. The message from a factor node to a connected variable node is based on all other variable nodes. For the matrix-wise version BiG-AMP, a variable node in the matrix-AMP is the element in the matrix. For the Bayesian tensor AMP,  a variable node is the spike vector in the tensor. Therefore, the variables in the factor graph can be divided into the considered variable node and other variable nodes, which is ``\emph{one step}'' relationship. 

In contrast, in our TeG-AMP, full tensor structures are used to construct the factor graph. For variable nodes in the TR-tensor, we classify their structure categories into: 
\begin{itemize}
    \item \ding{172} considered variable nodes in the retrieved vector; 
    \item \ding{173} other variable nodes in the retrieved vector; 
    \item \ding{174} vectors in the tensor other than the retrieved vector.
\end{itemize}
Hence, this is a ``\emph{higher order}'' relationship. This ``\emph{higher order}'' relationship that makes full use of the tensor structure also brings many challenges to the algorithm derivation, which will be introduced in the following subsections.
    
\end{remark}

\begin{figure}[!tb]
	\centering
	\subfloat{\includegraphics[width=3.2in]{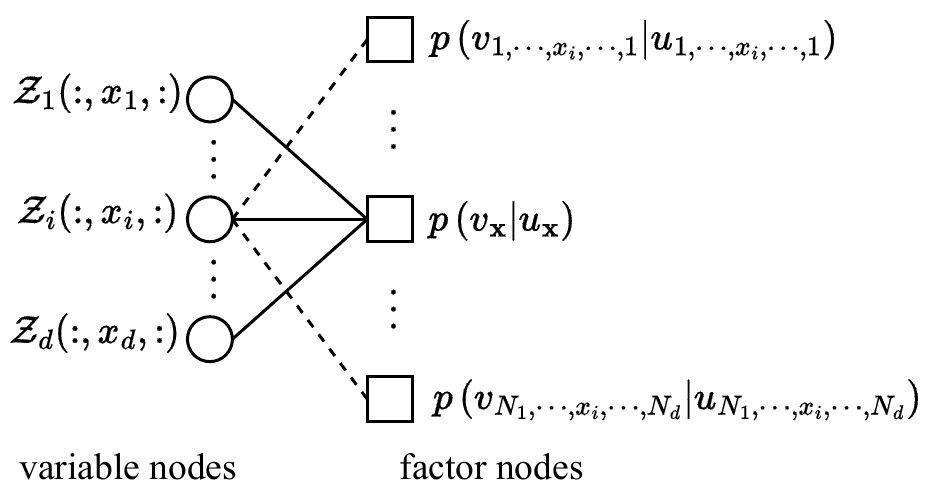}}
	\caption{An illustration of the factor-graph for tensor generalized inference problem based on TR decomposition. The factor nodes $\{ p({v_{\mathbf{x}} | u_{\mathbf{x}}}) \}$ are represented as `boxes', and the number of factor nodes is $\prod_{i}{N_i}$. The variable nodes $\{ \mathcal{Z}_i (:,x_i,:)\}$ are represented as `circles', and the number of variable nodes is $\sum_{i}{N_i}$. Each factor node $p\left({ v_{\mathbf{n}} | u_{\mathbf{n}}}\right)$ is connected to $d$ variable nodes $\{ \mathcal{Z}_{i}(:,x_i,:):i=1,...,d\}$, and each variable node $\mathcal{Z}_{i}(:,x_i,:)$ is connected to $\prod_{i'=1}^{d}{N_{i'}}/N_{i}$ factor nodes $\{p\left({ v_{\mathbf{x}'} | u_{\mathbf{x}'}}\right):x'_i= x_i\}$.}
	\label{figure:TR-AMP}
\end{figure}

\subsection{Sum-Product Algorithm}

 With loops, the updating rules of SPA are based on stipulating that the belief emitted by a variable node along a given edge of the graph is computed as the product of the incoming beliefs from all of the other edges, whereas the belief emitted by a factor node along a given edge is computed as the integral of the product of the factor corresponding to that node and the incoming beliefs on all other edges. The product of all the beliefs affecting a given variable node yields the posterior pdf for that variable \cite{parker2014bilinear}.

Based on the SPA in \cite{parker2014bilinear}, \cite{mackay2003information} and the factor-graph in Figure~\ref{figure:TR-AMP}, we derive the following updating rules for two types of messages (see Appendix~\ref{appendixA})
\begin{align} 
\label{sum-pro-1-0}
&p_{\mathbf{x} \rightarrow x_{i}}(\mathcal{Z}_{i}(:,x_{i},:)) \nonumber \\
=&~\int_{{\left\{\mathcal{Z}_{i'}(:,x_{i'},:)~:~i'\neq i\right \}}}    
\Bigg\{p\left(v_{\mathbf{x}}| \sum_{\mathbf{l}=1}^{\mathbf{r}} 
\prod_{i'=1}^{d} {\mathcal{Z}_{i'}(:,x_{i'},:)} \right) \nonumber \\
&~ \times \prod_{\{i':i' \neq i\}} p_{\mathbf{x} \leftarrow x_{i'}} \Bigg\},
\end{align}
and
\begin{multline}
   \label{sum-pro-2-0}
p_{\mathbf{x} \leftarrow x_{i}}(\mathcal{Z}_{i}(:,x_{i},:))  \\
=p\left(\mathcal{Z}_{i}(:,x_{i},:)\right) \prod_{\{\mathbf{x}':x'_i = x_{i},~ \mathbf{x}' \neq \mathbf{x}\}}p_{\mathbf{x}' \rightarrow x_{i}}, 
\end{multline}
where $\{\mathcal{Z}_{i'}(:,x_{i'},:):i'\neq i \}$ includes $d-1$ integral variables, and $x'_i$ is the $i$-th element of $\mathbf{x}'$. So, $\{\mathbf{x}':x'_i=x_i\}$ indicates $\{x_1',x_2',\cdots,x_i,\cdots,x_d'\}$, which is over ``all elements except for $x_i$''. We also indicate the mean and variance of the pdf $p_{\mathbf{x} \leftarrow x_i}(\mathcal{Z}_{i}(:,x_{i},:))$ by $\hat{\mathcal{Z}}_{i}^{\mathbf{x} \leftarrow x_i}(:,x_i,:)$ and $\nu^{\mathbf{x} \leftarrow x_i}$. In the following, we approximate the two messages in \eqref{sum-pro-1-0} and \eqref{sum-pro-2-0}.

\subsection{Message Approximation}

For high-dimensional inference problems, the exact implementation of the SPA is not practical, thus approximations of the SPA are needed. 
The derivations of our TeG-AMP are based on: 1) central limit theorem; 2) Taylor series arguments, which turn to be exact in the large system limits where $r_1,\cdots,r_d, d, N_1,\cdots, N_d \rightarrow \infty$ with fixed non-zero ratios between any two of them; 3) ignoring some terms whose correlations are disappeared to the others as $r_1,\cdots,r_d, d, N_1,\cdots, N_d \rightarrow \infty$.

In order to approximate, messages are converted to log-pdf forms. For example, the message $p_{\mathbf{x} \rightarrow x_i}$ is converted to $\frac{1}{C} \exp[p_{\mathbf{x} \rightarrow x_i}]$, where the scale factor $C$ assures that the integral of the pdf is equal to $1$. Hence, the updating rules for two types of messages in \eqref{sum-pro-1-0} and \eqref{sum-pro-2-0} are converted to
\begin{align} 
\label{sum-pro-1}
&p_{\mathbf{x} \rightarrow x_i}(\mathcal{Z}_{i}(:,x_{i},:)) \nonumber \\
=& ~ \log  
\int_{{\left\{\mathcal{Z}_{i'}(:,x_{i'},:):i'\neq i\right \}}}    
\Bigg\{p\left(v_{\mathbf{x}}| \sum_{\mathbf{l}=1}^{\mathbf{r}} 
\prod_{i'=1}^{d} \mathcal{Z}_{i'}(:,x_{i'},:) \right) \nonumber \\
&~ \times \prod_{\{i':i' \neq i\}}  \exp\left[p_{\mathbf{x} \leftarrow x_{i'}} \right] \Bigg\}+\text{const},
\end{align}
and
\begin{align}
\label{sum-pro-2}
p_{\mathbf{x} \leftarrow x_i}(\mathcal{Z}_{i}(:,x_{i},:))
=&~\text{const} + \log p\left(\mathcal{Z}_{i}(:,x_i,:)\right)\nonumber \\
&+ \sum_{\{\mathbf{x}':x'_i = x_i,~\mathbf{x}' \neq \mathbf{x}\}}p_{\mathbf{x}' \rightarrow x_i}.
\end{align}
Define the mean and variance of the pdf $\frac{1}{C} \exp [p_{\mathbf{x} \leftarrow x_i}(\mathcal{Z}_{i}(:,x_{i},:))]$ as $ \hat{\mathcal{Z}}_{i}^{\mathbf{x} \leftarrow x_i}(:,x_i,:)$ and $\nu^{\mathbf{x} \leftarrow x_i}$. Based on the factor-graph in Figure~\ref{figure:TR-AMP}, the estimate of $p( \mathcal{Z}_{i}(:,x_i,:) | \mathcal{V} )$ takes the form of $\frac{1}{C} \exp p_{x_i} (\cdot)$, where
\begin{align}
\label{sum-pro-3}
p_{x_i}\left(\mathcal{Z}_{i}(:,x_i,:) \right) &= \text{const} + \log p\left(\mathcal{Z}_{i}(:,x_i,:)\right) \nonumber \\
& + \sum_{\{\mathbf{x}':x'_i=x_i\}} p_{\mathbf{x} ' \rightarrow x_i} (\mathcal{Z}_{i}(:,x_i,:)).
\end{align}

We start by approximating the message $p_{\mathbf{x} \rightarrow x_i}(\mathcal{Z}_{i}(:,x_i,:))$. Denote the mean and variance of $\frac{1}{C} \exp [p_{x_i}\big(\mathcal{Z}_{i}(:,x_{i},:)\big)]$ as $ \hat{\mathcal{Z}}_{i}^{x_i}(:,x_i,:)$ and $\nu^{x_i}$, respectively.
Define ${{\mathcal{Z}}_{i, \mathbf{x}}^{\ell}(t)} \triangleq {{\mathcal{Z}}_{i, \mathbf{x}}^{\ell_i,\ell_{i+1}}(t)}$ and $\nu_{i,\mathbf{x} }^{\mathcal{Z},\ell} \triangleq \nu_{i,\mathbf{x} }^{\mathcal{Z},\ell_i, \ell_{i+1}}$ for short and define
\begin{align}
\label{p-x-1}
\hat{p}_{\mathbf{x} }(t) & \triangleq  \sum_{\mathbf l' = 1}^{\mathbf r}  
{\hat{\mathcal{Z}}_{1, \mathbf{x}}^{\ell'}(t)} \cdots {\hat{\mathcal{Z}}_{d, \mathbf{x}}^{\ell'}(t)},\\
\label{v-p-1}
{\nu}^{p}_{\mathbf{x} }(t) &\triangleq  \sum_{\substack{\mathbf l' = 1 }}^{\mathbf r}  
\sum_{ \substack {A \subset \mathbb{D} \\ A \neq \emptyset} } \left( 
\prod_{i' \in A} \nu_{i',\mathbf{x} }^{\mathcal{Z},\ell'}(t)   
\prod_{i'' \in \mathbb{D} \setminus A}  {\hat{\mathcal{Z}}_{i'',\mathbf{x} }^{\ell_{i''}', \ell_{i'' + 1}' } (t) }^2 \right),
\end{align}
and
\begin{align}
\label{hat-u-1}
\hat{u}_{\mathbf{x} }(t)  &\triangleq \mathbb{E} \left\{ \mathsf{u}_{\mathbf{x} } | \mathsf{p}_{\mathbf{x} } = \hat{p}_{\mathbf{x} }(t) ; \nu_{\mathbf{x} }^{p}(t)\right\}, \\
\label{nu-u-1}
 \nu_{\mathbf{x} }^{u}(t) & \triangleq \mathrm{Var}\left\{\mathsf{u}_{\mathbf{x} } | \mathsf{p}_{\mathbf{x} }=\hat{p}_{\mathbf{x} }(t) ; \nu_{\mathbf{x} }^{p}(t)\right \},
 \end{align}
computed according to the conditional pdf
\begin{equation}
\label{condtional-pdf-1}
p_{\mathsf{u}_{\mathbf{x} } | \mathsf{p}_{\mathbf{x} }} \left(u | \hat{p} ; \nu^{p} \right) \triangleq \frac{
p_{\mathsf{v}_{\mathbf{x} } | \mathsf{u}_{\mathbf{x} }} \left( v_{\mathbf{x} } | u_{\mathbf{x} } \right)
\mathcal{N}( u_{\mathbf{x} } ;  \hat{p}_{\mathbf{x} }(t), \nu_{\mathbf{x} }^p(t) )
}{\int_{u^{\prime}} p_{\mathsf{v}_{\mathbf{x} } | \mathsf{u}_{\mathbf{x} }} \left( v_{\mathbf{x} } | u' \right)
\mathcal{N}(u' ; \hat{p}_{\mathbf{x} }(t), \nu_{\mathbf{x} }^p(t))}.
 \end{equation}
 Define $\hat{\mathcal{Z}}^{\ell'}_{\setminus i, {\mathbf{x}}}(t)$ $\hat{\mathcal{Z}}^{\ell'}_{\setminus i, x}(t)$ and $\zeta_{x}(t)$ as given in Appendix~\ref{appendixB}. 
Assume that the size of $\mathcal{Z}$ is large enough, then the following approximation of $p_{\mathbf{x} \rightarrow [i,\ell_i,\ell_{i+1},x_i]}(\mathcal{Z}_{i}(\ell_i,x_i,\ell_{i+1}))$ in \eqref{sum-pro-1} holds:
\begin{align} 
\label{sum-pro-Taylor-1}
&~{ p_{\mathbf{x} \rightarrow [i,\ell_i,\ell_{i+1},x_i]} \left(t,\mathcal{Z}_{i}(\ell_i,x_i,\ell_{i+1}) \right)} 
\approx  {\text{const}}  \\
& + \Big[
 {\sum_{\mathbf{l}'_{\setminus i} = 1}^{\mathbf{r}_{\setminus i}}}
  {\hat{\mathcal{Z}}^{\ell'}_{\setminus i, \mathbf{x}}(t)}
\Big] \mathcal{Z}_i^{\ell} (x_i) 
 \hat{s}_{\mathbf{x} }(t) + 
 \frac{1}{2} {\omega}_{\mathbf{x} }(t){\mathcal{Z}_i^{\ell} (x_i) }^2 {\zeta}_{x}(t) \nonumber \\
 &-  \frac{ 1 }{2} \Big[ 
 {\sum_{\mathbf{l}'_{\setminus i} = 1}^{\mathbf{r}_{\setminus i}}}
 {\hat{\mathcal{Z}}^{\ell'}_{\setminus i, x}(t)}
\Big]^2  \left[\mathcal{Z}_i^{\ell} (x_i)^2 - 2 \hat{\mathcal{Z}}^{\ell}_{i,x_i}(t) \mathcal{Z}_i^{\ell} (x_i) \right] \nu_{\mathbf{x} }^s(t), \nonumber 
\end{align}
where $\mathbf r_{\setminus i} \triangleq  \mathbf r \backslash (r_i, r_{i+1})$, $\mathbf l' _{\setminus i} \triangleq \mathbf l' \backslash (\ell_i', \ell_{i+1}')$, and
\begin{align}
\label{hat-s-2}
\hat{s}_{\mathbf{x} }(t) &= \frac{1}{ \nu_{\mathbf{x} }^p(t) } \left(\hat{u}_{\mathbf{x} }(t)-\hat{p}_{\mathbf{x} }(t)\right), \\
\label{nu-s-2}
{\nu}_{\mathbf{x} }^{s}(t) &= \frac{1}{\nu_{\mathbf{x} }^{p}(t)}\left(1-\frac{\nu_{\mathbf{x} }^{u}(t)}{\nu_{\mathbf{x} }^{p}(t)}\right),
\\
\label{bias}
{\omega}_{\mathbf{x} }(t) &= {\hat{s}_{\mathbf{x} }(t)}^2  -  {\nu_{\mathbf{x} }^s(t)}.
\end{align}

\begin{remark}[\textbf{Structure}]
    $\hat{\mathcal{Z}}^{\ell}_{i,x_i}(t)$, ${\hat{\mathcal{Z}}^{\ell'}_{\backslash i, x}(t)}$, and ${\hat{\mathcal{Z}}^{\ell'}_{\backslash i, \mathbf{x}}(t)}$ in \eqref{sum-pro-Taylor-1} corresponds to the structures \ding{172}, \ding{173}, and \ding{174} in Remark~\ref{tensor-structure}, respectively. Full tensor structures is used. 
\end{remark}


Equation \eqref{sum-pro-Taylor-1} enables us to approximate the message $p_{\mathbf x \rightarrow [i,\ell_i,\ell_{i+1},x_i]} \left(t,\mathcal{Z}_{i}(\ell_i,x_i,\ell_{i+1}) \right)$ once \eqref{hat-s-2} and \eqref{nu-s-2} are obtained, which is derived in Appendix~\ref{appendixB}. In fact, \eqref{condtional-pdf-1} is TeG-AMP's the $t$-th iteration approximation to the true marginal posterior $p_{\mathsf{u}_{\mathbf{x} } | \mathsf V }(\cdot | \mathcal{V})$. We note that \eqref{condtional-pdf-1} can also be interpreted as the posterior pdf for $\mathsf{u}_{\mathbf{x} }$ given the likelihood $p_{\mathsf{v}_{\mathbf{x} } | \mathsf{u}_{\mathbf{x} } } \left( v_{\mathbf{x} } | \cdot \right)$ from \eqref{likelihood-decoupled} and the prior $\mathsf{u}_{\mathbf{x} } \sim \mathcal{N}( \hat{p}_{\mathbf{x} }(t) , \nu_{\mathbf{x} }^p(t) )$ that is implicitly assumed by the $t$-th iteration TeG-AMP.

We now turn to approximate the messages flowing from the variable nodes to the factor nodes. Assume that the size of $\mathcal{Z}$ is large enough,
$p_{\mathbf x \leftarrow [i,\ell_i, \ell_{i+1},x_i] } \left(t+1,\mathcal{Z}_{i}(\ell_i,x_i,\ell_{i+1}) \right)$ in \eqref{sum-pro-2} can be approximated by
\begin{align} 
\label{var-to-fac-1}
&p_{\mathbf x \leftarrow [i,\ell_i, \ell_{i+1},x_i] } \left(t+1,\mathcal{Z}_{i}(\ell_i,x_i,\ell_{i+1}) \right) \ {\approx}\ \text{const}\ + \nonumber \\
& \log\Big( p\left(\mathcal{Z}_{i}^{\ell}(x_i)\right) \mathcal{N}\left({\mathcal{Z}_i^{\ell} (x_i) };\hat{r}_{\mathbf{x} , \mathbf q^i},  \nu^r_{\mathbf{x} , \mathbf q^i} \right) \Big),
\end{align}
where ${{\mathcal{Z}}_{i}^{\ell}} \triangleq {{\mathcal{Z}}_{i}^{\ell_i,\ell_{i+1}}}$ and
\begin{align*}
\frac{1}{\nu^r_{\mathbf{x} ,\mathbf q^i}}  & \triangleq   \sum_{\substack{{ {\mathbf{x}':x'_i = x_i}}\\ \mathbf{x} ' \neq \mathbf{x} }}
\Bigg\{    
 \Big[ 
 {\sum_{\mathbf{l}'_{\setminus i} = 1}^{\mathbf{r}_{\setminus i}}}
 {\hat{\mathcal{Z}}^{\ell'}_{\setminus i, x'}(t)}
\Big]^2
  \nu_{\mathbf{x} '}^s(t) -  
  {\omega}_{\mathbf{x} '}(t) {\zeta}_{x}(t) \Bigg\} \\
\hat{r}_{\mathbf{x} , \mathbf q^i} &\triangleq \nu^r_{\mathbf{x} ,\mathbf q^i}  
 \sum_{\substack{{ {\mathbf{x}':x'_i = x_i}},\\ \mathbf{x} ' \neq \mathbf{x} }}
\Bigg\{  
\Big[
 {\sum_{\mathbf{l}'_{\setminus i} = 1}^{\mathbf{r}_{\setminus i}}}
 {\hat{\mathcal{Z}}^{\ell'}_{\setminus i, \mathbf{x}'}(t)}
\Big]
 \hat{s}_{\mathbf{x} '}(t) 
  \Bigg\} 
  \nonumber \\
 &~~~~~~~~~~~
 + \hat{\mathcal{Z}}^{\ell}_{i,x_i}(t)
 \Bigg\{ 1 + \nu^r_{\mathbf{x} ,\mathbf q^i}
 \sum_{\substack {{ {\mathbf{x}':x'_i = x_i}},\\ \mathbf{x} ' \neq \mathbf{x} }}
   {\omega}_{\mathbf{x} '}(t) {\zeta}_{x}(t) \Bigg\}.
\end{align*}
which is derived in Appendix~\ref{appendixC}.

The mean and variance of the pdf associated with the approximation in \eqref{var-to-fac-1} are
\begin{align}
\label{z-ix-mean-1}
& \hat{\mathcal{Z}}_{i,\mathbf{x} }^{\ell}(t+1) \triangleq g(\hat{r}_{\mathbf{x} , \mathbf q^i}, \nu^r_{\mathbf{x} ,\mathbf q^i}),\\
\label{var-z-ix-1}
& \nu_{i,\mathbf{x} }^{\mathcal{Z},\ell}(t+1) \triangleq \nu^r_{\mathbf{x} ,\mathbf q^i} g' (\hat{r}_{\mathbf{x} , \mathbf q^i}, \nu^r_{\mathbf{x} ,\mathbf q^i}) ,
\end{align}
respectively, where
\begin{align}    
g(\hat{r}  , 
   \nu^r) &\triangleq \frac{1}{C} 
\int_{x} x ~ p ( x )
  \mathcal{N}\left( x; \hat{r}, \nu^r \right), \nonumber \\   
C & \triangleq \int_{x}  p ( x )
  \mathcal{N}\left( x;  \hat{r}_{\mathbf{x} , \mathbf q^i}  ,  \nu^r_{\mathbf{x} ,\mathbf q^i}   \right) \nonumber
\end{align}  
   and $g'$ denotes the derivative of $g$ with respect to the first argument. 
   
\subsection{Completing the Loop of TeG-AMP Algorithm}

Based on the message approximation results, we next show how to complete the loop of TeG-AMP algorithm.
First, we have 
    $\hat{\mathcal{Z}}_{i,\mathbf{x} }^{\ell}(t+1)$ can be approximated by
    \begin{align} 
    \label{z-ix-mean-3}
&\hat{\mathcal{Z}}_{i,\mathbf{x} }^{\ell}(t+1)   =
 g 
 \left(
 \hat{r}_{\mathbf{x} , \mathbf q^i}  , 
   \nu^r_{\mathbf{x} ,\mathbf q^i}
   \right) \nonumber \\
{\approx}  &~ 
\hat{\mathcal{Z}}_{i,x_i}^{\ell}(t+1)  
-    \hat{s}_{\mathbf{x} }(t)  \nu_{i,x_i}^{{ {\mathcal{Z}}},\ell}(t+1)
\times 
  {\sum_{\mathbf{l}'_{\setminus i} = 1}^{\mathbf{r}_{\setminus i}}}
 {\hat{\mathcal{Z}}^{\ell'}_{\setminus i, x}(t)},
\end{align}
where $\hat{\mathcal{Z}}_{i,x_i}^{\ell} \triangleq  g
(\hat{r}_{\mathbf q^i} ,  \nu^r_{\mathbf q^i})$ and $\nu_{i,x_i}^{{ {\mathcal{Z}}},\ell}
 \triangleq \nu^r_{\mathbf q^i} 
  g'(\hat{r}_{\mathbf q^i}, \nu^r_{\mathbf q^i})$ with
\begin{align*}
\frac{1}{\nu^r_{\mathbf q^i}}  
&\triangleq  \sum_{{ {\mathbf{x}':x'_i = x_i}}}  
\Bigg\{    
 \Big[ 
 {\sum_{\mathbf{l}'_{\setminus i} = 1}^{\mathbf{r}_{\setminus i}}}
 {\hat{\mathcal{Z}}^{\ell'}_{\setminus i, x'}(t)}
\Big]^2
\nu_{\mathbf{x} '}^s(t) -  
  {\omega}_{\mathbf{x} '}(t) {\zeta}_{x}(t)
  \Bigg\}  , \\
\hat{r}_{\mathbf q^i} &\triangleq  \nu^r_{\mathbf q^i}  
 \sum_{{ {\mathbf{x}':x'_i = x_i}}}  
\Bigg\{  
\Big[
 {\sum_{\mathbf{l}'_{\setminus i} = 1}^{\mathbf{r}_{\setminus i}}}
 {\hat{\mathcal{Z}}^{\ell'}_{\setminus i, \mathbf{x}'}(t)}
\Big]
 \hat{s}_{\mathbf{x} '}(t) 
  \Bigg\}  
  \nonumber \\
 &~~~+ \hat{\mathcal{Z}}^{\ell}_{i,x_i}(t) 
 \Bigg\{ 1 + \nu^r_{\mathbf q^i}   
 \sum_{{ {\mathbf{x}':x'_i = x_i}}}     
   {\omega}_{\mathbf{x} '}(t) {\zeta}_{x}(t)
  \Bigg\},
\end{align*}
which is derived in Appendix~\ref{appendixD}.

\begin{remark}[\textbf{Structure}]
In \eqref{var-to-fac-1} and \eqref{z-ix-mean-3}, we distinguish between $\hat{r}_{\mathbf{x} , \mathbf q^i}$ and $\hat{r}_{\mathbf q^i}; {1}/{ \nu^r_{\mathbf{x} ,\mathbf q^i}}$ and ${1}/{\nu^r_{\mathbf q^i}}$, corresponding to exploiting the structures \ding{173} and \ding{174} in Remark~\ref{tensor-structure}. Similarly, distinguish $\sum_{\substack{{ {\mathbf{x}':x'_i = x_i}}, \mathbf{x} ' \neq \mathbf{x} }}$ and $\sum_{\substack{{ {\mathbf{x}':x'_i = x_i}}}}$ is required.
\end{remark}

The penultimate step in the derivation of TeG-AMP algorithm is to approximate some earlier steps that use $\hat{\mathcal{Z}}_{i,\mathbf{x} }^{\ell} (t)$ in place of $\hat{\mathcal{Z}}_{i,x_i}^{\ell} (t)$. We present the following results, which are derived in Appendix~\ref{appendixE}.

$\hat{p}_{\mathbf{x} }(t)$ in \eqref{p-x-1} can be approximated by
\begin{align} 
\label{p-x-2}
\hat{p}_{\mathbf{x} }(t) =&~ \sum_{\mathbf{l}' = 1}^{\mathbf{r}}  
{\hat{\mathcal{Z}}_{1, \mathbf{x}}^{\ell'}(t) }  
\cdots {\hat{\mathcal{Z}}_{d, \mathbf{x}}^{\ell'}(t) } 
 \nonumber \\
{\approx} &~ \bar p_{\mathbf{x} }(t) 
- \hat{s}_{\mathbf{x} }(t-1)  \bar\nu^p_{\mathbf{x} }(t), 
\end{align}
where
\begin{align} 
\label{bar-var-p-1}
\bar\nu^p_{\mathbf{x} }(t) =&~  \sum_{\mathbf{l}' = 1}^{\mathbf{r}}  
 \sum_{\substack{A \subsetneqq \mathbb{D} \\ A\neq \emptyset} }
 \Bigg\{ 
 \Big(
 \prod_{i \in A}  \hat{\mathcal{Z}}_{i, x_{i}}^{\ell'}(t) 
 \Big)^2
 \times \prod_{i'' \in \mathbb{D} \setminus A}
 \nu_{i'',x_{i''}}^{\mathcal{Z},\ell'}(t) \nonumber \\
 &
 \Big(
 - \hat{s}_{\mathbf{x} }(t-1) 
  \prod_{i' \in \mathbb{D}} 
 \hat{\mathcal{Z}}^{\ell'}_{i',x_{i'}}(t)
 \Big)^{d-|A|-1} 
 \Bigg\}. 
\end{align}
${\nu}^{p}_{\mathbf{x} }(t)$ in \eqref{v-p-1} can be approximated by
\begin{align} 
\label{v-p-2}
{\nu}^{p}_{\mathbf{x} }(t) =&~ \sum_{\substack{\mathbf{l}'= \mathbf{1} }}^{\mathbf r}  
\sum_{\substack {A \subset \mathbb{D} \\ A \neq \emptyset}}
\left( \prod_{i' \in A} \nu_{i',\mathbf{x} }^{{ {\mathcal{Z}}},\ell'}(t)    
\prod_{i'' \in \mathbb{D} \setminus A }  {\hat{\mathcal{Z}}_{i'',\mathbf{x} }^{\ell' } (t) }^2 \right) \nonumber \\
 {\approx}&~
 \sum_{\substack{\mathbf{l} = \mathbf{1}}}^{\mathbf r}  
\prod_{i \in \mathbb{D}} \nu_{i,x_{i}}^{{ {\mathcal{Z}}},\ell}(t) + \bar\nu^{p}_{\mathbf{x}}(t).
\end{align}
$\nu^r_{\mathbf q^i}$ and $\hat{r}_{\mathbf q^i}$ in \eqref{z-ix-mean-3} can be respectively approximated by
\begin{align} 
\label{var-r-approx-1}
&\frac{1}{
  \nu^r_{\mathbf q^i}}
  \approx
 \sum_{{ {\mathbf{x}':x'_i = x_i}}}  
\Bigg\{    
 \Big[ 
 {\sum_{\mathbf{l}'_{\setminus i} = 1}^{\mathbf{r}_{\setminus i}}}
 {\hat{\mathcal{Z}}^{\ell'}_{\setminus i, x'}(t)}
\Big]^2
\nu_{\mathbf{x} '}^s(t)  
\Bigg\},
\end{align}
and
\begin{align} 
\label{hat-r-3}
&\hat{r}_{\mathbf q^i}
~{\approx}~
\nu^r_{\mathbf q^i}  
 \sum_{{ {\mathbf{x}':x'_i = x_i}}}  
\Bigg\{  
\Bigg[ 
 {\sum_{\mathbf{l}'_{\setminus i} = 1}^{\mathbf{r}_{\setminus i}}}
 {\hat{\mathcal{Z}}^{\ell'}_{\setminus i, x'}(t)}
\Bigg]
 \hat{s}_{\mathbf{x} '}(t) 
  \Bigg\}  +
  \\
  & \hat{\mathcal{Z}}^{\ell}_{i,x_i}(t) \left\{ 1 -
\frac{
 \sum\limits_{{ {\mathbf{x}':x'_i = x_i}}}     
   {\nu_{\mathbf{x} '}^s(t)}  
 {\zeta}_{x}(t)}
 {\sum\limits_{{ {\mathbf{x}':x'_i = x_i}}}  
\Bigg\{    
 \left[ 
 {\sum_{\mathbf{l}'_{\setminus i} = 1}^{\mathbf{r}_{\setminus i}}}
 {\hat{\mathcal{Z}}^{\ell'}_{\setminus i, x'}(t)}
\right]^2
\nu_{\mathbf{x} '}^s(t)  
\Bigg\} 
}
\right \}.\nonumber 
\end{align}


\subsection{Approximated Posteriors}

The final step in the TeG-AMP derivation is to approximate the SPA posterior log-pdfs in \eqref{sum-pro-3}. Plugging \eqref{sum-pro-Taylor-1} into those expressions, we get
\begin{align} 
\label{fac-1}
&~ p_{x_i}(t+1,\mathcal{Z}_{i}^{\ell}(x_i))  \nonumber \\
\approx &~  \text{const}  + \log \Big(  p\left(\mathcal{Z}_{i}^{\ell}(x_i)\right) \mathcal{N}\left({ { {\mathcal{Z}}}_i^{\ell} (x_i) }; \hat{r}_{\mathbf q^i}  ,  \nu^r_{\mathbf q^i}  \right)   \Big),
\end{align}
using steps similar to \eqref{var-to-fac-1}. The associated pdfs are
\begin{align} 
\label{ass-pdf-1}
&~p
(\mathcal{Z}_{i}^{\ell} \left(x_{i}\right) | \hat{r}_{\mathbf q^i} ; \nu_{\mathbf q^i}^{r})
\nonumber \\
\triangleq &~ 
p
\left(  \mathcal{Z}_{i}^{\ell} \left(x_{i}\right)  \right) \times
\frac{
\mathcal{N}( \mathcal{Z}_{i}^{\ell} \left(x_{i}\right) ; \hat{r}_{\mathbf q^i} , \nu_{\mathbf q^i}^{r})}
{
\int_{z^{\prime}} 
p
( \mathcal{Z}^{\prime}) 
\mathcal{N}(\mathcal{Z}^{\prime} ; \hat{r}_{\mathbf q^i}, \nu_{\mathbf q^i}^{r})
},
\end{align}
which are the $t$-th iteration approximations to the true marginal posteriors $p(\mathcal{Z}_{i}^{\ell} \left(x_{i}\right) | \mathcal{V} )$. Note that $\hat{\mathcal{Z}}_{i,x_i}^{\ell}(t+1)$ and $ \nu_{i,x_i}^{\mathcal{Z},\ell}(t+1)$ are the mean and variance of the posterior pdf in \eqref{ass-pdf-1}, respectively. Hence, \eqref{ass-pdf-1} can be interpreted as the posterior pdf of ${\mathsf z}_{i}^{\ell}(x_i)$ given the observation $ \mathsf{r}_{\mathbf q^i} = \hat{r}_{\mathbf q^i}$ under the prior model ${\mathsf z}_{i}^{\ell}(x_i) \sim p_{{\mathsf z}_{i}^{\ell}(x_i)}$ and the likelihood model 
\begin{align} 
\label{fac-2}
&p
(\mathcal{Z}_{i}^{\ell} \left(x_{i}\right) | \hat{r}_{\mathbf q^i} ; \nu_{\mathbf q^i}^{r})
= \mathcal{N}( \mathcal{Z}_{i}^{\ell} \left(x_{i}\right) ; \hat{r}_{\mathbf q^i}, \nu_{\mathbf q^i}^{r})
\end{align}
implicitly assumed by the $t$-th iteration TeG-AMP.
\color{black}


\subsection{Algorithm \& Complexity \& Damping}

The TeG-AMP algorithm derived is summarized in Algorithm~\ref{tab:A1}, which is given in Appendix~\ref{appendixE1}. The complexity order is dominated by \eqref{bar-var-p-1}, as the traverse addition and multiplication. So, the complexity of Algorithm~\ref{tab:A1} is $O(d^4 \prod_{i = 1}^{d}{r_i})$. Since $r_i$ is generally small for low-rank problems, the complexity is still acceptable.

In addition, the approximations, e.g., 1) central limit theorem; 2) Taylor series argument; and 3) ignoring some terms, made in the TeG-AMP derivation were obtained in the large system limit. In practical applications, however, these dimensions are finite, and hence the algorithm may diverge.
In this case, we can apply adaptive damping technique to TeG-AMP as described in Appendix~\ref{appendixE}.

\section{Discussions \& Extension}

Note that the TR decomposition is the most extensive tensor decomposition and can include many other tensor decompositions. In this subsection, we present the relationship between TR-decomposition and other tensor decompositions to show that our TeG-AMP algorithm is a very general algorithm, basically applicable to all tensors.

\subsection{TeG-AMP for CP decomposition}

The CP decomposition aims to represent a $d$-order tensor $\mathcal{U}\in \mathbb{R}^{N_{1} \times \ldots \times N_{d}}$ by a sum of $r$ rank-one tensors, where the $\ell$-th rank-one tensor is generated by an outer product of $\mathbf{a}_{i}^{\ell} \in \mathbb{R}^{N_{i}},~i = 1,...,d$, given by
\begin{align} 
\label{CPD}
\mathcal{U}=\sum_{\ell=1}^{r} \mathbf{a}_{1}^{\ell} \otimes \mathbf{a}_{2}^{\ell} \ldots \otimes \mathbf{a}_{d}^{\ell},
\end{align}
or equivalently,
\begin{align} 
\label{CPD-2}
u_{x_{1}, x_{2}, \ldots, x_{d}}=\sum_{\ell=1}^{r} \mathbf{a}_{1}^{\ell}\left(x_{1}\right) \mathbf{a}_{2}^{\ell}\left(x_{2}\right) \ldots \mathbf{a}_{d}^{\ell}\left(x_{d}\right).
\end{align}
The number $r$ of rank-one tensors is defined as the CP-rank.

Follows the analysis in~\cite{zhao2016tensor}, by letting $\mathcal{Z}^{x_i}_i \in \mathbb{R}^{r_i \times r_{i+1}},~i=1,\cdots,d,~x_i = 1,\cdots,N_i$ be a diagonal matrix with vector $[\mathbf{a}_{i}^{1}\left(x_{i}\right),\mathbf{a}_{i}^{2}\left(x_{i}\right),\cdots,\mathbf{a}_{i}^{r}\left(x_{i}\right)]^T \in \mathbb{R}^{r}$ on the diagonal. We can rewrite \eqref{CPD-2} as
\begin{align} 
\label{CPD-3}
u_{x_{1}, x_{2}, \ldots, x_{d}}= \mathrm{Tr} \{ \mathcal{Z}^{x_1}_1 \mathcal{Z}^{x_2}_2\cdots \mathcal{Z}^{x_d}_d \}.
\end{align}
Comparing \eqref{CPD-3} with \eqref{TR-decomposition-1} it is straightforward to see that the CP decomposition is a special case of TR decomposition where $r_{1} = \cdots = r_d = r$. Hence, given a low CP-rank tensor, our TeG-AMP algorithm could be easily implemented by setting $r_{1} = \cdots = r_d = r$.

\subsection{TeG-AMP for TT decomposition}
For a $d$-th order tensor $\mathcal{U}\in \mathbb{R}^{N_{1} \times \ldots \times N_{d}}$, the tensor train decomposition aims to represent it by a sequence of cores $\mathbf{C}_1 \in \mathbb{R}^{N_1 \times r_1}$, $\mathbf{C}_d \in \mathbb{R}^{r_d \times N_d}$ and ${\mathcal{C}_i \in \mathbb{R}^{r_i \times N_{i} \times r_{i+1} },~i=2,\cdots,d-1 }$. In particular, the TT decomposition is the element-wise form is given by
\begin{align} 
\label{TT-1}
u_{x_{1}, x_{2}, \ldots, x_{d}}=  {(\mathbf{C}_1^{x_1})}^T \mathcal{C}^{x_2}_2 \mathcal{C}^{x_3}_3\cdots \mathcal{C}^{x_{d-1}}_{d-1}  \mathbf{C}_d^{x_d},
\end{align}
where $\mathbf{C}_1^{x_1}$ denotes the $x_1$-th row vector of $\mathbf{C}_1$, $\mathbf{C}_d^{x_d}$ denotes the $x_d$-th column vector of $\mathbf{C}_d$ and $\mathcal{C}^{x_i}_i$ denotes the $x_i$-th lateral slice matrix of the $i$-th latent tensor $\mathcal{C}_i$.

Comparing \eqref{TT-1} with \eqref{TR-decomposition-1} it is straightforward to see that the TT decomposition is a special case of TR decomposition where $r_{1} = r_{d+1} = 1$, i.e., the first and last cores are matrix. Hence, given a low TT-rank tensor, our TeG-AMP algorithm could be easily implemented by setting $r_{1} = r_{d+1} = 1$.

\subsection{TeG-AMP for Tucker decomposition}

For a $d$-th order tensor $\mathcal{U}\in \mathbb{R}^{N_{1} \times \ldots \times N_{d}}$, the Tucker decomposition aims to represent it by a multilinear product between a core tensor $\mathcal{G} \in \mathbb{R}^{r_{1} \times \ldots \times r_{d}}$ and factor matrices $\mathbf{M}_i \in \mathbb{R}^{N_i \times  r_{i}},~i = 1,\cdots,d $, given by
\begin{align} 
\label{Tucker-1}
\mathcal{U}=\mathcal{G} \times_{1} \mathbf{M}_1 \times_{2} \cdots \times_{d} \mathbf{M}_d,
\end{align}
where $\times_i$ denotes the mode-$i$ product.

Assuming that the core tensor $\mathcal{G}$ can be represented by a TR decomposition based on a sequence of latent tensors $\mathcal{Y}_i,~i=1,\cdots,d$, the Tucker decomposition \eqref{TT-1} in the element-wise form is then given by~\cite{zhao2016tensor}
\begin{align} 
\label{Tucker-2}
u_{x_{1}, x_{2}, \ldots, x_{d}}=  {\mathrm{Tr}} \left\{ \prod_{i=1}^{d} \left( \sum_{\ell_i =1}^{r_i} \mathbf{Y}_i^{\ell_i} \mathbf{M}_i(x_i, \ell_i) \right)\right\},
\end{align}
where $\mathbf{Y}_i^{\ell_i}$ denotes the $\ell_i$-th lateral slice matrix of the $i$-th latent tensor $\mathcal{Y}_i$ and $\mathbf{M}_i(x_i, \ell_i)$ denotes the $(x_i, \ell_i)$-th element of $i$-th factor matrices $\mathbf{M}_i$. Under this condition, the Tucker decomposition is a special case of TR decomposition, where each latent tensor is generated by
\begin{align} 
\label{Tucker-3}
\mathcal{Z}_i = \mathcal{Y}_i \times_2  \mathbf{M}_i,~i = 1,\cdots,d.
\end{align}

Hence, given a low Tucker-rank tensor, our TeG-AMP algorithm could be intuitively implemented by setting the ranks $r_{1},\cdots, r_{d+1}$ as the dimensions of the core tensor of the Tucker decomposition.

\subsection{Extension to TeS-AMP}

Although our TeG-AMP algorithm is applicable to CP decomposition, comparing \eqref{CPD-3} with \eqref{TR-decomposition-1}, we find that the CP decomposition is a special case of TR decomposition, and it is possible to simplify the TeG-AMP algorithm based on the CP decomposition model. We hence extend our TeG-AMP and propose the tensor simplified AMP (TeS-AMP) algorithm, which is presented in Appendix~\ref{appendixH}.

\begin{figure}[!b]
	\centering
  \vspace{-0.5cm} 
	\subfloat{\includegraphics[width=1.65in]{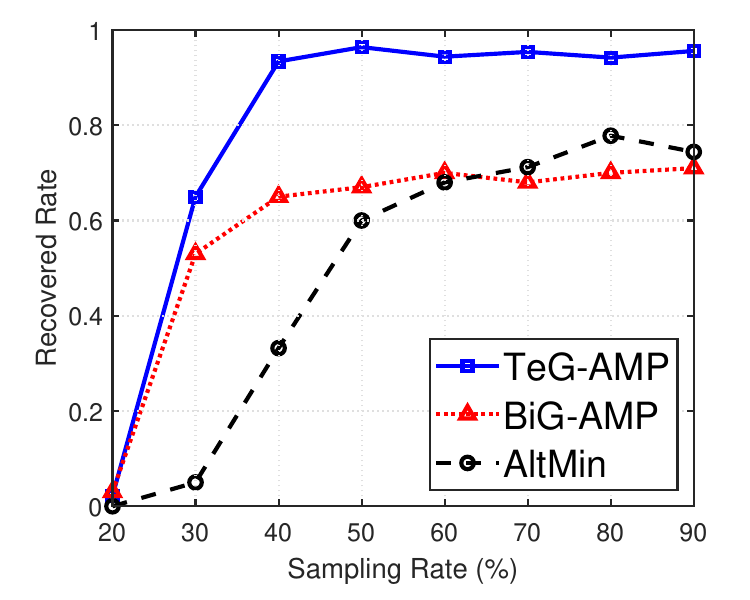}}
	\subfloat{\includegraphics[width=1.65in]{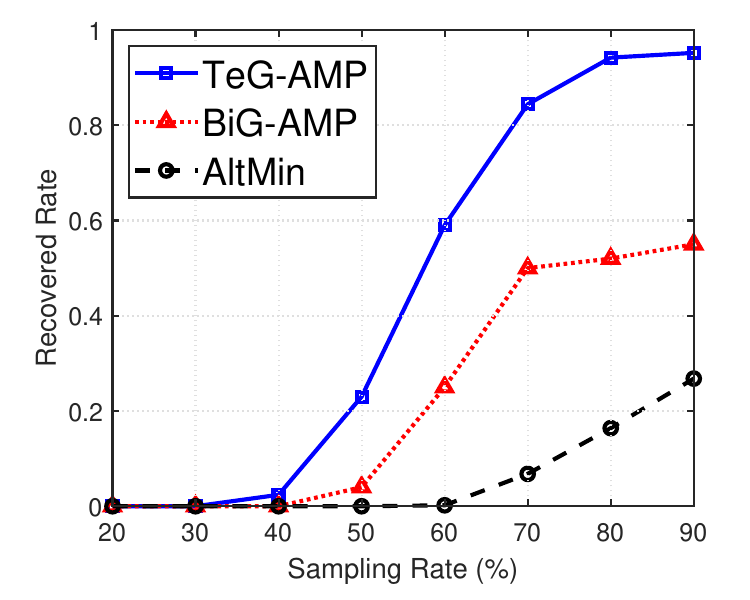}}
     \vspace{-0.3cm} 
     
        \subfloat{\includegraphics[width=1.65in]{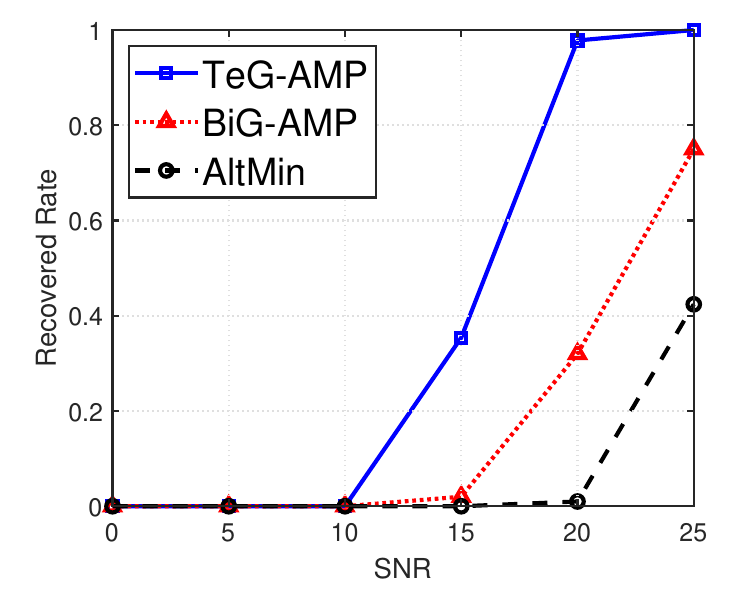}}
	\subfloat{\includegraphics[width=1.65in]{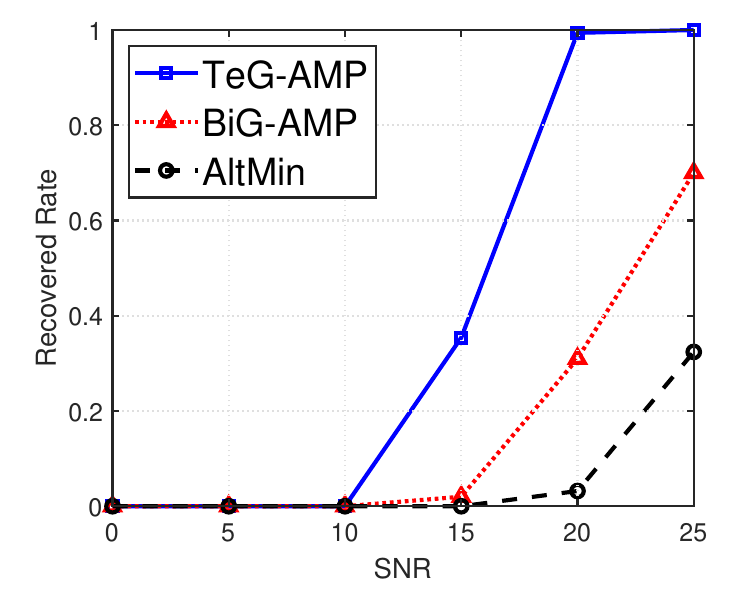}}
	
	\caption{Comparison results for TR-rank tensor $\mathcal{U} \in \mathbb{R}^{6 \times 7 \times 8}$. The sampling rate in noisy cases is $100\%$. \textbf{Upper-Left:} TR-rank $2,3,3$ in noiseless cases; \textbf{Upper-Right:} TR-rank $2,2,2$ in noiseless cases; \textbf{Below-Left:} TR-rank $2,3,3$ in noisy cases; \textbf{Below-Right:} TR-rank $2,2,2$ in noisy cases.}
        \label{figure:TRmodelNoiseless-main}
\end{figure}
\section{Experimental Results in Tensor Completion}

\begin{table*}[!htbp] 
\centering
\caption{NMSE of different digits. The best result has been bolded.}\label{table2}
\begin{tabular}{ccccccccccc}
\toprule
\multicolumn{2}{c}{Digit}&\multicolumn{1}{c}{\bsifamily{5}}& \multicolumn{1}{c}{\bsifamily{6}}& \multicolumn{1}{c}{\bsifamily{7}}& \multicolumn{1}{c}{\bsifamily{8}} \\
\rowcolor{gray!10}\multicolumn{2}{c}{Sampling Rate} & $20\%~~~ ~~~40\%~~~ ~~~60\%$ & $20\%~~~ ~~~40\%~~~ ~~~60\%$ & $20\%~~~ ~~~40\%~~~ ~~~60\%$ & $20\%~~~ ~~~40\%~~~ ~~~60\%$ \\
\midrule
\rowcolor{gray!10}\multicolumn{2}{c}{AltMin}& ${9.2409}, ~{9.8921}, ~7.1096$ & ${8.0302}, ~{7.4701}, ~3.4961$ & ${8.3517}, ~{7.7645}, ~5.6153$ & ${5.7151}, ~{5.3243}, ~5.1195$\\
\multicolumn{2}{c}{BiG-AMP} & ${2.3660}, ~{2.1803}, ~0.6384$ & ${2.5594}, ~{2.2070}, ~0.5171$ & ${6.9261}, ~{6.9849}, ~0.4866$ & ${2.1466}, ~{1.1154}, ~0.5249$\\
\rowcolor{gray!10}\multicolumn{2}{c}{\textbf{TeG-AMP}} & $\textbf{0.6986}, ~\textbf{0.4823}, ~\textbf{0.4020}$ & $\textbf{0.6422}, ~\textbf{0.4425}, ~\textbf{0.2839}$ & $\textbf{0.6166}, ~\textbf{0.4465}, ~\textbf{0.3113}$ & $\textbf{0.6148}, ~\textbf{0.4774}, ~\textbf{0.3311}$ \\
\bottomrule
\end{tabular}
\label{table_error}
\end{table*}

In this section, we present experimental results on simulated TR and CP tensor data as well as on the MNIST dataset. We compared TeG-AMP with the most typical machine-learning-based tensor completion algorithms. 
As for the Bayesian AMP algorithm in~\cite{kadmon2018statistical}, it cannot work in our scenario since it is only suitable for spiked-tensor decomposition.
Details of the experimental setup, baselines, and additional experiments can be found in Appendix~\ref{appendixG}.

\subsection{Simulation Results of Low TR-rank Tensor}


We first evaluate the performances of the proposed methods in noiseless cases. In the first example, we generate tensors $\mathcal{U} \in \mathbb{R}^{6 \times 7 \times 8}$ with rank $2,2,2$ and rank $2,3,3$ by generating its corresponding TR decomposition components randomly. Then, we sample each entry of $\mathcal{U}$ independently with probability $p_{\Omega}$, where we vary the value of $p_{\Omega}$ from $0.2$ to $0.9$. Due to the random nature of our tests, and in order to make the error curves more accurate and smooth, for each value of $p_{\Omega}$ we run $500$ random tests and calculate the average error for each of the four mentioned methods, where the error is defined as
$\epsilon= {\left\|\left(\mathcal{U}^{*} - \mathcal{U}\right)_{\bar{\Omega}}\right\|_{\mathcal{F}}} / {\left\| \mathcal{U}_{\bar{\Omega}}\right\|_{\mathcal{F}}}$,
where $\bar\Omega$ is the complement of $\Omega$, i.e., the set of indices corresponding to the non-observed entries. We take the test result as recovered when the error of one test is less than 0.01. The recovered rate is defined as the number of recovered test results divided by the number of tests.

\begin{figure}[!tb]
	\centering
	\subfloat{\includegraphics[width=3.25in]{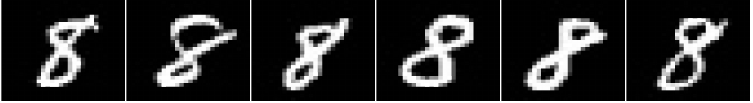}}
    \vspace{-0.3cm} 
    
	\subfloat{\includegraphics[width=3.25in]{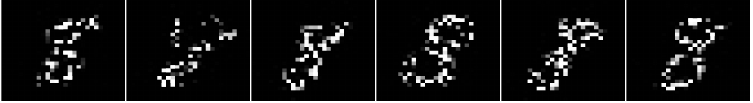}}
 \vspace{-0.3cm} 

	\subfloat{\includegraphics[width=3.25in]{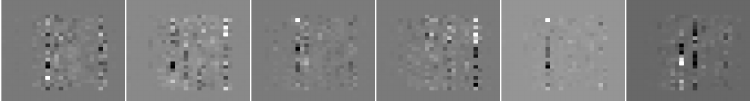}}
 \vspace{-0.3cm} 
	
	\subfloat{\includegraphics[width=3.25in]{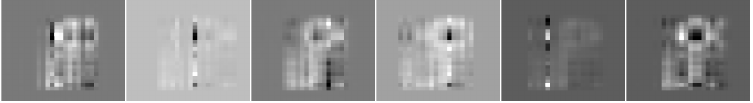}}
 \vspace{-0.3cm} 
		
	\subfloat{\includegraphics[width=3.25in]{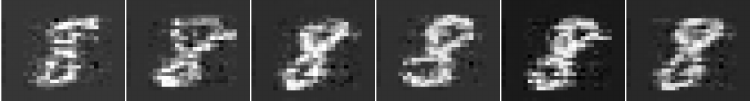}}
	
	\caption{MNIST digits with size $28\times 28 \times 6$ and TR rank $14\times 14 \times 6$. The sampling rate is $40\%$. \textbf{Line~1:} Ground truth; \textbf{Line~2:} Sampling Results; \textbf{Line~3:} Recovered digits via AltMin; \textbf{Line~4:} Recovered digits via BiG-AMP; \textbf{Line~5:} Recovered digits via TeG-AMP.}
	\label{figure:MovingChar}
\end{figure}

The error curves for the simulation of TR-rank are shown in Figure~\ref{figure:TRmodelNoiseless-main}. We can see that the proposed TeG-AMP method performs significantly better than AltMin and BiG-AMP methods. For not very low TR-rank tensor ($\mathcal{U} \in \mathbb{R}^{6 \times 7 \times 8}$ with rank $2,2,2$), other methods nearly become invalid while TeG-AMP still works well under the sampling rate of $40\%$. Moreover, we can find that the TeG-AMP is much more resistant to noise than other methodologies from the simulations in noisy cases.

\subsection{Experimental Results of MNIST Digits}

The recovered results of MNIST digits with size $28\times 28 \times 6$ and TR rank $14\times 14 \times 6$ are shown in Figure~\ref{figure:MovingChar} and Appendix~\ref{appendix-Results of MNIST digits}. The recover errors of TeG-AMP, BiG-AMP and AltMin methods are presented in Table~\ref{table2}. 
We can see that the proposed TeG-AMP method performs significantly better than AltMin and BiG-AMP methods since it takes full advantage of the tensor structure. The recovered digits are clearly visible. In contrast, the digits recovered by BiG-AMP and AltMin methods are drowned in noise.

\section{Conclusions}

In this paper, the derivation of TeG-AMP was presented, which is developed as an approximation of the sum-product belief propagation algorithm in high dimensions where the central limit theorem and Taylor series approximations are applicable under the assumption of statistically independent tensor entries. We also discussed in detail the direct applicability of TeG-AMP to low CP/TT-rank and some low Tucker-rank tensor inference problems, since it is developed based on the general TR decomposition model. In addition, we proposed a simplified version of TeG-AMP, named TeS-AMP, based on the CP decomposition model, which can be used for low CP-rank tensor interference problems. Moreover, in order to implement the proposed TeG-AMP under finite problem size conditions, an adaptive damping mechanism was given. At last, we presented the specialization of TeG-AMP to tensor completion problems, and compared the results of empirical analysis to state-of-the-art algorithms on such problems. Our numerical results showed that TeG-AMP achieves significant improvement as it takes full advantage of tensor structures.

\subsection{Limitations and Future works}

The disadvantage of our algorithm is that the complexity is higher compared to alternate minimization or gradient methods. 
The complexity order is dominated by calculating the approximated variances. 
Therefore, we propose a TeS-AMP algorithm to alleviate this problem. 
Future works will be to find ways to further simplify the computation of the mean and variance in the algorithm.

\bibliography{aaai24}


\newpage
\appendix
\onecolumn

\section{Derivation of Tensor Sum-Product Algorithm}\label{appendixA}

\subsection{Sum-product Algorithm Background}

\begin{definition}\label{def1}[Sum-product Update Rule \cite{kschischang2001factor}]
The message sent form a node $v$ on an edge $e$ is the product of the local function at $v$ with all messages received at $v$ on edges other than $e$, summarized for the variable associated with $e$.
\end{definition}

Define $\mu_{x\rightarrow f}(x)$ as the message from node $x$ to node $f$ in the operation of the sum-product algorithm, and define $\mu_{f\rightarrow x}(x)$ as the message from node $f$ to node $x$. For the variable node $x$, there is no local function to include, hence based on Definition~\ref{def1}, the massage $\mu_{x\rightarrow f}(x)$ equals all messages received at $x$ on edges other than $f$, which is given by
$$
\mu_{x\rightarrow f}(x) = \prod_{h \in n(x) \setminus \{f\}} \mu_{h \rightarrow x}(x),
$$
where $n(x)$ denotes the set of neighbors of node $x$ in a factor graph. Note that a variable node sends the \textbf{product} of messages received from its children.

For the factor node $f$ to variable node $x$, the local function at $f$ is given by
\begin{equation}
\label{eq1}
    \sum_{\sim \{x\}} f (n(f)),
\end{equation}
where $\sum_{\sim}$ denotes the ``not-sum'' or \emph{summary}. For example, denote $h$ as a function of three variables $x_1$, $x_2$, and $x_3$, then the ``summary for $x_2$'' is denoted by \cite{kschischang2001factor}
\begin{align}
\label{eq2}
    \sum_{\sim \{x_2\}} h(x_1,x_2,x_3) := \sum_{x_1 \in A_1 } \sum_{x_3 \in A_3} h(x_1, x_2, x_3).
\end{align}
In this way we have $f_i(x_i) = \sum_{\sim \{x_i\}} f(x_1,...,x_n)$ is the $i$-th marginal function associated with $f(x_1,...,x_n)$.

Then, based on Definition~\ref{def1}, we have $\mu_{f \rightarrow x}(x)$ is the product of the local function at $f$ with all messages received at $f$ on edges other than $x$
\begin{align}
    \mu_{f \rightarrow x}(x) &=  \left\{ \sum_{\sim \{x\}} f (n(f))  \right \} \prod_{y \in n(f) \setminus \{x\}} \mu_{y \rightarrow f}(y),  \nonumber \\
&= \sum_{\sim \{x\}} \left\{ f (n(f)) \prod_{y \in n(f) \setminus \{x\}}  \mu_{y \rightarrow f}(y) \right \},
\end{align}
where $n(f)$ is the set of arguments of the function $f$. Note that a factor node $f$ with parent $x$ forms the product of $f$ with the messages received from its children, and then operates on the result with a $\sum_{\sim \{x\}}$ \emph{summary} operator.

\subsection{Modeling MMSE Problem as Sum-product}

\begin{figure}[!htb]
	\centering
	\includegraphics[width=2.0in]{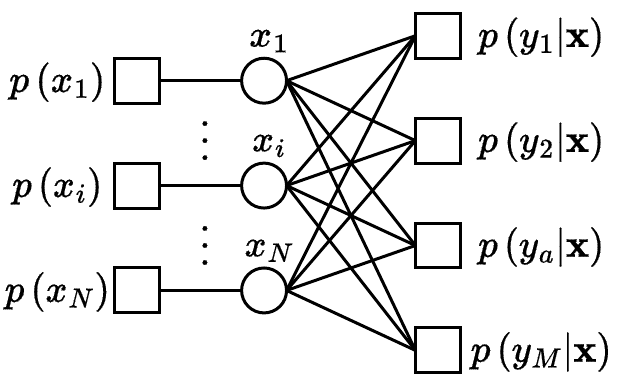}
	\caption{Factor graph of postulated posterior $p(x|y)$.}
	\label{figure:graph}
\end{figure}

Consider a problem that we want to estimate $x$ from its observation $y$. It is known that the MMSE estimate $\hat x$ is the mean of its posterior probability, i.e.,
\begin{equation}
\label{eq:MMSE}
    \hat x_{\text{MMSE}} = \int x p(x|y) \textrm{d} x.
\end{equation}
Using Bayes’ rules, the postulated posterior $p(x|y)$ is given by
\begin{equation}
\label{eqpxy}
    p(x|y) = \frac{1}{p(y)} p(x) p(y|x),
\end{equation}
where $p(y)$ is known. Hence, the factor graph of postulated posterior defined in \eqref{eq:MMSE} can be depicted in Figure~\ref{figure:graph} \cite{zou2022concise,kschischang2001factor,richardson2008modern}. Then, we present the corresponding sum-product algorithm.

For the variable node $x_i$, its local function is $p(x_i)$. Hence, based on Definition~\ref{def1}, the message from node $x_i$ to node $p(y_a|x)$ equals the product of the local function at $x_i$ with all messages received at $x_i$ on edges other than $y_a$, which is given by
\begin{equation}
\label{eqsp1}
    \mu_{i\rightarrow a}^{(t+1)} (x_i) = p(x_i) \prod_{b\neq a}^{M} \mu_{i\leftarrow b}^{(t)} (x_i).
\end{equation}

For the factor node $p(y_a|x)$, based on \eqref{eq1} and \eqref{eq2}, we have its local function is the $a$-th marginal function, which is given by
\begin{equation}
    \sum_{\sim x_i} p(y_a|x). \nonumber
\end{equation}
And for continuous $x$, we have the local function is given by
\begin{equation}
    \int p(y_a|x) \textrm{d} x_{\setminus i}, \nonumber
\end{equation}
where $x_{\setminus i}$ is $x$ expect $x_i$.

Then, based on Definition~\ref{def1}, we have $\mu_{i \leftarrow a}^{(t)}(x_i)$ is the product of the local function at $p(y_a|x)$ with all messages received at $p(y_a|x)$ on edges other than $x_i$, which is given by
\begin{align}
\label{eqsp2}
\mu_{i \leftarrow a}^{(t)}(x_i) &= \left\{ \int p(y_a|x) \textrm{d} x_{\setminus i}  \right\}  \prod_{j\neq i}^{N} \mu_{j \rightarrow a}^{(t)} (x_j) \nonumber \\
&=  \int p(y_a|x)    \prod_{j\neq i}^{N} \mu_{j \rightarrow a}^{(t)} (x_j)  \textrm{d} x_{\setminus i}.
\end{align}
The sum-product algorithm is based on the iterations \eqref{eqsp1} and \eqref{eqsp2}.

\subsection{Tensor Sum-product Derivation}

\begin{figure}[!htb]
	\centering
	\subfloat{\includegraphics[width=3.0in]{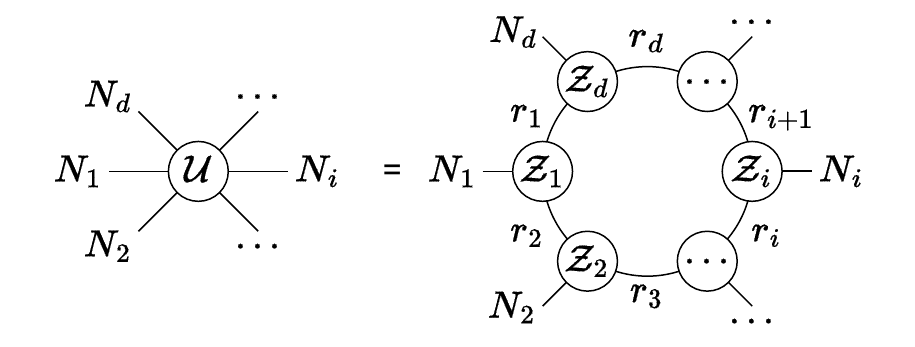}}
	\caption{A graphical representation of the TR decomposition.}
	\label{figure:TR-model}
\end{figure}

Recall that $\mathcal{V}$ can be represented as a likelihood function of the decoupled form
\begin{align} 
\label{likelihood-1-A}
p(\mathcal{V} | \mathcal{U})
=\prod_{\mathbf{x}} p\left(v_{\mathbf{x}} | u_{\mathbf{x}}\right),
\end{align}
where $\mathbf{x} = [x_1, x_2, \cdots, x_d]$. Such a tensor ring decomposition is illustrated graphically in Figure~\ref{figure:TR-model}. 

We assume that a prior of the tensor $\mathcal{U}$ is composed of independent priors of each element of each core
\begin{align}
\label{pdf-1}
p(\mathcal{U})
=&~ \prod_{i=1}^{d} p\left(\mathcal{Z}_{i}\right)\nonumber \\
=&~ \prod_{i=1}^{d} \prod_{\ell_i=1}^{r_{i}} \prod_{x_i=1}^{N_{i}} \prod_{\ell_{i+1}=1}^{r_{i+1}} p\left(\mathcal{Z}_{i}(r_i,x_i,r_{i+1})\right).
\end{align}

The goal is to compute the minimum mean-squared error (MMSE) estimates of $\{\mathcal{Z}_{i}\}_{i=1}^{d}$ from the observation ${\mathcal V}$ based on \eqref{likelihood-1-A} and \eqref{pdf-1}. With a given index $\mathbf{x} = [x_1, x_2, \cdots, x_d]$, according to Bayes' rule, the problem is equivalent to computing the posterior
\begin{align} 
\label{posterior-1-A}
p(u_{\mathbf{x}} | v_{\mathbf{x}}) \propto ~ p(v_{\mathbf{x}} | u_{\mathbf{x}}) \prod_{i=1}^{d} p\left(\mathcal{Z}_{i}(:,x_i,:) \right).
\end{align}
Similar with \eqref{eqpxy}, we can build a factor graph by setting $p(v_{\mathbf{x}} | u_{\mathbf{x}})$ as the factor node and $\mathcal{Z}_{i}(:,x_i,:)$ as the variable node, respectively.

Then, similar with \eqref{eqsp1}, the local function of $\mathcal{Z}_{i}(:,x_i,:)$ is $p\left(\mathcal{Z}_{i}(:,x_i,:)\right)$, hence the message from a variable node $\mathcal{Z}_{i}(:,x_i,:)$ to factor node $p(v_{\mathbf{x}} | u_{\mathbf{x}})$ is given by
\begin{align}
\label{sum-pro-2-A}
p_{\mathbf{x} \leftarrow x_i}(\mathcal{Z}_{i}(:,x_{i},:))
=&~p\left(\mathcal{Z}_{i}(:,x_i,:)\right) \prod_{\{\mathbf{x}':x'_i=x_i,~ \mathbf{x}' \neq \mathbf{x}\}}p_{\mathbf{x}' \rightarrow x_i},
\end{align}
where $\{\mathbf{x}':x'_i=x_i\}$ indicates $\{x_1',x_2',\cdots,x_i,\cdots,x_d'\}$, which is over ``all elements except for $x_i$".

On the other hand, similar with \eqref{eqsp2}, the local function as $p(v_{\mathbf{x}} | u_{\mathbf{x}})$ is the following marginal function
\begin{align} 
\int_{{\left\{\mathcal{Z}_{i'}(:,x_{i'},:)~:~i'\neq i\right \}}} p\left(v_{\mathbf{x}}|  \sum_{\mathbf{l}=1}^{\mathbf{r}}\prod_{i'=1}^{d} \mathcal{Z}_{i'}(:,x_{i'},:) \right),
\end{align}
where $\{\mathcal{Z}_{i'}(:,x_{i'},:):i'\neq i \}$ includes $d-1$ integral variables, and $x'_i$ is the $i$-th element of $\mathbf{x}'$ and
$$
u_{\mathbf{x}} = \sum_{\mathbf{l}=1}^{\mathbf{r}}\prod_{i'=1}^{d} \mathcal{Z}_{i'}(:,x_{i'},:).
$$
Hence, the message from the factor node $p(v_{\mathbf{x}} | u_{\mathbf{x}})$ to the variable node $\mathcal{Z}_{i}(:,x_i,:)$ is given by
\begin{align} 
\label{sum-pro-1-A}
p_{\mathbf{x} \rightarrow x_i}(\mathcal{Z}_{i}(:,x_{i},:)) 
= \int_{{\left\{\mathcal{Z}_{i'}(:,x_{i'},:)~:~i'\neq i\right \}}}    
\Bigg\{p\left(v_{\mathbf{x}}| \sum_{\mathbf{l}=1}^{\mathbf{r}} \prod_{i'=1}^{d} \mathcal{Z}_{i'}(:,x_{i'},:) \right) \times \prod_{\{i':i' \neq i\}} p_{\mathbf{x} \leftarrow x_{i'}} \Bigg\}.
\end{align}

\section{Derivation of Equation \eqref{sum-pro-Taylor-1}} \label{appendixB}

\textbf{Element-wise Result:}~
{\it
$p_{\mathbf{x} \rightarrow [i,\ell_i,\ell_{i+1},x_i]}(\mathcal{Z}_{i}(\ell_i,x_i,\ell_{i+1}))$ in \eqref{sum-pro-1} can be approximated by
\begin{align*} 
&{  p_{\mathbf{x} \rightarrow [i,\ell_i,\ell_{i+1},x_i]} \left(t,\mathcal{Z}_{i}(\ell_i,x_i,\ell_{i+1}) \right)} \\
{\approx} &~  {\text{const}} + 
\Big[
 {\sum_{\mathbf{l}'_{\setminus i} = 1}^{\mathbf{r}_{\setminus i}}}
  {\hat{\mathcal{Z}}^{\ell'}_{\setminus i, \mathbf{x}}(t)}
\Big] \mathcal{Z}_i^{\ell} (x_i) 
 \hat{s}_{\mathbf{x} }(t) + 
 \frac{1}{2} {\omega}_{\mathbf{x} }(t){\mathcal{Z}_i^{\ell} (x_i) }^2 {\zeta}_{x}(t) 
 -  \frac{ 1 }{2} \Big[ 
 {\sum_{\mathbf{l}'_{\setminus i} = 1}^{\mathbf{r}_{\setminus i}}}
 {\hat{\mathcal{Z}}^{\ell'}_{\setminus i, x}(t)}
\Big]^2  \left[\mathcal{Z}_i^{\ell} (x_i)^2 - 2 \hat{\mathcal{Z}}^{\ell}_{i,x_i}(t) \mathcal{Z}_i^{\ell} (x_i) \right] \nu_{\mathbf{x} }^s(t), \nonumber 
\end{align*}
where $\mathbf r_{\setminus i} \triangleq  \mathbf r \backslash (r_i, r_{i+1})$, $\mathbf l' _{\setminus i} \triangleq \mathbf l' \backslash (\ell_i', \ell_{i+1}')$,
\begin{align*}
\hat{s}_{\mathbf{x} }(t) &= \frac{1}{ \nu_{\mathbf{x} }^p(t) } \left(\hat{u}_{\mathbf{x} }(t)-\hat{p}_{\mathbf{x} }(t)\right), \\
{\nu}_{\mathbf{x} }^{s}(t) &= \frac{1}{\nu_{\mathbf{x} }^{p}(t)}\left(1-\frac{\nu_{\mathbf{x} }^{u}(t)}{\nu_{\mathbf{x} }^{p}(t)}\right)
\\
{\omega}_{\mathbf{x} }(t) &= {\hat{s}_{\mathbf{x} }(t)}^2  -  {\nu_{\mathbf{x} }^s(t)}, 
\end{align*}
 and
\begin{align*}
{  \hat{\mathcal{Z}}^{\ell'}_{\setminus i,{\mathbf{x} }}(t)} 
&\triangleq \hat{\mathcal{Z}}^{\ell'}_{1,\mathbf{x} }(t)
 \cdots 
\hat{\mathcal{Z}}^{\ell_{i-1}',\ell_i}_{i-1,\mathbf{x} }(t) 
\hat{\mathcal{Z}}^{\ell_{i+1},\ell_{i+2}'}_{i+1,\mathbf{x} }(t) \cdots
\hat{\mathcal{Z}}^{\ell'}_{d,\mathbf{x} }(t), \nonumber \\
{  \hat{\mathcal{Z}}^{\ell'}_{\setminus i, x}(t)} 
&\triangleq \hat{\mathcal{Z}}^{\ell'}_{1, x_1}(t)
 \cdots \hat{\mathcal{Z}}^{\ell_{i-1}',\ell_i}_{i-1, x_{i-1}}(t) 
\hat{\mathcal{Z}}^{\ell_{i+1},\ell_{i+2}'}_{i+1,x_{i+1}}(t)
\cdots
\hat{\mathcal{Z}}^{\ell'}_{d, x_d}(t), \nonumber \\
\zeta_{x}(t) &\triangleq  {\sum_{\mathbf{l}'_{\setminus i} = 1}^{\mathbf{r}_{\setminus i}}}
\sum_{ \substack {A \subset \mathbb{D} \setminus \{i\} \\ A \neq \emptyset} } \Bigg[
 \prod_{ \substack{i' \in A ,\ell_i'=\ell_i \\
\ell_{i+1}' = \ell_{i+1}
}}
 \nu_{i', x_{i'}}^{{ {\mathcal{Z}}},\ell'}(t) 
   \times 
   \prod_{ \substack{
    i'' \in \mathbb{D} \setminus (\{ i \} \cup A)\\
    \ell_i'=\ell_i, \\
    \ell_{i+1}' = \ell_{i+1}
    }} 
 {\hat{\mathcal{Z}}_{i'', x_{i''}}^{\ell_{i''}', \ell_{i'' + 1}' } (t) }^2 
 \Bigg]. \nonumber
\end{align*}
}

\emph{Derivation:}
Recall that the matrix form of the TR decomposition is given by 
\begin{align} 
u_{\mathbf{x}} 
&= \sum_{\mathbf{l}=1}^{\mathbf{r}} \prod_{i=1}^{d} \mathcal{Z}_{i}(r_i,x_{i},r_{i+1}), \nonumber
\end{align}
where $\mathbf{r} = [r_1, r_2,\cdots, r_d]^T$ and $\mathbf{l}=(\ell_{1}, \cdots, \ell_{d})$. 
For the convenience of later derivation, we give the following index form of the above matrix form as the following
\begin{align} 
\label{TR-decomposition-index-A}
u_{x_{1}, x_{2}, ..., x_{d}} = \sum_{\ell_{1}, \ldots, \ell_{d}=1}^{r_{1}, \ldots, r_{d}} \prod_{i=1}^{d} \mathcal{Z}_{i}^{\ell_{i}, \ell_{i+1} }(x_{i}),
\end{align}
where $\mathcal{Z}_{i}^{\ell_{i}, \ell_{i+1} } \in \mathbb{R}^{n_{i} \times 1}$, whose $x_i$-th element is $\mathcal{Z}_{i}^{\ell_{i}, \ell_{i+1} }(x_{i})$, denotes the $({\ell_{i}, \ell_{i+1} })$-th mode-2 fiber of latent tensors $\mathcal{Z}_{i}$, and $\ell_{d+1} = \ell_{1}$ due to the trace operation.

Note that we have
\begin{align} 
p(u_{\mathbf{x}} | v_{\mathbf{x}}) \propto ~ p(v_{\mathbf{x}} | u_{\mathbf{x}}) \prod_{i=1}^{d} p\left(\mathcal{Z}_{i}(:,x_i,:) \right), \nonumber
\end{align}
which yields
\begin{align} \label{posterior-A-expanded}
p
\left( {\{{\mathcal{Z}}_i^{\ell_i,\ell_{i+1}} \}_{i,\ell_i,\ell_{i+1}} } | \mathcal{V} \right)  
= &~ p 
\left( \mathcal{V} | {\{{\mathcal{Z}}_i^{\ell_i,\ell_{i+1}} \}_{i,\ell_i,\ell_{i+1}} } \right) 
p 
\left( {\{{\mathcal{Z}}_i^{\ell_i,\ell_{i+1}} \}_{i,\ell_i,\ell_{i+1}} }  \right)
/ p(\mathcal{V}) \nonumber \\
\propto &~ p 
\left( \mathcal{V} | \mathcal{U}  \right) 
p 
\left( {\{{\mathcal{Z}}_i^{\ell_i,\ell_{i+1}} \}_{i,\ell_i,\ell_{i+1}} }  \right) \nonumber \\
=&~ p 
\left( \mathcal{V} | \mathcal{U}  \right) 
\left[  \prod_{\ell_{1}, \ldots, \ell_{d}=1}^{r_{1}, \ldots, r_{d}} 
\prod_{i=1}^{d} p
\left( \mathcal{Z}_{i}^{\ell_{i}, \ell_{i+1} } \right)  \right]  \nonumber \\
=&~ \left[\prod_{x_{1},...,x_{d}=1}^{N_1,...,N_{d}} 
p
\left(v_{x_{1}, \ldots, x_{d}} | u_{x_{1}, \ldots, x_{d}}\right) \right ] \times \left[
 \prod_{\ell_{1}, \ldots, \ell_{d}=1}^{r_{1}, \ldots, r_{d}} \prod_{i=1}^{d} \prod_{x_i=1}^{N_i} 
 p
 \left( \mathcal{Z}_{i}^{\ell_{i}, \ell_{i+1} } (x_i)\right) 
 \right ].
\end{align}

We will see that the approximations are primarily based on the central limit theorem and Taylor series reasonings that turn to be exact in the large system limits where $r_1,...,r_d, d, N_1,..., N_d \rightarrow \infty$ with fixed non-zero ratios between every two of them. Because of the use of finite dimensions in the real world, we consider such reasonings as approximations. Particularly, our derivation will ignore some terms that disappear relative to others as $r_1,...,r_d, d, N_1,..., N_d \rightarrow \infty$.

We start by approximating the message $p_{\mathbf x \rightarrow [i,\ell_i,\ell_{i+1},x_i] }(t, \mathcal{Z}_{i}^{\ell_i,\ell_{i+1}}(x_i))$. Recall \eqref{sum-pro-1} as the following
\begin{align} 
p_{\mathbf{x} \rightarrow x_i}(\mathcal{Z}_{i}(:,x_{i},:)) 
= \log  
\int_{{\left\{\mathcal{Z}_{i'}(:,x_{i'},:):i'\neq i\right \}}}    
\Bigg\{p\left(v_{\mathbf{x}}| \sum_{\mathbf{l}=1}^{\mathbf{r}} 
\prod_{i'=1}^{d} \mathcal{Z}_{i'}(:,x_{i'},:) \right) \times \prod_{\{i':i' \neq i\}}  \exp\left[p_{\mathbf{x} \leftarrow x_{i'}} \right] \Bigg\}+\text{const},  \nonumber
\end{align}
which can be expanded based on \eqref{posterior-A-expanded} by
\begin{align}
\label{sum-pro-1-2}
p_{\mathbf x \rightarrow [i,\ell_i,\ell_{i+1},x_i] }(t, \mathcal{Z}_{i}^{\ell_i,\ell_{i+1}}(x_i)) 
&= \log 
\left( \int_{  \left\{ \mathcal{Z}_{i'}^{\ell_{i'}',\ell_{i'+1}'}(x_{i'}) \right \}_{i',\ell_{i'}',\ell_{i'+1}' }  \backslash  \mathcal{Z}_{i}^{\ell_i,\ell_{i+1}}(x_i) }    
p \left(  v_{\mathbf x}| \overbrace{  \sum_{\ell_1,...,\ell_d = 1}^{r_1,...,r_d}  \prod_{i=1}^{d}   \mathcal{Z}_{i}^{\ell_i,\ell_{i+1}}(x_i) }^{u_{\mathbf x}}  \right)   \right.  \nonumber \\
&\left.  \times  \prod_{ ( i',\ell_{i'}',\ell_{i'+1}' ) \neq (i,\ell_i,\ell_{i+1})}  
\exp\left\{ 
p_{\mathbf x \leftarrow [i',\ell_{i'}',\ell_{i'+1}',x_{i'}] }\left(t, \mathcal{Z}_{i'}^{\ell_{i'}',\ell_{i'+1}'}(x_{i'}) \right) \right\}  \right) + \text{const}.
\end{align}

For large dimensions, the central limit theorem motivates the treatment of $\mathsf{u}_{x_1,x_2,...,x_d}$, the random variable associated with the ${u}_{x_1,x_2,...,x_d}$ identified in \eqref{sum-pro-1-2}, conditioned on a function of ${\mathsf{Z}}_{i}^{\ell_i,\ell_{i+1}}(x_i) = \mathcal{Z}_{i}^{\ell_i,\ell_{i+1}}(x_i)$, as Gaussian and therefore perfectly characterized through a conditional mean and variance.
By defining the random variable $\tilde{\mathsf{Z}}^{\ell_i,\ell_{i+1}}_{i,\mathbf x} \triangleq {\mathsf{Z}}_{i}^{\ell_i,\ell_{i+1}}(x_i)  -  \hat{\mathcal{Z}}_{i, \mathbf{x}}^{\ell_i,\ell_{i+1}}(t)$ with zero mean where ${\mathsf{Z}}_{i}^{\ell_i,\ell_{i+1}}(x_i) \backsim \frac{1}{C} \exp( p_{\mathbf x \leftarrow [i,\ell_i,\ell_{i+1},x_i] }(t,\cdot ) )$, we can write
\begin{align} 
\label{u-x-1}
\mathsf{u}_{\mathbf x} &= \sum_{\ell_1',...,\ell_d' = 1}^{r_1,...,r_d}  \prod_{i=1}^{d}   \mathsf{Z}_{i}^{\ell_i',\ell_{i+1}'}(x_i)  
= \sum_{\ell_1',...,\ell_d' = 1}^{r_1,...,r_d}  \mathsf{Z}_{1}^{\ell_1',\ell_{2}'}(x_1)  \mathsf{Z}_{2}^{\ell_2',\ell_{3}'}(x_2) \cdots  \mathsf{Z}_{d}^{\ell_d',\ell_{1}'}(x_d)  \nonumber \\
& = \sum_{\substack{\ell_1',...,\ell_d' = 1 \\ 
\ell_i' \neq \ell_i ~\text{or}~ \ell_{i+1}' \neq \ell_{i+1} 
}}^{r_1,...,r_d}  
{ \left( \hat{\mathcal{Z}}_{1, \mathbf{x}}^{\ell_1',\ell_{2}'}(t)  +  \tilde{\mathsf{Z}}^{\ell_1',\ell_{2}'}_{1,\mathbf x}  \right) }  
\cdots  { \left( \hat{\mathcal{Z}}_{d, \mathbf{x}}^{\ell_d',\ell_{1}'}(t)  +  \tilde{\mathsf{Z}}^{\ell_d',\ell_{1}'}_{d,\mathbf x}  \right) } \nonumber \\
&~~~+ \sum_{\substack{\ell_1',...,\ell_{i-1}',\ell_{i+2}',...,\ell_d' = 1  }}^{r_1,...,r_{i-1},r_{i+2},...,r_d}
 { \left( \hat{\mathcal{Z}}_{1, \mathbf{x}}^{\ell_1',\ell_{2}'}(t)  +  \tilde{\mathsf{Z}}^{\ell_1',\ell_{2}'}_{1,\mathbf x}  \right) }
\cdots { \left( \hat{\mathcal{Z}}_{i-1, \mathbf{x}}^{\ell_{i-1}',\ell_{i}}(t)  +  \tilde{\mathsf{Z}}^{\ell_{i-1}',\ell_{i}}_{i-1,\mathbf x}  \right) }  
{\mathsf{Z}}_{i}^{\ell_i,\ell_{i+1}}(x_i) \nonumber \\
&~~~~~{ \left( \hat{\mathcal{Z}}_{i+1, \mathbf{x}}^{\ell_{i+1},\ell_{i+2}'}(t)  +  \tilde{\mathsf{Z}}^{\ell_{i+1},\ell_{i+2}'}_{i+1,\mathbf x}  \right) }
\cdots { \left( \hat{\mathcal{Z}}_{d, \mathbf{x}}^{\ell_{d}',\ell_{1}'}(t)  +  \tilde{\mathsf{Z}}^{\ell_{d}',\ell_{1}'}_{d,\mathbf x}  \right) }  .
\end{align}

Define $\hat{p}_{\mathbf x}(t)$ and $\hat{p}_{\mathbf x}^{\ell_i,\ell_{i+1}}(t)$ as below that will be used in our derivations.
 
\begin{align} 
\hat{p}_{\mathbf x}(t) & \triangleq  \sum_{\mathbf l' = 1}^{\mathbf r}  
{\hat{\mathcal{Z}}_{1, \mathbf{x}}^{\ell'}(t)} \cdots {\hat{\mathcal{Z}}_{d, \mathbf{x}}^{\ell'}(t)} =  \sum_{\ell_1',...,\ell_d' = 1}^{r_1,...,r_d}  
{  \hat{\mathcal{Z}}_{1, \mathbf{x}}^{\ell_1',\ell_{2}'}(t)  }  
\cdots  {  \hat{\mathcal{Z}}_{d, \mathbf{x}}^{\ell_d',\ell_{1}'}(t)  },
\\
\label{p-xl-1-A}
\hat{p}_{\mathbf x}^{\ell_i,\ell_{i+1}}(t) & \triangleq  \sum_{\substack{\ell_1',...,\ell_d' = 1 \\ 
\ell_i' \neq \ell_i ~\text{or}~ \ell_{i+1}' \neq \ell_{i+1} 
} }^{r_1,...,r_d}  
{  \hat{\mathcal{Z}}_{1, \mathbf{x}}^{\ell_1',\ell_{2}'}(t)  }  
\cdots  { \hat{\mathcal{Z}}_{d, \mathbf{x}}^{\ell_d',\ell_{1}'}(t)   }  ,
\end{align}
\color{black}
after which it is possible to see that
\begin{align}
\label{E-u-1}
& ~\mathbb{E}\{ \mathsf{u}_{\mathbf x} | {\mathsf{Z}}_{i}^{\ell_i,\ell_{i+1}}(x_i)  =  {\mathcal{Z}}_{i}^{\ell_i,\ell_{i+1}}(x_i)  \} \nonumber \\
=&~  \sum_{\substack{\ell_1',...,\ell_d' = 1 \\ 
\ell_i' \neq \ell_i ~\text{or}~ \ell_{i+1}' \neq \ell_{i+1} 
} }^{r_1,...,r_d}  
{  \hat{\mathcal{Z}}_{1, \mathbf{x}}^{\ell_1',\ell_{2}'}(t)  }  
\cdots  { \hat{\mathcal{Z}}_{d, \mathbf{x}}^{\ell_d',\ell_{1}'}(t)   }  
+ \sum_{\substack{\ell_1',...,\ell_{i-1}',\ell_{i+2}',...,\ell_d' = 1  }}^{r_1,...,r_{i-1},r_{i+2},...,r_d}
\hat{\mathcal{Z}}^{\ell_1',\ell_2'}_{1,\mathbf x}(t)
 \cdots \hat{\mathcal{Z}}^{\ell_{i-1}',\ell_i}_{i-1,\mathbf x}(t) 
 \mathcal{Z}_i^{\ell_i,\ell_{i+1}} (x_i)  
\hat{\mathcal{Z}}^{\ell_{i+1},\ell_{i+2}'}_{i+1,\mathbf x}(t) \cdots
\hat{\mathcal{Z}}^{\ell_{d}',\ell_1'}_{d,\mathbf x}(t)   \nonumber \\
=&~  \hat{p}_{\mathbf x}^{\ell_i,\ell_{i+1}}(t) 
 + \sum_{\substack{\ell_1',...,\ell_{i-1}',\ell_{i+2}',...,\ell_d' = 1  }}^{r_1,...,r_{i-1},r_{i+2},...,r_d}
\hat{\mathcal{Z}}^{\ell_1',\ell_2'}_{1,\mathbf x}(t)
 \cdots \hat{\mathcal{Z}}^{\ell_{i-1}',\ell_i}_{i-1,\mathbf x}(t) 
 \mathcal{Z}_i^{\ell_i,\ell_{i+1}} (x_i)  
\hat{\mathcal{Z}}^{\ell_{i+1},\ell_{i+2}'}_{i+1,\mathbf x}(t) \cdots
\hat{\mathcal{Z}}^{\ell_{d}',\ell_1'}_{d,\mathbf x}(t)   \nonumber \\
=&~  \hat{p}_{\mathbf x}(t) -
 \sum_{\substack{\ell_1',...,\ell_{i-1}',\ell_{i+2}',...,\ell_d' = 1  }}^{r_1,...,r_{i-1},r_{i+2},...,r_d}
\hat{\mathcal{Z}}^{\ell_1',\ell_2'}_{1,\mathbf x}(t)
 \cdots \hat{\mathcal{Z}}^{\ell_{i-1}',\ell_i}_{i-1,\mathbf x}(t) 
  \hat{\mathcal{Z}}^{\ell_{i},\ell_{i+1}}_{i,\mathbf x}(t) 
\hat{\mathcal{Z}}^{\ell_{i+1},\ell_{i+2}'}_{i+1,\mathbf x}(t) \cdots
\hat{\mathcal{Z}}^{\ell_{d}',\ell_1'}_{d,\mathbf x}(t)   \nonumber \\
&~~~~~~~~~~~~~~~~~~~~~~~~~~~~~~~~~~~~~~~~~~~~~~~~~~~~+ \sum_{\substack{\ell_1',...,\ell_{i-1}',\ell_{i+2}',...,\ell_d' = 1  }}^{r_1,...,r_{i-1},r_{i+2},...,r_d}
\hat{\mathcal{Z}}^{\ell_1',\ell_2'}_{1,\mathbf x}(t)
 \cdots \hat{\mathcal{Z}}^{\ell_{i-1}',\ell_i}_{i-1,\mathbf x}(t) 
 \mathcal{Z}_i^{\ell_i,\ell_{i+1}} (x_i)  
\hat{\mathcal{Z}}^{\ell_{i+1},\ell_{i+2}'}_{i+1,\mathbf x}(t) \cdots
\hat{\mathcal{Z}}^{\ell_{d}',\ell_1'}_{d,\mathbf x}(t)   \nonumber \\
=&~  \hat{p}_{\mathbf x}(t) 
+ \sum_{\substack{\ell_1',...,\ell_{i-1}',\ell_{i+2}',...,\ell_d' = 1  }}^{r_1,...,r_{i-1},r_{i+2},...,r_d}
\hat{\mathcal{Z}}^{\ell_1',\ell_2'}_{1,\mathbf x}(t)
 \cdots \hat{\mathcal{Z}}^{\ell_{i-1}',\ell_i}_{i-1,\mathbf x}(t) 
\left[ \mathcal{Z}_i^{\ell_i,\ell_{i+1}} (x_i) -  \hat{\mathcal{Z}}^{\ell_{i},\ell_{i+1}}_{i,\mathbf x}(t) \right] 
\hat{\mathcal{Z}}^{\ell_{i+1},\ell_{i+2}'}_{i+1,\mathbf x}(t) \cdots
\hat{\mathcal{Z}}^{\ell_{d}',\ell_1'}_{d,\mathbf x}(t)   \nonumber \\
=&~  \hat{p}_{\mathbf x}(t) 
+ \sum_{\substack{\ell_1',...,\ell_{i-1}',\ell_{i+2}',...,\ell_d' = 1  }}^{r_1,...,r_{i-1},r_{i+2},...,r_d}
\hat{\mathcal{Z}}^{\ell_1',\ell_2'}_{1,\mathbf x}(t)
 \cdots \hat{\mathcal{Z}}^{\ell_{i-1}',\ell_i}_{i-1,\mathbf x}(t) 
\Big[ \mathcal{Z}_i^{\ell_i,\ell_{i+1}} (x_i)   \nonumber \\
&~~~~~~~~~~~~~~~~~~~~~~~~~~~~~~~~~~~~~~~~~~~~~~~~~~~~~~~~~~~~~~~~~~~~~~~~~~~
-  \hat{\mathcal{Z}}^{\ell_{i},\ell_{i+1}}_{i,x_i}(t) +  \hat{\mathcal{Z}}^{\ell_{i},\ell_{i+1}}_{i, x_i}(t)
 -  \hat{\mathcal{Z}}^{\ell_{i},\ell_{i+1}}_{i,\mathbf x}(t) \Big]
\hat{\mathcal{Z}}^{\ell_{i+1},\ell_{i+2}'}_{i+1,\mathbf x}(t) \cdots
\hat{\mathcal{Z}}^{\ell_{d}',\ell_1'}_{d,\mathbf x}(t)   \nonumber \\
\overset{(a)}{\approx}&~  \hat{p}_{\mathbf x}(t)  +  \sum_{\substack{\ell_1',...,\ell_{i-1}',\ell_{i+2}',...,\ell_d' = 1  }}^{r_1,...,r_{i-1},r_{i+2},...,r_d}
\hat{\mathcal{Z}}^{\ell_1',\ell_2'}_{1,\mathbf x}(t)
 \cdots \hat{\mathcal{Z}}^{\ell_{i-1}',\ell_i}_{i-1,\mathbf x}(t) 
\left[ \mathcal{Z}_i^{\ell_i,\ell_{i+1}} (x_i) -  \hat{\mathcal{Z}}^{\ell_{i},\ell_{i+1}}_{i,x_i}(t)  \right] 
\hat{\mathcal{Z}}^{\ell_{i+1},\ell_{i+2}'}_{i+1,\mathbf x}(t) \cdots
\hat{\mathcal{Z}}^{\ell_{d}',\ell_1'}_{d,\mathbf x}(t)  ,
\end{align}
where in $(a)$ we neglected $\hat{\mathcal{Z}}^{\ell_{i},\ell_{i+1}}_{i, x_i}(t) -  \hat{\mathcal{Z}}^{\ell_{i},\ell_{i+1}}_{i,\mathbf x}(t)$, 
noting that we assumed that data size is large and 
$p_{\mathbf x \leftarrow [i,\ell_i,\ell_{i+1},x_i] } (t,\cdot)$
and $p_{ [i,\ell_i,\ell_{i+1},x_i] } (t,\cdot)$ differ by only one term (which vanishes relatively to the others in the large-system limit). See \cite{rangan2011generalized,parker2014bilinear,montanari2012graphical} for similar arguments.

Also, define ${\nu}^{p}_{\mathbf x}(t)$ and ${\nu}^{p,\ell_i,\ell_{i+1}}_{\mathbf x}(t)$ as below that will be used in our derivations
\begin{align} 
\label{v-p-1-A}
{\nu}^{p}_{\mathbf x}(t) &\triangleq  \sum_{\substack{\mathbf{l}' = 1 }}^{\mathbf r}  
\sum_{ \substack {A \subset \mathbb{D} \\ A \neq \emptyset} } \left( \prod_{i' \in A} \nu_{i',\mathbf x}^{\mathcal{Z},\ell'}(t)    \prod_{i'' \in \mathbb{D} \setminus A}  {\hat{\mathcal{Z}}_{i'',\mathbf x}^{\ell_{i''}', \ell_{i'' + 1}' } (t) }^2 \right) \nonumber \\
&= \sum_{\substack{\ell_1',...,\ell_d' = 1 }}^{r_1,...,r_d}  
\sum_{ \substack {A \subset \{1,...,d\} \\ A \neq \emptyset} }
\left( \prod_{i' \in A} \nu_{i',\mathbf x}^{\mathcal{Z},\ell_{i'}', \ell_{i'+1}'}(t)    \prod_{i'' \in \{1,...,d\} \setminus A }  {\hat{\mathcal{Z}}_{i'',\mathbf x}^{\ell_{i''}', \ell_{i'' + 1}' } (t) }^2 \right), 
\\
\label{v-pl-1-A}
{\nu}^{p,\ell_i,\ell_{i+1}}_{\mathbf x}(t) & \triangleq  \sum_{\substack{\ell_1',...,\ell_d' = 1 \\ 
\ell_i' \neq \ell_i ~\text{or}~ \ell_{i+1}' \neq \ell_{i+1} 
}}^{r_1,...,r_d}  
\sum_{ \substack {A \subset \{1,...,d\} \\ A \neq \emptyset} }
\left( \prod_{i' \in A} \nu_{i',\mathbf x}^{\mathcal{Z},\ell_{i'}', \ell_{i'+1}'}(t)    \prod_{i'' \in \{1,...,d\} \setminus A }  {\hat{\mathcal{Z}}_{i'',\mathbf x}^{\ell_{i''}', \ell_{i'' + 1}' } (t) }^2 \right) ,
\end{align}
after which it is possible to see that
\begin{align} 
& \text{Var}\{ \mathsf{u}_{\mathbf x} | {\mathsf{Z}}_{i}^{\ell_i,\ell_{i+1}}(x_i)  =  {\mathcal{Z}}_{i}^{\ell_i,\ell_{i+1}}(x_i)  \} \nonumber \\
=&~  \sum_{\substack{\ell_1',...,\ell_d' = 1 \\ 
\ell_i' \neq \ell_i ~\text{or}~ \ell_{i+1}' \neq \ell_{i+1} 
}}^{r_1,...,r_d}  
\sum_{ \substack {A \subset \{1,...,d\} \\ A \neq \emptyset} }
\left( 
\prod_{i' \in A} \nu_{i',\mathbf x}^{\mathcal{Z},\ell_{i'}', \ell_{i'+1}'}(t)    \prod_{i'' \in \{1,...,d\} \setminus A }  {\hat{\mathcal{Z}}_{i'',\mathbf x}^{\ell_{i''}', \ell_{i'' + 1}' } (t) }^2 \right) + \nonumber \\
& \sum_{\substack{\ell_1',...,\ell_{i-1}',\ell_{i+2}',...,\ell_d' = 1  }}^{r_1,...,r_{i-1},r_{i+2},...,r_d}
\sum_{ \substack {A \subset \{1,...,d\} \setminus \{i\} \\ A \neq \emptyset} }  
\Bigg[
 \prod_{ \substack{i' \in A ,\ell_i'=\ell_i \\
\ell_{i+1}' = \ell_{i+1}
}}
 \nu_{i',\mathbf x}^{\mathcal{Z},\ell_{i'}', \ell_{i'+1}'}(t) 
    \prod_{ \substack{
    i'' \in \{1,...,d\} \setminus (\{ i \} \cup A) \\
    \ell_i'=\ell_i,\ell_{i+1}' = \ell_{i+1}
    }} 
 {\hat{\mathcal{Z}}_{i'',\mathbf x}^{\ell_{i''}', \ell_{i'' + 1}' } (t) }^2 
 \Bigg]
  { \mathcal{Z}_i^{\ell_i, \ell_{i+1}} (x_i) }^2  \nonumber \\
=& ~ {\nu}^{p,\ell_i,\ell_{i+1}}_{\mathbf x}(t)  + \nonumber \\
& \sum_{\substack{\ell_1',...,\ell_{i-1}',\ell_{i+2}',...,\ell_d' = 1  }}^{r_1,...,r_{i-1},r_{i+2},...,r_d}
\sum_{ \substack {A \subset \{1,...,d\} \setminus \{i\} \\ A \neq \emptyset} }  
\Bigg[
 \prod_{ \substack{i' \in A ,\ell_i'=\ell_i \\
\ell_{i+1}' = \ell_{i+1}
}}
 \nu_{i',\mathbf x}^{\mathcal{Z},\ell_{i'}', \ell_{i'+1}'}(t) 
    \prod_{ \substack{
    i'' \in \{1,...,d\} \setminus (\{ i \} \cup A) \\
    \ell_i'=\ell_i,\ell_{i+1}' = \ell_{i+1}
    }} 
 {\hat{\mathcal{Z}}_{i'',\mathbf x}^{\ell_{i''}', \ell_{i'' + 1}' } (t) }^2 
 \Bigg]
  { \mathcal{Z}_i^{\ell_i, \ell_{i+1}} (x_i) }^2  \nonumber \\
  = &~ \nu_{\mathbf x}^p(t)  - \sum_{\substack{\ell_1',...,\ell_{i-1}',\ell_{i+2}',...,\ell_d' = 1  }}^{r_1,...,r_{i-1},r_{i+2},...,r_d}
\sum_{ \substack {A \subset \{1,...,d\}  \\ A \neq \emptyset} }  
\Bigg[
 \prod_{ \substack{i' \in A ,\ell_i'=\ell_i \\
\ell_{i+1}' = \ell_{i+1}
}}
 \nu_{i',\mathbf x}^{\mathcal{Z},\ell_{i'}', \ell_{i'+1}'}(t) 
    \prod_{ \substack{
    i'' \in \{1,...,d\} \setminus ( A) \\
    \ell_i'=\ell_i,\ell_{i+1}' = \ell_{i+1}
    }} 
 {\hat{\mathcal{Z}}_{i'',\mathbf x}^{\ell_{i''}', \ell_{i'' + 1}' } (t) }^2 
 \Bigg]
 + \nonumber \\
  & \sum_{\substack{\ell_1',...,\ell_{i-1}',\ell_{i+2}',...,\ell_d' = 1  }}^{r_1,...,r_{i-1},r_{i+2},...,r_d}
\sum_{ \substack {A \subset \{1,...,d\} \setminus \{i\} \\ A \neq \emptyset} }  
\Bigg[
 \prod_{ \substack{i' \in A ,\ell_i'=\ell_i \\
\ell_{i+1}' = \ell_{i+1}
}}
 \nu_{i',\mathbf x}^{\mathcal{Z},\ell_{i'}', \ell_{i'+1}'}(t) 
    \prod_{ \substack{
    i'' \in \{1,...,d\} \setminus (\{ i \} \cup A) \\
    \ell_i'=\ell_i,\ell_{i+1}' = \ell_{i+1}
    }} 
 {\hat{\mathcal{Z}}_{i'',\mathbf x}^{\ell_{i''}', \ell_{i'' + 1}' } (t) }^2 
 \Bigg]
  { \mathcal{Z}_i^{\ell_i, \ell_{i+1}} (x_i) }^2  \nonumber \\
      = &~ \nu_{\mathbf x}^p(t) 
 + 
   \sum_{\substack{\ell_1',...,\ell_{i-1}',\ell_{i+2}',...,\ell_d' = 1  }}^{r_1,...,r_{i-1},r_{i+2},...,r_d}
\sum_{ \substack {A \subset \{1,...,d\} \setminus \{i\} \\ A \neq \emptyset} } 
\Bigg\{  \nonumber \\
& ~~~~~~~~~~\Bigg[
 \prod_{ \substack{i' \in A ,\ell_i'=\ell_i \\
\ell_{i+1}' = \ell_{i+1}
}}
 \nu_{i',\mathbf x}^{\mathcal{Z},\ell_{i'}', \ell_{i'+1}'}(t) 
    \prod_{ \substack{
    i'' \in \{1,...,d\} \setminus (\{ i \} \cup A) \\
    \ell_i'=\ell_i,\ell_{i+1}' = \ell_{i+1}
    }} 
 {\hat{\mathcal{Z}}_{i'',\mathbf x}^{\ell_{i''}', \ell_{i'' + 1}' } (t) }^2 
 \Bigg]
  \left[ {  \mathcal{Z}_i^{\ell_i, \ell_{i+1}} (x_i) }^2 - { \hat{\mathcal{Z}}_{i,x_i}^{\ell_i, \ell_{i+1}} (t) }^2  \right]
  \Bigg\}
  \nonumber 
\\
  & + 
   \sum_{\substack{\ell_1',...,\ell_{i-1}',\ell_{i+2}',...,\ell_d' = 1  }}^{r_1,...,r_{i-1},r_{i+2},...,r_d}
\sum_{ \substack {A \subset \{1,...,d\} \setminus \{i\} \\ A \neq \emptyset} } 
\Bigg[
 \prod_{ \substack{i' \in A ,\ell_i'=\ell_i \\
\ell_{i+1}' = \ell_{i+1}
}}
 \nu_{i',\mathbf x}^{\mathcal{Z},\ell_{i'}', \ell_{i'+1}'}(t) 
    \prod_{ \substack{
    i'' \in \{1,...,d\} \setminus (\{ i \} \cup A) \\
    \ell_i'=\ell_i,\ell_{i+1}' = \ell_{i+1}
    }} 
 {\hat{\mathcal{Z}}_{i'',\mathbf x}^{\ell_{i''}', \ell_{i'' + 1}' } (t) }^2 
 \Bigg]
  { \hat{\mathcal{Z}}_{i,x_i}^{\ell_i, \ell_{i+1}} (t) }^2  
  \nonumber \\
 &  - \sum_{\substack{\ell_1',...,\ell_{i-1}',\ell_{i+2}',...,\ell_d' = 1  }}^{r_1,...,r_{i-1},r_{i+2},...,r_d}
\sum_{ \substack {A \subset \{1,...,d\}  \\ A \neq \emptyset} }  
\Bigg[
 \prod_{ \substack{i' \in A ,\ell_i'=\ell_i \\
\ell_{i+1}' = \ell_{i+1}
}}
 \nu_{i',\mathbf x}^{\mathcal{Z},\ell_{i'}', \ell_{i'+1}'}(t) 
    \prod_{ \substack{
    i'' \in \{1,...,d\} \setminus (A) \\
    \ell_i'=\ell_i,\ell_{i+1}' = \ell_{i+1}
    }} 
 {\hat{\mathcal{Z}}_{i'',\mathbf x}^{\ell_{i''}', \ell_{i'' + 1}' } (t) }^2 
 \Bigg]  \nonumber \\
    = &~ \nu_{\mathbf x}^p(t) 
 + 
   \sum_{\substack{\ell_1',...,\ell_{i-1}',\ell_{i+2}',...,\ell_d' = 1  }}^{r_1,...,r_{i-1},r_{i+2},...,r_d}
\sum_{ \substack {A \subset \{1,...,d\} \setminus \{i\} \\ A \neq \emptyset} } 
\Bigg\{  \nonumber \\
& ~~~~~~~~~~\Bigg[
 \prod_{ \substack{i' \in A ,\ell_i'=\ell_i \\
\ell_{i+1}' = \ell_{i+1}
}}
 \nu_{i',\mathbf x}^{\mathcal{Z},\ell_{i'}', \ell_{i'+1}'}(t) 
    \prod_{ \substack{
    i'' \in \{1,...,d\} \setminus (\{ i \} \cup A) \\
    \ell_i'=\ell_i,\ell_{i+1}' = \ell_{i+1}
    }} 
 {\hat{\mathcal{Z}}_{i'',\mathbf x}^{\ell_{i''}', \ell_{i'' + 1}' } (t) }^2 
 \Bigg]
  \left[ {  \mathcal{Z}_i^{\ell_i, \ell_{i+1}} (x_i) }^2 - { \hat{\mathcal{Z}}_{i,x_i}^{\ell_i, \ell_{i+1}} (t) }^2  \right]
  \Bigg\}
  \nonumber \\
  & + 
   \sum_{\substack{\ell_1',...,\ell_{i-1}',\ell_{i+2}',...,\ell_d' = 1  }}^{r_1,...,r_{i-1},r_{i+2},...,r_d}
\sum_{ \substack {A \subset \{1,...,d\} \setminus \{i\} \\ A \neq \emptyset} } 
\Bigg\{  \nonumber \\
& ~~~~~~~~~~
\Bigg[
 \prod_{ \substack{i' \in A ,\ell_i'=\ell_i \\
\ell_{i+1}' = \ell_{i+1}
}}
 \nu_{i',\mathbf x}^{\mathcal{Z},\ell_{i'}', \ell_{i'+1}'}(t) 
    \prod_{ \substack{
    i'' \in \{1,...,d\} \setminus (\{ i \} \cup A) \\
    \ell_i'=\ell_i,\ell_{i+1}' = \ell_{i+1}
    }} 
 {\hat{\mathcal{Z}}_{i'',\mathbf x}^{\ell_{i''}', \ell_{i'' + 1}' } (t) }^2 
 \Bigg]
  \left[ { \hat{\mathcal{Z}}_{i,x_i}^{\ell_i, \ell_{i+1}} (t) }^2  - { \hat{\mathcal{Z}}_{i,\mathbf{x}}^{\ell_i, \ell_{i+1}} (t) }^2 \right]
  \Bigg\}
  \nonumber \\
 &  - \underbrace{
 \sum_{\substack{\ell_1',...,\ell_{i-1}',\ell_{i+2}',...,\ell_d' = 1  }}^{r_1,...,r_{i-1},r_{i+2},...,r_d}
\sum_{ \substack {A \subset \{1,...,d\} \setminus \{i\}  \\ A \neq \emptyset} }  
\Bigg[
 \prod_{ \substack{i' \in A ,\ell_i'=\ell_i \\
\ell_{i+1}' = \ell_{i+1}
}}
 \nu_{i',\mathbf x}^{\mathcal{Z},\ell_{i'}', \ell_{i'+1}'}(t) 
    \prod_{ \substack{
    i'' \in \{1,...,d\} \setminus (\{ i \} \cup A) \\
    \ell_i'=\ell_i,\ell_{i+1}' = \ell_{i+1}
    }} 
 {\hat{\mathcal{Z}}_{i'',\mathbf x}^{\ell_{i''}', \ell_{i'' + 1}' } (t) }^2 
 \Bigg]   
 \nu_{i,\mathbf x}^{\mathcal{Z},\ell_{i}, \ell_{i+1}}(t) 
 }_{(b)}  \nonumber \\
\overset{(a)}{\approx} &~ \nu_{\mathbf x}^p(t) 
 + 
   \sum_{\substack{\ell_1',...,\ell_{i-1}',\ell_{i+2}',...,\ell_d' = 1  }}^{r_1,...,r_{i-1},r_{i+2},...,r_d}
\sum_{ \substack {A \subset \{1,...,d\} \setminus \{i\} \\ A \neq \emptyset} } 
\Bigg\{  \nonumber \\
& ~~~~~~~~~~\Bigg[
 \prod_{ \substack{i' \in A ,\ell_i'=\ell_i \\
\ell_{i+1}' = \ell_{i+1}
}}
 \nu_{i',\mathbf x}^{\mathcal{Z},\ell_{i'}', \ell_{i'+1}'}(t) 
    \prod_{ \substack{
    i'' \in \{1,...,d\} \setminus (\{ i \} \cup A) \\
    \ell_i'=\ell_i,\ell_{i+1}' = \ell_{i+1}
    }} 
 {\hat{\mathcal{Z}}_{i'',\mathbf x}^{\ell_{i''}', \ell_{i'' + 1}' } (t) }^2 
 \Bigg]
  \left[ {  \mathcal{Z}_i^{\ell_i, \ell_{i+1}} (x_i) }^2 - { \hat{\mathcal{Z}}_{i,x_i}^{\ell_i, \ell_{i+1}} (t) }^2  \right]
  \Bigg\}, \label{Var-u-1}
\end{align}
where in $(a)$ we neglected $\hat{\mathcal{Z}}^{\ell_{i},\ell_{i+1}}_{i, x_i}(t) -  \hat{\mathcal{Z}}^{\ell_{i},\ell_{i+1}}_{i,\mathbf x}(t)$ with similar arguments as before, and neglected term $(b)$ as it is small compared to $\nu_{\mathbf x}^p(t) $ as we assume $r_1,...,r_d$ are large.

With this conditional-Gaussian approximation, \eqref{sum-pro-1-2} becomes
\begin{align} 
\label{sum-pro-1-3}
&~p_{\mathbf x \rightarrow [i,\ell_i,\ell_{i+1},x_i] }(t, \mathcal{Z}_{i}^{\ell_i,\ell_{i+1}}(x_i))  \nonumber \\
=&~ \log \int_{u_{\mathbf x}} 
p
\left(v_{\mathbf x} | u_{\mathbf x} \right) \times
\mathcal{N}\left(\mathsf u_{\mathbf x} ; \mathbb{E}\left\{\mathsf{u}_{\mathbf x} |  {\mathsf{Z}}_{i}^{\ell_i,\ell_{i+1}}(x_i)  =  {\mathcal{Z}}_{i}^{\ell_i,\ell_{i+1}}(x_i)  \right\}, \operatorname{Var}\left\{\mathsf{u}_{\mathbf x} | {\mathsf{Z}}_{i}^{\ell_i,\ell_{i+1}}(x_i)  =  {\mathcal{Z}}_{i}^{\ell_i,\ell_{i+1}}(x_i)   \right\}\right) + \text{const} \nonumber \\
 =&~ 
 H_{\mathbf x} \left( \mathbb{E}\left\{\mathsf{u}_{\mathbf x} |  {\mathsf{Z}}_{i}^{\ell_i,\ell_{i+1}}(x_i)  =  {\mathcal{Z}}_{i}^{\ell_i,\ell_{i+1}}(x_i)  \right\}, \operatorname{Var}\left\{\mathsf{u}_{\mathbf x} | {\mathsf{Z}}_{i}^{\ell_i,\ell_{i+1}}(x_i)  =  {\mathcal{Z}}_{i}^{\ell_i,\ell_{i+1}}(x_i)   \right\}    ; v_{\mathbf x}  \right) + \text{const}
\end{align}
in terms of the function
\begin{align} 
\label{H-1}
H_{\mathbf x} \left(\hat{p}, \nu^{p} ; v\right)  \triangleq  \log \int_{u} p(v | u) \mathcal{N}\left(u ; \hat{p}, \nu^{p}\right).
\end{align}

Rewriting \eqref{sum-pro-1-3} using a Taylor series expansion in ${ \mathcal{Z}_i^{\ell_i, \ell_{i+1}} (x_i) }$ about the point ${ \hat{\mathcal{Z}}_{i,x_i}^{\ell_i, \ell_{i+1}} (t) }$, using \eqref{E-u-1} and \eqref{Var-u-1} we obtain
\begin{align} 
&p_{\mathbf x \rightarrow [i,\ell_i,\ell_{i+1},x_i] }(t, \mathcal{Z}_{i}^{\ell_i,\ell_{i+1}}(x_i)) \nonumber \\
\approx &~  \text{const} + H_{\mathbf x}( \hat{p}_{\mathbf x} (t) , \nu^p_{\mathbf x} (t) ; v_{\mathbf x} )  \nonumber \\
&+  \left[
\sum_{\substack{\ell_1',...,\ell_{i-1}',\ell_{i+2}',...,\ell_d' = 1  }}^{r_1,...,r_{i-1},r_{i+2},...,r_d}
\hat{\mathcal{Z}}^{\ell_1',\ell_2'}_{1,\mathbf x}(t)
 \cdots \hat{\mathcal{Z}}^{\ell_{i-1}',\ell_i}_{i-1,\mathbf x}(t) 
\left[ \mathcal{Z}_i^{\ell_i,\ell_{i+1}} (x_i) -  \hat{\mathcal{Z}}^{\ell_{i},\ell_{i+1}}_{i,x_i}(t)  \right] 
\hat{\mathcal{Z}}^{\ell_{i+1},\ell_{i+2}'}_{i+1,\mathbf x}(t) \cdots
\hat{\mathcal{Z}}^{\ell_{d}',\ell_1'}_{d,\mathbf x}(t)
\right]
 \nonumber \\
&~ \times H_{\mathbf x}'( \hat{p}_{\mathbf x} (t) , \nu^p_{\mathbf x} (t) ; v_{\mathbf x} ) 
 \nonumber \\
 &+ \frac{1}{2} \left[ 
\sum_{\substack{\ell_1',...,\ell_{i-1}',\ell_{i+2}',...,\ell_d' = 1  }}^{r_1,...,r_{i-1},r_{i+2},...,r_d}
\hat{\mathcal{Z}}^{\ell_1',\ell_2'}_{1,\mathbf x}(t)
 \cdots \hat{\mathcal{Z}}^{\ell_{i-1}',\ell_i}_{i-1,\mathbf x}(t) 
\left[ \mathcal{Z}_i^{\ell_i,\ell_{i+1}} (x_i) -  \hat{\mathcal{Z}}^{\ell_{i},\ell_{i+1}}_{i,x_i}(t)  \right] 
\hat{\mathcal{Z}}^{\ell_{i+1},\ell_{i+2}'}_{i+1,\mathbf x}(t) \cdots
\hat{\mathcal{Z}}^{\ell_{d}',\ell_1'}_{d,\mathbf x}(t)
\right]^2
 \nonumber \\
 &~ \times H_{\mathbf x}''( \hat{p}_{\mathbf x} (t) , \nu^p_{\mathbf x} (t) ; v_{\mathbf x} ) \nonumber \\
& +  \dot H_{\mathbf x}( \hat{p}_{\mathbf x} (t) , \nu^p_{\mathbf x} (t) ; v_{\mathbf x} )  \times  
 \sum_{\substack{\ell_1',...,\ell_{i-1}',\ell_{i+2}',...,\ell_d' = 1  }}^{r_1,...,r_{i-1},r_{i+2},...,r_d}
\sum_{ \substack {A \subset \{1,...,d\} \setminus \{i\} \\ A \neq \emptyset} } 
\Bigg\{  \nonumber \\
& ~~~~~~~~~~\Bigg[
 \prod_{ \substack{i' \in A ,\ell_i'=\ell_i \\
\ell_{i+1}' = \ell_{i+1}
}}
 \nu_{i',\mathbf x}^{\mathcal{Z},\ell_{i'}', \ell_{i'+1}'}(t) 
    \prod_{ \substack{
    i'' \in \{1,...,d\} \setminus (\{ i \} \cup A) \\
    \ell_i'=\ell_i,\ell_{i+1}' = \ell_{i+1}
    }} 
 {\hat{\mathcal{Z}}_{i'',\mathbf x}^{\ell_{i''}', \ell_{i'' + 1}' } (t) }^2 
 \Bigg]
  \left[ {  \mathcal{Z}_i^{\ell_i, \ell_{i+1}} (x_i) }^2 - { \hat{\mathcal{Z}}_{i,x_i}^{\ell_i, \ell_{i+1}} (t) }^2  \right]
  \Bigg\}    \nonumber     \\
\label{sum-pro-Taylor-1-A}
\overset{(a)}{=} &~  \text{const} + \nonumber \\  
& \left[
\sum_{\substack{\ell_1',...,\ell_{i-1}',\ell_{i+2}',...,\ell_d' = 1  }}^{r_1,...,r_{i-1},r_{i+2},...,r_d}
\hat{\mathcal{Z}}^{\ell_1',\ell_2'}_{1,\mathbf x}(t)
 \cdots \hat{\mathcal{Z}}^{\ell_{i-1}',\ell_i}_{i-1,\mathbf x}(t) 
\left[ \mathcal{Z}_i^{\ell_i,\ell_{i+1}} (x_i) -  \hat{\mathcal{Z}}^{\ell_{i},\ell_{i+1}}_{i,x_i}(t)  \right] 
\hat{\mathcal{Z}}^{\ell_{i+1},\ell_{i+2}'}_{i+1,\mathbf x}(t) \cdots
\hat{\mathcal{Z}}^{\ell_{d}',\ell_1'}_{d,\mathbf x}(t)
\right]
 \hat{s}_{\mathbf x}(t) \nonumber \\
 &+ \frac{1}{2} \left[ 
\sum_{\substack{\ell_1',...,\ell_{i-1}',\ell_{i+2}',...,\ell_d' = 1  }}^{r_1,...,r_{i-1},r_{i+2},...,r_d}
\hat{\mathcal{Z}}^{\ell_1',\ell_2'}_{1,\mathbf x}(t)
 \cdots \hat{\mathcal{Z}}^{\ell_{i-1}',\ell_i}_{i-1,\mathbf x}(t) 
\left[ \mathcal{Z}_i^{\ell_i,\ell_{i+1}} (x_i) -  \hat{\mathcal{Z}}^{\ell_{i},\ell_{i+1}}_{i,x_i}(t)  \right] 
\hat{\mathcal{Z}}^{\ell_{i+1},\ell_{i+2}'}_{i+1,\mathbf x}(t) \cdots
\hat{\mathcal{Z}}^{\ell_{d}',\ell_1'}_{d,\mathbf x}(t)
\right]^2 \nonumber \\
 &~ \times \left( - \nu_{\mathbf x}^s(t) \right)  \nonumber \\
& + 
 \frac{1}{2} \left [  {\hat{s}_{\mathbf x}(t)}^2  -  {\nu_{\mathbf x}^s(t)} \right]    \times
  \sum_{\substack{\ell_1',...,\ell_{i-1}',\ell_{i+2}',...,\ell_d' = 1  }}^{r_1,...,r_{i-1},r_{i+2},...,r_d}
\sum_{ \substack {A \subset \{1,...,d\} \setminus \{i\} \\ A \neq \emptyset} } 
\Bigg\{  \nonumber \\
& ~~~~~~~~~~\Bigg[
 \prod_{ \substack{i' \in A ,\ell_i'=\ell_i \\
\ell_{i+1}' = \ell_{i+1}
}}
 \nu_{i',\mathbf x}^{\mathcal{Z},\ell_{i'}', \ell_{i'+1}'}(t) 
    \prod_{ \substack{
    i'' \in \{1,...,d\} \setminus (\{ i \} \cup A) \\
    \ell_i'=\ell_i,\ell_{i+1}' = \ell_{i+1}
    }} 
 {\hat{\mathcal{Z}}_{i'',\mathbf x}^{\ell_{i''}', \ell_{i'' + 1}' } (t) }^2 
 \Bigg]
  \left[ {\mathcal{Z}_i^{\ell_i, \ell_{i+1}} (x_i) }^2 - { \hat{\mathcal{Z}}_{i,x_i}^{\ell_i, \ell_{i+1}} (t) }^2  \right]
  \Bigg\}    
  \nonumber \\
 \overset{(b)}{=} &~  \text{const} +  
\left[
\sum_{\substack{\ell_1',...,\ell_{i-1}',\ell_{i+2}',...,\ell_d' = 1  }}^{r_1,...,r_{i-1},r_{i+2},...,r_d}
\hat{\mathcal{Z}}^{\ell_1',\ell_2'}_{1,\mathbf x}(t)
 \cdots \hat{\mathcal{Z}}^{\ell_{i-1}',\ell_i}_{i-1,\mathbf x}(t) 
\hat{\mathcal{Z}}^{\ell_{i+1},\ell_{i+2}'}_{i+1,\mathbf x}(t) \cdots
\hat{\mathcal{Z}}^{\ell_{d}',\ell_1'}_{d,\mathbf x}(t)
\right] \mathcal{Z}_i^{\ell_i,\ell_{i+1}} (x_i) 
 \hat{s}_{\mathbf x}(t) \nonumber \\
 &+ \frac{1}{2} \left[ 
\sum_{\substack{\ell_1',...,\ell_{i-1}',\ell_{i+2}',...,\ell_d' = 1  }}^{r_1,...,r_{i-1},r_{i+2},...,r_d}
\hat{\mathcal{Z}}^{\ell_1',\ell_2'}_{1,\mathbf x}(t)
 \cdots \hat{\mathcal{Z}}^{\ell_{i-1}',\ell_i}_{i-1,\mathbf x}(t) 
\hat{\mathcal{Z}}^{\ell_{i+1},\ell_{i+2}'}_{i+1,\mathbf x}(t) \cdots
\hat{\mathcal{Z}}^{\ell_{d}',\ell_1'}_{d,\mathbf x}(t)
\right]^2 
\nonumber \\
&~\times \left[ \mathcal{Z}_i^{\ell_i,\ell_{i+1}} (x_i)^2 - 2 \hat{\mathcal{Z}}^{\ell_{i},\ell_{i+1}}_{i,x_i}(t) \mathcal{Z}_i^{\ell_i,\ell_{i+1}} (x_i) \right] 
 \left( - \nu_{\mathbf x}^s(t) \right)  \nonumber \\
& + 
 \frac{1}{2} \left [  {\hat{s}_{\mathbf x}(t)}^2  -  {\nu_{\mathbf x}^s(t)} \right]     \left[ {\mathcal{Z}_i^{\ell_i, \ell_{i+1}} (x_i) }^2  \right] \times  \nonumber \\
&  \sum_{\substack{\ell_1',...,\ell_{i-1}',\ell_{i+2}',...,\ell_d' = 1  }}^{r_1,...,r_{i-1},r_{i+2},...,r_d}
\sum_{ \substack {A \subset \{1,...,d\} \setminus \{i\} \\ A \neq \emptyset} } 
\Bigg[
 \prod_{ \substack{i' \in A ,\ell_i'=\ell_i \\
\ell_{i+1}' = \ell_{i+1}
}}
 \nu_{i',\mathbf x}^{\mathcal{Z},\ell_{i'}', \ell_{i'+1}'}(t) 
    \prod_{ \substack{
    i'' \in \{1,...,d\} \setminus (\{ i \} \cup A) \\
    \ell_i'=\ell_i,\ell_{i+1}' = \ell_{i+1}
    }} 
 {\hat{\mathcal{Z}}_{i'',\mathbf x}^{\ell_{i''}', \ell_{i'' + 1}' } (t) }^2 
 \Bigg]
  \nonumber \\
   \overset{(c)}{\approx} &~  \text{const} +  
\left[
\sum_{\substack{\ell_1',...,\ell_{i-1}',\ell_{i+2}',...,\ell_d' = 1  }}^{r_1,...,r_{i-1},r_{i+2},...,r_d}
\hat{\mathcal{Z}}^{\ell_1',\ell_2'}_{1,\mathbf x}(t)
 \cdots \hat{\mathcal{Z}}^{\ell_{i-1}',\ell_i}_{i-1,\mathbf x}(t) 
\hat{\mathcal{Z}}^{\ell_{i+1},\ell_{i+2}'}_{i+1,\mathbf x}(t) \cdots
\hat{\mathcal{Z}}^{\ell_{d}',\ell_1'}_{d,\mathbf x}(t)
\right] \mathcal{Z}_i^{\ell_i,\ell_{i+1}} (x_i) 
 \hat{s}_{\mathbf x}(t) \nonumber \\
 &+ \frac{1}{2} \left[ 
\sum_{\substack{\ell_1',...,\ell_{i-1}',\ell_{i+2}',...,\ell_d' = 1  }}^{r_1,...,r_{i-1},r_{i+2},...,r_d}
\hat{\mathcal{Z}}^{\ell_1',\ell_2'}_{1, x_1}(t)
 \cdots \hat{\mathcal{Z}}^{\ell_{i-1}',\ell_i}_{i-1, x_{i-1}}(t) 
\hat{\mathcal{Z}}^{\ell_{i+1},\ell_{i+2}'}_{i+1,x_{i+1}}(t) \cdots
\hat{\mathcal{Z}}^{\ell_{d}',\ell_1'}_{d, x_d}(t)
\right]^2 
\nonumber \\
&~\times \left[ \mathcal{Z}_i^{\ell_i,\ell_{i+1}} (x_i)^2 - 2 \hat{\mathcal{Z}}^{\ell_{i},\ell_{i+1}}_{i,x_i}(t) \mathcal{Z}_i^{\ell_i,\ell_{i+1}} (x_i) \right] 
 \left( - \nu_{\mathbf x}^s(t) \right)  \nonumber \\
& + 
 \frac{1}{2} \left [  {\hat{s}_{\mathbf x}(t)}^2  -  {\nu_{\mathbf x}^s(t)} \right]     \left[ {\mathcal{Z}_i^{\ell_i, \ell_{i+1}} (x_i) }^2  \right] \times \nonumber \\
 & \sum_{\substack{\ell_1',...,\ell_{i-1}',\ell_{i+2}',...,\ell_d' = 1  }}^{r_1,...,r_{i-1},r_{i+2},...,r_d}
\sum_{ \substack {A \subset \{1,...,d\} \setminus \{i\} \\ A \neq \emptyset} }  
\Bigg[
 \prod_{ \substack{i' \in A ,\ell_i'=\ell_i \\
\ell_{i+1}' = \ell_{i+1}
}}
 \nu_{i', x_{i'}}^{\mathcal{Z},\ell_{i'}', \ell_{i'+1}'}(t) 
    \prod_{ \substack{
    i'' \in \{1,...,d\} \setminus (\{i \} \cup A) \\
    \ell_i'=\ell_i,\ell_{i+1}' = \ell_{i+1}
    }} 
 {\hat{\mathcal{Z}}_{i'', x_{i''}}^{\ell_{i''}', \ell_{i'' + 1}' } (t) }^2 
 \Bigg],
\end{align}
where $(a)$ follows from putting some constants without $\mathcal{Z}_i^{\ell_i,\ell_{i+1}} (x_i)$ in ``const'' and from below definitions
\begin{align} 
\label{hat-s-1}
\hat{s}_{\mathbf x}(t) &\triangleq  H_{\mathbf x}'( \hat{p}_{\mathbf x} (t) , \nu^p_{\mathbf x} (t) ; v_{\mathbf x} ) \\
\label{nu-s-1}
\nu_{\mathbf x}^s(t)  & \triangleq  - H_{\mathbf x}''( \hat{p}_{\mathbf x} (t) , \nu^p_{\mathbf x} (t) ; v_{\mathbf x} )
\end{align}
together with the fact that according to~\cite{parker2014bilinear} the following general relationship holds
\begin{align} 
\label{dot-H-1}
{\omega}_{\mathbf{x} }(t) \triangleq
\dot H_{\mathbf x}( \hat{p}_{\mathbf x} (t) , \nu^p_{\mathbf x} (t) ; v_{\mathbf x} )  
= \frac{1}{2} \left [ H_{\mathbf x}'( \hat{p}_{\mathbf x} (t) , \nu^p_{\mathbf x} (t) ; v_{\mathbf x} )^2  +  H_{\mathbf x}''( \hat{p}_{\mathbf x} (t) , \nu^p_{\mathbf x} (t) ; v_{\mathbf x} )  \right ]    
=  \frac{1}{2} \left [  {\hat{s}_{\mathbf x}(t)}^2  -  {\nu_{\mathbf x}^s(t)} \right],      
\end{align}
and $(b)$ follows from putting some constants without $\mathcal{Z}_i^{\ell_i,\ell_{i+1}} (x_i)$ in ``const''. Moreover, $(c)$ follows as we replace $\hat{\mathcal{Z}}_{i,\mathbf x}^{\ell_i,\ell_{i+1}} (t)$ and ${\nu}_{i,\mathbf x}^{\mathcal{Z},\ell_i,\ell_{i+1}} (t)$ by $\hat{\mathcal{Z}}_{i, x_i}^{\ell_i,\ell_{i+1}} (t)$ and ${\nu}_{i, x_i}^{\mathcal{Z},\ell_i,\ell_{i+1}} (t)$, respectively, since recall that we assumed that data size is large and $p_{\mathbf x \leftarrow [i,\ell_i,\ell_{i+1},x_i] } (t,\cdot)$ and $p_{ [i,\ell_i,\ell_{i+1},x_i] } (t,\cdot)$ differ by only one term (which vanishes relative to the others in the large-system limit). See \cite{rangan2011generalized,parker2014bilinear,montanari2012graphical} for similar arguments.

It is argued in \cite{rangan2011generalized,parker2014bilinear} that given \eqref{hat-s-1}-\eqref{nu-s-1} and \eqref{H-1} the followings hold
\begin{align} 
\label{hat-s-2-A}
\hat{s}_{\mathbf x}(t) &= \frac{1}{ \nu_{\mathbf x}^p(t) } \left(\hat{u}_{\mathbf x}(t)-\hat{p}_{\mathbf x}(t)\right) \\
\label{nu-s-2-A}
{\nu}_{\mathbf x}^{s}(t) &= \frac{1}{\nu_{\mathbf x}^{p}(t)}\left(1-\frac{\nu_{\mathbf x}^{u}(t)}{\nu_{\mathbf x}^{p}(t)}\right),
\end{align}
for the conditional mean and variance
\begin{align} 
\label{hat-u-1-A}
\hat{u}_{\mathbf x}(t)  &\triangleq \mathbb{E} \left\{ \mathsf{u}_{\mathbf x} | \mathsf{p}_{\mathbf x} = \hat{p}_{\mathbf x}(t) ; \nu_{\mathbf x}^{p}(t)\right\} \\
\label{nu-u-1-A}
 \nu_{\mathbf x}^{u}(t) & \triangleq \mathrm{Var}\left\{\mathsf{u}_{\mathbf x} | \mathsf{p}_{\mathbf x}=\hat{p}_{\mathbf x}(t) ; \nu_{\mathbf x}^{p}(t)\right \}, 
 \end{align}
computed according to the conditional pdf
\begin{align} 
\label{condtional-pdf-1-A}
p_{\mathsf{u}_{\mathbf x} | \mathsf{p}_{\mathbf x} } \left( u | \hat{p} ; \nu^{p} \right) \triangleq   \frac{
p_{\mathsf{v}_{\mathbf x} | \mathsf{u}_{\mathbf x} } \left( v_{\mathbf x} | u_{\mathbf x} \right)
\mathcal{N}( u_{\mathbf x} ;  \hat{p}_{\mathbf x}(t) , \nu_{\mathbf x}^p(t) )
}{\int_{u^{\prime}}  p_{\mathsf{v}_{\mathbf x} | \mathsf{u}_{\mathbf x} } \left( v_{\mathbf x} | u' \right)
\mathcal{N}( u' ;  \hat{p}_{\mathbf x}(t) , \nu_{\mathbf x}^p(t) )   }.
 \end{align}

In fact, \eqref{condtional-pdf-1-A} is TeG-AMP's the $t$-th iteration approximation to the true marginal posterior $p_{\mathsf{u}_{\mathbf x} | \mathsf V }(\cdot | \mathcal{V})$. We note that \eqref{condtional-pdf-1-A} can also be interpreted as the posterior pdf for $\mathsf{u}_{\mathbf x}$ given the likelihood $p_{\mathsf{v}_{\mathbf x} | \mathsf{u}_{\mathbf x} } \left( v_{\mathbf x} | \cdot \right)$ from \eqref{likelihood-decoupled} and the prior $\mathsf{u}_{\mathbf x} \sim \mathcal{N}( \hat{p}_{\mathbf x}(t) , \nu_{\mathbf x}^p(t) )$ that is implicitly assumed by the $t$-th iteration TeG-AMP.

As a result, we have
\begin{align*}
    &~p_{\mathbf{x} \rightarrow [i,\ell_i,\ell_{i+1},x_i]} \left(t,\mathcal{Z}_{i}(\ell_i,x_i,\ell_{i+1}) \right) \\
{\approx} &~  \text{const} +  
\left[
\sum_{\substack{\ell_1',...,\ell_{i-1}',\ell_{i+2}',...,\ell_d' = 1  }}^{r_1,...,r_{i-1},r_{i+2},...,r_d}
\hat{\mathcal{Z}}^{\ell_1',\ell_2'}_{1,\mathbf x}(t)
 \cdots \hat{\mathcal{Z}}^{\ell_{i-1}',\ell_i}_{i-1,\mathbf x}(t) 
\hat{\mathcal{Z}}^{\ell_{i+1},\ell_{i+2}'}_{i+1,\mathbf x}(t) \cdots
\hat{\mathcal{Z}}^{\ell_{d}',\ell_1'}_{d,\mathbf x}(t)
\right] \mathcal{Z}_i^{\ell_i,\ell_{i+1}} (x_i) 
 \hat{s}_{\mathbf x}(t) \nonumber \\
 &+ \frac{1}{2} \left[ 
\sum_{\substack{\ell_1',...,\ell_{i-1}',\ell_{i+2}',...,\ell_d' = 1  }}^{r_1,...,r_{i-1},r_{i+2},...,r_d}
\hat{\mathcal{Z}}^{\ell_1',\ell_2'}_{1, x_1}(t)
 \cdots \hat{\mathcal{Z}}^{\ell_{i-1}',\ell_i}_{i-1, x_{i-1}}(t) 
\hat{\mathcal{Z}}^{\ell_{i+1},\ell_{i+2}'}_{i+1,x_{i+1}}(t) \cdots
\hat{\mathcal{Z}}^{\ell_{d}',\ell_1'}_{d, x_d}(t)
\right]^2 
\nonumber \\
&~\times \left[ \mathcal{Z}_i^{\ell_i,\ell_{i+1}} (x_i)^2 - 2 \hat{\mathcal{Z}}^{\ell_{i},\ell_{i+1}}_{i,x_i}(t) \mathcal{Z}_i^{\ell_i,\ell_{i+1}} (x_i) \right] 
 \left( - \nu_{\mathbf x}^s(t) \right)  \nonumber \\
& + 
 \frac{1}{2} \left [  {\hat{s}_{\mathbf x}(t)}^2  -  {\nu_{\mathbf x}^s(t)} \right]     \left[ {\mathcal{Z}_i^{\ell_i, \ell_{i+1}} (x_i) }^2  \right] \times \nonumber \\
 & \sum_{\substack{\ell_1',...,\ell_{i-1}',\ell_{i+2}',...,\ell_d' = 1  }}^{r_1,...,r_{i-1},r_{i+2},...,r_d}
\sum_{ \substack {A \subset \{1,...,d\} \setminus \{i\} \\ A \neq \emptyset} }  
\Bigg[
 \prod_{ \substack{i' \in A ,\ell_i'=\ell_i \\
\ell_{i+1}' = \ell_{i+1}
}}
 \nu_{i', x_{i'}}^{\mathcal{Z},\ell_{i'}', \ell_{i'+1}'}(t) 
    \prod_{ \substack{
    i'' \in \{1,...,d\} \setminus (\{i \} \cup A) \\
    \ell_i'=\ell_i,\ell_{i+1}' = \ell_{i+1}
    }} 
 {\hat{\mathcal{Z}}_{i'', x_{i''}}^{\ell_{i''}', \ell_{i'' + 1}' } (t) }^2 
 \Bigg] \\
= &~ {\text{const}} + 
\Big[
 {\sum_{\mathbf{l}'_{\setminus i} = 1}^{\mathbf{r}_{\setminus i}}}
  {\hat{\mathcal{Z}}^{\ell'}_{\setminus i, \mathbf{x}}(t)}
\Big] \mathcal{Z}_i^{\ell} (x_i) 
 \hat{s}_{\mathbf{x} }(t) + 
 \frac{1}{2} {\omega}_{\mathbf{x} }(t){\mathcal{Z}_i^{\ell} (x_i) }^2 {\zeta}_{x}(t) 
 - \frac{ 1 }{2} \Big[ 
 {\sum_{\mathbf{l}'_{\setminus i} = 1}^{\mathbf{r}_{\setminus i}}}
 {\hat{\mathcal{Z}}^{\ell'}_{\setminus i, x}(t)}
\Big]^2  \left[\mathcal{Z}_i^{\ell} (x_i)^2 - 2 \hat{\mathcal{Z}}^{\ell}_{i,x_i}(t) \mathcal{Z}_i^{\ell} (x_i) \right] \nu_{\mathbf{x} }^s(t), \nonumber 
\end{align*}
where ${{\mathcal{Z}}_{i, \mathbf{x}}^{\ell}(t)} \triangleq {{\mathcal{Z}}_{i, \mathbf{x}}^{\ell_i,\ell_{i+1}}(t)}$,
${\hat{\mathcal{Z}}^{\ell}_{i,x_i}(t)} \triangleq {\hat{\mathcal{Z}}_{i, x_i}^{\ell_i,\ell_{i+1}}(t)}$, $\nu_{i, x_i}^{\mathcal{Z},\ell} \triangleq \nu_{i,x_i }^{\mathcal{Z},\ell_i, \ell_{i+1}}$,
$\mathbf r_{\setminus i} \triangleq  \mathbf r \backslash (r_i, r_{i+1})$, $\mathbf l' _{\setminus i} \triangleq \mathbf l' \backslash (\ell_i', \ell_{i+1}')$ and
\begin{align*}
{  \hat{\mathcal{Z}}^{\ell'}_{\setminus i,{\mathbf{x} }}(t)} 
& = \hat{\mathcal{Z}}^{\ell'}_{1,\mathbf{x} }(t)
 \cdots 
\hat{\mathcal{Z}}^{\ell_{i-1}',\ell_i}_{i-1,\mathbf{x} }(t) 
\hat{\mathcal{Z}}^{\ell_{i+1},\ell_{i+2}'}_{i+1,\mathbf{x} }(t) \cdots
\hat{\mathcal{Z}}^{\ell'}_{d,\mathbf{x} }(t), \nonumber \\
{  \hat{\mathcal{Z}}^{\ell'}_{\setminus i, x}(t)} 
&= \hat{\mathcal{Z}}^{\ell'}_{1, x_1}(t)
 \cdots \hat{\mathcal{Z}}^{\ell_{i-1}',\ell_i}_{i-1, x_{i-1}}(t) 
\hat{\mathcal{Z}}^{\ell_{i+1},\ell_{i+2}'}_{i+1,x_{i+1}}(t)
\cdots
\hat{\mathcal{Z}}^{\ell'}_{d, x_d}(t), \nonumber \\
\zeta_{x}(t) &=  {\sum_{\mathbf{l}'_{\setminus i} = 1}^{\mathbf{r}_{\setminus i}}}
\sum_{ \substack {A \subset \mathbb{D} \setminus \{i\} \\ A \neq \emptyset} } \Bigg[
 \prod_{ \substack{i' \in A ,\ell_i'=\ell_i \\
\ell_{i+1}' = \ell_{i+1}
}}
 \nu_{i', x_{i'}}^{{ {\mathcal{Z}}},\ell'}(t) 
   \times  
\prod_{ \substack{
    i'' \in \mathbb{D} \setminus (\{ i \} \cup A)\\
    \ell_i'=\ell_i, 
    \ell_{i+1}' = \ell_{i+1}
    }} 
 {\hat{\mathcal{Z}}_{i'', x_{i''}}^{\ell_{i''}', \ell_{i'' + 1}' } (t) }^2 
 \Bigg]. \nonumber
\end{align*}
Then we complete the derivation.
\section{Derivation of Equation \eqref{var-to-fac-1}} \label{appendixC}

\textbf{Element-wise Result:}~
{\it
$p_{\mathbf x \leftarrow [i,\ell_i, \ell_{i+1},x_i] } \left(t+1,\mathcal{Z}_{i}(\ell_i,x_i,\ell_{i+1}) \right)$ in \eqref{sum-pro-2} can be approximated by
\begin{align*} 
&p_{\mathbf x \leftarrow [i,\ell_i, \ell_{i+1},x_i] } \left(t+1,\mathcal{Z}_{i}(\ell_i,x_i,\ell_{i+1}) \right) \ {\approx}\ \text{const}\ + 
\log\Big( p\left(\mathcal{Z}_{i}^{\ell}(x_i)\right) \mathcal{N}\left({\mathcal{Z}_i^{\ell} (x_i) };\hat{r}_{\mathbf{x} , \mathbf q^i},  \nu^r_{\mathbf{x} , \mathbf q^i} \right) \Big),
\end{align*}
where ${{\mathcal{Z}}_{i}^{\ell}} \triangleq {{\mathcal{Z}}_{i}^{\ell_i,\ell_{i+1}}}$ and
\begin{align*}
\frac{1}{\nu^r_{\mathbf{x} ,\mathbf q^i}}  & \triangleq   \sum_{\substack{{ {\mathbf{x}':x'_i = x_i}},\\ \mathbf{x} ' \neq \mathbf{x} }}
\Bigg\{    
 \Big[ 
 {\sum_{\mathbf{l}'_{\setminus i} = 1}^{\mathbf{r}_{\setminus i}}}
 {\hat{\mathcal{Z}}^{\ell'}_{\setminus i, x'}(t)}
\Big]^2
  \nu_{\mathbf{x} '}^s(t) -  
  {\omega}_{\mathbf{x} '}(t) {\zeta}_{x}(t) \Bigg\}, \\
\hat{r}_{\mathbf{x} , \mathbf q^i} &\triangleq \nu^r_{\mathbf{x} ,\mathbf q^i}  
 \sum_{\substack{{ {\mathbf{x}':x'_i = x_i}},\\ \mathbf{x} ' \neq \mathbf{x} }}
\Bigg\{  
\Big[
 {\sum_{\mathbf{l}'_{\setminus i} = 1}^{\mathbf{r}_{\setminus i}}}
 {\hat{\mathcal{Z}}^{\ell'}_{\setminus i, \mathbf{x}'}(t)}
\Big]
 \hat{s}_{\mathbf{x} '}(t) 
  \Bigg\} 
 + \hat{\mathcal{Z}}^{\ell}_{i,x_i}(t)
 \Bigg\{ 1 + \nu^r_{\mathbf{x} ,\mathbf q^i}
 \sum_{\substack {{ {\mathbf{x}':x'_i = x_i}},\\ \mathbf{x} ' \neq \mathbf{x} }}
   {\omega}_{\mathbf{x} '}(t) {\zeta}_{x}(t) \Bigg\}.
\end{align*}
}


\emph{Derivation:}
Recall \eqref{sum-pro-2} as
\begin{align*}
p_{\mathbf{x} \leftarrow x_i}(\mathcal{Z}_{i}(:,x_{i},:))
=&~\text{const} + \log p\left(\mathcal{Z}_{i}(:,x_i,:)\right)
+ \sum_{\{\mathbf{x}':x'_i = x_i,~\mathbf{x}' \neq \mathbf{x}\}}p_{\mathbf{x}' \rightarrow x_i},
\end{align*}
which can be expanded as the following element-wise version
\begin{align} 
\label{sum-pro-2-A_2}
 p_{\mathbf x \leftarrow [i,\ell_i,\ell_{i+1},x_i] }(t+1,\mathcal{Z}_{i}^{\ell_i,\ell_{i+1}}(x_i)) 
= \log p
\left(\mathcal{Z}_{i}^{\ell_i,\ell_{i+1}}(x_i)\right)  + \sum_{ {\{\mathbf{x}':x'_i = x_i,~\mathbf{x}' \neq \mathbf{x}\}}}   
p_{\mathbf x' \rightarrow [i,\ell_i,\ell_{i+1}, x_i] } (t, \mathcal{Z}_{i}^{\ell_i,\ell_{i+1}}(x_i) ) + \text{const}. 
\end{align}

Starting with \eqref{sum-pro-2-A_2} and plugging in \eqref{sum-pro-Taylor-1} we obtain
\begin{align} 
&~ p_{\mathbf x \leftarrow [i,\ell_i,\ell_{i+1},x_i] }(t+1,\mathcal{Z}_{i}^{\ell_i,\ell_{i+1}}(x_i))  \nonumber \\
=&~ \log p\left(\mathcal{Z}_{i}^{\ell_i,\ell_{i+1}}(x_i)\right)  + \sum_{ {\{\mathbf{x}':x'_i = x_i,~\mathbf{x}' \neq \mathbf{x}\}}}   p_{\mathbf x' \rightarrow [i,\ell_i,\ell_{i+1}, x_i] } (t, \mathcal{Z}_{i}^{\ell_i,\ell_{i+1}}(x_i) ) + \text{const}  \nonumber \\
 \overset{(a)}{\approx} &~ \text{const}  + 
 \log p\left(\mathcal{Z}_{i}^{\ell_i,\ell_{i+1}}(x_i)\right)  
 +  \sum_{ {\{\mathbf{x}':x'_i = x_i,~\mathbf{x}' \neq \mathbf{x}\}}}  
\Bigg\{    
 \nonumber \\
& 
\left[
\sum_{\substack{\ell_1',...,\ell_{i-1}',\ell_{i+2}',...,\ell_d' = 1  }}^{r_1,...,r_{i-1},r_{i+2},...,r_d}
\hat{\mathcal{Z}}^{\ell_1',\ell_2'}_{1,\mathbf x'}(t)
 \cdots \hat{\mathcal{Z}}^{\ell_{i-1}',\ell_i}_{i-1,\mathbf x'}(t) 
\hat{\mathcal{Z}}^{\ell_{i+1},\ell_{i+2}'}_{i+1,\mathbf x'}(t) \cdots
\hat{\mathcal{Z}}^{\ell_{d}',\ell_1'}_{d,\mathbf x'}(t)
\right] 
 \mathcal{Z}_i^{\ell_i,\ell_{i+1}} (x_i) 
 \hat{s}_{\mathbf x'}(t) 
 \nonumber \\
 &+ \frac{1}{2} 
 \left[ 
\sum_{\substack{\ell_1',...,\ell_{i-1}',\ell_{i+2}',...,\ell_d' = 1  }}^{r_1,...,r_{i-1},r_{i+2},...,r_d}
\hat{\mathcal{Z}}^{\ell_1',\ell_2'}_{1, x_1'}(t)
 \cdots \hat{\mathcal{Z}}^{\ell_{i-1}',\ell_i}_{i-1, x_{i-1}'}(t) 
\hat{\mathcal{Z}}^{\ell_{i+1},\ell_{i+2}'}_{i+1,x_{i+1}'}(t) \cdots
\hat{\mathcal{Z}}^{\ell_{d}',\ell_1'}_{d, x_d'}(t)
\right]^2 
\nonumber \\
&~\times \left[ \mathcal{Z}_i^{\ell_i,\ell_{i+1}} (x_i)^2 - 2 \hat{\mathcal{Z}}^{\ell_{i},\ell_{i+1}}_{i,x_i}(t) \mathcal{Z}_i^{\ell_i,\ell_{i+1}} (x_i) \right] 
 \left( - \nu_{\mathbf x'}^s(t) \right)  
 + 
 \frac{1}{2} \left [  {\hat{s}_{\mathbf x'}(t)}^2  -  {\nu_{\mathbf x'}^s(t)} \right]     \left[ {  \mathcal{Z}_i^{\ell_i, \ell_{i+1}} (x_i) }^2  \right] \times \nonumber \\
 & \sum_{\substack{\ell_1',...,\ell_{i-1}',\ell_{i+2}',...,\ell_d' = 1  }}^{r_1,...,r_{i-1},r_{i+2},...,r_d}
\sum_{ \substack {A \subset \{1,...,d\} \setminus \{i\} \\ A \neq \emptyset} } 
\Bigg[
 \prod_{ \substack{i' \in A ,\ell_i'=\ell_i \\
\ell_{i+1}' = \ell_{i+1}
}}
 \nu_{i', x_{i'}'}^{\mathcal{Z},\ell_{i'}', \ell_{i'+1}'}(t) 
    \prod_{ \substack{
    i'' \in \{1,...,d\} \setminus (\{ i \} \cup A) \\
    \ell_i'=\ell_i,\ell_{i+1}' = \ell_{i+1}
    }} 
 {\hat{\mathcal{Z}}_{i'', x_{i''}'}^{\ell_{i''}', \ell_{i'' + 1}' } (t) }^2 
 \Bigg]
 \Bigg\} \nonumber \\
 %
 %
= &~ \text{const}  + \log p\left(\mathcal{Z}_{i}^{\ell_i,\ell_{i+1}}(x_i)\right) +  
  { \mathcal{Z}_i^{\ell_i, \ell_{i+1}} (x_i) }^2  
  \sum_{ {\{\mathbf{x}':x'_i = x_i,~\mathbf{x}' \neq \mathbf{x}\}}}  
\Bigg\{     \nonumber \\
& 
\frac{1}{2} \left[ 
\sum_{\substack{\ell_1',...,\ell_{i-1}',\ell_{i+2}',...,\ell_d' = 1  }}^{r_1,...,r_{i-1},r_{i+2},...,r_d}
\hat{\mathcal{Z}}^{\ell_1',\ell_2'}_{1, x_1'}(t)
 \cdots \hat{\mathcal{Z}}^{\ell_{i-1}',\ell_i}_{i-1, x_{i-1}'}(t) 
\hat{\mathcal{Z}}^{\ell_{i+1},\ell_{i+2}'}_{i+1,x_{i+1}'}(t) \cdots
\hat{\mathcal{Z}}^{\ell_{d}',\ell_1'}_{d, x_d'}(t)
\right]^2
 \left( - \nu_{\mathbf x'}^s(t) \right)    
 +
 \frac{1}{2}    
  \left [  {\hat{s}_{\mathbf x'}(t)}^2  -  {\nu_{\mathbf x'}^s(t)} \right]  \times  \nonumber \\
 & \sum_{\substack{\ell_1',...,\ell_{i-1}',\ell_{i+2}',...,\ell_d' = 1  }}^{r_1,...,r_{i-1},r_{i+2},...,r_d}
\sum_{ \substack {A \subset \{1,...,d\} \setminus \{i\} \\ A \neq \emptyset} } 
\Bigg[
 \prod_{ \substack{i' \in A ,\ell_i'=\ell_i \\
\ell_{i+1}' = \ell_{i+1}
}}
 \nu_{i', x_{i'}'}^{\mathcal{Z},\ell_{i'}', \ell_{i'+1}'}(t) 
    \prod_{ \substack{
    i'' \in \{1,...,d\} \setminus (\{ i \} \cup A) \\
    \ell_i'=\ell_i,\ell_{i+1}' = \ell_{i+1}
    }} 
 {\hat{\mathcal{Z}}_{i'', x_{i''}'}^{\ell_{i''}', \ell_{i'' + 1}' } (t) }^2 
 \Bigg]
  \Bigg\}  
 \nonumber \\
%
\label{var-to-fac-1-A}
&  + \mathcal{Z}_i^{\ell_i,\ell_{i+1}} (x_i)  \sum_{ {\{\mathbf{x}':x'_i = x_i,~\mathbf{x}' \neq \mathbf{x}\}}}  
\Bigg\{    
\left[
\sum_{\substack{\ell_1',...,\ell_{i-1}',\ell_{i+2}',...,\ell_d' = 1  }}^{r_1,...,r_{i-1},r_{i+2},...,r_d}
\hat{\mathcal{Z}}^{\ell_1',\ell_2'}_{1,\mathbf x'}(t)
 \cdots \hat{\mathcal{Z}}^{\ell_{i-1}',\ell_i}_{i-1,\mathbf x'}(t) 
\hat{\mathcal{Z}}^{\ell_{i+1},\ell_{i+2}'}_{i+1,\mathbf x'}(t) \cdots
\hat{\mathcal{Z}}^{\ell_{d}',\ell_1'}_{d,\mathbf x'}(t)
\right] 
 \hat{s}_{\mathbf x'}(t) 
\nonumber \\
&
 -  \left[ 
\sum_{\substack{\ell_1',...,\ell_{i-1}',\ell_{i+2}',...,\ell_d' = 1  }}^{r_1,...,r_{i-1},r_{i+2},...,r_d}
\hat{\mathcal{Z}}^{\ell_1',\ell_2'}_{1, x_1'}(t)
 \cdots \hat{\mathcal{Z}}^{\ell_{i-1}',\ell_i}_{i-1, x_{i-1}'}(t) 
\hat{\mathcal{Z}}^{\ell_{i+1},\ell_{i+2}'}_{i+1,x_{i+1}'}(t) \cdots
\hat{\mathcal{Z}}^{\ell_{d}',\ell_1'}_{d, x_d'}(t)
\right]^2 
\hat{\mathcal{Z}}^{\ell_{i},\ell_{i+1}}_{i,x_i}(t)
 \left( - \nu_{\mathbf x'}^s(t) \right) 
 \Bigg\} \nonumber \\
 %
 \overset{(b)}{=}&~  \text{const}  + \log p\left(\mathcal{Z}_{i}^{\ell_i,\ell_{i+1}}(x_i)\right)  
 -  \frac{1}{2\nu^r_{\mathbf x,[i,\ell_i,\ell_{i+1},x_i]}} \left( { \mathcal{Z}_i^{\ell_i, \ell_{i+1}} (x_i) }   -   \hat{r}_{\mathbf x, [i,\ell_i,\ell_{i+1},x_i]} \right)^2  \nonumber \\
  =&~  \text{const}  + \log \left(  p\left(\mathcal{Z}_{i}^{\ell_i,\ell_{i+1}}(x_i)\right)  
  \mathcal{N}\left( { \mathcal{Z}_i^{\ell_i, \ell_{i+1}} (x_i) };  \hat{r}_{\mathbf x, [i,\ell_i,\ell_{i+1},x_i]}  ,  \nu^r_{\mathbf x,[i,\ell_i,\ell_{i+1},x_i]}  \right)   \right),
\end{align}
where $(a)$ follows from \eqref{sum-pro-Taylor-1}, $(b)$ follows from the following definitions and ${\hat{r}_{\mathbf x, [i,\ell_i,\ell_{i+1},x_i]}^2}/{2\nu^r_{\mathbf x,[i,\ell_i,\ell_{i+1},x_i]}}$ is a constant without $\mathcal{Z}_i^{\ell_i,\ell_{i+1}} (x_i)$:
\begin{align} 
\label{1-var-r-1-A}
 \frac{1}{\nu^r_{\mathbf x,[i,\ell_i,\ell_{i+1},x_i]}}   
\triangleq&~   \sum_{ {\{\mathbf{x}':x'_i = x_i,~\mathbf{x}' \neq \mathbf{x}\}}}  
\Bigg\{    
 \left[ 
\sum_{\substack{\ell_1',...,\ell_{i-1}',\ell_{i+2}',...,\ell_d' = 1  }}^{r_1,...,r_{i-1},r_{i+2},...,r_d}
\hat{\mathcal{Z}}^{\ell_1',\ell_2'}_{1, x_1'}(t)
 \cdots \hat{\mathcal{Z}}^{\ell_{i-1}',\ell_i}_{i-1, x_{i-1}'}(t) 
\hat{\mathcal{Z}}^{\ell_{i+1},\ell_{i+2}'}_{i+1,x_{i+1}'}(t) \cdots
\hat{\mathcal{Z}}^{\ell_{d}',\ell_1'}_{d, x_d'}(t)
\right]^2
  \nu_{\mathbf x'}^s(t)     \nonumber \\
&-  
  \left [  {\hat{s}_{\mathbf x'}(t)}^2  -  {\nu_{\mathbf x'}^s(t)} \right]  \times  \nonumber \\
 & \sum_{\substack{\ell_1',...,\ell_{i-1}',\ell_{i+2}',...,\ell_d' = 1  }}^{r_1,...,r_{i-1},r_{i+2},...,r_d}
\sum_{ \substack {A \subset \{1,...,d\} \setminus \{i\} \\ A \neq \emptyset} } 
\Bigg[
 \prod_{ \substack{i' \in A ,\ell_i'=\ell_i \\
\ell_{i+1}' = \ell_{i+1}
}}
 \nu_{i', x_{i'}'}^{\mathcal{Z},\ell_{i'}', \ell_{i'+1}'}(t) 
    \prod_{ \substack{
    i'' \in \{1,...,d\} \setminus (\{ i \} \cup A) \\
    \ell_i'=\ell_i,\ell_{i+1}' = \ell_{i+1}
    }} 
 {\hat{\mathcal{Z}}_{i'', x_{i''}'}^{\ell_{i''}', \ell_{i'' + 1}' } (t) }^2 
 \Bigg]
  \Bigg\}  ,
\end{align}
and
\begin{align} 
&~~~~~\hat{r}_{\mathbf x, [i,\ell_i,\ell_{i+1},x_i]}  \nonumber \\
&\triangleq \nu^r_{\mathbf x,[i,\ell_i,\ell_{i+1},x_i]}  
 \sum_{ {\{\mathbf{x}':x'_i = x_i,~\mathbf{x}' \neq \mathbf{x}\}}}  
\Bigg\{    
\left[
\sum_{\substack{\ell_1',...,\ell_{i-1}',\ell_{i+2}',...,\ell_d' = 1  }}^{r_1,...,r_{i-1},r_{i+2},...,r_d}
\hat{\mathcal{Z}}^{\ell_1',\ell_2'}_{1,\mathbf x'}(t)
 \cdots \hat{\mathcal{Z}}^{\ell_{i-1}',\ell_i}_{i-1,\mathbf x'}(t) 
\hat{\mathcal{Z}}^{\ell_{i+1},\ell_{i+2}'}_{i+1,\mathbf x'}(t) \cdots
\hat{\mathcal{Z}}^{\ell_{d}',\ell_1'}_{d,\mathbf x'}(t)
\right] 
 \hat{s}_{\mathbf x'}(t) 
\nonumber \\
&
~~~ -  \left[ 
\sum_{\substack{\ell_1',...,\ell_{i-1}',\ell_{i+2}',...,\ell_d' = 1  }}^{r_1,...,r_{i-1},r_{i+2},...,r_d}
\hat{\mathcal{Z}}^{\ell_1',\ell_2'}_{1, x_1'}(t)
 \cdots \hat{\mathcal{Z}}^{\ell_{i-1}',\ell_i}_{i-1, x_{i-1}'}(t) 
\hat{\mathcal{Z}}^{\ell_{i+1},\ell_{i+2}'}_{i+1,x_{i+1}'}(t) \cdots
\hat{\mathcal{Z}}^{\ell_{d}',\ell_1'}_{d, x_d'}(t)
\right]^2 
\hat{\mathcal{Z}}^{\ell_{i},\ell_{i+1}}_{i,x_i}(t)
 \left( - \nu_{\mathbf x'}^s(t) \right) 
 \Bigg\} \nonumber \\
  \label{hat-r-1-A}
 &= \nu^r_{\mathbf x,[i,\ell_i,\ell_{i+1},x_i]}  
 \sum_{ {\{\mathbf{x}':x'_i = x_i,~\mathbf{x}' \neq \mathbf{x}\}}}  
\Bigg\{  
\left[
\sum_{\substack{\ell_1',...,\ell_{i-1}',\ell_{i+2}',...,\ell_d' = 1  }}^{r_1,...,r_{i-1},r_{i+2},...,r_d}
\hat{\mathcal{Z}}^{\ell_1',\ell_2'}_{1,\mathbf x'}(t)
 \cdots \hat{\mathcal{Z}}^{\ell_{i-1}',\ell_i}_{i-1,\mathbf x'}(t) 
\hat{\mathcal{Z}}^{\ell_{i+1},\ell_{i+2}'}_{i+1,\mathbf x'}(t) \cdots
\hat{\mathcal{Z}}^{\ell_{d}',\ell_1'}_{d,\mathbf x'}(t)
\right]
 \hat{s}_{\mathbf x'}(t) 
  \Bigg\}  
  \nonumber \\
 &~~~+ \hat{\mathcal{Z}}^{\ell_{i},\ell_{i+1}}_{i,x_i}(t) 
 \Bigg\{ 1 + \nu^r_{\mathbf x,[i,\ell_i,\ell_{i+1},x_i]}   
 \sum_{ {\{\mathbf{x}':x'_i = x_i,~\mathbf{x}' \neq \mathbf{x}\}}}     
   \left [  {\hat{s}_{\mathbf x'}(t)}^2  -  {\nu_{\mathbf x'}^s(t)} \right]  \times
   \nonumber \\
 &~~~~\sum_{\substack{\ell_1',...,\ell_{i-1}',\ell_{i+2}',...,\ell_d' = 1  }}^{r_1,...,r_{i-1},r_{i+2},...,r_d}
\sum_{ \substack {A \subset \{1,...,d\} \setminus \{i\} \\ A \neq \emptyset} } 
\Bigg[
 \prod_{ \substack{i' \in A ,\ell_i'=\ell_i \\
\ell_{i+1}' = \ell_{i+1}
}}
 \nu_{i', x_{i'}'}^{\mathcal{Z},\ell_{i'}', \ell_{i'+1}'}(t) 
    \prod_{ \substack{
    i'' \in \{1,...,d\} \setminus (\{ i \} \cup A) \\
    \ell_i'=\ell_i,\ell_{i+1}' = \ell_{i+1}
    }} 
 {\hat{\mathcal{Z}}_{i'', x_{i''}'}^{\ell_{i''}', \ell_{i'' + 1}' } (t) }^2 
 \Bigg]   
  \Bigg\}.
\end{align}
Denote $\mathbf q^i \triangleq [i,\ell_i,\ell_{i+1},x_i]$, we have
\begin{align*} 
&p_{\mathbf x \leftarrow [i,\ell_i, \ell_{i+1},x_i] } \left(t+1,\mathcal{Z}_{i}(\ell_i,x_i,\ell_{i+1}) \right) \ {\approx}\ \text{const}\ + 
\log\Big( p\left(\mathcal{Z}_{i}^{\ell}(x_i)\right) \mathcal{N}\left({\mathcal{Z}_i^{\ell} (x_i) };\hat{r}_{\mathbf{x} , \mathbf q^i},  \nu^r_{\mathbf{x} , \mathbf q^i} \right) \Big),
\end{align*}
where ${{\mathcal{Z}}_{i}^{\ell}} \triangleq {{\mathcal{Z}}_{i}^{\ell_i,\ell_{i+1}}}$ and
\begin{align*}
\frac{1}{\nu^r_{\mathbf{x} ,\mathbf q^i}}  & \triangleq   \sum_{\substack{{ {\mathbf{x}':x'_i = x_i}},\\ \mathbf{x} ' \neq \mathbf{x} }}
\Bigg\{    
 \Big[ 
 {\sum_{\mathbf{l}'_{\setminus i} = 1}^{\mathbf{r}_{\setminus i}}}
 {\hat{\mathcal{Z}}^{\ell'}_{\setminus i, x'}(t)}
\Big]^2
  \nu_{\mathbf{x} '}^s(t) -  
  {\omega}_{\mathbf{x} '}(t) {\zeta}_{x}(t) \Bigg\}, \\
\hat{r}_{\mathbf{x} , \mathbf q^i} &\triangleq \nu^r_{\mathbf{x} ,\mathbf q^i}  
 \sum_{\substack{{ {\mathbf{x}':x'_i = x_i}},\\ \mathbf{x} ' \neq \mathbf{x} }}
\Bigg\{  
\Big[
 {\sum_{\mathbf{l}'_{\setminus i} = 1}^{\mathbf{r}_{\setminus i}}}
 {\hat{\mathcal{Z}}^{\ell'}_{\setminus i, \mathbf{x}'}(t)}
\Big]
 \hat{s}_{\mathbf{x} '}(t) 
  \Bigg\} 
 + \hat{\mathcal{Z}}^{\ell}_{i,x_i}(t)
 \Bigg\{ 1 + \nu^r_{\mathbf{x} ,\mathbf q^i}
 \sum_{\substack {{ {\mathbf{x}':x'_i = x_i}},\\ \mathbf{x} ' \neq \mathbf{x} }}
   {\omega}_{\mathbf{x} '}(t) {\zeta}_{x}(t) \Bigg\}.
\end{align*}
Then we complete the derivation.

\section{Derivation of Equation \eqref{z-ix-mean-3}}\label{appendixD}

\textbf{Element-wise Result:}~
{\it
    $\hat{\mathcal{Z}}_{i,\mathbf{x} }^{\ell}(t+1)$ can be approximated by
    \begin{align*}
\hat{\mathcal{Z}}_{i,\mathbf{x} }^{\ell}(t+1)   =
 g 
 \left(
 \hat{r}_{\mathbf{x} , \mathbf q^i}  , 
   \nu^r_{\mathbf{x} ,\mathbf q^i}
   \right) 
\approx 
\hat{\mathcal{Z}}_{i,x_i}^{\ell}(t+1)  
-    \hat{s}_{\mathbf{x} }(t)  \nu_{i,x_i}^{{ {\mathcal{Z}}},\ell}(t+1)
\times 
  {\sum_{\mathbf{l}'_{\setminus i} = 1}^{\mathbf{r}_{\setminus i}}}
 {\hat{\mathcal{Z}}^{\ell'}_{\setminus i, x}(t)},
\end{align*}
where $\hat{\mathcal{Z}}_{i,x_i}^{\ell} \triangleq  g
(\hat{r}_{\mathbf q^i} ,  \nu^r_{\mathbf q^i})$ and $\nu_{i,x_i}^{{ {\mathcal{Z}}},\ell}
 \triangleq \nu^r_{\mathbf q^i} 
  g'(\hat{r}_{\mathbf q^i}, \nu^r_{\mathbf q^i})$ with
\begin{align*}
\frac{1}{\nu^r_{\mathbf q^i}}  
&\triangleq  \sum_{{ {\mathbf{x}':x'_i = x_i}}}  
\Bigg\{    
 \Big[ 
 {\sum_{\mathbf{l}'_{\setminus i} = 1}^{\mathbf{r}_{\setminus i}}}
 {\hat{\mathcal{Z}}^{\ell'}_{\setminus i, x'}(t)}
\Big]^2
\nu_{\mathbf{x} '}^s(t) -  
  {\omega}_{\mathbf{x} '}(t) {\zeta}_{x}(t)
  \Bigg\}  , \\
\hat{r}_{\mathbf q^i} &\triangleq  \nu^r_{\mathbf q^i}  
 \sum_{{ {\mathbf{x}':x'_i = x_i}}}  
\Bigg\{  
\Big[
 {\sum_{\mathbf{l}'_{\setminus i} = 1}^{\mathbf{r}_{\setminus i}}}
 {\hat{\mathcal{Z}}^{\ell'}_{\setminus i, \mathbf{x}'}(t)}
\Big]
 \hat{s}_{\mathbf{x} '}(t) 
  \Bigg\}  
 + \hat{\mathcal{Z}}^{\ell}_{i,x_i}(t) 
 \Bigg\{ 1 + \nu^r_{\mathbf q^i}   
 \sum_{{ {\mathbf{x}':x'_i = x_i}}}     
   {\omega}_{\mathbf{x} '}(t) {\zeta}_{x}(t)
  \Bigg\}.
\end{align*}
}


\emph{Derivation:}
Recall that the mean and variance of the pdf associated with the $ p_{\mathbf x \leftarrow [i,\ell_i,\ell_{i+1},x_i] }(t+1,\cdot)$ approximation in \eqref{var-to-fac-1} can be expanded by
\begin{align} 
\label{z-ix-mean-1-A}
\hat{\mathcal{Z}}_{i,\mathbf x}^{\ell_i,\ell_{i+1}}(t+1) \triangleq 
\underbrace{\frac{1}{C} 
\int_{x} x ~ p ( x )
  \mathcal{N}\left( x;  \hat{r}_{\mathbf x, [i,\ell_i,\ell_{i+1},x_i]}  ,  \nu^r_{\mathbf x,[i,\ell_i,\ell_{i+1},x_i]}   \right) 
  }_{ 
  \triangleq ~ g(\hat{r}_{\mathbf x, [i,\ell_i,\ell_{i+1},x_i]}  , 
   \nu^r_{\mathbf x,[i,\ell_i,\ell_{i+1},x_i]}) 
  },
\end{align}
and
\begin{align} 
\label{var-z-ix-1-A}
\nu_{i,\mathbf x}^{\mathcal{Z},\ell_{i}, \ell_{i+1}}(t+1) \triangleq
\underbrace{
\frac{1}{C} \int_{x} | x - \hat{\mathcal{Z}}_{i,\mathbf x}^{\ell_i,\ell_{i+1}}(t+1) |^2 ~
 p ( x )
  \mathcal{N}
  \left( x;  
  \hat{r}_{\mathbf x, [i,\ell_i,\ell_{i+1},x_i]}  , 
   \nu^r_{\mathbf x,[i,\ell_i,\ell_{i+1},x_i]}  
    \right)
   }_{ 
  \triangleq ~  \nu^r_{\mathbf x,[i,\ell_i,\ell_{i+1},x_i]}  
  g'
  (\hat{r}_{\mathbf x, [i,\ell_i,\ell_{i+1},x_i]}  ,
    \nu^r_{\mathbf x,[i,\ell_i,\ell_{i+1},x_i]}) 
  },
\end{align}
where here $C = \int_{x}  p ( x )
  \mathcal{N}\left( x;  \hat{r}_{\mathbf x, [i,\ell_i,\ell_{i+1},x_i]}  ,  \nu^r_{\mathbf x,[i,\ell_i,\ell_{i+1},x_i]}   \right)$ and $g'$ denotes the derivative of $g$ with respect to the first argument. The fact that \eqref{z-ix-mean-1-A} and  \eqref{var-z-ix-1-A} are related through a derivative was shown in~\cite{rangan2011generalized}.

We now derive approximations of $\hat{\mathcal{Z}}^{\ell_i,\ell_{i+1}}_{i,\mathbf x}$ and $\nu_{i,\mathbf x}^{\mathcal{Z},\ell_{i}, \ell_{i+1}}$ that avoid the dependence on the destination node $u_{\mathbf x}$. For this, we introduce $\mathbf x$-invariant versions of $ \nu^r_{\mathbf x,[i,\ell_i,\ell_{i+1},x_i]}$ and $\hat{r}_{\mathbf x, [i,\ell_i,\ell_{i+1},x_i]}$ as follows
\begin{align} 
\label{1-var-r-2-A}
\frac{1}{\nu^r_{[i,\ell_i,\ell_{i+1},x_i]}}    
\triangleq&~   \sum_{ {\mathbf{x}':x'_i = x_i}}  
\Bigg\{    
 \left[ 
\sum_{\substack{\ell_1',...,\ell_{i-1}',\ell_{i+2}',...,\ell_d' = 1  }}^{r_1,...,r_{i-1},r_{i+2},...,r_d}
\hat{\mathcal{Z}}^{\ell_1',\ell_2'}_{1, x_1'}(t)
 \cdots \hat{\mathcal{Z}}^{\ell_{i-1}',\ell_i}_{i-1, x_{i-1}'}(t) 
\hat{\mathcal{Z}}^{\ell_{i+1},\ell_{i+2}'}_{i+1,x_{i+1}'}(t) \cdots
\hat{\mathcal{Z}}^{\ell_{d}',\ell_1'}_{d, x_d'}(t)
\right]^2
\nu_{\mathbf x'}^s(t)     \nonumber \\
&-  
  \left [  {\hat{s}_{\mathbf x'}(t)}^2  -  {\nu_{\mathbf x'}^s(t)} \right]  \times  \nonumber \\
 & \sum_{\substack{\ell_1',...,\ell_{i-1}',\ell_{i+2}',...,\ell_d' = 1  }}^{r_1,...,r_{i-1},r_{i+2},...,r_d}
\sum_{ \substack {A \subset \{1,...,d\} \setminus \{i\} \\ A \neq \emptyset} } 
\Bigg[
 \prod_{ \substack{i' \in A ,\ell_i'=\ell_i \\
\ell_{i+1}' = \ell_{i+1}
}}
 \nu_{i', x_{i'}'}^{\mathcal{Z},\ell_{i'}', \ell_{i'+1}'}(t) 
    \prod_{ \substack{
    i'' \in \{1,...,d\} \setminus (\{ i \} \cup A) \\
    \ell_i'=\ell_i,\ell_{i+1}' = \ell_{i+1}
    }} 
 {\hat{\mathcal{Z}}_{i'', x_{i''}'}^{\ell_{i''}', \ell_{i'' + 1}' } (t) }^2 
 \Bigg]
  \Bigg\}  ,
\end{align}
and
\begin{align} 
\label{hat-r-2-A}
\hat{r}_{[i,\ell_i,\ell_{i+1},x_i]}
&\triangleq  \nu^r_{[i,\ell_i,\ell_{i+1},x_i]}  
 \sum_{ {\mathbf{x}':x'_i = x_i}}  
\Bigg\{  
\left[
\sum_{\substack{\ell_1',...,\ell_{i-1}',\ell_{i+2}',...,\ell_d' = 1  }}^{r_1,...,r_{i-1},r_{i+2},...,r_d}
\hat{\mathcal{Z}}^{\ell_1',\ell_2'}_{1,\mathbf x'}(t)
 \cdots \hat{\mathcal{Z}}^{\ell_{i-1}',\ell_i}_{i-1,\mathbf x'}(t) 
\hat{\mathcal{Z}}^{\ell_{i+1},\ell_{i+2}'}_{i+1,\mathbf x'}(t) \cdots
\hat{\mathcal{Z}}^{\ell_{d}',\ell_1'}_{d,\mathbf x'}(t)
\right]
 \hat{s}_{\mathbf x'}(t) 
  \Bigg\}  
  \nonumber \\
 &~~~+ \hat{\mathcal{Z}}^{\ell_{i},\ell_{i+1}}_{i,x_i}(t) 
 \Bigg\{ 1 + \nu^r_{[i,\ell_i,\ell_{i+1},x_i]}   
 \sum_{ {\mathbf{x}':x'_i = x_i}}     
   \left [  {\hat{s}_{\mathbf x'}(t)}^2  -  {\nu_{\mathbf x'}^s(t)} \right]  \times
   \nonumber \\
 &~~~\sum_{\substack{\ell_1',...,\ell_{i-1}',\ell_{i+2}',...,\ell_d' = 1  }}^{r_1,...,r_{i-1},r_{i+2},...,r_d}
\sum_{ \substack {A \subset \{1,...,d\} \setminus \{i\} \\ A \neq \emptyset} } 
\Bigg[
 \prod_{ \substack{i' \in A ,\ell_i'=\ell_i \\
\ell_{i+1}' = \ell_{i+1}
}}
 \nu_{i', x_{i'}'}^{\mathcal{Z},\ell_{i'}', \ell_{i'+1}'}(t) 
    \prod_{ \substack{
    i'' \in \{1,...,d\} \setminus (\{ i \} \cup A) \\
    \ell_i'=\ell_i,\ell_{i+1}' = \ell_{i+1}
    }} 
 {\hat{\mathcal{Z}}_{i'', x_{i''}'}^{\ell_{i''}', \ell_{i'' + 1}' } (t) }^2 
 \Bigg]   
  \Bigg\}.
\end{align}

\begin{remark}
\label{remark1}
Comparing \eqref{1-var-r-2-A}-\eqref{hat-r-2-A} with \eqref{1-var-r-1-A}-\eqref{hat-r-1-A} suggest that $\nu^r_{\mathbf x, [i,\ell_i,\ell_{i+1},x_i]} - \nu^r_{[i,\ell_i,\ell_{i+1},x_i]}$ is relatively negligible and hence each $ \hat{r}_{\mathbf x, [i,\ell_i,\ell_{i+1},x_i]}$ can be approximated by the term 
\begin{align} 
\label{term-ne-1-A}
\hat{r}_{[i,\ell_i,\ell_{i+1},x_i]} - \nu^r_{[i,\ell_i,\ell_{i+1},x_i]} \times 
\left[
 \sum_{\substack{\ell_1',...,\ell_{i-1}',\ell_{i+2}',...,\ell_d' = 1  }}^{r_1,...,r_{i-1},r_{i+2},...,r_d}
\hat{\mathcal{Z}}^{\ell_1',\ell_2'}_{1,x_1}(t)
 \cdots \hat{\mathcal{Z}}^{\ell_{i-1}',\ell_i}_{i-1,x_{i-1}}(t) 
\hat{\mathcal{Z}}^{\ell_{i+1},\ell_{i+2}'}_{i+1,x_{i+1}}(t) \cdots
\hat{\mathcal{Z}}^{\ell_{d}',\ell_1'}_{d,x_d}(t)
\right ]
\times \hat{s}_{\mathbf x}(t),
\end{align}
\end{remark}

Hence, \eqref{z-ix-mean-1-A} implies the following approximation, we have
\begin{align} 
&\hat{\mathcal{Z}}_{i,\mathbf x}^{\ell_i,\ell_{i+1}}(t+1)   \nonumber \\
= &~
 g 
 \left(
 \hat{r}_{\mathbf x, [i,\ell_i,\ell_{i+1},x_i]}  , 
   \nu^r_{\mathbf x,[i,\ell_i,\ell_{i+1},x_i]}
   \right) \nonumber \\
\label{z-ix-mean-2-A}
   \overset{(a)}{\approx} &~
   g
    \Bigg(
 \hat{r}_{[i,\ell_i,\ell_{i+1},x_i]} 
 - \nu^r_{[i,\ell_i,\ell_{i+1},x_i]} 
 \nonumber \\
&
\times
  \Big[ 
   \sum_{\substack{\ell_1',...,\ell_{i-1}',\ell_{i+2}',...,\ell_d' = 1  }}^{r_1,...,r_{i-1},r_{i+2},...,r_d}
\hat{\mathcal{Z}}^{\ell_1',\ell_2'}_{1,x_1}(t) 
 \cdots \hat{\mathcal{Z}}^{\ell_{i-1}',\ell_i}_{i-1,x_{i-1}}(t) 
\hat{\mathcal{Z}}^{\ell_{i+1},\ell_{i+2}'}_{i+1,x_{i+1}}(t) \cdots
\hat{\mathcal{Z}}^{\ell_{d}',\ell_1'}_{d,x_d}(t)
\Big]
  \hat{s}_{\mathbf x}(t) , 
   \nu^r_{[i,\ell_i,\ell_{i+1},x_i]}
   \Bigg)  \\
   \label{z-ix-mean-3-A}
 \overset{(b)}{\approx} &~
 g
    \left(
 \hat{r}_{[i,\ell_i,\ell_{i+1},x_i]} ,  \nu^r_{[i,\ell_i,\ell_{i+1},x_i]} \right)
 - g'
    \left(
 \hat{r}_{[i,\ell_i,\ell_{i+1},x_i]} ,  \nu^r_{[i,\ell_i,\ell_{i+1},x_i]} \right)  \nonumber \\
 &  \times
 \nu^r_{[i,\ell_i,\ell_{i+1},x_i]} 
  \Big[ 
 \sum_{\substack{\ell_1',...,\ell_{i-1}',\ell_{i+2}',...,\ell_d' = 1  }}^{r_1,...,r_{i-1},r_{i+2},...,r_d}
\hat{\mathcal{Z}}^{\ell_1',\ell_2'}_{1,x_1}(t)
 \cdots \hat{\mathcal{Z}}^{\ell_{i-1}',\ell_i}_{i-1,x_{i-1}}(t) 
\hat{\mathcal{Z}}^{\ell_{i+1},\ell_{i+2}'}_{i+1,x_{i+1}}(t) \cdots
\hat{\mathcal{Z}}^{\ell_{d}',\ell_1'}_{d,x_d}(t)
\Big]
  \hat{s}_{\mathbf x}(t)   \nonumber \\
\overset{(c)}{=}  &~ 
\hat{\mathcal{Z}}_{i,x_i}^{\ell_i,\ell_{i+1}}(t+1)  
-    \hat{s}_{\mathbf x}(t)  \nu_{i,x_i}^{\mathcal{Z},\ell_{i}, \ell_{i+1}}(t+1)  \nonumber \\
&
\times \Big[ 
 \sum_{\substack{\ell_1',...,\ell_{i-1}',\ell_{i+2}',...,\ell_d' = 1  }}^{r_1,...,r_{i-1},r_{i+2},...,r_d}
\hat{\mathcal{Z}}^{\ell_1',\ell_2'}_{1,x_1}(t)
 \cdots \hat{\mathcal{Z}}^{\ell_{i-1}',\ell_i}_{i-1,x_{i-1}}(t) 
\hat{\mathcal{Z}}^{\ell_{i+1},\ell_{i+2}'}_{i+1,x_{i+1}}(t) \cdots
\hat{\mathcal{Z}}^{\ell_{d}',\ell_1'}_{d,x_d}(t)
\Big].
\end{align}
Above, $(a)$ follows from Remark~\ref{remark1}; $(b)$ follows from  a Taylor series expansion in the first argument of \eqref{z-ix-mean-2-A} about the point $\hat{r}_{[i,\ell_i,\ell_{i+1},x_i]} $; and $(c)$ follows by applying the definitions
\begin{align} 
\label{z-x-i-mean-3-A}
\hat{\mathcal{Z}}_{i,x_i}^{\ell_i,\ell_{i+1}}(t+1) \triangleq  g
    \left(
 \hat{r}_{[i,\ell_i,\ell_{i+1},x_i]} ,  \nu^r_{[i,\ell_i,\ell_{i+1},x_i]} \right),
\end{align}
and 
\begin{align} 
\label{var-z-x-i-3-A}
 \nu_{i,x_i}^{\mathcal{Z},\ell_{i}, \ell_{i+1}}(t+1) \triangleq  \nu^r_{[i,\ell_i,\ell_{i+1},x_i]} 
  g'
    \left(
 \hat{r}_{[i,\ell_i,\ell_{i+1},x_i]} ,  \nu^r_{[i,\ell_i,\ell_{i+1},x_i]} \right),
\end{align}
which match \eqref{z-ix-mean-1-A}-\eqref{var-z-ix-1-A} without the $\mathbf x$-dependence.

\begin{remark}
\label{remark2}
Likewise, taking Taylor series expansions of $ g'$ in \eqref{var-z-ix-1-A} about the point $\hat{r}_{[i,\ell_i,\ell_{i+1},x_i]} $ in the first argument and about the point $\nu^{r}_{[i,\ell_i,\ell_{i+1},x_i]} $ in the second argument and then comparing the result with \eqref{var-z-x-i-3-A} confirms that $\nu_{i,\mathbf x}^{\mathcal{Z},\ell_{i}, \ell_{i+1}}(t) -  \nu_{i,x_i}^{\mathcal{Z},\ell_{i}, \ell_{i+1}}(t)$ is negligible.
\end{remark}

As a result, we have
\begin{align*}
\hat{\mathcal{Z}}_{i,\mathbf{x} }^{\ell}(t+1)   =
 g 
 \left(
 \hat{r}_{\mathbf{x} , \mathbf q^i}  , 
   \nu^r_{\mathbf{x} ,\mathbf q^i}
   \right) 
\approx 
\hat{\mathcal{Z}}_{i,x_i}^{\ell}(t+1)  
-    \hat{s}_{\mathbf{x} }(t)  \nu_{i,x_i}^{{ {\mathcal{Z}}},\ell}(t+1)
\times 
  {\sum_{\mathbf{l}'_{\setminus i} = 1}^{\mathbf{r}_{\setminus i}}}
 {\hat{\mathcal{Z}}^{\ell'}_{\setminus i, x}(t)},
\end{align*}
where $\hat{\mathcal{Z}}_{i,x_i}^{\ell} \triangleq  g
(\hat{r}_{\mathbf q^i} ,  \nu^r_{\mathbf q^i})$ and $\nu_{i,x_i}^{{ {\mathcal{Z}}},\ell}
 \triangleq \nu^r_{\mathbf q^i} 
  g'(\hat{r}_{\mathbf q^i}, \nu^r_{\mathbf q^i})$ with
\begin{align*}
\frac{1}{\nu^r_{\mathbf q^i}}  
&\triangleq  \sum_{{ {\mathbf{x}':x'_i = x_i}}}  
\Bigg\{    
 \Big[ 
 {\sum_{\mathbf{l}'_{\setminus i} = 1}^{\mathbf{r}_{\setminus i}}}
 {\hat{\mathcal{Z}}^{\ell'}_{\setminus i, x'}(t)}
\Big]^2
\nu_{\mathbf{x} '}^s(t) -  
  {\omega}_{\mathbf{x} '}(t) {\zeta}_{x}(t)
  \Bigg\}  , \\
\hat{r}_{\mathbf q^i} &\triangleq  \nu^r_{\mathbf q^i}  
 \sum_{{ {\mathbf{x}':x'_i = x_i}}}  
\Bigg\{  
\Big[
 {\sum_{\mathbf{l}'_{\setminus i} = 1}^{\mathbf{r}_{\setminus i}}}
 {\hat{\mathcal{Z}}^{\ell'}_{\setminus i, \mathbf{x}'}(t)}
\Big]
 \hat{s}_{\mathbf{x} '}(t) 
  \Bigg\}  
 + \hat{\mathcal{Z}}^{\ell}_{i,x_i}(t) 
 \Bigg\{ 1 + \nu^r_{\mathbf q^i}   
 \sum_{{ {\mathbf{x}':x'_i = x_i}}}     
   {\omega}_{\mathbf{x} '}(t) {\zeta}_{x}(t)
  \Bigg\}.
\end{align*}
Then we complete the derivation.
\section{Derivation of Equations \eqref{p-x-2}, \eqref{v-p-2} and \eqref{var-r-approx-1} }\label{appendixE}
\subsection{Derivation of Equation \eqref{p-x-2}}

\textbf{Element-wise Result:}~
{\it
$\hat{p}_{\mathbf{x} }(t)$ in \eqref{p-x-1} can be approximated by
\begin{align*} 
\hat{p}_{\mathbf{x} }(t) =&~ \sum_{\mathbf{l}' = 1}^{\mathbf{r}}  
{\hat{\mathcal{Z}}_{1, \mathbf{x}}^{\ell'}(t) }  
\cdots {\hat{\mathcal{Z}}_{d, \mathbf{x}}^{\ell'}(t) } 
\approx \bar p_{\mathbf{x} }(t) 
- \hat{s}_{\mathbf{x} }(t-1)  \bar\nu^p_{\mathbf{x} }(t), 
\end{align*}
where $\bar p_{\mathbf x}(t) \triangleq \sum_{\mathbf{l}' = 1}^{\mathbf{r}}
{\hat{\mathcal{Z}}_{1, x}^{\ell'}(t) }  
\cdots {\hat{\mathcal{Z}}_{d, x}^{\ell'}(t) }$ and
\begin{align*} 
\bar\nu^p_{\mathbf{x} }(t) =&~  \sum_{\mathbf{l}' = 1}^{\mathbf{r}}  
 \sum_{\substack{A \subsetneqq \mathbb{D} \\ A\neq \emptyset} }
 \Bigg\{ 
 \Big(
 \prod_{i \in A}  \hat{\mathcal{Z}}_{i, x_{i}}^{\ell'}(t) 
 \Big)^2
 \times \prod_{i'' \in \mathbb{D} \setminus A}
 \nu_{i'',x_{i''}}^{\mathcal{Z},\ell'}(t) 
 \Big(
 - \hat{s}_{\mathbf{x} }(t-1) 
  \prod_{i' \in \mathbb{D}} 
 \hat{\mathcal{Z}}^{\ell'}_{i',x_{i'}}(t)
 \Big)^{d-|A|-1} 
 \Bigg\}. 
\end{align*}
}

\emph{Derivation:}
Plug \eqref{z-ix-mean-3} into \eqref{p-x-1} to obtain
\begin{align} 
\label{p-x-2-A}
\hat{p}_{\mathbf x}(t) 
& \overset{(a)}{=} \sum_{\ell_1',...,\ell_d' = 1}^{r_1,...,r_d}  
{  \hat{\mathcal{Z}}_{1, \mathbf{x}}^{\ell_1',\ell_{2}'}(t)  }  
\cdots  {  \hat{\mathcal{Z}}_{d, \mathbf{x}}^{\ell_d',\ell_{1}'}(t)  } 
 \nonumber \\
& \overset{(b)}{\approx}
 \underbrace{
\sum_{\ell_1',...,\ell_d' = 1}^{r_1,...,r_d}  
{  \hat{\mathcal{Z}}_{1, x_1}^{\ell_1',\ell_{2}'}(t)  }  
\cdots  {  \hat{\mathcal{Z}}_{d, x_d}^{\ell_d',\ell_{1}'}(t)  }
}_{
\bar p_{\mathbf x}(t)}
+
 \sum_{\ell_1',...,\ell_d' = 1}^{r_1,...,r_d}  
 \sum_{\substack{ {A\subsetneqq \{1,...,d\}} } }
 \Bigg\{ 
 \prod_{i' \in A}  \hat{\mathcal{Z}}_{i', x_{i'}}^{\ell_{i'}',\ell_{i'+1}'}(t) 
  \prod_{i'' \in \{1,...,d\}\setminus A}
   \hat{s}_{\mathbf x}(t-1)  \nu_{i'',x_{i''}}^{\mathcal{Z},\ell_{i''}', \ell_{i''+1}'}(t)
   \nonumber \\
 & ~~~ \Big[- 
  \sum_{\ell_1'',...,\ell_{i''-1}'',\ell_{i''+2}'',...,\ell_d'' = 1 }^{r_1,...,r_{i-1},r_{i+2},...,r_d}
 \hat{\mathcal{Z}}^{\ell_1'',\ell_2''}_{1,x_1}(t-1)
 \cdots 
 \hat{\mathcal{Z}}^{\ell_{i''-1}'',\ell_{i''}'}_{i''-1,x_{i''-1}}(t-1) 
\hat{\mathcal{Z}}^{\ell_{i''+1}',\ell_{i''+2}''}_{i''+1,x_{i''+1}}(t-1) 
\cdots
\hat{\mathcal{Z}}^{\ell_{d}'',\ell_1''}_{d,x_{d}}(t-1) 
\Big]
\Bigg\}  \nonumber \\
& \overset{(c)}{\approx}  \bar p_{\mathbf x}(t) 
+
 \sum_{\ell_1',...,\ell_d' = 1}^{r_1,...,r_d}  
 \sum_{\substack{ {A\subsetneqq \{1,...,d\}} \\ A\neq \emptyset} }
 \Bigg\{ 
 \prod_{i' \in A}  \hat{\mathcal{Z}}_{i', x_{i'}}^{\ell_{i'}',\ell_{i'+1}'}(t) 
  \prod_{i'' \in \{1,...,d\}\setminus A} 
  \hat{s}_{\mathbf x}(t-1)  \nu_{i'',x_{i''}}^{\mathcal{Z},\ell_{i''}', \ell_{i''+1}'}(t)
  \nonumber \\
 & ~~~ \Big[- 
  \sum_{\ell_1'',...,\ell_{i''-1}'',\ell_{i''+2}'',...,\ell_d'' = 1 }^{r_1,...,r_{i-1},r_{i+2},...,r_d}
 \hat{\mathcal{Z}}^{\ell_1'',\ell_2''}_{1,x_1}(t-1)
 \cdots 
 \hat{\mathcal{Z}}^{\ell_{i''-1}'',\ell_{i''}'}_{i''-1,x_{i''-1}}(t-1) 
\hat{\mathcal{Z}}^{\ell_{i''+1}',\ell_{i''+2}''}_{i''+1,x_{i''+1}}(t-1) 
\cdots
\hat{\mathcal{Z}}^{\ell_{d}'',\ell_1''}_{d,x_{d}}(t-1) 
\Big]
\Bigg\}  \nonumber \\
& \overset{(d)}{\approx}  \bar p_{\mathbf x}(t) 
+
 \sum_{\ell_1',...,\ell_d' = 1}^{r_1,...,r_d}  
 \sum_{\substack{ {A\subsetneqq \{1,...,d\}} \\ A\neq \emptyset} }
 \Bigg\{ 
 \prod_{i' \in A}  \hat{\mathcal{Z}}_{i', x_{i'}}^{\ell_{i'}',\ell_{i'+1}'}(t) 
  \prod_{i'' \in \{1,...,d\}\setminus A} 
  \hat{s}_{\mathbf x}(t-1)  \nu_{i'',x_{i''}}^{\mathcal{Z},\ell_{i''}', \ell_{i''+1}'}(t)
  \nonumber \\
 & ~~~ \Big[- 
  \sum_{\ell_1'',...,\ell_{i''-1}'',\ell_{i''+2}'',...,\ell_d'' = 1 }^{r_1,...,r_{i-1},r_{i+2},...,r_d}
 \hat{\mathcal{Z}}^{\ell_1'',\ell_2''}_{1,x_1}(t)
 \cdots 
 \hat{\mathcal{Z}}^{\ell_{i''-1}'',\ell_{i''}'}_{i''-1,x_{i''-1}}(t) 
\hat{\mathcal{Z}}^{\ell_{i''+1}',\ell_{i''+2}''}_{i''+1,x_{i''+1}}(t) 
\cdots
\hat{\mathcal{Z}}^{\ell_{d}'',\ell_1''}_{d,x_{d}}(t) 
\Big]
\Bigg\}  \nonumber \\
& \overset{(e)}{\approx}  \bar p_{\mathbf x}(t) 
-
\hat{s}_{\mathbf x}(t-1)  \bar\nu^p_{\mathbf x}(t)
, 
\end{align}
where $(a)$ follows from \eqref{p-x-1}; $(b)$ follows from \eqref{z-ix-mean-3}; $(c)$ follows noting that we assumed data size is large and equations on two side of this differ by only one term (which vanishes relative to the others in the large-system limit); for $(d)$ we used $ \hat{\mathcal{Z}}^{\ell_{i''}'',\ell_{i''+1}'}_{i'',x_{i''}}(t) $ in place of $ \hat{\mathcal{Z}}^{\ell_{i''}'',\ell_{i''+1}'}_{i'',x_{i''}}(t-1) $; and for $(e)$ we follows Remark~\ref{remark3} and the following definition: 
\begin{align} 
\label{bar-var-p-1-A}
\bar\nu^p_{\mathbf x}(t) =&~  \sum_{\ell_1',...,\ell_d' = 1}^{r_1,...,r_d}  
 \sum_{\substack{ {A\subsetneqq \{1,...,d\}} \\ A\neq \emptyset} }
 \Bigg\{ 
 \Big(
 \prod_{i \in A}  \hat{\mathcal{Z}}_{i, x_{i}}^{\ell_{i}',\ell_{i+1}'}(t) 
 \Big)^2
 \times \nonumber \\
 &~~~~~~~~~~~~~~~
 \Big(
 - \hat{s}_{\mathbf x}(t-1) 
  \prod_{i' \in \{1,...,d\}} 
 \hat{\mathcal{Z}}^{\ell_{i'}',\ell_{i'+1}'}_{i',x_{i'}}(t)
 \Big)^{d-|A|-1}
  \prod_{i'' \in \{1,...,d\}\setminus A}
 \nu_{i'',x_{i''}}^{\mathcal{Z},\ell_{i''}', \ell_{i''+1}'}(t)
 \Bigg\}. 
\end{align}

\begin{remark}
\label{remark3}
We observed that the following term is small
\begin{align} 
\label{remark3-term-1}
&- \sum_{\ell_1',...,\ell_d' = 1}^{r_1,...,r_d}  
 \sum_{\substack{ {A\subsetneqq \{1,...,d\}} \\ A\neq \emptyset} }
 \Bigg\{ 
 \prod_{i' \in A}  \hat{\mathcal{Z}}_{i', x_{i'}}^{\ell_{i'}',\ell_{i'+1}'}(t) 
  \prod_{i'' \in \{1,...,d\}\setminus A} 
  \hat{s}_{\mathbf x}(t-1)  \nu_{i'',x_{i''}}^{\mathcal{Z},\ell_{i''}', \ell_{i''+1}'}(t)
  \nonumber \\
 & ~~~ \Big[- 
  \sum_{\ell_1'',...,\ell_{i''-1}'',\ell_{i''+2}'',...,\ell_d'' = 1 }^{r_1,...,r_{i-1},r_{i+2},...,r_d}
 \hat{\mathcal{Z}}^{\ell_1'',\ell_2''}_{1,x_1}(t)
 \cdots 
 \hat{\mathcal{Z}}^{\ell_{i''-1}'',\ell_{i''}'}_{i''-1,x_{i''-1}}(t) 
\hat{\mathcal{Z}}^{\ell_{i''+1}',\ell_{i''+2}''}_{i''+1,x_{i''+1}}(t) 
\cdots
\hat{\mathcal{Z}}^{\ell_{d}'',\ell_1''}_{d,x_{d}}(t) 
\Big]
\Bigg\} \nonumber \\
&+ \sum_{\ell_1',...,\ell_d' = 1}^{r_1,...,r_d}  
 \sum_{\substack{ {A\subsetneqq \{1,...,d\}} \\ A\neq \emptyset} }
 \Bigg\{ 
 \prod_{i' \in A}  \hat{\mathcal{Z}}_{i', x_{i'}}^{\ell_{i'}',\ell_{i'+1}'}(t) 
  \prod_{i'' \in \{1,...,d\}\setminus A} 
  \hat{s}_{\mathbf x}(t-1)  \nu_{i'',x_{i''}}^{\mathcal{Z},\ell_{i''}', \ell_{i''+1}'}(t)
  \nonumber \\
 & ~~~ \Big[-
 \hat{\mathcal{Z}}^{\ell_1',\ell_2'}_{1,x_1}(t)
 \cdots 
 \hat{\mathcal{Z}}^{\ell_{i''-1}',\ell_{i''}'}_{i''-1,x_{i''-1}}(t) 
\hat{\mathcal{Z}}^{\ell_{i''+1}',\ell_{i''+2}'}_{i''+1,x_{i''+1}}(t) 
\cdots
\hat{\mathcal{Z}}^{\ell_{d}',\ell_1'}_{d,x_{d}}(t) 
\Big]
\Bigg\}
\nonumber \\
=&~
- \sum_{\ell_1',...,\ell_d' = 1}^{r_1,...,r_d}  
 \sum_{\substack{ {A\subsetneqq \{1,...,d\}} \\ A\neq \emptyset} }
 \Bigg\{ 
 \prod_{i' \in A}  \hat{\mathcal{Z}}_{i', x_{i'}}^{\ell_{i'}',\ell_{i'+1}'}(t) 
  \prod_{i'' \in \{1,...,d\}\setminus A} 
  \hat{s}_{\mathbf x}(t-1)  \nu_{i'',x_{i''}}^{\mathcal{Z},\ell_{i''}', \ell_{i''+1}'}(t)
  \nonumber \\
 & ~~~ \Big[- 
  \sum_{\ell_1'',...,\ell_{i''-1}'',\ell_{i''+2}'',...,\ell_d'' = 1 }^{r_1,...,r_{i-1},r_{i+2},...,r_d}
 \hat{\mathcal{Z}}^{\ell_1'',\ell_2''}_{1,x_1}(t)
 \cdots 
 \hat{\mathcal{Z}}^{\ell_{i''-1}'',\ell_{i''}'}_{i''-1,x_{i''-1}}(t) 
\hat{\mathcal{Z}}^{\ell_{i''+1}',\ell_{i''+2}''}_{i''+1,x_{i''+1}}(t) 
\cdots
\hat{\mathcal{Z}}^{\ell_{d}'',\ell_1''}_{d,x_{d}}(t) 
\Big]
\Bigg\}
\nonumber \\
& + \hat{s}_{\mathbf x}(t-1)  \bar\nu^p_{\mathbf x}(t).
\end{align}
\end{remark}
Hence, we have
$\hat{p}_{\mathbf{x} }(t)$ in \eqref{p-x-1} can be approximated by
\begin{align*} 
\hat{p}_{\mathbf{x} }(t) =&~ \sum_{\mathbf{l}' = 1}^{\mathbf{r}}  
{\hat{\mathcal{Z}}_{1, \mathbf{x}}^{\ell'}(t) }  
\cdots {\hat{\mathcal{Z}}_{d, \mathbf{x}}^{\ell'}(t) } 
\approx \bar p_{\mathbf{x} }(t) 
- \hat{s}_{\mathbf{x} }(t-1)  \bar\nu^p_{\mathbf{x} }(t), 
\end{align*}
where $\bar p_{\mathbf x}(t) \triangleq \sum_{\mathbf{l}' = 1}^{\mathbf{r}}
{\hat{\mathcal{Z}}_{1, x}^{\ell'}(t) }  
\cdots {\hat{\mathcal{Z}}_{d, x}^{\ell'}(t) }$ and
\begin{align*} 
\bar\nu^p_{\mathbf{x} }(t) =&~  \sum_{\mathbf{l}' = 1}^{\mathbf{r}}  
 \sum_{\substack{A \subsetneqq \mathbb{D} \\ A\neq \emptyset} }
 \Bigg\{ 
 \Big(
 \prod_{i \in A}  \hat{\mathcal{Z}}_{i, x_{i}}^{\ell'}(t) 
 \Big)^2
 \times \prod_{i'' \in \mathbb{D} \setminus A}
 \nu_{i'',x_{i''}}^{\mathcal{Z},\ell'}(t) 
 \Big(
 - \hat{s}_{\mathbf{x} }(t-1) 
  \prod_{i' \in \mathbb{D}} 
 \hat{\mathcal{Z}}^{\ell'}_{i',x_{i'}}(t)
 \Big)^{d-|A|-1} 
 \Bigg\}. 
\end{align*}
Then we complete the derivation. Note that we neglected small terms in Remark~\ref{remark3}, we observed that the approximations work well. If under specific applications this term was not negligible, it can simply be included back in the algorithm.

\subsection{Derivation of Equation  \eqref{v-p-2}  }

\textbf{Element-wise Result:}~
{\it
${\nu}^{p}_{\mathbf{x} }(t)$ in \eqref{v-p-1} can be approximated by
\begin{align*} 
{\nu}^{p}_{\mathbf{x} }(t) =&~ \sum_{\substack{\mathbf{l}'= \mathbf{1} }}^{\mathbf r}  
\sum_{\substack {A \subset \mathbb{D} \\ A \neq \emptyset}}
\left( \prod_{i' \in A} \nu_{i',\mathbf{x} }^{{ {\mathcal{Z}}},\ell'}(t)    
\prod_{i'' \in \mathbb{D} \setminus A }  {\hat{\mathcal{Z}}_{i'',\mathbf{x} }^{\ell' } (t) }^2 \right) \nonumber \\
 {\approx}&~
 \sum_{\substack{\mathbf{l} = \mathbf{1}}}^{\mathbf r}  
\prod_{i \in \mathbb{D}} \nu_{i,x_{i}}^{{ {\mathcal{Z}}},\ell}(t) + \bar\nu^{p}_{\mathbf{x}}(t).
\end{align*}
}

    \emph{Derivation:}
Plug \eqref{z-ix-mean-3} and $ \nu_{i,\mathbf x}^{\mathcal{Z},\ell_{i}, \ell_{i+1}}(t)  \approx  \nu_{i, x_i}^{\mathcal{Z},\ell_{i}, \ell_{i+1}}(t) $ into \eqref{v-p-1}, which yields
\begin{align} 
\label{v-p-2-A}
&~~~~{\nu}^{p}_{\mathbf x}(t) \nonumber \\
&  =  
\sum_{\substack{\ell_1',...,\ell_d' = 1 }}^{r_1,...,r_d}  
\sum_{ \substack {A \subset \{1,...,d\} \\ A \neq \emptyset} }
\left( \prod_{i' \in A} \nu_{i',\mathbf x}^{\mathcal{Z},\ell_{i'}', \ell_{i'+1}'}(t)    
\prod_{i'' \in \{1,...,d\} \setminus A }  {\hat{\mathcal{Z}}_{i'',\mathbf x}^{\ell_{i''}', \ell_{i'' + 1}' } (t) }^2 \right) \nonumber \\
& \approx
\sum_{\substack{\ell_1',...,\ell_d' = 1 }}^{r_1,...,r_d}  
\sum_{ \substack {A \subset \{1,...,d\} \\ A \neq \emptyset} }
\left( \prod_{i' \in A} \nu_{i',x_{i'}}^{\mathcal{Z},\ell_{i'}', \ell_{i'+1}'}(t)    
\prod_{i'' \in \{1,...,d\} \setminus A }  
{\hat{\mathcal{Z}}_{i'',\mathbf x}^{\ell_{i''}', \ell_{i'' + 1}' } (t) }^2 \right) \nonumber \\
& \overset{(a)}{\approx}
\sum_{\substack{\ell_1,...,\ell_d = 1 }}^{r_1,...,r_d}  
\prod_{i \in \{1,...,d\}} \nu_{i,x_{i}}^{\mathcal{Z},\ell_{i}, \ell_{i+1}}(t)
+
\sum_{\substack{\ell_1',...,\ell_d' = 1 }}^{r_1,...,r_d}  
\sum_{ \substack {A \subset \{1,...,d\} \\ A \neq \{1,...,d\} \\ A \neq \emptyset } }
\Bigg( \prod_{i' \in A} \nu_{i',x_{i'}}^{\mathcal{Z},\ell_{i'}', \ell_{i'+1}'}(t)    
\prod_{i'' \in \{1,...,d\} \setminus A }  
\Bigg[ \hat{\mathcal{Z}}_{i'',x_{i''}}^{\ell_{i''}',\ell_{i''+1}'}(t) - 
 \nonumber \\
&~~~
 \Big[ 
   \sum_{\ell_1'',...,\ell_{i''-1}'',\ell_{i''+2}'',...,\ell_d'' = 1 }^{r_1,...,r_{i-1},r_{i+2},...,r_d}
 \hat{\mathcal{Z}}^{\ell_1'',\ell_2''}_{1,x_1}(t-1)
 \cdots 
 \hat{\mathcal{Z}}^{\ell_{i''-1}'',\ell_{i''}'}_{i''-1,x_{i''-1}}(t-1) 
\hat{\mathcal{Z}}^{\ell_{i''+1}',\ell_{i''+2}''}_{i''+1,x_{i''+1}}(t-1) 
\cdots
\hat{\mathcal{Z}}^{\ell_{d}'',\ell_1''}_{d,x_{d}}(t-1) 
\Big]
\nonumber \\
&~~~\times
\hat{s}_{\mathbf x}(t-1)  \nu_{i'',x_{i''}}^{\mathcal{Z},\ell_{i''}', \ell_{i''+1}'}(t) 
\Bigg]^2
\Bigg)
 \nonumber \\
& \overset{(b)}{\approx}
\sum_{\substack{\ell_1,...,\ell_d = 1 }}^{r_1,...,r_d}  
\prod_{i \in \{1,...,d\}} \nu_{i,x_{i}}^{\mathcal{Z},\ell_{i}, \ell_{i+1}}(t)
+
\sum_{\substack{\ell_1',...,\ell_d' = 1 }}^{r_1,...,r_d}  
\sum_{ \substack {A \subset \{1,...,d\} \\ A \neq \{1,...,d\} \\ A \neq \emptyset } }
\Bigg( \prod_{i' \in A} \nu_{i',x_{i'}}^{\mathcal{Z},\ell_{i'}', \ell_{i'+1}'}(t)    
\prod_{i'' \in \{1,...,d\} \setminus A }  
\Bigg[ \hat{\mathcal{Z}}_{i'',x_{i''}}^{\ell_{i''}',\ell_{i''+1}'}(t) - 
 \nonumber \\
&~~~
 \Big[ 
   \sum_{\ell_1'',...,\ell_{i''-1}'',\ell_{i''+2}'',...,\ell_d'' = 1 }^{r_1,...,r_{i-1},r_{i+2},...,r_d}
 \hat{\mathcal{Z}}^{\ell_1'',\ell_2''}_{1,x_1}(t)
 \cdots 
 \hat{\mathcal{Z}}^{\ell_{i''-1}'',\ell_{i''}'}_{i''-1,x_{i''-1}}(t) 
\hat{\mathcal{Z}}^{\ell_{i''+1}',\ell_{i''+2}''}_{i''+1,x_{i''+1}}(t) 
\cdots
\hat{\mathcal{Z}}^{\ell_{d}'',\ell_1''}_{d,x_{d}}(t) 
\Big]
\hat{s}_{\mathbf x}(t-1)  \nu_{i'',x_{i''}}^{\mathcal{Z},\ell_{i''}', \ell_{i''+1}'}(t) 
\Bigg]^2
\Bigg)
\nonumber \\
 & \overset{(c)}{\approx}
 \sum_{\substack{\ell_1,...,\ell_d = 1 }}^{r_1,...,r_d}  
\prod_{i \in \{1,...,d\}} \nu_{i,x_{i}}^{\mathcal{Z},\ell_{i}, \ell_{i+1}}(t) + \bar\nu^{p}_{\mathbf{x}}(t),
\end{align}
where $(a)$ follows from \eqref{z-ix-mean-3}, for $(b)$ we used  $ \hat{\mathcal{Z}}^{\ell_{i''}'',\ell_{i''+1}'}_{i'',x_{i''}}(t) $ in place of $ \hat{\mathcal{Z}}^{\ell_{i''}'',\ell_{i''+1}'}_{i'',x_{i''}}(t-1) $ and $(c)$ follows from Remark~\ref{remarkvp}.

\begin{remark}
\label{remarkvp}
We observed that the following approximations hold
\begin{align} 
\label{remarkvp-term-1}
&\sum_{\substack{\ell_1',...,\ell_d' = 1 }}^{r_1,...,r_d}  
\sum_{ \substack {A \subset \{1,...,d\} \\ A \neq \{1,...,d\} \\ A \neq \emptyset } }
\Bigg( \prod_{i' \in A} \nu_{i',x_{i'}}^{\mathcal{Z},\ell_{i'}', \ell_{i'+1}'}(t)    
\prod_{i'' \in \{1,...,d\} \setminus A }  
\Bigg[ \hat{\mathcal{Z}}_{i'',x_{i''}}^{\ell_{i''}',\ell_{i''+1}'}(t) - 
 \nonumber \\
&~~~
 \Big[ 
   \sum_{\ell_1'',...,\ell_{i''-1}'',\ell_{i''+2}'',...,\ell_d'' = 1 }^{r_1,...,r_{i-1},r_{i+2},...,r_d}
 \hat{\mathcal{Z}}^{\ell_1'',\ell_2''}_{1,x_1}(t)
 \cdots 
 \hat{\mathcal{Z}}^{\ell_{i''-1}'',\ell_{i''}'}_{i''-1,x_{i''-1}}(t) 
\hat{\mathcal{Z}}^{\ell_{i''+1}',\ell_{i''+2}''}_{i''+1,x_{i''+1}}(t) 
\cdots
\hat{\mathcal{Z}}^{\ell_{d}'',\ell_1''}_{d,x_{d}}(t) 
\Big]
\hat{s}_{\mathbf x}(t-1)  \nu_{i'',x_{i''}}^{\mathcal{Z},\ell_{i''}', \ell_{i''+1}'}(t) 
\Bigg]^2
\Bigg)
\nonumber \\
=
&\sum_{\substack{\ell_1',...,\ell_d' = 1 }}^{r_1,...,r_d}  
\sum_{ \substack {A \subset \{1,...,d\} \\ A \neq \{1,...,d\} \\ A \neq \emptyset } }
\Bigg( \prod_{i' \in  \{1,...,d\} \setminus A} \nu_{i',x_{i'}}^{\mathcal{Z},\ell_{i'}', \ell_{i'+1}'}(t)    
\prod_{i'' \in A}  
\Bigg[ \hat{\mathcal{Z}}_{i'',x_{i''}}^{\ell_{i''}',\ell_{i''+1}'}(t) - 
 \nonumber \\
&~~~
 \Big[ 
   \sum_{\ell_1'',...,\ell_{i''-1}'',\ell_{i''+2}'',...,\ell_d'' = 1 }^{r_1,...,r_{i-1},r_{i+2},...,r_d}
 \hat{\mathcal{Z}}^{\ell_1'',\ell_2''}_{1,x_1}(t)
 \cdots 
 \hat{\mathcal{Z}}^{\ell_{i''-1}'',\ell_{i''}'}_{i''-1,x_{i''-1}}(t) 
\hat{\mathcal{Z}}^{\ell_{i''+1}',\ell_{i''+2}''}_{i''+1,x_{i''+1}}(t) 
\cdots
\hat{\mathcal{Z}}^{\ell_{d}'',\ell_1''}_{d,x_{d}}(t) 
\Big]
\hat{s}_{\mathbf x}(t-1)  \nu_{i'',x_{i''}}^{\mathcal{Z},\ell_{i''}', \ell_{i''+1}'}(t) 
\Bigg]^2
\Bigg)
\nonumber \\
\overset{(a)}{\approx}
&\sum_{\substack{\ell_1',...,\ell_d' = 1 }}^{r_1,...,r_d}  
\sum_{ \substack {A \subset \{1,...,d\} \\ A \neq \{1,...,d\} \\ A \neq \emptyset } }
\Bigg( \prod_{i' \in  \{1,...,d\} \setminus A} \nu_{i',x_{i'}}^{\mathcal{Z},\ell_{i'}', \ell_{i'+1}'}(t)    
\prod_{i'' \in A}  
\hat{\mathcal{Z}}_{i'',x_{i''}}^{\ell_{i''}',\ell_{i''+1}'}(t)
\prod_{i'' \in A}  
\Bigg[ \hat{\mathcal{Z}}_{i'',x_{i''}}^{\ell_{i''}',\ell_{i''+1}'}(t) - 
 \nonumber \\
&~~~
2 \Big[ 
   \sum_{\ell_1'',...,\ell_{i''-1}'',\ell_{i''+2}'',...,\ell_d'' = 1 }^{r_1,...,r_{i-1},r_{i+2},...,r_d}
 \hat{\mathcal{Z}}^{\ell_1'',\ell_2''}_{1,x_1}(t)
 \cdots 
 \hat{\mathcal{Z}}^{\ell_{i''-1}'',\ell_{i''}'}_{i''-1,x_{i''-1}}(t) 
\hat{\mathcal{Z}}^{\ell_{i''+1}',\ell_{i''+2}''}_{i''+1,x_{i''+1}}(t) 
\cdots
\hat{\mathcal{Z}}^{\ell_{d}'',\ell_1''}_{d,x_{d}}(t) 
\Big]
\hat{s}_{\mathbf x}(t-1)  \nu_{i'',x_{i''}}^{\mathcal{Z},\ell_{i''}', \ell_{i''+1}'}(t) 
\Bigg]
\Bigg)
\nonumber \\
\overset{(b)}{\approx}
& \frac{1}{\hat{s}_{\mathbf x}(t-1)} 
\sum_{\ell_1',...,\ell_d' = 1}^{r_1,...,r_d}  
 \sum_{\substack{ {A\subsetneqq \{1,...,d\}} \\ A\neq \emptyset} }
 \Bigg\{ 
 \prod_{i' \in A}  \hat{\mathcal{Z}}_{i', x_{i'}}^{\ell_{i'}',\ell_{i'+1}'}(t) 
  \prod_{i'' \in \{1,...,d\}\setminus A} 
  \hat{s}_{\mathbf x}(t-1)  \nu_{i'',x_{i''}}^{\mathcal{Z},\ell_{i''}', \ell_{i''+1}'}(t)
  \nonumber \\
 & ~~~ \Big[-
 \hat{\mathcal{Z}}^{\ell_1',\ell_2'}_{1,x_1}(t)
 \cdots 
 \hat{\mathcal{Z}}^{\ell_{i''-1}',\ell_{i''}'}_{i''-1,x_{i''-1}}(t) 
\hat{\mathcal{Z}}^{\ell_{i''+1}',\ell_{i''+2}'}_{i''+1,x_{i''+1}}(t) 
\cdots
\hat{\mathcal{Z}}^{\ell_{d}',\ell_1'}_{d,x_{d}}(t) 
\Big]
\Bigg\},
\end{align}
where in $(a)$ we neglect the term 
\begin{align}
\Big[ 
   \sum_{\ell_1'',...,\ell_{i''-1}'',\ell_{i''+2}'',...,\ell_d'' = 1 }^{r_1,...,r_{i-1},r_{i+2},...,r_d}
 \hat{\mathcal{Z}}^{\ell_1'',\ell_2''}_{1,x_1}(t)
 \cdots 
 \hat{\mathcal{Z}}^{\ell_{i''-1}'',\ell_{i''}'}_{i''-1,x_{i''-1}}(t) 
\hat{\mathcal{Z}}^{\ell_{i''+1}',\ell_{i''+2}''}_{i''+1,x_{i''+1}}(t) 
\cdots
\hat{\mathcal{Z}}^{\ell_{d}'',\ell_1''}_{d,x_{d}}(t) 
\Big]
\hat{s}_{\mathbf x}(t-1)  \nu_{i'',x_{i''}}^{\mathcal{Z},\ell_{i''}', \ell_{i''+1}'}(t) 
\Big]^2 \nonumber
\end{align}
with similar arguments as before, and $(b)$ follows from that we observed the following approximation holds:
\begin{align} 
&\prod_{i'' \in A}  
\Bigg[ \hat{\mathcal{Z}}_{i'',x_{i''}}^{\ell_{i''}',\ell_{i''+1}'}(t) - 
2 \Big[ 
   \sum_{\ell_1'',...,\ell_{i''-1}'',\ell_{i''+2}'',...,\ell_d'' = 1 }^{r_1,...,r_{i-1},r_{i+2},...,r_d}
 \hat{\mathcal{Z}}^{\ell_1'',\ell_2''}_{1,x_1}(t)
 \cdots 
 \hat{\mathcal{Z}}^{\ell_{i''-1}'',\ell_{i''}'}_{i''-1,x_{i''-1}}(t) 
\hat{\mathcal{Z}}^{\ell_{i''+1}',\ell_{i''+2}''}_{i''+1,x_{i''+1}}(t) 
\cdots
\hat{\mathcal{Z}}^{\ell_{d}'',\ell_1''}_{d,x_{d}}(t) 
\Big]
 \nonumber \\
&~~~
\times
\hat{s}_{\mathbf x}(t-1)  \nu_{i'',x_{i''}}^{\mathcal{Z},\ell_{i''}', \ell_{i''+1}'}(t) 
\Bigg] 
 \nonumber \\
\approx 
& \frac{1}{\hat{s}_{\mathbf x}(t-1)} 
\prod_{i'' \in \{1,...,d\}\setminus A} 
  \hat{s}_{\mathbf x}(t-1)
\Big[-
 \hat{\mathcal{Z}}^{\ell_1',\ell_2'}_{1,x_1}(t)
 \cdots 
 \hat{\mathcal{Z}}^{\ell_{i''-1}',\ell_{i''}'}_{i''-1,x_{i''-1}}(t) 
\hat{\mathcal{Z}}^{\ell_{i''+1}',\ell_{i''+2}'}_{i''+1,x_{i''+1}}(t) 
\cdots
\hat{\mathcal{Z}}^{\ell_{d}',\ell_1'}_{d,x_{d}}(t) 
\Big]
\Bigg\}.
\end{align}
\end{remark}
We observed that the above approximation holds well since the difference between the left-hand side and the right-hand side is small, note that if under specific applications the above approximation was not applicable, it can simply be included back in the algorithm. 

As a result, we have
\begin{align*} 
{\nu}^{p}_{\mathbf{x} }(t) =&~ \sum_{\substack{\mathbf{l}'= \mathbf{1} }}^{\mathbf r}  
\sum_{\substack {A \subset \mathbb{D} \\ A \neq \emptyset}}
\left( \prod_{i' \in A} \nu_{i',\mathbf{x} }^{{ {\mathcal{Z}}},\ell'}(t)    
\prod_{i'' \in \mathbb{D} \setminus A }  {\hat{\mathcal{Z}}_{i'',\mathbf{x} }^{\ell' } (t) }^2 \right) 
 \approx
 \sum_{\substack{\mathbf{l} = \mathbf{1}}}^{\mathbf r}  
\prod_{i \in \mathbb{D}} \nu_{i,x_{i}}^{{ {\mathcal{Z}}},\ell}(t) + \bar\nu^{p}_{\mathbf{x}}(t).
\end{align*}

Then we complete the derivation.

\subsection{Derivation of Equation \eqref{var-r-approx-1} }

\textbf{Element-wise Result:}~
{\it
$\nu^r_{\mathbf q^i}$ and $\hat{r}_{\mathbf q^i}$ in \eqref{z-ix-mean-3} can be respectively approximated by
\begin{align*}
&\frac{1}{
  \nu^r_{\mathbf q^i}}
  \approx
 \sum_{{ {\mathbf{x}':x'_i = x_i}}}  
\Bigg\{    
 \Big[ 
 {\sum_{\mathbf{l}'_{\setminus i} = 1}^{\mathbf{r}_{\setminus i}}}
 {\hat{\mathcal{Z}}^{\ell'}_{\setminus i, x'}(t)}
\Big]^2
\nu_{\mathbf{x} '}^s(t)  
\Bigg\},
\end{align*}
and
\begin{align*} 
&\hat{r}_{\mathbf q^i}
~{\approx}~
\nu^r_{\mathbf q^i}  
 \sum_{{ {\mathbf{x}':x'_i = x_i}}}  
\Bigg\{  
\Bigg[ 
 {\sum_{\mathbf{l}'_{\setminus i} = 1}^{\mathbf{r}_{\setminus i}}}
 {\hat{\mathcal{Z}}^{\ell'}_{\setminus i, x'}(t)}
\Bigg]
 \hat{s}_{\mathbf{x} '}(t) 
  \Bigg\}  +
 \hat{\mathcal{Z}}^{\ell}_{i,x_i}(t) \left\{ 1 -
\frac{
 \sum\limits_{{ {\mathbf{x}':x'_i = x_i}}}     
   {\nu_{\mathbf{x} '}^s(t)}  
 {\zeta}_{x}(t)}
 {\sum\limits_{{ {\mathbf{x}':x'_i = x_i}}}  
\Bigg\{    
 \left[ 
 {\sum_{\mathbf{l}'_{\setminus i} = 1}^{\mathbf{r}_{\setminus i}}}
 {\hat{\mathcal{Z}}^{\ell'}_{\setminus i, x'}(t)}
\right]^2
\nu_{\mathbf{x} '}^s(t)  
\Bigg\} 
}
\right \}.\nonumber 
\end{align*}
}

\emph{Derivation:}
Plug \eqref{z-ix-mean-3} into $\hat{r}_{\mathbf q^i}$ to obtain
\begin{align} 
\label{hat-r-3-A}
&~~~~~~\hat{r}_{[i,\ell_i,\ell_{i+1},x_i]}  \nonumber \\
 &= \nu^r_{[i,\ell_i,\ell_{i+1},x_i]}  
 \sum_{{ {\mathbf{x}':x'_i = x_i}}}  
\Bigg\{  
\Bigg[
\sum_{\substack{\ell_1',...,\ell_{i-1}',\ell_{i+2}',...,\ell_d' = 1  }}^{r_1,...,r_{i-1},r_{i+2},...,r_d}
\hat{\mathcal{Z}}^{\ell_1',\ell_2'}_{1,\mathbf x'}(t)
 \cdots \hat{\mathcal{Z}}^{\ell_{i-1}',\ell_i}_{i-1,\mathbf x'}(t) 
\hat{\mathcal{Z}}^{\ell_{i+1},\ell_{i+2}'}_{i+1,\mathbf x'}(t) \cdots
\hat{\mathcal{Z}}^{\ell_{d}',\ell_1'}_{d,\mathbf x'}(t)
\Bigg]
 \hat{s}_{\mathbf x'}(t) 
  \Bigg\}  
  \nonumber \\
 &~~~+ \hat{\mathcal{Z}}^{\ell_{i},\ell_{i+1}}_{i,x_i}(t) 
 \Bigg\{ 1 + \nu^r_{[i,\ell_i,\ell_{i+1},x_i]}   
 \sum_{{ {\mathbf{x}':x'_i = x_i}}}     
   \left [  {\hat{s}_{\mathbf x'}(t)}^2  -  {\nu_{\mathbf x'}^s(t)} \right]  \times
   \nonumber \\
 &~~~\sum_{\substack{\ell_1',...,\ell_{i-1}',\ell_{i+2}',...,\ell_d' = 1  }}^{r_1,...,r_{i-1},r_{i+2},...,r_d}
\sum_{ \substack {A \subset \{1,...,d\} \setminus \{i\} \\ A \neq \emptyset} } 
\Bigg[
 \prod_{ \substack{i' \in A ,\ell_i'=\ell_i \\
\ell_{i+1}' = \ell_{i+1}
}}
 \nu_{i', x_{i'}'}^{\mathcal{Z},\ell_{i'}', \ell_{i'+1}'}(t) 
    \prod_{ \substack{
    i'' \in \{1,...,d\} \setminus (\{ i \} \cup A) \\
    \ell_i'=\ell_i,\ell_{i+1}' = \ell_{i+1}
    }} 
 {\hat{\mathcal{Z}}_{i'', x_{i''}'}^{\ell_{i''}', \ell_{i'' + 1}' } (t) }^2 
 \Bigg]   
  \Bigg\} 
  \nonumber \\
& \overset{(a)}{\approx}  
\nu^r_{[i,\ell_i,\ell_{i+1},x_i]}  
 \sum_{{ {\mathbf{x}':x'_i = x_i}}}  
\Bigg\{  
\Bigg[ 
\sum_{\substack{\ell_1',...,\ell_{i-1}',\ell_{i+2}',...,\ell_d' = 1  }}^{r_1,...,r_{i-1},r_{i+2},...,r_d}
\hat{\mathcal{Z}}^{\ell_1',\ell_2'}_{1,x_1'}(t)
 \cdots \hat{\mathcal{Z}}^{\ell_{i-1}',\ell_i}_{i-1,x_{i-1}'}(t) 
\hat{\mathcal{Z}}^{\ell_{i+1},\ell_{i+2}'}_{i+1,x_{i+1}'}(t) \cdots
\hat{\mathcal{Z}}^{\ell_{d}',\ell_1'}_{d,x_d'}(t)
\Bigg]
 \hat{s}_{\mathbf x'}(t) 
  \Bigg\}  
  \nonumber \\
 &~~~+ \hat{\mathcal{Z}}^{\ell_{i},\ell_{i+1}}_{i,x_i}(t) 
 \Bigg\{ 1 -
  \nu^r_{[i,\ell_i,\ell_{i+1},x_i]}   
 \sum_{{ {\mathbf{x}':x'_i = x_i}}}     
   {\nu_{\mathbf x'}^s(t)}  \times
   \nonumber \\
 &~~~\sum_{\substack{\ell_1',...,\ell_{i-1}',\ell_{i+2}',...,\ell_d' = 1  }}^{r_1,...,r_{i-1},r_{i+2},...,r_d}
\sum_{ \substack {A \subset \{1,...,d\} \setminus \{i\} \\ A \neq \emptyset} } 
\Bigg[
 \prod_{ \substack{i' \in A ,\ell_i'=\ell_i \\
\ell_{i+1}' = \ell_{i+1}
}}
 \nu_{i', x_{i'}'}^{\mathcal{Z},\ell_{i'}', \ell_{i'+1}'}(t) 
    \prod_{ \substack{
    i'' \in \{1,...,d\} \setminus (\{ i \} \cup A) \\
    \ell_i'=\ell_i,\ell_{i+1}' = \ell_{i+1}
    }} 
 {\hat{\mathcal{Z}}_{i'', x_{i''}'}^{\ell_{i''}', \ell_{i'' + 1}' } (t) }^2 
 \Bigg]   
  \Bigg\} 
  \nonumber \\
& \overset{(b)}{\approx} 
\nu^r_{[i,\ell_i,\ell_{i+1},x_i]}  
 \sum_{{ {\mathbf{x}':x'_i = x_i}}}  
\Bigg\{  
\Bigg[ 
\sum_{\substack{\ell_1',...,\ell_{i-1}',\ell_{i+2}',...,\ell_d' = 1  }}^{r_1,...,r_{i-1},r_{i+2},...,r_d}
\hat{\mathcal{Z}}^{\ell_1',\ell_2'}_{1,x_1'}(t)
 \cdots \hat{\mathcal{Z}}^{\ell_{i-1}',\ell_i}_{i-1,x_{i-1}'}(t) 
\hat{\mathcal{Z}}^{\ell_{i+1},\ell_{i+2}'}_{i+1,x_{i+1}'}(t) \cdots
\hat{\mathcal{Z}}^{\ell_{d}',\ell_1'}_{d,x_d'}(t)
\Bigg]
 \hat{s}_{\mathbf x'}(t) 
  \Bigg\}  
  \nonumber \\
    &+ \hat{\mathcal{Z}}^{\ell_{i},\ell_{i+1}}_{i,x_i}(t) \times
  \nonumber \\
 &
 \left\{ 1 -
\frac{
 \sum\limits_{{ {\mathbf{x}':x'_i = x_i}}}     
   {\nu_{\mathbf x'}^s(t)}  
  \sum\limits_{\substack{\ell_1',...,\ell_{i-1}',\ell_{i+2}',...,\ell_d' = 1  }}^{r_1,...,r_{i-1},r_{i+2},...,r_d}
\sum\limits_{ \substack {A \subset \{1,...,d\} \\  \setminus \{i\}, A \neq \emptyset} } 
\Bigg[
 \prod\limits_{ \substack{i' \in A ,\ell_i'=\ell_i \\
\ell_{i+1}' = \ell_{i+1}
}}
 \nu_{i', x_{i'}'}^{\mathcal{Z},\ell_{i'}', \ell_{i'+1}'}(t) 
    \prod\limits_{ \substack{
    i'' \in \{1,...,d\} \setminus (\{ i \} \cup A) \\
    \ell_i'=\ell_i,\ell_{i+1}' = \ell_{i+1}
    }} 
 {\hat{\mathcal{Z}}_{i'', x_{i''}'}^{\ell_{i''}', \ell_{i'' + 1}' } (t) }^2 
 \Bigg]      
  }{
 \sum\limits_{{ {\mathbf{x}':x'_i = x_i}}}  
\Bigg\{    
 \left[ 
\sum\limits_{\substack{\ell_1',...,\ell_{i-1}',\ell_{i+2}',...,\ell_d' = 1  }}^{r_1,...,r_{i-1},r_{i+2},...,r_d}
\hat{\mathcal{Z}}^{\ell_1',\ell_2'}_{1, x_1'}(t)
 \cdots \hat{\mathcal{Z}}^{\ell_{i-1}',\ell_i}_{i-1, x_{i-1}'}(t) 
\hat{\mathcal{Z}}^{\ell_{i+1},\ell_{i+2}'}_{i+1,x_{i+1}'}(t) \cdots
\hat{\mathcal{Z}}^{\ell_{d}',\ell_1'}_{d, x_d'}(t)
\right]^2
\nu_{\mathbf x'}^s(t)  
\Bigg\} 
}
\right \},
\end{align}
where $(a)$ follows from neglecting the below term with the same arguments as before.
\begin{align} 
\label{nu-r-approx-2-TR}
&\nu^r_{[i,\ell_i,\ell_{i+1},x_i]}  
 \sum_{{ {\mathbf{x}':x'_i = x_i}}}  
\Bigg\{  
\left[
\sum_{\substack{\ell_1',...,\ell_{i-1}',\ell_{i+2}',...,\ell_d' = 1  }}^{r_1,...,r_{i-1},r_{i+2},...,r_d}
\hat{\mathcal{Z}}^{\ell_1',\ell_2'}_{1,\mathbf x'}(t)
 \cdots \hat{\mathcal{Z}}^{\ell_{i-1}',\ell_i}_{i-1,\mathbf x'}(t) 
\hat{\mathcal{Z}}^{\ell_{i+1},\ell_{i+2}'}_{i+1,\mathbf x'}(t) \cdots
\hat{\mathcal{Z}}^{\ell_{d}',\ell_1'}_{d,\mathbf x'}(t)
\right]
 \hat{s}_{\mathbf x'}(t) 
  \Bigg\}  
  \nonumber \\
 &- \nu^r_{[i,\ell_i,\ell_{i+1},x_i]}  
 \sum_{{ {\mathbf{x}':x'_i = x_i}}}  
\Bigg\{  
\Bigg[ 
\sum_{\substack{\ell_1',...,\ell_{i-1}',\ell_{i+2}',...,\ell_d' = 1  }}^{r_1,...,r_{i-1},r_{i+2},...,r_d}
\hat{\mathcal{Z}}^{\ell_1',\ell_2'}_{1,x_1'}(t)
 \cdots \hat{\mathcal{Z}}^{\ell_{i-1}',\ell_i}_{i-1,x_{i-1}'}(t) 
\hat{\mathcal{Z}}^{\ell_{i+1},\ell_{i+2}'}_{i+1,x_{i+1}'}(t) \cdots
\hat{\mathcal{Z}}^{\ell_{d}',\ell_1'}_{d,x_d'}(t)
\Bigg]
 \hat{s}_{\mathbf x'}(t) 
  \Bigg\}  
    \nonumber \\
 &+ \hat{\mathcal{Z}}^{\ell_{i},\ell_{i+1}}_{i,x_i}(t) 
 \Bigg\{  \nu^r_{[i,\ell_i,\ell_{i+1},x_i]}   
 \sum_{{ {\mathbf{x}':x'_i = x_i}}}     
   \left [  {\hat{s}_{\mathbf x'}(t)}^2  \right]  \times
   \nonumber \\
 &\sum_{\substack{\ell_1',...,\ell_{i-1}',\ell_{i+2}',...,\ell_d' = 1  }}^{r_1,...,r_{i-1},r_{i+2},...,r_d}
\sum_{ \substack {A \subset \{1,...,d\} \setminus \{i\} \\ A \neq \emptyset} } 
\Bigg[
 \prod_{ \substack{i' \in A ,\ell_i'=\ell_i \\
\ell_{i+1}' = \ell_{i+1}
}}
 \nu_{i', x_{i'}'}^{\mathcal{Z},\ell_{i'}', \ell_{i'+1}'}(t) 
    \prod_{ \substack{
    i'' \in \{1,...,d\} \setminus (\{ i \} \cup A) \\
    \ell_i'=\ell_i,\ell_{i+1}' = \ell_{i+1}
    }} 
 {\hat{\mathcal{Z}}_{i'', x_{i''}'}^{\ell_{i''}', \ell_{i'' + 1}' } (t) }^2 
 \Bigg]   
  \Bigg\} 
  \nonumber \\
\overset{(x)}{\approx} &~
-\nu^r_{[i,\ell_i,\ell_{i+1},x_i]}  
 \sum_{{ {\mathbf{x}':x'_i = x_i}}}  
\Bigg\{ \hat{s}_{\mathbf x'}(t)   
\sum_{\substack{\ell_1',...,\ell_{i-1}',\ell_{i+2}',...,\ell_d' = 1  }}^{r_1,...,r_{i-1},r_{i+2},...,r_d} 
\Bigg[
 \sum_{\substack{A\subset \{1,...,d\} \setminus \{i\} \\ A\neq\{1,...,d\} \setminus \{i\} } }
 \Bigg\{ 
 \nonumber \\
 &~\prod_{\substack{i' \in A, \ell_i' = \ell_i  \\  \ell_{i+1}' = \ell_{i+1} }} 
  \hat{\mathcal{Z}}_{i', x_{i'}}^{\ell_{i'}',\ell_{i'+1}'}(t) 
  \prod_{ \substack{ i'' \in \{1,...,d\}\setminus (\{i\} \cup A)  \\     \ell_i' = \ell_i,  \ell_{i+1}' = \ell_{i+1}        }}
   \hat{s}_{\mathbf x}(t-1)  \nu_{i'',x_{i''}}^{\mathcal{Z},\ell_{i''}', \ell_{i''+1}'}(t)
   \nonumber \\
 & ~~~ \Big[- 
  \sum_{\ell_1'',...,\ell_{i''-1}'',\ell_{i''+2}'',...,\ell_d'' = 1 }^{r_1,...,r_{i-1},r_{i+2},...,r_d}
 \hat{\mathcal{Z}}^{\ell_1'',\ell_2''}_{1,x_1}(t)
 \cdots 
 \hat{\mathcal{Z}}^{\ell_{i''-1}'',\ell_{i''}'}_{i''-1,x_{i''-1}}(t) 
\hat{\mathcal{Z}}^{\ell_{i''+1}',\ell_{i''+2}''}_{i''+1,x_{i''+1}}(t) 
\cdots
\hat{\mathcal{Z}}^{\ell_{d}'',\ell_1''}_{d,x_{d}}(t) 
\Big]
\Bigg\}
\Bigg]
\Bigg\}
\nonumber \\
&+ \hat{\mathcal{Z}}^{\ell_{i},\ell_{i+1}}_{i,x_i}(t) 
 \Bigg\{  \nu^r_{[i,\ell_i,\ell_{i+1},x_i]}   
 \sum_{{ {\mathbf{x}':x'_i = x_i}}}     
   \left [  {\hat{s}_{\mathbf x'}(t)}^2  \right]  \times
   \nonumber \\
 &\sum_{\substack{\ell_1',...,\ell_{i-1}',\ell_{i+2}',...,\ell_d' = 1  }}^{r_1,...,r_{i-1},r_{i+2},...,r_d}
\sum_{ \substack {A \subset \{1,...,d\} \setminus \{i\} \\ A \neq \emptyset} } 
\Bigg[
 \prod_{ \substack{i' \in A ,\ell_i'=\ell_i \\
\ell_{i+1}' = \ell_{i+1}
}}
 \nu_{i', x_{i'}'}^{\mathcal{Z},\ell_{i'}', \ell_{i'+1}'}(t) 
    \prod_{ \substack{
    i'' \in \{1,...,d\} \setminus (\{ i \} \cup A) \\
    \ell_i'=\ell_i,\ell_{i+1}' = \ell_{i+1}
    }} 
 {\hat{\mathcal{Z}}_{i'', x_{i''}'}^{\ell_{i''}', \ell_{i'' + 1}' } (t) }^2 
 \Bigg]   
  \Bigg\} ,
\end{align}
where $(x)$ follows Remark~\eqref{remark-4}. We observed that the term \eqref{nu-r-approx-2-TR} is small, if under specific applications this term was not negligible, it can simply be included back in the algorithm.

And for $(b)$, we had the approximation:
\begin{align} 
\label{var-r-approx-1-A}
\frac{1}{
  \nu^r_{[i,\ell_i,\ell_{i+1},x_i]}   
  }
  \approx
 \sum_{{ {\mathbf{x}':x'_i = x_i}}}  
\Bigg\{    
 \left[ 
\sum_{\substack{\ell_1',...,\ell_{i-1}',\ell_{i+2}',...,\ell_d' = 1  }}^{r_1,...,r_{i-1},r_{i+2},...,r_d}
\hat{\mathcal{Z}}^{\ell_1',\ell_2'}_{1, x_1'}(t)
 \cdots \hat{\mathcal{Z}}^{\ell_{i-1}',\ell_i}_{i-1, x_{i-1}'}(t) 
\hat{\mathcal{Z}}^{\ell_{i+1},\ell_{i+2}'}_{i+1,x_{i+1}'}(t) \cdots
\hat{\mathcal{Z}}^{\ell_{d}',\ell_1'}_{d, x_d'}(t)
\right]^2
\nu_{\mathbf x'}^s(t)  
\Bigg\} ,
\end{align}
by neglecting ${\hat{s}_{\mathbf x'}(t)}^2 - \nu^s_{\mathbf x'}(t)$ terms  as explained in Appendix B in~\cite{parker2014bilinear}.

\begin{remark}
\label{remark-4}
The following holds:
\begin{align} 
\label{remark-3-1}
&\nu^r_{[i,\ell_i,\ell_{i+1},x_i]}  
 \sum_{{ {\mathbf{x}':x'_i = x_i}}}  
\Bigg\{  
\left[
\sum_{\substack{\ell_1',...,\ell_{i-1}',\ell_{i+2}',...,\ell_d' = 1  }}^{r_1,...,r_{i-1},r_{i+2},...,r_d}
\hat{\mathcal{Z}}^{\ell_1',\ell_2'}_{1,\mathbf x'}(t)
 \cdots \hat{\mathcal{Z}}^{\ell_{i-1}',\ell_i}_{i-1,\mathbf x'}(t) 
\hat{\mathcal{Z}}^{\ell_{i+1},\ell_{i+2}'}_{i+1,\mathbf x'}(t) \cdots
\hat{\mathcal{Z}}^{\ell_{d}',\ell_1'}_{d,\mathbf x'}(t)
\right]
\hat{s}_{\mathbf x'}(t) 
\Bigg\}  \nonumber \\
\overset{(z)}{\approx} &~
\nu^r_{[i,\ell_i,\ell_{i+1},x_i]}  
 \sum_{{ {\mathbf{x}':x'_i = x_i}}}  
\Bigg\{ \hat{s}_{\mathbf x'}(t)   
\sum_{\substack{\ell_1',...,\ell_{i-1}',\ell_{i+2}',...,\ell_d' = 1  }}^{r_1,...,r_{i-1},r_{i+2},...,r_d} 
\Bigg[
{  \hat{\mathcal{Z}}_{1, x_1}^{\ell_1',\ell_{2}'}(t)  }  
\cdots 
{  \hat{\mathcal{Z}}_{i-1, x_{i-1}}^{\ell_{i-1}',\ell_{i}}(t)  }
{  \hat{\mathcal{Z}}_{i+1, x_{i+1}}^{\ell_{i+1},\ell_{i+2}'}(t)  }
\cdots
{  \hat{\mathcal{Z}}_{d, x_d}^{\ell_d',\ell_{1}'}(t)  }
\nonumber \\
&+
 \sum_{\substack{A\subset \{1,...,d\} \setminus \{i\} \\ A\neq\{1,...,d\} \setminus \{i\} } }
 \Bigg\{ 
 \prod_{\substack{i' \in A, \ell_i' = \ell_i  \\  \ell_{i+1}' = \ell_{i+1} }} 
  \hat{\mathcal{Z}}_{i', x_{i'}}^{\ell_{i'}',\ell_{i'+1}'}(t) 
  \prod_{ \substack{ i'' \in \{1,...,d\}\setminus (\{i\} \cup A)  \\     \ell_i' = \ell_i,  \ell_{i+1}' = \ell_{i+1}        }}
   \hat{s}_{\mathbf x}(t-1)  \nu_{i'',x_{i''}}^{\mathcal{Z},\ell_{i''}', \ell_{i''+1}'}(t)
   \nonumber \\
 & ~~~ \Big[- 
  \sum_{\ell_1'',...,\ell_{i''-1}'',\ell_{i''+2}'',...,\ell_d'' = 1 }^{r_1,...,r_{i-1},r_{i+2},...,r_d}
 \hat{\mathcal{Z}}^{\ell_1'',\ell_2''}_{1,x_1}(t)
 \cdots 
 \hat{\mathcal{Z}}^{\ell_{i''-1}'',\ell_{i''}'}_{i''-1,x_{i''-1}}(t) 
\hat{\mathcal{Z}}^{\ell_{i''+1}',\ell_{i''+2}''}_{i''+1,x_{i''+1}}(t) 
\cdots
\hat{\mathcal{Z}}^{\ell_{d}'',\ell_1''}_{d,x_{d}}(t) 
\Big]
\Bigg\}
\Bigg]
\Bigg\},
\end{align}
 where $(z)$ follows from
\begin{align} 
\label{remark-3-1_2}
&\hat{\mathcal{Z}}^{\ell_1',\ell_2'}_{1,\mathbf x'}(t)
 \cdots \hat{\mathcal{Z}}^{\ell_{i-1}',\ell_i}_{i-1,\mathbf x'}(t) 
\hat{\mathcal{Z}}^{\ell_{i+1},\ell_{i+2}'}_{i+1,\mathbf x'}(t) \cdots
\hat{\mathcal{Z}}^{\ell_{d}',\ell_1'}_{d,\mathbf x'}(t) \nonumber \\
 \overset{(a)}{\approx}&~
{  \hat{\mathcal{Z}}_{1, x_1}^{\ell_1',\ell_{2}'}(t)  }  
\cdots 
{  \hat{\mathcal{Z}}_{i-1, x_{i-1}}^{\ell_{i-1}',\ell_{i}}(t)  }
{  \hat{\mathcal{Z}}_{i+1, x_{i+1}}^{\ell_{i+1},\ell_{i+2}'}(t)  }
\cdots
{  \hat{\mathcal{Z}}_{d, x_d}^{\ell_d',\ell_{1}'}(t)  }
\nonumber \\
&+
 \sum_{\substack{A\subset \{1,...,d\} \setminus \{i\} \\ A\neq\{1,...,d\} \setminus \{i\} } }
 \Bigg\{ 
 \prod_{\substack{i' \in A, \ell_i' = \ell_i  \\  \ell_{i+1}' = \ell_{i+1} }} 
  \hat{\mathcal{Z}}_{i', x_{i'}}^{\ell_{i'}',\ell_{i'+1}'}(t) 
  \prod_{ \substack{ i'' \in \{1,...,d\}\setminus (\{i\} \cup A)  \\     \ell_i' = \ell_i,  \ell_{i+1}' = \ell_{i+1}        }}
   \hat{s}_{\mathbf x}(t-1)  \nu_{i'',x_{i''}}^{\mathcal{Z},\ell_{i''}', \ell_{i''+1}'}(t)
   \nonumber \\
 & ~~~ \Big[- 
  \sum_{\ell_1'',...,\ell_{i''-1}'',\ell_{i''+2}'',...,\ell_d'' = 1 }^{r_1,...,r_{i-1},r_{i+2},...,r_d}
 \hat{\mathcal{Z}}^{\ell_1'',\ell_2''}_{1,x_1}(t-1)
 \cdots 
 \hat{\mathcal{Z}}^{\ell_{i''-1}'',\ell_{i''}'}_{i''-1,x_{i''-1}}(t-1) 
\hat{\mathcal{Z}}^{\ell_{i''+1}',\ell_{i''+2}''}_{i''+1,x_{i''+1}}(t-1) 
\cdots
\hat{\mathcal{Z}}^{\ell_{d}'',\ell_1''}_{d,x_{d}}(t-1) 
\Big]
\Bigg\}  \nonumber \\
\overset{(b)}{\approx}&~
{  \hat{\mathcal{Z}}_{1, x_1}^{\ell_1',\ell_{2}'}(t)  }  
\cdots 
{  \hat{\mathcal{Z}}_{i-1, x_{i-1}}^{\ell_{i-1}',\ell_{i}}(t)  }
{  \hat{\mathcal{Z}}_{i+1, x_{i+1}}^{\ell_{i+1},\ell_{i+2}'}(t)  }
\cdots
{  \hat{\mathcal{Z}}_{d, x_d}^{\ell_d',\ell_{1}'}(t)  }
\nonumber \\
&+
 \sum_{\substack{A\subset \{1,...,d\} \setminus \{i\} \\ A\neq\{1,...,d\} \setminus \{i\} } }
 \Bigg\{ 
 \prod_{\substack{i' \in A, \ell_i' = \ell_i  \\  \ell_{i+1}' = \ell_{i+1} }} 
  \hat{\mathcal{Z}}_{i', x_{i'}}^{\ell_{i'}',\ell_{i'+1}'}(t) 
  \prod_{ \substack{ i'' \in \{1,...,d\}\setminus (\{i\} \cup A)  \\     \ell_i' = \ell_i,  \ell_{i+1}' = \ell_{i+1}        }}
   \hat{s}_{\mathbf x}(t-1)  \nu_{i'',x_{i''}}^{\mathcal{Z},\ell_{i''}', \ell_{i''+1}'}(t)
   \nonumber \\
 & ~~~ \Big[- 
  \sum_{\ell_1'',...,\ell_{i''-1}'',\ell_{i''+2}'',...,\ell_d'' = 1 }^{r_1,...,r_{i-1},r_{i+2},...,r_d}
 \hat{\mathcal{Z}}^{\ell_1'',\ell_2''}_{1,x_1}(t)
 \cdots 
 \hat{\mathcal{Z}}^{\ell_{i''-1}'',\ell_{i''}'}_{i''-1,x_{i''-1}}(t) 
\hat{\mathcal{Z}}^{\ell_{i''+1}',\ell_{i''+2}''}_{i''+1,x_{i''+1}}(t) 
\cdots
\hat{\mathcal{Z}}^{\ell_{d}'',\ell_1''}_{d,x_{d}}(t) 
\Big]
\Bigg\},
\end{align}
where $(a)$ follows from \eqref{z-ix-mean-3} and in $(b)$ we used $ \hat{\mathcal{Z}}^{\ell_{i''}'',\ell_{i''+1}'}_{i'',x_{i''}}(t) $ in place of $ \hat{\mathcal{Z}}^{\ell_{i''}'',\ell_{i''+1}'}_{i'',x_{i''}}(t-1) $.
\end{remark}

As a result, we have
\begin{align*}
&\frac{1}{
  \nu^r_{\mathbf q^i}}
  \approx
 \sum_{{ {\mathbf{x}':x'_i = x_i}}}  
\Bigg\{    
 \Big[ 
 {\sum_{\mathbf{l}'_{\setminus i} = 1}^{\mathbf{r}_{\setminus i}}}
 {\hat{\mathcal{Z}}^{\ell'}_{\setminus i, x'}(t)}
\Big]^2
\nu_{\mathbf{x} '}^s(t)  
\Bigg\},
\end{align*}
and
\begin{align*} 
&\hat{r}_{\mathbf q^i}
~{\approx}~
\nu^r_{\mathbf q^i}  
 \sum_{{ {\mathbf{x}':x'_i = x_i}}}  
\Bigg\{  
\Bigg[ 
 {\sum_{\mathbf{l}'_{\setminus i} = 1}^{\mathbf{r}_{\setminus i}}}
 {\hat{\mathcal{Z}}^{\ell'}_{\setminus i, x'}(t)}
\Bigg]
 \hat{s}_{\mathbf{x} '}(t) 
  \Bigg\}  +
 \hat{\mathcal{Z}}^{\ell}_{i,x_i}(t) \left\{ 1 -
\frac{
 \sum\limits_{{ {\mathbf{x}':x'_i = x_i}}}     
   {\nu_{\mathbf{x} '}^s(t)}  
 {\zeta}_{x}(t)}
 {\sum\limits_{{ {\mathbf{x}':x'_i = x_i}}}  
\Bigg\{    
 \left[ 
 {\sum_{\mathbf{l}'_{\setminus i} = 1}^{\mathbf{r}_{\setminus i}}}
 {\hat{\mathcal{Z}}^{\ell'}_{\setminus i, x'}(t)}
\right]^2
\nu_{\mathbf{x} '}^s(t)  
\Bigg\} 
}
\right \}.\nonumber 
\end{align*}
Then we complete the derivation.
\section{Algorithm Summary and Complexity Analysis} \label{appendixE1}

The TeG-AMP algorithm derived is summarized in Algorithm~\ref{tab:A1}. There, we have included a maximum number of iterations $T_{\text{max}}$, and a stopping condition in Algorithm~\ref{tab:A1} based on the change in the residual and a user-defined parameter $\tau_{\text{threshold}}$. We have also written the algorithm in a general form that allows the use of complex-valued quantities, in which case $\mathcal{N}$ in \eqref{condtional-pdf-1} and \eqref{ass-pdf-1} would be circular complex Gaussian. For ease of interpretation, Algorithm~\ref{tab:A1} does not include the important damping modifications that will be detailed in Appendix~\ref{appendixE}. Suggestions for the initializations in Algorithm~\ref{tab:A1} will be given in the following. Note that TeG-AMP avoids the use of SVD or QR decompositions, lending itself to simple and potentially parallel implementations.

\begin{algorithm}[h]
	\caption{Tensor Generalized Approximate Message Passing (TeG-AMP) Algorithm.}\label{tab:A1}

\textbf{Initialization:} $p_{\mathsf{u}_{\mathbf{x} } | \mathsf{p}_{\mathbf{x} } } \left( u | \hat{p} ; \nu^{p} \right) = \frac{
p_{\mathsf{v}_{\mathbf{x} } | \mathsf{u}_{\mathbf{x} }} \left( v_{\mathbf{x} } | u \right)
\mathcal{N}(u; \hat{p}(t), \nu^p(t) )
}{\int_{u^{\prime}}  p_{\mathsf{v}_{\mathbf{x} } | \mathsf{u}_{\mathbf{x} } } \left( v_{\mathbf{x} } | u' \right)
\mathcal{N}(u'; \hat{p}(t), \nu^p(t) ) }$; 
 $p_{\mathsf{Z}_{i}^{\ell} \left(x_{i}\right) | \mathsf{r}_{\mathbf q^i}}
\left(z | \hat{r} ; \nu^{r}\right)
= 
\frac{p
\left(z \right) 
\mathcal{N}\left(z ; \hat{r}, \nu^{r}\right)}
{\int_{z^{\prime}} p
\left( z' \right) 
\mathcal{N}\left(z'; \hat{r}, \nu^{r}\right)}$; 
 $\forall i, x_i, \hat{s}_{\mathbf{x} }(0)=0$;
 $\forall i, \ell_i, x_i$, \text{choose} $\hat{\mathcal{Z}}^{\ell}_{i,x_i}(1)$, $\nu^{\mathcal{Z},\ell}_{i,x_i}(1)$. 

~~1: \textbf{for} $t=1,\cdots,T_{\text{max}}$ \textbf{do}

~~2: ~~~~~~~~$\bar{p}_{\mathbf{x} }(t) \leftarrow \sum_{\mathbf l' = 1}^{\mathbf r}  
{\hat{\mathcal{Z}}_{1, x_1}^{\ell'}(t)}  
\cdots {\hat{\mathcal{Z}}_{d, x_d}^{\ell'}(t)}$ [according to \eqref{p-x-2}]

~~3: ~~~~~~~~Update $\bar\nu^p_{\mathbf{x} }(t) $ [according to \eqref{bar-var-p-1}]

~~4: ~~~~~~~~Update $\hat{p}_{\mathbf{x}}(t)$ [according to \eqref{p-x-2}]

~~5: ~~~~~~~~Update $\nu^{p}_{\mathbf{x}}(t)$ [according to \eqref{v-p-2}]

~~6: ~~~~~~~~$\nu_{\mathbf{x} }^{u}(t) \leftarrow \mathrm{Var}\left\{\mathsf{u}_{\mathbf{x} } | \mathsf{p}_{\mathbf{x} }=\hat{p}_{\mathbf{x} }(t) ; \nu_{\mathbf{x} }^{p}(t)\right \}$ [according to \eqref{nu-u-1}]

~~7: ~~~~~~~~$\hat{u}_{\mathbf{x} }(t) \leftarrow  \mathbb{E} \left\{ \mathsf{u}_{\mathbf{x} } | \mathsf{p}_{\mathbf{x} } = \hat{p}_{\mathbf{x} }(t) ; \nu_{\mathbf{x} }^{p}(t)\right\}$ [according to \eqref{hat-u-1}]

~~8: ~~~~~~~~$\hat{s}_{\mathbf{x} }(t) \leftarrow \frac{1}{ \nu_{\mathbf{x} }^p(t) } \left(\hat{u}_{\mathbf{x} }(t)-\hat{p}_{\mathbf{x} }(t)\right)$ [according to \eqref{hat-s-2}]
 
~~9: ~~~~~~~~${\nu}_{\mathbf{x} }^{s}(t) \leftarrow \frac{1}{\nu_{\mathbf{x} }^{p}(t)}\left(1-\frac{\nu_{\mathbf{x} }^{u}(t)}{\nu_{\mathbf{x} }^{p}(t)}\right)$ [according to \eqref{nu-s-2}]

10: ~~~~~~~~Update $\nu^r_{\mathbf q^i} $ [according to \eqref{var-r-approx-1}]

11: ~~~~~~~~Update $\hat{r}_{\mathbf q^i}$ [according to \eqref{hat-r-3}]

12: ~~~~~~~~$\nu_{i,x_i}^{{\mathcal{Z}},\ell}(t+1) \leftarrow \mathrm{Var}\left\{\mathsf{Z}_{i,x_i}^{\ell} | \mathsf{r}_{\mathbf q^i} \leftarrow \hat{r}_{\mathbf q^i} ; \nu_{\mathbf q^i}^{r} \right \}$ [according to \eqref{ass-pdf-1}]

13: ~~~~~~~~$\hat{\mathcal{Z}}_{i,x_i}^{\ell}(t+1) \leftarrow \mathbb{E}\left\{\mathsf{Z}_{i,x_i}^{\ell} | \mathsf{r}_{\mathbf q^i}=\hat{r}_{\mathbf q^i} ; \nu_{\mathbf q^i}^{r} \right \}$ [according to \eqref{ass-pdf-1}]

14: ~~~~~~~~\textbf{If} $\sum_{\mathbf{x}} |\bar{p}_{\mathbf{x}}(t) - \bar{p}_{\mathbf{x}}(t-1) |^2 \leq \tau_{\text{threshold} } \sum_{\mathbf{x}} |\bar{p}_{\mathbf{x}}(t)|^2$ \textbf{then}

15: ~~~~~~~~ ~~~~~~~~\textbf{stop}
 
16: \textbf{end for} 
\end{algorithm}

The steps in Algorithm~\ref{tab:A1} can be interpreted as follows. Steps 2-3 compute a {\em plug-in} estimate $\bar{\mathcal{P}}$ of the tensor ${\mathcal{U}}$ and a corresponding set of element-wise variances $\{\bar\nu^p_{\mathbf{x}}\}$. Steps 4-5 can be seen as a correction (see  \cite{rangan2011generalized,parker2014bilinear,montanari2012graphical} for discussions in the contexts of BiG-AMP, G-AMP and
AMP, respectively) to obtain the corresponding quantities $\hat{\mathcal{P}}$ and $\{\nu^p_{\mathbf{x}}\}$. Using these quantities, steps 6-7 compute the approximate marginal posterior means $\hat{\mathcal{U}}$ and variances of $\{\nu^u_{\mathbf{x}}\}$.
Steps 8-9 then use these posterior moments to compute the scaled residual $\hat{\mathcal{S}}$ and a set of residual-variances $\{\nu^s_{\mathbf{x}}\}$. 

Steps 10-11 then use the residual terms $\hat{\mathcal{S}}$ and $\{\nu^s_{\mathbf{x}}\}$ to compute $\hat{\mathcal{R}}$ and $\{\nu^r_{\mathbf q^i}\}$, where $\hat{r}_{\mathbf q^i}$ can be interpreted as a $\{\nu^r_{\mathbf q^i}\}$-variance-AWGN corrupted observation of the true $\mathsf{Z}^{\ell_i,\ell_{i+1}}_{i}(x_i)$. Finally, steps 12-13 merge these AWGN-corrupted observations with the priors $p_{\mathsf{Z}^{\ell}_{i}(x_i)}$ to produce the posterior means $\hat{\mathcal{Z}}$ and variances $\nu^{\mathcal{Z},\ell}_{i,x_i}$.

The complexity order is dominated by \eqref{bar-var-p-1} [in steps 3], as the traverse addition and multiplication. So, the complexity of Algorithm~\ref{tab:A1} is $O(d^4 \prod_{i = 1}^{d}{r_i})$. $r_i$ is generally small for low-rank problems. Therefore, the complexity is still acceptable.
\section{Adaptive Damping} \label{appendixF}

The approximations, e.g., 1) central limit theorem; 2) Taylor series argument; and 3) ignoring some terms, made in the TeG-AMP derivation were obtained in the large system limit. In practical applications, however, these dimensions are finite, and hence the algorithm may diverge. The use of damping with G-AMP and BiG-AMP achieves convergence with arbitrary matrices~\cite{rangan2011generalized,parker2014bilinear,parker2014bilinear2}. Here, we introduce to incorporate damping into TeG-AMP. Moreover, we adapt the damping of these variables to ensure that a particular cost criterion decreases monotonically, as described in the sequel. The specific damping strategy that we adopt is similar to those described in~\cite{parker2014bilinear,schniter2014compressive}.

\subsection{Damping}

In TeG-AMP, the $t$-th iteration damping factor $\beta(t)\in(0,1]$ is used to slow the evolution of certain variables, namely $\bar\nu^p_{\mathbf x}$, $\nu^p_{\mathbf x}$, $\hat{s}_{\mathbf x}$, $\nu_{\mathbf x}^s$ and $\hat{z}^{\ell}_{i,x_i}$. To do this, steps 3, 5, 8, and 9 in Algorithm~\ref{tab:A1} are replaced with
\begin{align}
\label{damping-1}
\bar\nu^p_{\mathbf x}(t) &= (1-\beta(t)) \bar\nu^p_{\mathbf x}(t-1) \nonumber \\
&~~~~~
+ \beta(t) \sum_{\mathbf l' = 1}^{\mathbf r}  
 \sum_{\substack{A\subsetneqq \mathbb{D} \\ A\neq \emptyset} }
 \Bigg\{ 
 \Big(
 \prod_{i \in A}  \hat{\mathcal{Z}}_{i, x_{i}}^{\ell'}(t) 
 \Big)^2
 \prod_{i'' \in \mathbb{D}\setminus A}
 \nu_{i'',x_{i''}}^{\mathcal{Z},\ell'}(t)
 \Big(
 - \hat{s}_{\mathbf x}(t-1) 
  \prod_{i' \in \mathbb{D}} 
 \hat{\mathcal{Z}}^{\ell'}_{i',x_{i'}}(t)
 \Big)^{d-|A|-1}
 \Bigg\}, 
 \\
 \label{damping-2}
 \nu^p_{\mathbf{x}} (t) &= \beta(t) \left(  \bar\nu^p_{\mathbf x}(t)  +  \sum_{\substack{\mathbf l= 1 }}^{\mathbf r}  
\prod_{i \in \mathbb{D}} \nu_{i,x_{i}}^{\mathcal{Z},\ell}(t)  \right) 
+ (1- \beta(t) )  \nu^p_{\mathbf{x}} (t-1), \\
 \label{damping-3}
\hat{s}_{\mathbf x}(t) &= \beta(t)  \frac{1}{ \nu_{\mathbf x}^p(t) } \left(\hat{u}_{\mathbf x}(t)-\hat{p}_{\mathbf x}(t)\right) + (1- \beta(t) ) \hat{s}_{\mathbf x}(t-1)  , \\
 \label{damping-4}
{\nu}_{\mathbf x}^{s}(t) &=  \beta(t)  \frac{1}{\nu_{\mathbf x}^{p}(t)}\left(1-\frac{\nu_{\mathbf x}^{u}(t)}{\nu_{\mathbf x}^{p}(t)}\right) + (1- \beta(t) ) {\nu}_{\mathbf x}^{s}(t-1), 
\end{align}
and the following are inserted between step 9 and step 10 in Algorithm~\ref{tab:A1}:
\begin{align}
 \label{damping-5}
{\bar{\mathcal{Z}}^{\ell}_{i,x_i}}(t) &=  \beta(t) {\hat{\mathcal{Z}}^{\ell}_{i,x_i}}(t) + (1- \beta(t) ) {\bar{\mathcal{Z}}^{\ell}_{i,x_i}}(t-1).
\end{align}
The newly defined state variables ${\bar{\mathcal{Z}}^{\ell}_{i,x_i}}(t)$ are then used in place of ${\hat{\mathcal{Z}}^{\ell}_{i,x_i}}(t)$ in steps 10 and 11 of Algorithm~\ref{tab:A1}. When $\beta(t) = 1$, the damping has no effect, whereas when $\beta(t) = 0$, all quantities become frozen in $t$.

\subsection{Adaptive Damping}

The idea behind adaptive damping is to monitor a chosen cost criterion $J(t)$ and decrease $\beta(t)$ when the cost has not decreased sufficiently relative to $\{J(\tau)\}_{\tau = t-1-T}^{t-1}$ for some step window $T\geq0$. This mechanism allows the cost criterion to increase over short intervals of $T$ iterations and in this sense is similar to the procedure used in~\cite{wright2009sparse}. We now describe how the cost criterion $J(t)$ is constructed, building on ideas in~\cite{rangan2016fixed}. The same approach is used in the BiG-AMP in~\cite{parker2014bilinear,parker2014bilinear2}.

For fixed observations $\mathcal{V}$, the joint posterior pdf solves the KL-divergence minimization problem
\begin{align}
\label{KL-min}
p_{\mathsf{Z} | \mathsf{V}}=\arg \min _{b_{\mathsf{Z}}} D\left(b_{\mathsf{Z}} \| p_{\mathsf{Z} | \mathsf{V}}\right).
\end{align}

The factorized form \eqref{posterior-1} of the posterior allows writing
\begin{align} 
\label{KL-min-2}
D\left(b_{\mathsf{Z}} \| p_{\mathsf{Z} | \mathsf{V}}\right)-p_{\mathsf{V}}(\mathcal{V}) 
=&~ \int_{\mathsf{Z}} b_{\mathsf{Z}}(\mathcal{A}) \log \frac{b_{\mathsf{Z}}(\mathcal{Z})}{p_{\mathsf{V} | \mathsf{U}}(\mathcal{V} | \mathcal{U}) p_{\mathsf{Z}}(\mathcal{Z})} \nonumber \\
=&~ D\left(b_{\mathsf{Z}} \| p_{\mathsf{Z}}\right)-\int_{\mathsf{Z}} b_{\mathsf{Z}}(\mathcal{Z}) \log p_{\mathsf{V} | \mathsf{U}}(\mathcal{V} | \mathcal{U}).
\end{align}

Equations \eqref{KL-min} and \eqref{KL-min-2} then imply that
\begin{align} 
\label{KL-min-3}
 p_{\mathsf{Z} | \mathsf{V}} &=\arg \min _{b_{\mathsf{Z}}} J\left(b_{\mathsf{Z}}\right),  \\
 \label{KL-min-4}
  J\left(b_{\mathsf{Z}}\right) & \triangleq D\left(b_{\mathsf{Z}} \| p_{\mathsf{Z}}\right)-\mathbb{E}_{b_{\mathsf{Z}}}\left\{\log p_{\mathsf{V} | \mathsf{U} }(\mathcal{V} | \mathcal{U})\right\}.
\end{align}

To decide whether a given time-$t$ TeG-AMP approximation $b_{\mathsf{Z}}(t)$ of the joint posterior $ p_{\mathsf{Z} | \mathsf{V}} $ is preferred than the previous approximation $b_{\mathsf{Z}}(t-1)$, one could essentially plug the posterior approximation expressions \eqref{condtional-pdf-1} into \eqref{KL-min-3} and then check whether $J(b_{\mathsf{Z}}(t)) < J(b_{\mathsf{Z}}(t-1))$. But, since the expectation in \eqref{KL-min-3} is difficult to evaluate, we approximate the cost \eqref{KL-min-3} by using, in place of $\mathcal{U}$, an independent Gaussian tensor whose component means and variances are matched to those of $\mathcal{U}$. Taking the joint TeG-AMP posterior approximation $b_{\mathsf{Z}}(t)$ to be the product of the marginals from \eqref{condtional-pdf-1}, the resulting component means and variances are
\begin{align}
\mathbb{E}_{b_{\mathsf{Z}}(t)}\left\{u_{\mathbf{x}}\right\} 
&=\sum_{\mathbf l=1}^{\mathbf r} 
\mathbb{E}_{b_{\mathsf{Z}}(t)}\left\{\mathsf{Z}_{1, x_{1}}^{\ell}\right\} \cdots \mathbb{E}_{b_{\mathsf{Z}}(t)}\left\{\mathsf{Z}_{d, x_{d}}^{\ell}\right\} 
  \nonumber \\
 &=
\sum_{\mathbf l=1}^{\mathbf r} 
\hat{\mathcal{Z}}_{1, x_{1}}^{\ell}(t) \cdots \hat{\mathcal{Z}}_{d,x_{d}}^{\ell}(t) 
=
\bar{p}_{\mathbf{x}}(t),
\nonumber \\ 
\operatorname{Var}_{b_{\mathsf{Z}}(t)}\left\{u_{\mathbf{x}}\right\} 
&=\sum_{\mathbf l'=1}^{\mathbf r} 
\sum_{\substack{A \subset \mathbb{D} \\ A \neq \emptyset}}
\Bigg(\prod_{i^{\prime} \in A} \nu_{i^{\prime}, \mathbf{x} }^{\mathcal{Z}, \ell' }(t)
\prod_{i^{\prime \prime} \in \mathbb{D} \backslash A} 
\hat{\mathcal{Z}}_{i^{\prime \prime}, \mathbf{x}}^{ \ell' }(t)^{2} \Bigg) .
\end{align}

To simplify the expressions, we define
\begin{align}
\mathcal{N}_b \triangleq \mathcal{N}\left(\mathbb{E}_{b_{\mathsf{Z}}(t)}\left\{u_{\mathbf{x}} \right\} ; \mathrm{Var}_{b_{\mathsf{Z}}(t)} \left\{u_{\mathbf{x}} \right\}\right).
\end{align}
The approximate $t$-th iteration cost becomes
\begin{align} 
\label{hat-J-t-1}
\hat{J}(t)
&=\sum_{ \substack{1 \leq i \leq d \\ 1 \leq \ell_i \leq r_i \\ 1 \leq x_i \leq n_i }} 
D \Big(p_{\mathsf{Z}_{i, x_{i}}^{\ell} | r_{\mathbf q^i}}\big[\cdot | \hat{r}_{\mathbf q^i}(t) ;\nu_{\mathbf q^i}^{r}(t)\big] \| p_{\mathsf{Z}_{i, x_{i}}^{\ell}}(\cdot) \Big) 
- \sum_{\mathbf{x}} \mathbb{E}_{\mathsf{u}_{\mathbf{x}}  \sim \mathcal{N}_b} {\left\{ \log p_{\mathsf{v}_{\mathbf{x}} | \mathsf{u}_{\mathbf{x}} } \left(v_{\mathbf{x}} | u_{\mathbf{x}}\right)  \right\} } 
\nonumber \\
&\overset{(a)}{=}
\sum_{ \substack{1 \leq i \leq d \\ 1 \leq \ell_i \leq r_i \\ 1 \leq x_i \leq n_i }} 
\int_{\mathcal{Z}_{i}^{\ell}\left(x_{i}\right)} 
\mathcal{N}\left(\mathcal{Z}_{i}^{\ell} \left(x_{i}\right); 
\hat{\mathcal{Z}}_{i, x_{i}}^{\ell}(t),
\nu_{i, x_{i}}^{\mathcal{Z}, \ell}(t)\right)
\times\log
\frac{
\mathcal{N}\left(\mathcal{Z}_{i}^{\ell} \left(x_{i}\right); 
\hat{\mathcal{Z}}_{i, x_{i}}^{\ell}(t),
\nu_{i, x_{i}}^{\mathcal{Z}, \ell}(t)\right)
}{
\mathcal{N}\left(
\mathcal{Z}_{i}^{\ell} \left(x_{i}\right) ; 
\hat{\mathcal{Z}}_{\text{prior}},
\nu_{\text{prior}}^{\mathcal{Z}} 
\right)
}
\nonumber \\
&~~~~ - \sum_{\mathbf{x}} \mathbb{E}_{\mathsf{u}_{\mathbf{x}}  \sim \mathcal{N}_b} {\left\{ \log p_{\mathsf{v}_{\mathbf{x}} | \mathsf{u}_{\mathbf{x}} } \left(v_{\mathbf{x}} | u_{\mathbf{x}}\right)  \right\} } ,
\end{align}
where $(a)$ leverages the Gaussianity of the approximated posterior of $\mathsf{Z}_{i, x_{i}}^{\ell}$, $\hat{\mathcal{Z}}_{\text{prior}}$ and $\nu_{\text{prior}}^{\mathcal{Z}}$ represent the assumed prior mean and variance.

Intuitively, the first term in \eqref{hat-J-t-1} penalizes the deviation between the (TeG-AMP approximated) posterior and the assumed prior on $\mathcal{Z}$, and the second term rewards highly likely estimates $\mathcal{U}$.

\section{Tensor Simplified AMP based on CP Decomposition} \label{appendixH}

Although our TeG-AMP algorithm is applicable to CP decomposition, comparing \eqref{CPD-3} with \eqref{TR-decomposition-1} we can find that the CP decomposition is a very simple special case of TR decomposition, it is possible to simplify the TeG-AMP algorithm based on the CP decomposition model. In this subsection, we present the tensor simplified AMP (TeS-AMP) algorithm. That is, estimate $\mathbf{a}_{i}^{\ell}\left(x_{i}\right) \in \mathbb{R}$ for $i=1, \ldots, d, x_{i}=1, \ldots, n_{i}$ and $\ell=1, \ldots, r$ from the observation of some entries of $\mathcal{V}$ that is statistically coupled to $\mathcal{U}$, and is of the same dimensions as $\mathcal{U}$, with some potentially missing entries. In doing so, we consider $\mathbf{a}_{i}^{\ell}\left(x_{i}\right) \in \mathbb{R}$ for $i=1, \ldots, d, x_{i}=1, \ldots, n_{i}$ and $\ell=1, \ldots, r$ as realizations of independent random entries $\mathsf{a}_{i}^{\ell}\left(x_{i}\right)$ with known separable pdfs (or pmfs in the case of discrete models)
\begin{align} 
\label{separable-pdf-cp-1}
p_{\mathsf{a}}(\mathbf{a})=\prod_{\ell=1}^{r} \prod_{i=1}^{d} p_{\mathsf{a}_{i}^{\ell}}
\left(\mathbf{a}_{i}^{\ell}\right)
=\prod_{\ell=1}^{r} \prod_{i=1}^{d} \prod_{x_{i}=1}^{n_{i}} p_{\mathsf{a}_{i}^{\ell} 
\left(x_{i}\right)}\left(\mathbf{a}_{i}^{\ell}\left(x_{i}\right)\right)
\end{align}
and we similarly assume that the likelihood function of $\mathcal{V}$ is separable and known, i.e.,
\begin{align} 
\label{TeS-likelihood-1}
p_{\mathsf{V} | \mathsf{U} }(\mathcal{V} | \mathcal{U})
=\prod_{x_{1}=1}^{n_{1}} \ldots \prod_{x_{d}=1}^{n_{d}} 
p_{\mathsf{v}_{\mathbf{x}}  | \mathsf{u}_{\mathbf{x}} } \left(v_{\mathbf{x}} | u_{\mathbf{x}}\right) .
\end{align}

For the statistical model \eqref{separable-pdf-cp-1}-\eqref{TeS-likelihood-1}, the posterior distribution is:
\begin{align} 
& ~p_{\mathsf{a}| \mathsf{V}} (\mathbf{a} | \mathcal{V}) 
= p_{\left\{\mathsf{a}_{i}^{\ell}\right\}_{\ell,i} | \mathsf{V}} 
(\{\mathbf{a}_{i}^{\ell} \}_{\ell, i} | \mathcal{V} ) 
\nonumber \\
\overset{(a)}{=} &~
p_{\mathsf{V} | \{\mathsf{a}_{i}^{\ell} \}_{\ell, i}}
(\mathcal{V} | \{ \mathbf{a}_{i}^{\ell} \}_{\ell, i})
p_{\{\mathsf{a}_{i}^{\ell}\}_{\ell, i}}
(\{\mathbf{a}_{i}^{\ell}\}_{\ell, i}) / p_{\mathsf{V}}(\mathcal{V}) 
\propto 
p_{\mathsf{V} | \mathsf{U}}(\mathcal{V} | \mathcal{U}) 
p_{\{\mathsf{a}_{i}^{\ell}\}_{\ell, i}}
(\{\mathbf{a}_{i}^{\ell}\}_{\ell, i})  \\
=&~
\left[p_{\mathsf{V} | \mathsf{U}}(\mathcal{V} | \mathcal{U})\right]
\left[\prod_{\ell=1}^{r} \prod_{i=1}^{d} p_{\mathsf{a}_{i}^{\ell}}
\left(\mathbf{a}_{i}^{\ell}\right)\right] 
\nonumber \\
=&~
\left[\prod_{x_{1}=1}^{n_{1}} \ldots \prod_{x_{d}=1}^{n_{d}}
p_{\mathsf{v}_{\mathbf{x}} | \mathsf{u}_{\mathbf{x}} }
\left(v_{\mathbf{x}} | 
\sum_{\ell=1}^{r} 
\mathbf{a}_{1}^{\ell}\left(x_{1}\right) 
\mathbf{a}_{2}^{\ell}\left(x_{2}\right)
\ldots 
\mathbf{a}_{d}^{\ell}\left(x_{d}\right)
\right)
\right] 
\times 
\left[\prod_{\ell=1}^{r} \prod_{i=1}^{d} \prod_{x_{i}=1}^{n_{i}} 
p_{\mathsf{a}_{i}^{\ell}\left(x_{i}\right)}
\left(\mathbf{a}_{i}^{\ell}\left(x_{i}\right)\right)
\right],
\end{align}
where $(a)$ follows from Bayes' rule.

In this section, we present TeS-AMP to compute the MMSE estimates of $\mathbf{a}=\{\mathbf{a}_{i}^{\ell}\}_{\ell, i}$, i.e., the means of the marginal posteriors $p_{\mathsf{a}_{i}^{\ell}\left(x_{i}\right) | \mathsf{V}}(\cdot | \mathcal{V})$, for all of the triplets $\{\ell,i,x_i\}$.

\subsection{TeS-AMP Loopy Belief Propagation and SPA}

\begin{figure}[!tb]
	\centering
		
	\subfloat{\includegraphics[width=6.5in]{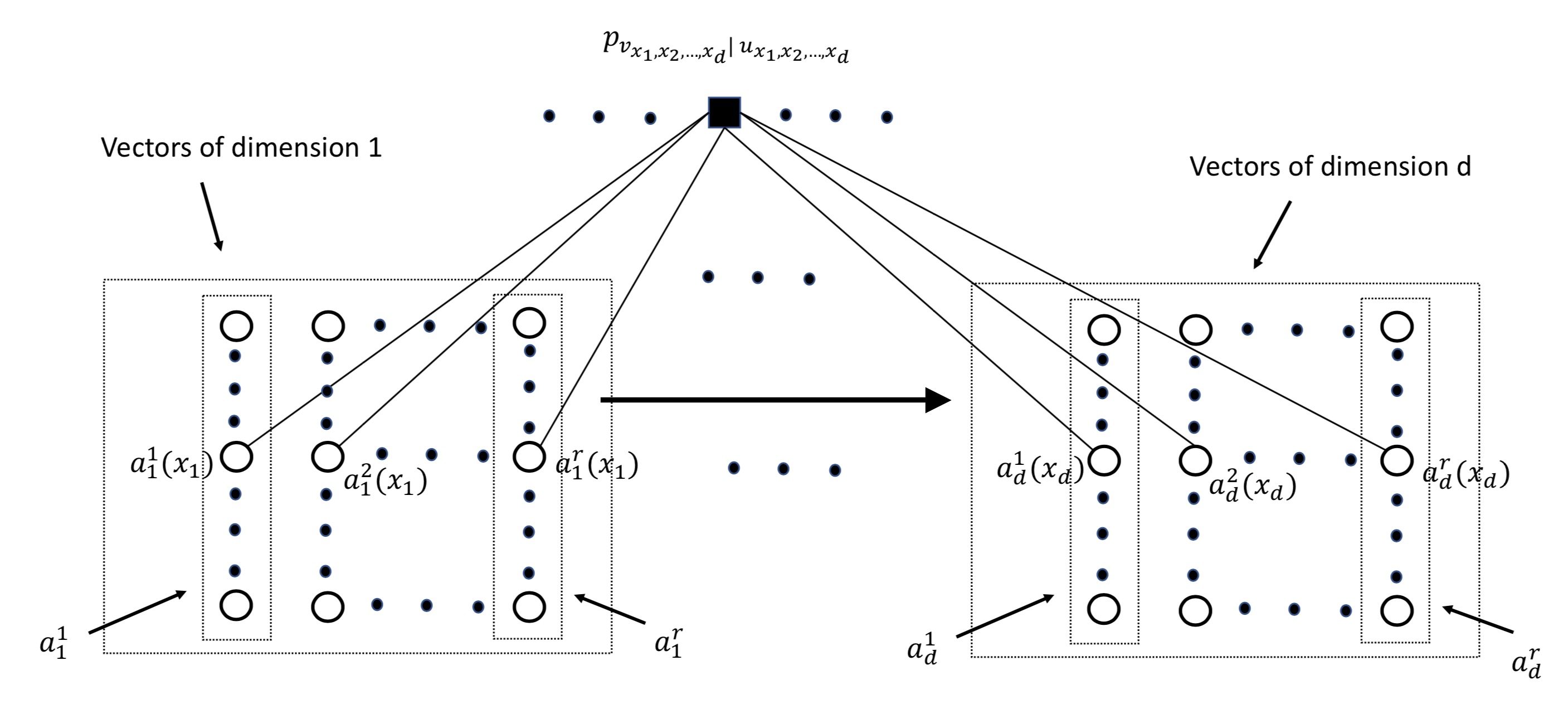}}
	
	\caption{An illustration of the factor-graph for tensor inference problem based on CP decomposition.}
	\label{figure:TeS-AMP}
\end{figure}

Similar with TeG-AMP, the TeS-AMP incorporates approximations to the SPA on the factor-graph in Figure~\ref{figure:TeS-AMP}. The messages that will be used in TeS-AMP are specified in Table II.

\begin{table}[!htb]
\caption{TeS-AMP SPA message definitions at iteration $t$}
\label{tab:runtime}
\centering
\begin{tabular}{c|c}  
\toprule
${{p}}_{\mathbf x \rightarrow [\ell,i,x_i] } (t,\cdot)$  & SPA message from node $\mathsf a_i^{\ell}(x_i)$ to node $p_{\mathsf v_{\mathbf x} | \mathsf u_{\mathbf x} }$  \\
${{p}}_{\mathbf x \leftarrow [\ell,i,x_i] } (t,\cdot)$    & SPA message from node $p_{\mathsf v_{\mathbf x} | \mathsf u_{\mathbf x} }$  to node  $\mathsf a_i^{\ell}(x_i)$  \\
${{p}}_{ [\ell,i,x_i] } (t,\cdot)$     & SPA-approximated log posterior pdf of message $\mathsf a_i^{\ell}(x_i)$ \\
\bottomrule
\end{tabular}
\end{table}

Similar to TeG-AMP, by applying the SPA to the factor-graph of the problem, we derive the following update rules for the messages in Table II.
\begin{align} 
\label{TeS-AMP-sum-pro-1}
&~~{{p}}_{\mathbf x \rightarrow [\ell,i,x_i] }\left( t, \mathbf{a}_{i}^{\ell}(x_{i}) \right) \nonumber \\
=&~ \log \left( \int_{\left\{\mathbf{a}_{i}^{\ell^{\prime}}(x_{i^{\prime}})\right\}_{i^{\prime}, \ell^{\prime}} \backslash \mathbf{a}_{i}^{\ell}\left(x_{i}\right)}
p_{\mathsf{v}_{\mathbf x}|\mathsf{u}_{\mathbf x}}  \left(  v_{\mathbf x}| 
\sum_{\ell=1}^{r} \mathbf{a}_{1}^{\ell}(x_{1}) \mathbf{a}_{2}^{\ell}(x_{2}) \ldots \mathbf{a}_{d}^{\ell}(x_{d})  
\right)   \right. \nonumber \\
&~\left.  \times  \prod_{\left(\ell^{\prime}, i^{\prime}\right) \neq(\ell, i)}  \exp\left\{ {{p}}_{\mathbf x \leftarrow [\ell',i',x_{i'}] }
\left(t, \mathbf{a}_{i^{\prime}}^{\ell^{\prime}}(x_{i^{\prime}}) \right) \right\}     \right) + \text{const},
\end{align}
and
\begin{align} 
\label{TeS-AMP-sum-pro-2}
{{p}}_{\mathbf x \leftarrow [\ell,i,x_i] }\left( t+1,\mathbf{a}_{i}^{\ell}(x_{i}) \right) &= \log p_{\mathsf{a}_{i}^{\ell}(x_i)}\left(\mathbf{a}_{i}^{\ell}(x_{i})\right)  + \sum_{\mathbf{x}':x'_i=x_i,~ \mathbf x' \neq \mathbf x}   {{p}}_{\mathbf x_i' \rightarrow [\ell,i, x_i] } (t, \mathbf{a}_{i}^{\ell}(x_{i}) ) + \text{const}.
\end{align}
In the following, we indicate the mean and variance of the pdf 
$\frac{1}{C} \exp {{p}}_{\mathbf{x} \leftarrow [\ell,i,x_i] } (t,\cdot)$ by $ \hat{\mathbf{a}}_{i, \mathbf{x}}^{\ell}(t)$ and $\nu_{i, \mathbf{x}}^{\mathbf{a}, \ell}(t)$, respectively. For the log-posterior, the SPA implies~\cite{kschischang2001factor}:
\begin{align} 
\label{TeS-AMP-sum-pro-3}
{{p}}_{[\ell,i,x_i]}\left( t+1,\mathbf{a}_{i}^{\ell}(x_{i}) \right) &= \left[ \log p_{\mathsf{a}_{i}^{\ell}(x_i)}\left(\mathbf{a}_{i}^{\ell}(x_{i})\right)\right] + \sum_{\mathbf{x}':x'_i=x_i}   {{p}}_{\mathbf x_i' \rightarrow [\ell,i, x_i] } (t, \mathbf{a}_{i}^{\ell}(x_{i}) ) + \text{const},
\end{align}
and we indicate the mean and variance of $\frac{1}{C} \exp {{p}}_{ [\ell,i,x_i] } (t,\cdot)$ by $ \hat{\mathbf{a}}_{i, x_i}^{\ell_i}(t)$ and $\nu_{i, x_i}^{\mathbf a, \ell}(t)$, respectively. Next, we use AMP approximations for the SPA updates \eqref{TeS-AMP-sum-pro-1}-\eqref{TeS-AMP-sum-pro-3}.

\subsection{Approximated Factor-to-Variable Messages}

We start by approximating the message ${{p}}_{\mathbf{x} \rightarrow[\ell, i, x_{i}] }(t, \mathbf{a}_{i}^{\ell}(x_{i}))$. Defining the random variable $\tilde{\mathsf{a}}^{\ell}_{i,\mathbf x} \triangleq {\mathsf{a}}_{i}^{\ell}(x_i)  -  \hat{\mathbf{a}}_{i, \mathbf{x}}^{\ell}(t)$ with zero mean where ${\mathsf{a}}_{i}^{\ell}(x_i) \backsim \frac{1}{C} \exp( {{p}}_{\mathbf x \leftarrow [\ell,i,x_i] }(t,\cdot ) )$, we can write
\begin{align} 
\label{TeS-u-x-1}
\mathsf{u}_{\mathbf x} &= 
\sum_{\ell' = 1}^{r}  \mathsf{a}_{1}^{\ell'}(x_1)  \mathsf{a}_{2}^{\ell'}(x_2) \cdots  \mathsf{a}_{d}^{\ell'}(x_d)  \nonumber \\
& = \sum_{\substack{\ell^{\prime}=1 \\ \ell' \neq \ell}}^{r}
\left(\hat{\mathbf{a}}_{1,\mathbf{x}}^{\ell^{\prime}}(t)
+\tilde{\mathsf{a}}_{1,\mathbf{x}}^{\ell^{\prime}}\right) 
\ldots
\left(\hat{\mathbf{a}}_{d,\mathbf{x}}^{\ell'}(t)
+\tilde{\mathsf{a}}_{d,\mathbf{x}}^{\ell^{\prime}}\right)
\nonumber \\
&~~~+ 
 { \left( \hat{\mathbf{a}}_{1, \mathbf{x}}^{\ell}(t)  +  \tilde{\mathsf{a}}^{\ell}_{1,\mathbf x}  \right) }
\cdots { \left( \hat{\mathbf{a}}_{i-1, \mathbf{x}}^{\ell}(t)  +  \tilde{\mathsf{a}}^{\ell}_{i-1,\mathbf x}  \right) }  
{\mathsf{a}}_{i}^{\ell}(x_i) { \left( \hat{\mathbf{a}}_{i+1, \mathbf{x}}^{\ell}(t)  +  \tilde{\mathsf{a}}^{\ell}_{i+1,\mathbf x}  \right) }
\cdots { \left( \hat{\mathbf{a}}_{d, \mathbf{x}}^{\ell}(t)  +  \tilde{\mathsf{a}}^{\ell}_{d,\mathbf x}  \right) }  .
\end{align}

Define $\hat{q}_{\mathbf x}(t)$ and $\hat{q}_{\mathbf x}^{\ell}(t)$ as below that will be used in our derivations.
\begin{align} 
\label{TeS-q-x-1}
\hat{q}_{\mathbf x}(t) & \triangleq  \sum_{\ell' = 1}^{r}  
{  \hat{\mathbf{a}}_{1, \mathbf{x}}^{\ell'}(t)  }  
\cdots  {  \hat{\mathbf{a}}_{d, \mathbf{x}}^{\ell'}(t)  }  , \\
\label{TeS-q-xl-1}
\hat{q}_{\mathbf x}^{\ell}(t) & \triangleq  \sum_{\substack{\ell' = 1 \\ 
\ell' \neq \ell 
} }^{r}  
{  \hat{\mathbf{a}}_{1, \mathbf{x}}^{\ell'}(t)  }  
\cdots  { \hat{\mathbf{a}}_{d, \mathbf{x}}^{\ell'}(t)   }  ,
\end{align}
after which it is possible to see that
\begin{align} 
\label{TeS-E-u-1}
&~ \mathbb{E}\{ \mathsf{u}_{\mathbf x} | {\mathsf{a}}_{i}^{\ell}(x_i)  =  {\mathbf{a}}_{i}^{\ell}(x_i)  \} \nonumber \\
=&~  \hat{q}_{\mathbf x}^{\ell}(t) 
 + 
\hat{\mathbf{a}}_{1,\mathbf{x}}^{\ell}(t) \ldots \hat{\mathbf{a}}_{i-1,\mathbf{x}}^{\ell}(t) 
\mathbf{a}_{i}^{\ell}\left(x_{i}\right) 
\hat{\mathbf{a}}_{i+1,\mathbf{x}}^{\ell}(t) \ldots \hat{\mathbf{a}}_{d,\mathbf{x}}^{\ell}(t)
   \nonumber \\
\overset{(a)}{\approx}&~  \hat{q}_{\mathbf x}(t) 
 + 
\hat{\mathbf{a}}_{1,\mathbf{x}}^{\ell}(t) \ldots \hat{\mathbf{a}}_{i-1,\mathbf{x}}^{\ell}(t) 
\left[ \mathbf{a}_{i}^{\ell}\left(x_{i}\right) - \hat{\mathbf{a}}_{i,x_{i}}^{\ell}(t) \right]
\hat{\mathbf{a}}_{i+1,\mathbf{x}}^{\ell}(t) \ldots \hat{\mathbf{a}}_{d,\mathbf{x}}^{\ell}(t) ,
\end{align}
where $(a)$ follows from similar approximations used in \eqref{E-u-1}. Since we assumed that data size is large and ${{p}}_{\mathbf x \leftarrow [\ell,i,x_i] } (t,\cdot)$ and ${{p}}_{ [\ell,i,x_i] } (t,\cdot)$ differ by only one term (which vanishes relative to the others in the large-system limit). 

Also, define ${\nu}^{q}_{\mathbf x}(t)$ and ${\nu}^{q,\ell}_{\mathbf x}(t)$ as below that will be used in our derivations
\begin{align} 
\label{TeS-v-p-1}
{\nu}^{q}_{\mathbf x}(t) &  \triangleq  \sum_{\substack{\ell' = 1 }}^{r}  
\sum_{ \substack {A \subset \{1,...,d\} \\ A \neq \emptyset} }
\left( \prod_{i' \in A} \nu_{i',\mathbf x}^{\mathbf a,\ell'}(t)    \prod_{i'' \in \{1,...,d\} \setminus A }  {\hat{\mathbf a}_{i'',\mathbf x}^{\ell'} (t) }^2 \right) , \\
\label{TeS-v-pl-1}
{\nu}^{q,\ell}_{\mathbf x}(t) & \triangleq  \sum_{\substack{\ell' = 1 \\ 
\ell' \neq \ell 
} }^{r}
\sum_{ \substack {A \subset \{1,...,d\} \\ A \neq \emptyset} }
\left( \prod_{i' \in A} \nu_{i',\mathbf x}^{\mathbf a,\ell'}(t)    \prod_{i'' \in \{1,...,d\} \setminus A }  {\hat{\mathbf a}_{i'',\mathbf x}^{\ell'} (t) }^2 \right)  ,
\end{align}
after which it is possible to see that
\begin{align} 
\label{TeS-Var-u-1}
& \text{Var}\{ \mathsf{u}_{\mathbf x} | {\mathsf{a}}_{i}^{\ell}(x_i)  =  {\mathbf{a}}_{i}^{\ell}(x_i)  \} \nonumber \\
=& ~ {\nu}^{q,\ell}_{\mathbf x}(t)  + 
\sum_{ \substack {A \subset \{1,...,d\} \setminus \{i\} \\ A \neq \emptyset} }  
\Bigg[
 \prod_{ \substack{i' \in A 
}}
 \nu_{i',\mathbf x}^{\mathbf a,\ell}(t) 
    \prod_{ \substack{
    i'' \in \{1,...,d\} \setminus (\{ i \} \cup A) 
    }} 
 {\hat{\mathbf a}_{i'',\mathbf x}^{\ell} (t) }^2 
 \Bigg]
  { \mathbf a_i (x_i) }^2  \nonumber \\
\overset{(a)}{\approx}& ~ {\nu}^{q,\ell}_{\mathbf x}(t)  + 
\sum_{ \substack {A \subset \{1,...,d\} \setminus \{i\} \\ A \neq \emptyset} }  
\Bigg[
 \prod_{ \substack{i' \in A 
}}
 \nu_{i',\mathbf x}^{\mathbf a,\ell}(t) 
    \prod_{ \substack{
    i'' \in \{1,...,d\} \setminus (\{ i \} \cup A) 
    }} 
 {\hat{\mathbf a}_{i'',\mathbf x}^{\ell} (t) }^2 
 \Bigg]
 \left[ { \mathbf a_i^{\ell} (x_i) }^2 - { \hat{\mathbf{a}}_{i,x_i}^{\ell} (t) }^2 \right],
\end{align}
where $(a)$ follows from similar approximations used in \eqref{Var-u-1}.

With this conditional-Gaussian approximation, \eqref{TeS-AMP-sum-pro-1} becomes
\begin{align} 
\label{TeS-AMP-sum-pro-1-2}
&~{{p}}_{\mathbf x \rightarrow [\ell,i,x_i] }(t, \mathbf{a}_{i}^{\ell}(x_i))  \nonumber \\
=&~ \log \int_{u_{\mathbf x}} p_{\mathsf v_{\mathbf x} | \mathsf{u}_{\mathbf x}}\left(v_{\mathbf x} | u_{\mathbf x} \right) \times
 \mathcal{N}\left(\mathsf u_{\mathbf x} ; \mathbb{E}\left\{\mathsf{u}_{\mathbf x} |  {\mathsf{a}}_{i}^{\ell}(x_i)  =  {\mathbf{a}}_{i}^{\ell}(x_i)  \right\}, \operatorname{Var}\left\{\mathsf{u}_{\mathbf x} | {\mathsf{a}}_{i}^{\ell}(x_i)  =  {\mathbf{a}}_{i}^{\ell}(x_i)   \right\}\right) + \text{const} \nonumber \\
 =&~ H_{\mathbf x} \left( \mathbb{E}\left\{\mathsf{u}_{\mathbf x} |  {\mathsf{a}}_{i}^{\ell}(x_i)  =  {\mathbf{a}}_{i}^{\ell}(x_i)  \right\}, \operatorname{Var}\left\{\mathsf{u}_{\mathbf x} | {\mathsf{a}}_{i}^{\ell}(x_i)  =  {\mathbf{a}}_{i}^{\ell}(x_i)   \right\}    ; v_{\mathbf x}  \right) + \text{const},
\end{align}
in terms of the function 
\begin{align} 
\label{H-1_2}
H_{\mathbf x} \left(\hat{q}, \nu^{q} ; v\right)  \triangleq  \log \int_{u} p_{\mathsf{v}_{\mathbf x} | \mathsf{u}_{\mathbf x}}(v | u) \mathcal{N}\left(u ; \hat{q}, \nu^{q}\right).
\end{align}

Rewriting \eqref{TeS-AMP-sum-pro-1-2} using a Taylor series expansion in ${ \mathbf a_i^{\ell} (x_i) }$ about the point ${ \hat{\mathbf{a}}_{i,x_i}^{\ell} (t) }$, using \eqref{TeS-E-u-1} and \eqref{TeS-Var-u-1} we obtain
\begin{align} 
\label{TeS-sum-pro-Taylor-1}
&~{{p}}_{\mathbf x \rightarrow [\ell,i,x_i] }(t, \mathbf{a}_{i}^{\ell}(x_i)) \nonumber \\
\approx &~  \text{const} + H_{\mathbf x}( \hat{q}_{\mathbf x} (t) , \nu^q_{\mathbf x} (t) ; v_{\mathbf x} )  \nonumber \\
&+  \left[
\hat{\mathbf{a}}_{1,\mathbf{x}}^{\ell}(t) \ldots \hat{\mathbf{a}}_{i-1,\mathbf{x}}^{\ell}(t) 
\left[ \mathbf{a}_{i}^{\ell}\left(x_{i}\right) - \hat{\mathbf{a}}_{i,x_{i}}^{\ell}(t) \right]
\hat{\mathbf{a}}_{i+1,\mathbf{x}}^{\ell}(t) \ldots \hat{\mathbf{a}}_{d,\mathbf{x}}^{\ell}(t)
\right]
\times H_{\mathbf x}'( \hat{q}_{\mathbf x} (t) , \nu^q_{\mathbf x} (t) ; v_{\mathbf x} ) 
 \nonumber \\
 &+ \frac{1}{2} \left[ 
\hat{\mathbf{a}}_{1,\mathbf{x}}^{\ell}(t) \ldots \hat{\mathbf{a}}_{i-1,\mathbf{x}}^{\ell}(t) 
\left[ \mathbf{a}_{i}^{\ell}\left(x_{i}\right) - \hat{\mathbf{a}}_{i,x_{i}}^{\ell}(t) \right]
\hat{\mathbf{a}}_{i+1,\mathbf{x}}^{\ell}(t) \ldots \hat{\mathbf{a}}_{d,\mathbf{x}}^{\ell}(t)
\right]^2
\times H_{\mathbf x}''( \hat{q}_{\mathbf x} (t) , \nu^q_{\mathbf x} (t) ; v_{\mathbf x} ) \nonumber \\
& +  \dot H_{\mathbf x}( \hat{p}_{\mathbf x} (t) , \nu^p_{\mathbf x} (t) ; v_{\mathbf x} )    
\sum_{ \substack {A \subset \{1,...,d\} \setminus \{i\} \\ A \neq \emptyset} }  
\Bigg[
 \prod_{ \substack{i' \in A 
}}
 \nu_{i',\mathbf x}^{\mathbf a,\ell}(t) 
    \prod_{ \substack{
    i'' \in \{1,...,d\} \setminus (\{ i \} \cup A) 
    }} 
 {\hat{\mathbf a}_{i'',\mathbf x}^{\ell} (t) }^2 
 \Bigg]
 \left[ { \mathbf a_i^{\ell} (x_i) }^2 - { \hat{\mathbf{a}}_{i,x_i}^{\ell} (t) }^2 \right]   \nonumber \\     
\overset{(a)}{\approx}&~  \text{const}  +  \left[
\hat{\mathbf{a}}_{1,\mathbf{x}}^{\ell}(t) \ldots \hat{\mathbf{a}}_{i-1,\mathbf{x}}^{\ell}(t) 
\hat{\mathbf{a}}_{i+1,\mathbf{x}}^{\ell}(t) \ldots \hat{\mathbf{a}}_{d,\mathbf{x}}^{\ell}(t)
\right]
 \mathbf{a}_{i}^{\ell}\left(x_{i}\right) 
\hat{s}_{q,\mathbf x}(t)
 \nonumber \\
 &+ \frac{1}{2} \left[ 
\hat{\mathbf{a}}_{1,\mathbf{x}}^{\ell}(t) \ldots \hat{\mathbf{a}}_{i-1,\mathbf{x}}^{\ell}(t) 
\hat{\mathbf{a}}_{i+1,\mathbf{x}}^{\ell}(t) \ldots \hat{\mathbf{a}}_{d,\mathbf{x}}^{\ell}(t)
\right]^2
\left[ \mathbf{a}_{i}^{\ell}\left(x_{i}\right)^2 - 2\hat{\mathbf{a}}_{i,x_{i}}^{\ell}(t)  \mathbf{a}_{i}^{\ell}\left(x_{i}\right) \right]
(-\nu_{q,\mathbf x}^s(t)) \nonumber \\
& +   \frac{1}{2} \left [  {\hat{s}_{q,\mathbf x}(t)}^2  -  {\nu_{q,\mathbf x}^s(t)} \right]    
 { \mathbf a_i^{\ell} (x_i) }^2  
\sum_{ \substack {A \subset \{1,...,d\} \setminus \{i\} \\ A \neq \emptyset} }  
\Bigg(
 \prod_{ \substack{i' \in A 
}}
 \nu_{i',\mathbf x}^{\mathbf a,\ell}(t) 
    \prod_{ \substack{
    i'' \in \{1,...,d\} \setminus (\{ i \} \cup A) 
    }} 
 {\hat{\mathbf a}_{i'',\mathbf x}^{\ell} (t) }^2 
 \Bigg),
\end{align}
where (a) follows from similar approximations used in \eqref{sum-pro-Taylor-1-A} and below definitions
\begin{align} 
\label{TeS-hat-s-1}
\hat{s}_{q,\mathbf x}(t) &\triangleq  H_{\mathbf x}'( \hat{q}_{\mathbf x} (t) , \nu^q_{\mathbf x} (t) ; v_{\mathbf x} ) \\
\label{TeS-nu-s-1}
\nu_{q,\mathbf x}^s(t)  & \triangleq  - H_{\mathbf x}''( \hat{q}_{\mathbf x} (t) , \nu^q_{\mathbf x} (t) ; v_{\mathbf x} ).
\end{align}

Similarly, given \eqref{TeS-hat-s-1}-\eqref{TeS-nu-s-1} and \eqref{H-1_2} the followings hold
\begin{align} 
\label{TeS-hat-s-2}
\hat{s}_{q,\mathbf x}(t) &= \frac{1}{ \nu_{\mathbf x}^q(t) } \left(\hat{u}_{\mathbf x}(t)-\hat{q}_{\mathbf x}(t)\right), \\
\label{TeS-nu-s-2}
{\nu}_{q,\mathbf x}^{s}(t) &= \frac{1}{\nu_{\mathbf x}^{q}(t)}\left(1-\frac{\nu_{\mathbf x}^{u}(t)}{\nu_{\mathbf x}^{q}(t)}\right),
\end{align}
for the conditional mean and variance
\begin{align} 
\label{TeS-hat-u-1}
\hat{u}_{q,\mathbf x}(t)  &\triangleq \mathbb{E} \left\{ \mathsf{u}_{\mathbf x} | \mathsf{q}_{\mathbf x} = \hat{q}_{\mathbf x}(t) ; \nu_{\mathbf x}^{q}(t)\right\}, \\
\label{TeS-nu-u-1}
 \nu_{q,\mathbf x}^{u}(t) & \triangleq \mathrm{Var}\left\{\mathsf{u}_{\mathbf x} | \mathsf{q}_{\mathbf x}=\hat{q}_{\mathbf x}(t) ; \nu_{\mathbf x}^{q}(t)\right \}, 
 \end{align}
computed according to the conditional pdf
\begin{align} 
\label{TeS-condtional-pdf-1}
p_{\mathsf{u}_{\mathbf x} | \mathsf{q}_{\mathbf x} } \left( u | \hat{q} ; \nu^{q} \right) \triangleq   \frac{
p_{\mathsf{v}_{\mathbf x} | \mathsf{u}_{\mathbf x} } \left( v_{\mathbf x} | u_{\mathbf x} \right)
\mathcal{N}( u_{\mathbf x} ;  \hat{q}_{\mathbf x}(t) , \nu_{\mathbf x}^q(t) )
}{\int_{u^{\prime}}  p_{\mathsf{v}_{\mathbf x} | \mathsf{u}_{\mathbf x} } \left( v_{\mathbf x} | u' \right)
\mathcal{N}( u' ;  \hat{q}_{\mathbf x}(t) , \nu_{\mathbf x}^q(t) )   }.
 \end{align}

The same, \eqref{TeS-condtional-pdf-1} is TeS-AMP's the $t$-th iteration approximation to the true marginal posterior $p_{\mathsf{u}_{\mathbf x} | \mathsf V }(\cdot | \mathcal{V})$. We note that \eqref{TeS-condtional-pdf-1} can also be interpreted as the posterior pdf for $\mathsf{u}_{\mathbf x}$ given the likelihood $p_{\mathsf{v}_{\mathbf x} | \mathsf{u}_{\mathbf x} } \left( v_{\mathbf x} | \cdot \right)$ from \eqref{TeS-likelihood-1} and the prior $\mathsf{u}_{\mathbf x} \sim \mathcal{N}( \hat{q}_{\mathbf x}(t) , \nu_{\mathbf x}^q(t) )$ that is implicitly assumed by the $t$-th iteration TeS-AMP.

\subsection{Approximated Variable-to-Factor Messages}

We now turn to approximating the messages flowing from the variable nodes to the factor nodes. Starting with \eqref{TeS-AMP-sum-pro-2} and plugging in \eqref{TeS-sum-pro-Taylor-1} we obtain
\begin{align} 
\label{TeS-var-to-fac-1}
&~ {{p}}_{\mathbf x \leftarrow [\ell,i,x_i] }(t+1,\mathbf{a}_{i}^{\ell}(x_i))  \nonumber \\
=&~ \log p_{\mathsf{a}_{i}^{\ell}(x_i)}\left(\mathbf{a}_{i}^{\ell}(x_i)\right)  + \sum_{\mathbf{x}':x'_i=x_i,~ \mathbf x' \neq \mathbf x}   {{p}}_{\mathbf x_i' \rightarrow [\ell,i,x_i] } (t, \mathbf{a}_{i}^{\ell}(x_i) ) + \text{const}  \nonumber \\
 \overset{(a)}{\approx} &~ \text{const}  + 
 \log p_{\mathsf{a}_{i}^{\ell}(x_i)}\left(\mathbf{a}_{i}^{\ell}(x_i)\right)  
 +  \sum_{\mathbf{x}':x'_i=x_i,~ \mathbf x' \neq \mathbf x}  
\Bigg\{    
\left[
\hat{\mathbf{a}}_{1,\mathbf{x}}^{\ell}(t) \ldots \hat{\mathbf{a}}_{i-1,\mathbf{x}}^{\ell}(t) 
\hat{\mathbf{a}}_{i+1,\mathbf{x}}^{\ell}(t) \ldots \hat{\mathbf{a}}_{d,\mathbf{x}}^{\ell}(t)
\right]
 \mathbf{a}_{i}^{\ell}\left(x_{i}\right) 
\hat{s}_{q,\mathbf x}(t)
 \nonumber \\
 &+ \frac{1}{2} \left[ 
\hat{\mathbf{a}}_{1,\mathbf{x}}^{\ell}(t) \ldots \hat{\mathbf{a}}_{i-1,\mathbf{x}}^{\ell}(t) 
\hat{\mathbf{a}}_{i+1,\mathbf{x}}^{\ell}(t) \ldots \hat{\mathbf{a}}_{d,\mathbf{x}}^{\ell}(t)
\right]^2
\left[ \mathbf{a}_{i}^{\ell}\left(x_{i}\right)^2 - 2\hat{\mathbf{a}}_{i,x_{i}}^{\ell}(t)  \mathbf{a}_{i}^{\ell}\left(x_{i}\right) \right]
(-\nu_{q,\mathbf x}^s(t)) \nonumber \\
& +   \frac{1}{2} \left [  {\hat{s}_{q,\mathbf x}(t)}^2  -  {\nu_{q,\mathbf x}^s(t)} \right]    
 { \mathbf a_i^{\ell} (x_i) }^2  
\sum_{ \substack {A \subset \{1,...,d\} \setminus \{i\} \\ A \neq \emptyset} }  
\Bigg(
 \prod_{ \substack{i' \in A 
}}
 \nu_{i',\mathbf x}^{\mathbf a,\ell}(t) 
    \prod_{ \substack{
    i'' \in \{1,...,d\} \setminus (\{ i \} \cup A) 
    }} 
 {\hat{\mathbf a}_{i'',\mathbf x}^{\ell} (t) }^2 
 \Bigg)
\Bigg\} \nonumber \\
 \overset{(b)}{=}&~  \text{const}  + \log p_{\mathsf{a}_{i}^{\ell}(x_i)}\left(\mathbf{a}_{i}^{\ell}(x_i)\right)  
 -  \frac{1}{2\nu^r_{\mathbf x,[\ell,i,x_i]}} \left( { \mathbf a_i^{\ell} (x_i) }   -   \hat{r}_{\mathbf x, [\ell,i,x_i]} \right)^2  \nonumber \\
  =&~  \text{const}  + \log \left(  p_{\mathsf{a}_{i}^{\ell}(x_i)}(\mathbf{a}_{i}^{\ell}(x_i))  
  \mathcal{N}\left( { \mathbf a_i^{\ell} (x_i) };  \hat{r}_{\mathbf x, [\ell,i,x_i]}  ,  \nu^r_{\mathbf x,[\ell,i,x_i]}  \right)   \right),
\end{align}
where $(a)$ follows from \eqref{TeS-sum-pro-Taylor-1}, $(b)$ follows from similar derivations used in \eqref{var-to-fac-1-A} and the following definitions:
\begin{align} 
\label{TeS-1-var-r-1}
\frac{1}{\nu^r_{\mathbf x,[\ell,i,x_i]}}  
&\triangleq   \sum_{\mathbf{x}':x'_i=x_i,~ \mathbf x' \neq \mathbf x}  
\Bigg\{    
 \left[ 
\hat{\mathbf{a}}^{\ell}_{1, x_1'}(t)
 \cdots \hat{\mathbf{a}}^{\ell}_{i-1, x_{i-1}'}(t) 
\hat{\mathbf{a}}^{\ell}_{i+1,x_{i+1}'}(t) \cdots
\hat{\mathbf{a}}^{\ell}_{d, x_d'}(t)
\right]^2
  \nu_{q,\mathbf x'}^s(t)     \nonumber \\
&~~~~~ -  
  \left [  {\hat{s}_{q,\mathbf x'}(t)}^2  -  {\nu_{q,\mathbf x'}^s(t)} \right]    
\sum_{ \substack {A \subset \{1,...,d\} \setminus \{i\} \\ A \neq \emptyset} } 
\Bigg[
 \prod_{ \substack{i' \in A 
}}
 \nu_{i', x_{i'}'}^{\mathbf a,\ell}(t) 
    \prod_{ \substack{
    i'' \in \{1,...,d\} \setminus (\{ i \} \cup A)
    }} 
 {\hat{\mathbf a}_{i'', x_{i''}'}^{\ell} (t) }^2 
 \Bigg]
  \Bigg\}  ,
\end{align}
and
\begin{align} 
\label{TeS-hat-r-1}
\hat{r}_{\mathbf x, [\ell,i,x_i]}  =&~ \nu^r_{\mathbf x,[\ell,i,x_i]}  
 \sum_{\mathbf{x}':x'_i=x_i,~ \mathbf x' \neq \mathbf x}  
\Bigg\{  
\left[
\hat{\mathbf{a}}^{\ell}_{1,\mathbf x'}(t)
 \cdots \hat{\mathbf{a}}^{\ell}_{i-1,\mathbf x'}(t) 
\hat{\mathbf{a}}^{\ell}_{i+1,\mathbf x'}(t) \cdots
\hat{\mathbf{a}}^{\ell}_{d,\mathbf x'}(t)
\right]
 \hat{s}_{q,\mathbf x'}(t) 
  \Bigg\}  
  \nonumber \\
 &+ \hat{\mathbf{a}}^{\ell}_{i,x_i}(t) 
 \Bigg\{ 1 + \nu^r_{\mathbf x,[\ell,i,x_i]}   
 \sum_{\mathbf{x}':x'_i=x_i,~ \mathbf x' \neq \mathbf x}     
   \left [  {\hat{s}_{q,\mathbf x'}(t)}^2  -  {\nu_{q,\mathbf x'}^s(t)} \right]  \times
   \nonumber \\
 &~~~~~~~~~~~~~~
\sum_{ \substack {A \subset \{1,...,d\} \setminus \{i\} \\ A \neq \emptyset} } 
\Bigg[
 \prod_{ \substack{i' \in A 
}}
 \nu_{i', x_{i'}'}^{\mathbf a,\ell}(t) 
    \prod_{ \substack{
    i'' \in \{1,...,d\} \setminus (\{ i \} \cup A)
    }} 
 {\hat{\mathbf a}_{i'', x_{i''}'}^{\ell} (t) }^2 
 \Bigg] 
  \Bigg\}.
\end{align}

The mean and variance of the pdf associated with the $ {{p}}_{\mathbf x \leftarrow [\ell,i,x_i] }(t+1,\cdot)$ approximation in \eqref{TeS-var-to-fac-1} are
\begin{align} 
\label{TeS-z-ix-mean-1}
\hat{\mathbf a}_{i,\mathbf x}^{\ell}(t+1) \triangleq 
\underbrace{\frac{1}{C} 
\int_{x} x ~ p_{\mathsf{a}_{i}^{\ell}(x_i)} ( x )
  \mathcal{N}\left( x;  \hat{r}_{\mathbf x, [\ell,i,x_i]}  ,  \nu^r_{\mathbf x,[\ell,i,x_i]}   \right) 
  }_{ 
  \triangleq ~ g_{\mathsf{a}_{i}^{\ell}(x_i)}(\hat{r}_{\mathbf x, [\ell,i,x_i]}  , 
   \nu^r_{\mathbf x,[\ell,i,x_i]}) 
  },
\end{align}
and
\begin{align} 
\label{TeS-var-z-ix-1}
\nu_{i,\mathbf x}^{\mathbf a,\ell}(t+1) \triangleq
\underbrace{
\frac{1}{C} \int_{x} | x - \hat{\mathbf a}_{i,\mathbf x}^{\ell}(t+1) |^2 ~
 p_{\mathsf{a}_{i}^{\ell}(x_i)} ( x )
  \mathcal{N}
  \left( x;  
  \hat{r}_{\mathbf x, [\ell,i,x_i]}  , 
   \nu^r_{\mathbf x,[\ell,i,x_i]}  
    \right)
   }_{ 
  \triangleq ~  \nu^r_{\mathbf x,[\ell,i,x_i]}  
  g_{\mathsf{a}_{i}^{\ell}(x_i)}'
  (\hat{r}_{\mathbf x, [\ell,i,x_i]}  ,
    \nu^r_{\mathbf x,[\ell,i,x_i]}) 
  },
\end{align}
where here $C = \int_{x}  p_{\mathsf{a}_{i}^{\ell}(x_i)} ( x )
  \mathcal{N}\left( x;  \hat{r}_{\mathbf x, [\ell,i,x_i]}  ,  \nu^r_{\mathbf x,[\ell,i,x_i]}   \right)$ and $g_{\mathsf{a}_{i}^{\ell}(x_i)}'$ denotes the derivative of $g_{\mathsf{a}_{i}^{\ell}(x_i)}$ with respect to the first argument. The fact that \eqref{TeS-z-ix-mean-1} and  \eqref{TeS-var-z-ix-1} are related through a derivative was shown in~\cite{rangan2011generalized}.

To derive approximations of $\hat{\mathbf a}_{i,\mathbf x}^{\ell}$ and $\nu_{i,\mathbf x}^{\mathbf a,\ell}$ that avoid the dependence on the destination node $u_{\mathbf{x}}$, we similarly introduce ${\mathbf{x}}$-invariant versions of $\hat{r}_{\mathbf x,[\ell,i,x_i]} $ and ${\nu^r_{\mathbf x,[\ell,i,x_i]}}$ as follows
\begin{align} 
\label{TeS-1-var-r-2}
\frac{1}{\nu^r_{[\ell,i,x_i]}}  
\triangleq &~   \sum_{\mathbf{x}':x'_i=x_i}  
\Bigg\{    
 \left[ 
\hat{\mathbf{a}}^{\ell}_{1, x_1'}(t)
 \cdots \hat{\mathbf{a}}^{\ell}_{i-1, x_{i-1}'}(t) 
\hat{\mathbf{a}}^{\ell}_{i+1,x_{i+1}'}(t) \cdots
\hat{\mathbf{a}}^{\ell}_{d, x_d'}(t)
\right]^2
  \nu_{q,\mathbf x'}^s(t)     \nonumber \\
&~ -  
  \left [  {\hat{s}_{q,\mathbf x'}(t)}^2  -  {\nu_{q,\mathbf x'}^s(t)} \right]    
\sum_{ \substack {A \subset \{1,...,d\} \setminus \{i\} \\ A \neq \emptyset} } 
\Bigg[
 \prod_{ \substack{i' \in A 
}}
 \nu_{i', x_{i'}'}^{\mathbf a,\ell}(t) 
    \prod_{ \substack{
    i'' \in \{1,...,d\} \setminus (\{ i \} \cup A)
    }} 
 {\hat{\mathbf a}_{i'', x_{i''}'}^{\ell} (t) }^2 
 \Bigg]
  \Bigg\}  ,
\end{align}
and
\begin{align} 
\label{TeS-hat-r-2}
\hat{r}_{[\ell,i,x_i]}  =&~ \nu^r_{\mathbf x,[\ell,i,x_i]}  
 \sum_{\mathbf{x}':x'_i=x_i}  
\Bigg\{  
\left[
\hat{\mathbf{a}}^{\ell}_{1,\mathbf x'}(t)
 \cdots \hat{\mathbf{a}}^{\ell}_{i-1,\mathbf x'}(t) 
\hat{\mathbf{a}}^{\ell}_{i+1,\mathbf x'}(t) \cdots
\hat{\mathbf{a}}^{\ell}_{d,\mathbf x'}(t)
\right]
 \hat{s}_{q,\mathbf x'}(t) 
  \Bigg\}  
  \nonumber \\
 &+ \hat{\mathbf{a}}^{\ell}_{i,x_i}(t) 
 \Bigg\{ 1 + \nu^r_{\mathbf x,[\ell,i,x_i]}   
 \sum_{\mathbf{x}':x'_i=x_i,~ \mathbf x' \neq \mathbf x}     
   \left [  {\hat{s}_{q,\mathbf x'}(t)}^2  -  {\nu_{q,\mathbf x'}^s(t)} \right]  \times
   \nonumber \\
 &~~~~~~~~~~~~~~
\sum_{ \substack {A \subset \{1,...,d\} \setminus \{i\} \\ A \neq \emptyset} } 
\Bigg[
 \prod_{ \substack{i' \in A 
}}
 \nu_{i', x_{i'}'}^{\mathbf a,\ell}(t) 
    \prod_{ \substack{
    i'' \in \{1,...,d\} \setminus (\{ i \} \cup A)
    }} 
 {\hat{\mathbf a}_{i'', x_{i''}'}^{\ell} (t) }^2 
 \Bigg] 
  \Bigg\}.
\end{align}

\begin{remark}
\label{TeS-remark1}
Comparing \eqref{TeS-1-var-r-2}-\eqref{TeS-hat-r-2} with \eqref{TeS-1-var-r-1}-\eqref{TeS-hat-r-1} suggest that ${\nu^r_{\mathbf x,[\ell,i,x_i]}} - {\nu^r_{[\ell,i,x_i]}}$ is relatively negligible and hence each $ \hat{r}_{\mathbf x, [\ell,i,x_i]}$ can be approximated by the term 
\begin{align} 
\label{TeS-term}
\hat{r}_{[\ell,i,x_i]} - \nu^r_{[\ell,i,x_i]}  
\left[
\hat{\mathbf{a}}^{\ell}_{1,x_1}(t)
 \cdots \hat{\mathbf{z}}^{\ell}_{i-1,x_{i-1}}(t) 
\hat{\mathbf{a}}^{\ell}_{i+1,x_{i+1}}(t) \cdots
\hat{\mathbf{a}}^{\ell}_{d,x_d}(t)
\right ]
 \hat{s}_{q,\mathbf x}(t),
\end{align}
so that \eqref{TeS-z-ix-mean-1} implies the following approximation.
\end{remark}

Similarly with \eqref{z-ix-mean-3-A}, by applying Remark~\ref{TeS-remark1} and Taylor series expansion in the first argument about the point $\hat{r}_{[\ell,i,x_i]} $ we have
\begin{align} 
\label{TeS-hat-z-1}
&\hat{\mathbf a}_{i,\mathbf x}^{\ell}(t+1)  =
 g_{\mathsf{a}_{i}^{\ell}(x_i)} 
 \left(
 \hat{r}_{\mathbf x, [\ell,i,x_i]}  , 
   \nu^r_{\mathbf x,[\ell,i,x_i]}
   \right) \nonumber \\
\approx &~ 
\hat{\mathbf a}_{i,x_i}^{\ell}(t+1)  
-    \hat{s}_{q,\mathbf x}(t)  \nu_{i,x_i}^{\mathbf a,\ell}(t+1) 
 \Big[ 
\hat{\mathbf{a}}^{\ell}_{1,x_1}(t)
 \cdots \hat{\mathbf{a}}^{\ell}_{i-1,x_{i-1}}(t) 
\hat{\mathbf{a}}^{\ell}_{i+1,x_{i+1}}(t) \cdots
\hat{\mathbf{a}}^{\ell}_{d,x_d}(t)
\Big],
\end{align}
where
\begin{align} 
\label{TeS-z-x-i-mean-3}
\hat{\mathbf a}_{i,x_i}^{\ell}(t+1) \triangleq  g_{\mathsf{a}_{i}^{\ell}(x_i)}
    \left(
 \hat{r}_{[\ell,i,x_i]} ,  \nu^r_{[\ell,i,x_i]} \right),
\end{align}
and 
\begin{align} 
\label{TeS-var-z-x-i-3}
 \nu_{i,x_i}^{\mathbf a,\ell}(t+1) \triangleq  \nu^r_{[\ell,i,x_i]} 
  g_{\mathsf{a}_{i}^{\ell}(x_i)}'
    \left(
 \hat{r}_{[\ell,i,x_i]} ,  \nu^r_{[\ell,i,x_i]} \right),
\end{align}
which match \eqref{TeS-z-ix-mean-1}-\eqref{TeS-var-z-ix-1} without the $\mathbf x$-dependence.

\begin{remark}
\label{remark2_2}
Likewise, taking Taylor series expansions of $ g_{\mathsf{a}_{i}^{\ell}(x_i)}'$ in \eqref{TeS-var-z-ix-1} about the point $\hat{r}_{[\ell,i,x_i]} $ in the first argument and about the point $\nu^{r}_{[\ell,i,x_i]} $ in the second argument and then comparing the result with \eqref{TeS-var-z-x-i-3} confirms that $\nu_{i,\mathbf x}^{\mathbf a,\ell}(t) -  \nu_{i,x_i}^{\mathbf a,\ell}(t)$ is negligible.
\end{remark}

\subsection{Completing the Loop of TeS-AMP Algorithm}

The penultimate step in the derivation of TeS-AMP algorithm is to approximate some earlier steps that use $\hat{\mathbf a}_{i,\mathbf x}^{\ell} (t)$ in place of $\hat{\mathbf a}_{i,x_i}^{\ell} (t)$. Similarly, we plug \eqref{TeS-z-x-i-mean-3} into \eqref{TeS-q-x-1} to obtain
\begin{align} 
\label{TeS-p-x-2}
&~~~~~~~\hat{q}_{\mathbf x}(t) = \sum_{\ell' = 1}^{r}  
{  \hat{\mathbf{a}}_{1, \mathbf{x}}^{\ell'}(t)  }  
\cdots  {  \hat{\mathbf{a}}_{d, \mathbf{x}}^{\ell'}(t)  } 
 \nonumber \\
& \overset{(a)}{\approx}
 \underbrace{
\sum_{\ell' = 1}^{r}  
{  \hat{\mathbf{a}}_{1, x_1}^{\ell'}(t)  }  
\cdots  {  \hat{\mathbf{a}}_{d, x_d}^{\ell'}(t)  }
}_{
\bar q_{\mathbf x}(t)}
+
 \sum_{\ell' = 1}^{r}  
 \sum_{\substack{A\subsetneqq \{1,...,d\} } }
 \Bigg\{ 
 \prod_{i' \in A}  \hat{\mathbf{a}}_{i', x_{i'}}^{\ell'}(t) 
  \prod_{i'' \in \{1,...,d\}\setminus A}
   \hat{s}_{\mathbf x}(t-1)  \nu_{i'',x_{i''}}^{\mathbf a,\ell'}(t)
   \nonumber \\
 & ~~~~~~ \Big[- 
 \hat{\mathbf{a}}^{\ell'}_{1,x_1}(t-1)
 \cdots 
 \hat{\mathbf{a}}^{\ell'}_{i''-1,x_{i''-1}}(t-1) 
\hat{\mathbf{a}}^{\ell'}_{i''+1,x_{i''+1}}(t-1) 
\cdots
\hat{\mathbf{a}}^{\ell'}_{d,x_{d}}(t-1) 
\Big]
\Bigg\}  \nonumber \\
& \overset{(b)}{\approx}  \bar q_{\mathbf x}(t) 
+
 \sum_{\ell' = 1}^{r}  
 \sum_{\substack{A\subsetneqq \{1,...,d\} \\ A\neq \emptyset} }
 \Bigg\{ 
 \prod_{i' \in A}  \hat{\mathbf{a}}_{i', x_{i'}}^{\ell'}(t) 
  \prod_{i'' \in \{1,...,d\}\setminus A} 
  \hat{s}_{\mathbf x}(t-1)  \nu_{i'',x_{i''}}^{\mathbf a,\ell'}(t)
  \nonumber \\
 & ~~~~~~ \Big[- 
 \hat{\mathbf{a}}^{\ell'}_{1,x_1}(t)
 \cdots 
 \hat{\mathbf{a}}^{\ell'}_{i''-1,x_{i''-1}}(t) 
\hat{\mathbf{a}}^{\ell'}_{i''+1,x_{i''+1}}(t) 
\cdots
\hat{\mathbf{a}}^{\ell'}_{d,x_{d}}(t) 
\Big]
\Bigg\}  \nonumber \\
& \overset{(c)}{=}  \bar q_{\mathbf x}(t) 
-
\hat{s}_{\mathbf x}(t-1)  \bar\nu^q_{\mathbf x}(t)
, 
\end{align}
where $(a)$ follows from \eqref{TeS-z-x-i-mean-3}; $(b)$ follows noting that we assumed data size is large and equations on two side of this differ by only one term (which vanishes relative to the others in the large-system limit) and we used $ \hat{\mathbf{a}}^{\ell'}_{i'',x_{i''}}(t) $ in place of $ \hat{\mathbf{a}}^{\ell'}_{i'',x_{i''}}(t-1) $; and $(c)$ follows from the following definition: 
\begin{align} 
\label{TeS-bar-var-p-1}
\bar\nu^q_{\mathbf x}(t) =&~  \sum_{\ell' = 1}^{r}  
 \sum_{\substack{A\subsetneqq \{1,...,d\} \\ A\neq \emptyset} }
 \Bigg\{ 
 \Big(
 \prod_{i \in A}  \hat{\mathbf{a}}_{i, x_{i}}^{\ell'}(t) 
 \Big)^2
 \Big(
 - \hat{s}_{\mathbf x}(t-1) 
  \prod_{i' \in \{1,...,d\}} 
 \hat{\mathbf{a}}^{\ell'}_{i',x_{i'}}(t)
 \Big)^{d-|A|-1}
  \prod_{i'' \in \{1,...,d\}\setminus A}
 \nu_{i'',x_{i''}}^{\mathbf a,\ell'}(t)
 \Bigg\}. 
\end{align}
Note that different with $\bar\nu^p_{\mathbf x}(t)$ in TeG-AMP, which uses the approximation in Remark~\ref{remark3}, the $\bar\nu^q_{\mathbf x}(t)$ in the last step in \eqref{TeS-p-x-2} is an exact result.

Next we plug \eqref{TeS-z-x-i-mean-3} and $ \nu_{i,\mathbf x}^{\mathbf a,\ell}(t)  \approx  \nu_{i, x_i}^{\mathbf a,\ell}(t) $ into \eqref{TeS-v-p-1}, which yields
\begin{align} 
\label{TeS-v-p-2}
{\nu}^{q}_{\mathbf x}(t)  \approx &~
\sum_{\substack{\ell'=1}}^{r}  
\sum_{ \substack {A \subset \{1,...,d\} \\ A \neq \emptyset} }
\left( \prod_{i' \in A} \nu_{i',x_{i'}}^{\mathbf a,\ell'}(t)    
\prod_{i'' \in \{1,...,d\} \setminus A }  
{\hat{\mathbf a}_{i'',\mathbf x}^{\ell' } (t) }^2 \right) \nonumber \\
\overset{(a)}{\approx} &~
\sum_{\substack{\ell = 1 }}^{r}  
\prod_{i \in \{1,...,d\}} \nu_{i,x_{i}}^{\mathbf a,\ell}(t)
+
\sum_{\substack{\ell' = 1 }}^{r}  
\sum_{ \substack {A\subsetneqq \{1,...,d\} \\ A \neq \emptyset } }
\Bigg( \prod_{i' \in A} \nu_{i',x_{i'}}^{\mathbf a,\ell'}(t)    
\prod_{i'' \in \{1,...,d\} \setminus A }  
\Bigg[ \hat{\mathbf a}_{i'',x_{i''}}^{\ell'}(t) - 
 \nonumber \\
&~
 \Big[ 
 \hat{\mathbf{a}}^{\ell'}_{1,x_1}(t)
 \cdots 
 \hat{\mathbf{a}}^{\ell'}_{i''-1,x_{i''-1}}(t) 
\hat{\mathbf{a}}^{\ell'}_{i''+1,x_{i''+1}}(t) 
\cdots
\hat{\mathbf{a}}^{\ell'}_{d,x_{d}}(t) 
\Big]
\hat{s}_{\mathbf x}(t-1)  \nu_{i'',x_{i''}}^{\mathbf a,\ell'}(t) 
\Bigg]^2
\Bigg),
\nonumber \\
\overset{(b)}{\approx} &~
\sum_{\substack{\ell = 1 }}^{r}  
\prod_{i \in \{1,...,d\}} \nu_{i,x_{i}}^{\mathbf a,\ell}(t)
+
\bar\nu^q_{\mathbf x}(t)
\end{align}
where $(a)$ follows from \eqref{TeS-z-x-i-mean-3} and that we used  $\hat{\mathbf{a}}^{\ell'}_{i'',x_{i''}}(t) $ in place of $ \hat{\mathbf{a}}^{\ell'}_{i'',x_{i''}}(t-1)$, and $(b)$ follows from Remark~\ref{remarkTeSvp}.

\begin{remark}
\label{remarkTeSvp}
We observed that the following approximation holds
\begin{align} 
&\sum_{\substack{\ell' = 1 }}^{r}  
\sum_{ \substack {A\subsetneqq \{1,...,d\} \\ A \neq \emptyset } }
\Bigg( \prod_{i' \in A} \nu_{i',x_{i'}}^{\mathbf a,\ell'}(t)    
\prod_{i'' \in \{1,...,d\} \setminus A }  
\Bigg[ \hat{\mathbf a}_{i'',x_{i''}}^{\ell'}(t) - 
 \nonumber \\
&~~~
 \Big[ 
 \hat{\mathbf{a}}^{\ell'}_{1,x_1}(t)
 \cdots 
 \hat{\mathbf{a}}^{\ell'}_{i''-1,x_{i''-1}}(t) 
\hat{\mathbf{a}}^{\ell'}_{i''+1,x_{i''+1}}(t) 
\cdots
\hat{\mathbf{a}}^{\ell'}_{d,x_{d}}(t) 
\Big]
\hat{s}_{\mathbf x}(t-1)  \nu_{i'',x_{i''}}^{\mathbf a,\ell'}(t) 
\Bigg]^2
\Bigg)
\nonumber \\
\approx
&\frac{1}{\hat{s}_{\mathbf x}(t-1) } \sum_{\ell' = 1}^{r}  
 \sum_{\substack{A\subsetneqq \{1,...,d\} \\ A\neq \emptyset} }
 \Bigg\{ 
 \prod_{i' \in A}  \hat{\mathbf{a}}_{i', x_{i'}}^{\ell'}(t) 
  \prod_{i'' \in \{1,...,d\}\setminus A} 
  \hat{s}_{\mathbf x}(t-1)  \nu_{i'',x_{i''}}^{\mathbf a,\ell'}(t)
  \nonumber \\
 & ~~~ \Big[- 
 \hat{\mathbf{a}}^{\ell'}_{1,x_1}(t)
 \cdots 
 \hat{\mathbf{a}}^{\ell'}_{i''-1,x_{i''-1}}(t) 
\hat{\mathbf{a}}^{\ell'}_{i''+1,x_{i''+1}}(t) 
\cdots
\hat{\mathbf{a}}^{\ell'}_{d,x_{d}}(t) 
\Big]
\Bigg\} ,
\end{align}
where the approximation follows from similar derivations and approximations used in Remark~\ref{remarkvp}.
\end{remark}

Similarly we plug \eqref{TeS-z-x-i-mean-3} into \eqref{TeS-hat-r-2} to obtain
\begin{align} 
\label{TeS-hat-r-3}
&~\hat{r}_{[\ell,i,x_i]}  \nonumber \\
 \overset{(a)}{\approx}   &~
\nu^r_{[\ell,i,x_i]}  
 \sum_{\mathbf{x}':x'_i=x_i}  
\Bigg\{  
\Bigg[ 
\hat{\mathbf{a}}^{\ell}_{1,x_1'}(t)
 \cdots \hat{\mathbf{a}}^{\ell}_{i-1,x_{i-1}'}(t) 
\hat{\mathbf{a}}^{\ell}_{i+1,x_{i+1}'}(t) \cdots
\hat{\mathbf{a}}^{\ell}_{d,x_d'}(t)
\Bigg]
 \hat{s}_{\mathbf x'}(t) 
  \Bigg\}  
  \nonumber \\
 &~ + \hat{\mathbf{a}}^{\ell}_{i,x_i}(t) 
 \Bigg\{ 1 -
  \nu^r_{[\ell,i,x_i]}   
 \sum_{\mathbf{x}':x'_i=x_i}     
   {\nu_{\mathbf x'}^s(t)}  
\sum_{ \substack {A \subset \{1,...,d\} \setminus \{i\} \\ A \neq \emptyset} } 
\Bigg[
 \prod_{ \substack{i' \in A 
}}
 \nu_{i', x_{i'}'}^{\mathbf a,\ell}(t) 
    \prod_{ \substack{
    i'' \in \{1,...,d\} \setminus (\{ i \} \cup A) 
    }} 
 {\hat{\mathbf a}_{i'', x_{i''}'}^{\ell } (t) }^2 
 \Bigg]   
  \Bigg\} 
  \nonumber \\
 \overset{(b)}{\approx} &~
\nu^r_{[\ell,i,x_i]}  
 \sum_{\mathbf{x}':x'_i=x_i}  
\Bigg\{  
\Bigg[ 
\hat{\mathbf{a}}^{\ell}_{1,x_1'}(t)
 \cdots \hat{\mathbf{a}}^{\ell}_{i-1,x_{i-1}'}(t) 
\hat{\mathbf{a}}^{\ell}_{i+1,x_{i+1}'}(t) \cdots
\hat{\mathbf{a}}^{\ell}_{d,x_d'}(t)
\Bigg]
 \hat{s}_{\mathbf x'}(t) 
  \Bigg\}  
  \nonumber \\
    & ~ + \hat{\mathbf{a}}^{\ell}_{i,x_i}(t) 
 \left\{ 1 -
\frac{
 \sum\limits_{\mathbf{x}':x'_i=x_i}     
   {\nu_{\mathbf x'}^s(t)}  
\sum_{ \substack {A \subset \{1,...,d\} \\  \setminus \{i\}, A \neq \emptyset} } 
\Bigg[
 \prod\limits_{ \substack{i' \in A 
}}
 \nu_{i', x_{i'}'}^{\mathbf a,\ell}(t) 
    \prod\limits_{ \substack{
    i'' \in \{1,...,d\} \setminus (\{ i \} \cup A) 
    }} 
 {\hat{\mathbf a}_{i'', x_{i''}'}^{\ell } (t) }^2 
 \Bigg]      
  }{
 \sum\limits_{\mathbf{x}':x'_i=x_i}  
\Bigg\{    
 \left[ 
\hat{\mathbf{a}}^{\ell}_{1, x_1'}(t)
 \cdots \hat{\mathbf{a}}^{\ell}_{i-1, x_{i-1}'}(t) 
\hat{\mathbf{a}}^{\ell}_{i+1,x_{i+1}'}(t) \cdots
\hat{\mathbf{a}}^{\ell}_{d, x_d'}(t)
\right]^2
\nu_{\mathbf x'}^s(t)  
\Bigg\} 
}
\right \},
\end{align}
where $(a)$ follows from neglecting the below term with the same arguments as before.
\begin{align} 
\label{nu-r-approx-2}
&\nu^r_{[\ell,i,x_i]}  
 \sum_{\mathbf{x}':x'_i=x_i}  
\Bigg\{  
\left[
\hat{\mathbf{a}}^{\ell}_{1,\mathbf x'}(t)
 \cdots \hat{\mathbf{a}}^{\ell}_{i-1,\mathbf x'}(t) 
\hat{\mathbf{a}}^{\ell}_{i+1,\mathbf x'}(t) \cdots
\hat{\mathbf{a}}^{\ell}_{d,\mathbf x'}(t)
\right]
 \hat{s}_{\mathbf x'}(t) 
  \Bigg\}  
  \nonumber \\
 &- \nu^r_{[\ell,i,x_i]}  
 \sum_{\mathbf{x}':x'_i=x_i}  
\Bigg\{  
\Bigg[ 
\hat{\mathbf{a}}^{\ell}_{1,x_1'}(t)
 \cdots \hat{\mathbf{a}}^{\ell}_{i-1,x_{i-1}'}(t) 
\hat{\mathbf{a}}^{\ell}_{i+1,x_{i+1}'}(t) \cdots
\hat{\mathbf{a}}^{\ell}_{d,x_d'}(t)
\Bigg]
 \hat{s}_{\mathbf x'}(t) 
  \Bigg\}  
    \nonumber \\
 &+ \hat{\mathbf{a}}^{\ell}_{i,x_i}(t) 
 \Bigg\{  \nu^r_{[\ell,i,x_i]}   
 \sum_{\mathbf{x}':x'_i=x_i}     
  {\hat{s}_{\mathbf x'}(t)}^2  
\sum_{ \substack {A \subset \{1,...,d\} \setminus \{i\} \\ A \neq \emptyset} } 
\Bigg[
 \prod_{ \substack{i' \in A 
}}
 \nu_{i', x_{i'}'}^{\mathbf a,\ell'}(t) 
    \prod_{ \substack{
    i'' \in \{1,...,d\} \setminus (\{ i \} \cup A) 
    }} 
 {\hat{\mathbf a}_{i'', x_{i''}'}^{\ell' } (t) }^2 
 \Bigg]   
  \Bigg\} 
  \nonumber \\
\overset{(a)}{\approx} &~
-\nu^r_{[\ell,i,x_i]}  
 \sum_{\mathbf{x}':x'_i=x_i}  
\Bigg\{ \hat{s}_{\mathbf x'}(t)^2   
\Bigg[
 \sum_{\substack{A\subsetneqq \{1,...,d\} \setminus \{i\} } }
 \hat{\mathbf{a}}_{i, x_{i}'}^{\ell}(t)
\nonumber \\
& 
\left(
 \prod_{\substack{ i' \in \{1,...,d\}\setminus (\{i\} \cup A)        }} 
  \hat{\mathbf{a}}_{i', x_{i'}'}^{\ell}(t) 
  \right)^2
  \left(
  -\hat{s}_{\mathbf x'}(t-1)   
  \prod_{ \substack{i''' \in \{1,...,d\} }}
  \hat{\mathbf{a}}_{i''', x_{i'''}'}^{\ell}(t) 
  \right)^{|A|-1}
  \prod_{ \substack{i'' \in A }}
   \nu_{i'',x_{i''}'}^{\mathbf a,\ell}(t)
\Bigg]
\Bigg\}
\nonumber \\
&+ \hat{\mathbf{a}}^{\ell}_{i,x_i}(t) 
\nu^r_{[\ell,i,x_i]} 
 \Bigg\{    
 \sum_{\mathbf{x}':x'_i=x_i}     
 {\hat{s}_{\mathbf x'}(t)}^2   
\sum_{ \substack {A \subset \{1,...,d\} \setminus \{i\} \\ A \neq \emptyset} } 
\Bigg[
 \prod_{ \substack{i' \in A 
}}
 \nu_{i', x_{i'}'}^{\mathbf a,\ell}(t) 
    \prod_{ \substack{
    i'' \in \{1,...,d\} \setminus (\{ i \} \cup A) 
    }} 
 {\hat{\mathbf a}_{i'', x_{i''}'}^{\ell } (t) }^2 
 \Bigg]   
  \Bigg\} 
  \nonumber \\
{\approx} &~
\nu^r_{[\ell,i,x_i]}  \hat{\mathbf{a}}^{\ell}_{i,x_i}(t) 
 \sum_{\mathbf{x}':x'_i=x_i}  
\Bigg\{ \hat{s}_{\mathbf x'}(t)^2   
\Bigg[
 \sum_{\substack{A\subsetneqq \{1,...,d\} \setminus \{i\} } }
   \prod_{ \substack{i'' \in A }}
   \nu_{i'',x_{i''}'}^{\mathbf a,\ell}(t)
   \left(
 \prod_{\substack{ i' \in \{1,...,d\}\setminus (\{i\} \cup A)        }} 
  \hat{\mathbf{a}}_{i', x_{i'}'}^{\ell}(t) 
  \right)^2
  \Bigg]
\nonumber \\
&~
\left(
  1-
  \left(
  -\hat{s}_{\mathbf x'}(t-1)   
  \prod_{ \substack{i''' \in \{1,...,d\} }}
  \hat{\mathbf{a}}_{i''', x_{i'''}'}^{\ell}(t) 
  \right)^{|A|-1}
  \right)
\Bigg\}
 ,
\end{align}
where $(a)$ follows Remark~\eqref{TeS-remark-4} and we used $\hat{s}_{\mathbf x'}(t)$ in replace of $\hat{s}_{\mathbf x'}(t-1)$. We observed that the term \eqref{nu-r-approx-2} is small. If under specific applications this terms was not negligible, it can simply be included back in the algorithm. And for $(b)$, we had the approximation:
\begin{align} 
\label{TeS-var-r-approx-1}
\frac{1}{
  \nu^r_{[\ell,i,x_i]}   
  }
  \approx
 \sum_{\mathbf{x}':x'_i=x_i}  
\Bigg\{    
 \left[ 
\hat{\mathbf{a}}^{\ell}_{1, x_1'}(t)
 \cdots \hat{\mathbf{a}}^{\ell}_{i-1, x_{i-1}'}(t) 
\hat{\mathbf{a}}^{\ell}_{i+1,x_{i+1}'}(t) \cdots
\hat{\mathbf{a}}^{\ell}_{d, x_d'}(t)
\right]^2
\nu_{\mathbf x'}^s(t)  
\Bigg\} ,
\end{align}
by neglecting ${\hat{s}_{q,\mathbf x'}(t)}^2 - \nu^s_{q,\mathbf x'}(t)$ terms in \eqref{TeS-hat-r-2} as explained in Appendix B in~\cite{parker2014bilinear}.

\begin{remark}
\label{TeS-remark-4}
The following holds:
\begin{align} 
\label{TeS-remark-3-1_3}
&~\nu^r_{[\ell,i,x_i]}  
 \sum_{\mathbf{x}':x'_i=x_i}  
\Bigg\{  
\left[
\hat{\mathbf{a}}^{\ell}_{1,\mathbf x'}(t)
 \cdots \hat{\mathbf{a}}^{\ell}_{i-1,\mathbf x'}(t) 
\hat{\mathbf{a}}^{\ell}_{i+1,\mathbf x'}(t) \cdots
\hat{\mathbf{a}}^{\ell}_{d,\mathbf x'}(t)
\right]
\hat{s}_{\mathbf x'}(t) 
\Bigg\}  \nonumber \\
\overset{(a)}{\approx} &~
\nu^r_{[\ell,i,x_i]}  
 \sum_{\mathbf{x}':x'_i=x_i}  
\Bigg\{ \hat{s}_{\mathbf x'}(t)   
\Bigg[
{  \hat{\mathbf{a}}_{1, x_1'}^{\ell}(t)  }  
\cdots 
{  \hat{\mathbf{a}}_{i-1, x_{i-1}'}^{\ell}(t)  }
{  \hat{\mathbf{a}}_{i+1, x_{i+1}'}^{\ell}(t)  }
\cdots
{  \hat{\mathbf{a}}_{d, x_d'}^{\ell}(t)  }
-\hat{s}_{\mathbf x'}(t-1)
 \sum_{\substack{A\subsetneqq \{1,...,d\} \setminus \{i\} } }
 \hat{\mathbf{a}}_{i, x_{i}'}^{\ell}(t)
\nonumber \\
& 
\left(
 \prod_{\substack{ i' \in \{1,...,d\}\setminus (\{i\} \cup A)        }} 
  \hat{\mathbf{a}}_{i', x_{i'}'}^{\ell}(t) 
  \right)^2
  \left(
  -\hat{s}_{\mathbf x'}(t-1)   
  \prod_{ \substack{i''' \in \{1,...,d\} }}
  \hat{\mathbf{a}}_{i''', x_{i'''}'}^{\ell}(t) 
  \right)^{|A|-1}
  \prod_{ \substack{i'' \in A }}
   \nu_{i'',x_{i''}'}^{\mathbf a,\ell}(t)
\Bigg]
\Bigg\},
\end{align}
 where $(a)$ follows from
\begin{align} 
\label{remark-3-1_4}
&\hat{\mathbf{a}}^{\ell}_{1,\mathbf x'}(t)
 \cdots \hat{\mathbf{a}}^{\ell}_{i-1,\mathbf x'}(t) 
\hat{\mathbf{a}}^{\ell}_{i+1,\mathbf x'}(t) \cdots
\hat{\mathbf{a}}^{\ell}_{d,\mathbf x'}(t) \nonumber \\
 \overset{(a)}{\approx}&~
{  \hat{\mathbf{a}}_{1, x_1'}^{\ell}(t)  }  
\cdots 
{  \hat{\mathbf{a}}_{i-1, x_{i-1}'}^{\ell}(t)  }
{  \hat{\mathbf{a}}_{i+1, x_{i+1}'}^{\ell}(t)  }
\cdots
{  \hat{\mathbf{a}}_{d, x_d'}^{\ell}(t)  }
\nonumber \\
&+
 \sum_{\substack{A\subsetneqq \{1,...,d\} \setminus \{i\} } }
 \Bigg\{ 
 \prod_{\substack{i' \in A }} 
  \hat{\mathbf{a}}_{i', x_{i'}'}^{\ell}(t) 
  \prod_{ \substack{ i'' \in \{1,...,d\}\setminus (\{i\} \cup A)         }}
   \hat{s}_{\mathbf x'}(t-1)  \nu_{i'',x_{i''}'}^{\mathbf a,\ell}(t)
   \nonumber \\
 & ~~~ \Big[- 
 \hat{\mathbf{a}}^{\ell}_{1,x_1'}(t-1)
 \cdots 
 \hat{\mathbf{a}}^{\ell}_{i''-1,x_{i''-1}'}(t-1) 
\hat{\mathbf{a}}^{\ell}_{i''+1,x_{i''+1}'}(t-1) 
\cdots
\hat{\mathbf{a}}^{\ell}_{d,x_{d}'}(t-1) 
\Big]
\Bigg\}  \nonumber \\
\overset{(b)}{\approx}&~
{  \hat{\mathbf{a}}_{1, x_1'}^{\ell}(t)  }  
\cdots 
{  \hat{\mathbf{a}}_{i-1, x_{i-1}'}^{\ell}(t)  }
{  \hat{\mathbf{a}}_{i+1, x_{i+1}'}^{\ell}(t)  }
\cdots
{  \hat{\mathbf{a}}_{d, x_d'}^{\ell}(t)  }
-\hat{s}_{\mathbf x'}(t-1)
 \sum_{\substack{A\subsetneqq \{1,...,d\} \setminus \{i\} } }
 \hat{\mathbf{a}}_{i, x_{i}'}^{\ell}(t)
\nonumber \\
& 
\left(
 \prod_{\substack{ i' \in \{1,...,d\}\setminus (\{i\} \cup A)        }} 
  \hat{\mathbf{a}}_{i', x_{i'}'}^{\ell}(t) 
  \right)^2
  \left(
  -\hat{s}_{\mathbf x'}(t-1)   
  \prod_{ \substack{i''' \in \{1,...,d\} }}
  \hat{\mathbf{a}}_{i''', x_{i'''}'}^{\ell}(t) 
  \right)^{|A|-1}
  \prod_{ \substack{i'' \in A }}
   \nu_{i'',x_{i''}'}^{\mathbf a,\ell}(t),
\end{align}
where $(a)$ follows from \eqref{TeS-z-x-i-mean-3} and in $(b)$ we used $ \hat{\mathbf{a}}^{\ell}_{i'',x_{i''}'}(t) $ in place of $ \hat{\mathbf{a}}^{\ell}_{i'',x_{i''}'}(t-1) $.
\end{remark}

\subsection{Approximated Posteriors}

The final step in the TeS-AMP derivation is also to approximate the SPA posterior log-pdfs in \eqref{TeS-AMP-sum-pro-3}. Plugging \eqref{TeS-sum-pro-Taylor-1} into those expressions, we get
\begin{align} 
\label{TeS-fac-1}
 {{p}}_{[\ell,i,x_i] }(t+1,\mathbf{a}_{i}^{\ell}(x_i))  \approx   \text{const}  + \log \left(  p_{\mathsf{a}_{i}^{\ell}(x_i)}\left(\mathbf{a}_{i}^{\ell}(x_i)\right)  
\mathcal{N}\left( { \mathbf a_i^{\ell} (x_i) };  \hat{r}_{[\ell,i,x_i]}  ,  \nu^r_{[\ell,i,x_i]}  \right)   \right),
\end{align}
using steps similar to \eqref{TeS-var-to-fac-1}. The associated pdfs are
\begin{align} 
\label{TeS-ass-pdf-1}
&p_{\mathsf{a}_{i}^{\ell} \left(x_{i}\right) | \mathsf{r}_{\left[\ell, i, x_{i} \right]}}
\left(\mathbf{a}_{i}^{\ell} \left(x_{i}\right) | \hat{r}_{[\ell,i,x_i]} ; \nu_{[\ell,i,x_i]}^{r}\right)
\triangleq 
\frac{
p_{   \mathsf{a}_{i}^{\ell} \left(x_{i}\right)   }
\left(  \mathbf{a}_{i}^{\ell} \left(x_{i}\right)  \right) 
\mathcal{N}\left( \mathbf{a}_{i}^{\ell} \left(x_{i}\right) ; \hat{r}_{[\ell,i,x_i]} , \nu_{[\ell,i,x_i]}^{r}\right)
}
{
\int_{a^{\prime}} 
p_{   \mathsf{a}_{i}^{\ell} \left(x_{i}\right)   }
\left( \mathbf{a}^{\prime}\right) 
\mathcal{N}\left(\mathbf{a}^{\prime} ; \hat{r}_{[\ell,i,x_i]}, \nu_{[\ell,i,x_i]}^{r}\right)
}
\end{align}
which are the $t$-th iteration approximations to the true marginal posteriors $p_{\mathsf{a}_{i}^{\ell} \left(x_{i}\right) | \mathsf{V} }
\left(\mathbf{a}_{i}^{\ell} \left(x_{i}\right) |  \mathcal{V} \right)$.

Note also that $\hat{\mathbf a}_{i,x_i}^{\ell}(t+1)$ and $ \nu_{i,x_i}^{\mathbf a,\ell}(t+1)$ from \eqref{TeS-z-x-i-mean-3}-\eqref{TeS-var-z-x-i-3} are the mean and variance, respectively, of the posterior pdf in \eqref{TeS-ass-pdf-1}, and \eqref{TeS-ass-pdf-1} is interpreted as the posterior pdf of ${\mathsf a}_{i}^{\ell}(x_i)$ given the observation $ \mathsf{r}_{\left[\ell, i, x_{i} \right]} = \hat{r}_{[\ell,i,x_i]}$ under the prior model ${\mathsf a}_{i}^{\ell}(x_i) \sim p_{{\mathsf a}_{i}^{\ell}(x_i)}$ and the likelihood model 
\begin{align} 
\label{TeS-fac-2}
&p_{\mathsf{a}_{i}^{\ell} \left(x_{i}\right) | \mathsf{r}_{\left[\ell, i, x_{i} \right]}}
\left(\mathbf{a}_{i}^{\ell} \left(x_{i}\right) | \hat{r}_{[\ell,i,x_i]} ; \nu_{[\ell,i,x_i]}^{r}\right) = \mathcal{N}\left( \mathbf{a}_{i}^{\ell} \left(x_{i}\right) ; \hat{r}_{[\ell,i,x_i]} , \nu_{[\ell,i,x_i]}^{r}\right)
\end{align}
implicitly assumed by the $t$-th iteration TeS-AMP. To this end, we complete the derivation of TeS-AMP.

The TeS-AMP algorithm derived is summarized in Algorithm~\ref{tab:A2}, where we have also written the algorithm in a general form that allows the use of complex-valued quantities. Comparing with TeG-AMP in Algorithm~1, the computational load of TeS-AMP is reduced, making it more suitable for real-time implementation when a tensor is low CP-rank.

\begin{algorithm}[!ht] \small
	\caption{Tensor Simplified Approximate Message Passing (TeS-AMP) Algorithm.}	\label{tab:A2}
\textbf{Initialization}: $p_{\mathsf{u}_{\mathbf x} | \mathsf{q}_{\mathbf x} } \left( u | \hat{q} ; \nu^{q} \right) =   \frac{
p_{\mathsf{v}_{\mathbf x} | \mathsf{u}_{\mathbf x} } \left( v_{\mathbf x} | u \right)
\mathcal{N}( u ;  \hat{q}(t) , \nu^q(t) )
}{\int_{u^{\prime}}  p_{\mathsf{v}_{\mathbf x} | \mathsf{u}_{\mathbf x} } \left( v_{\mathbf x} | u' \right)
\mathcal{N}( u' ;  \hat{q}(t) , \nu^q(t) )   }$; 
 $p_{\mathsf{a}_{i}^{\ell} \left(x_{i}\right) | \mathsf{r}_{\left[\ell, i, x_{i} \right]}}
\left(a | \hat{r} ; \nu^{r}\right)
=
\frac{p
\left(  a  \right) 
\mathcal{N}\left(a ; \hat{r} , \nu^{r}\right)}
{\int_{a^{\prime}} p
\left( a'  \right) 
\mathcal{N}\left(a' ; \hat{r} , \nu^{r}\right)
}$. $\forall i, x_i,~ \hat{s}_{\mathbf x}(0)=0$;~ $\forall i, \ell, x_i$, \text{choose} $\hat{\mathbf{a}}^{\ell}_{i,x_i}(1)$, $\nu^{\mathbf{a},\ell}_{i,x_i}(1)$.

~~1: \textbf{for} $t=1,\cdots,T_{\text{max}}$ \textbf{do}

~~2: ~~~~~~~~ $\bar{q}_{\mathbf x}(t) \leftarrow \sum_{\ell' = 1}^{r}  
{  \hat{\mathbf{a}}_{1, x_1}^{\ell'}(t)  }  
\cdots  {  \hat{\mathbf{a}}_{d, x_d}^{\ell'}(t)  }$ [according to \eqref{TeS-p-x-2}]

~~3: ~~~~~~~~ Update $\bar\nu^q_{\mathbf x}(t) $ [according to \eqref{TeS-bar-var-p-1}]

~~4: ~~~~~~~~ Update $\hat{q}_{\mathbf{x}}(t)$ [according to \eqref{TeS-p-x-2}]

~~5: ~~~~~~~~ Update $\nu^{q}_{\mathbf{x}}(t)$ [according to \eqref{TeS-v-p-2}]

~~6: ~~~~~~~~ $\nu_{q,\mathbf x}^{u}(t)  \leftarrow \mathrm{Var}\left\{\mathsf{u}_{\mathbf x} | \mathsf{q}_{\mathbf x}=\hat{q}_{\mathbf x}(t) ; \nu_{\mathbf x}^{q}(t)\right \}$ [according to \eqref{TeS-nu-u-1}]

~~7: ~~~~~~~~ $\hat{u}_{q,\mathbf x}(t)  \leftarrow \mathbb{E} \left\{ \mathsf{u}_{\mathbf x} | \mathsf{q}_{\mathbf x} = \hat{q}_{\mathbf x}(t) ; \nu_{\mathbf x}^{q}(t)\right\}$ [according to \eqref{TeS-hat-u-1}]

~~8: ~~~~~~~~ $\hat{s}_{q,\mathbf x}(t) \leftarrow \frac{1}{ \nu_{\mathbf x}^q(t) } \left(\hat{u}_{\mathbf x}(t)-\hat{q}_{\mathbf x}(t)\right)$ [according to \eqref{TeS-hat-s-2}]
 
~~9: ~~~~~~~~ ${\nu}_{q,\mathbf x}^{s}(t) \leftarrow \frac{1}{\nu_{\mathbf x}^{q}(t)}\left(1-\frac{\nu_{\mathbf x}^{u}(t)}{\nu_{\mathbf x}^{q}(t)}\right)$ [according to \eqref{TeS-nu-s-2}]

10: ~~~~~~~~ Update $\nu^r_{[\ell, i, x_{i}]} $ [according to \eqref{TeS-var-r-approx-1}]

11: ~~~~~~~~ Update $\hat{r}_{[\ell, i, x_{i}]}$ [according to \eqref{TeS-hat-r-3}]

12: ~~~~~~~~ $\nu_{i,x_i}^{\mathbf a,\ell}(t+1) \leftarrow \mathrm{Var}\left\{\mathsf{a}_{i,x_i}^{\ell} | \mathsf{r}_{[\ell, i, x_{i}]}=\hat{r}_{[\ell, i, x_{i}]} ; \nu_{[\ell, i, x_{i}]}^{r} \right \}$ [according to \eqref{TeS-ass-pdf-1}]

13: ~~~~~~~~ $\hat{\mathbf a}_{i,x_i}^{\ell}(t+1) \leftarrow \mathbb{E}\left\{\mathsf{a}_{i,x_i}^{\ell} | \mathsf{r}_{[\ell, i, x_{i}]}=\hat{r}_{[\ell, i, x_{i}]} ; \nu_{[\ell, i, x_{i}]}^{r} \right \}$ [according to \eqref{TeS-ass-pdf-1}]

14: ~~~~~~~~ \textbf{if} $\sum_{\mathbf{x}} |\bar{q}_{\mathbf{x}}(t) - \bar{q}_{\mathbf{x}}(t-1) |^2 \leq \tau_{\text{threshold} } \sum_{\mathbf{x}} |\bar{q}_{\mathbf{x}}(t)|^2$ \textbf{then}

15: ~~~~~~~~ ~~~~~~~~\textbf{stop}
 
16: \textbf{end for}
\end{algorithm}

\subsection{Damping \& Adaptive Damping for TeS-AMP}

Similarly, to adopt a damping strategy, we can also use a damping factor $\beta(t)\in(0,1]$ at the $t$-th iteration to slow the evolution of certain variables in TeS-AMP. To do this, steps 3, 5, 8, and 9 in Algorithm~\ref{tab:A2} are replaced with
\begin{align}
\label{TeS-damping-1}
\bar\nu^q_{\mathbf x}(t) &= (1-\beta(t)) \bar\nu^q_{\mathbf x}(t-1) + \beta(t) \sum_{\ell' = 1}^{r}  
 \sum_{\substack{A\subsetneqq \{1,...,d\} \\ A\neq \emptyset} }
 \Bigg\{ 
 \Big(
 \prod_{i \in A}  \hat{\mathbf{a}}_{i, x_{i}}^{\ell'}(t) 
 \Big)^2
 \times \nonumber \\
 &~~~~~~~~~~~~~~~
 \Big(
 - \hat{s}_{\mathbf x}(t-1) 
  \prod_{i' \in \{1,...,d\}} 
 \hat{\mathbf{a}}^{\ell'}_{i',x_{i'}}(t)
 \Big)^{d-|A|-1}
  \prod_{i'' \in \{1,...,d\}\setminus A}
 \nu_{i'',x_{i''}}^{\mathbf a,\ell'}(t)
 \Bigg\},   \\
 \label{TeS-damping-2}
 \nu^q_{\mathbf{x}} (t) &= \beta(t) \left(  \bar\nu^q_{\mathbf x}(t)  +  \sum_{\substack{\ell = 1 }}^{r}  
\prod_{i \in \{1,...,d\}} \nu_{i,x_{i}}^{\mathbf a,\ell}(t)  \right) + (1- \beta(t) )  \nu^q_{\mathbf{x}} (t-1), \\
 \label{TeS-damping-3}
\hat{s}_{\mathbf x}(t) &= \beta(t)  \frac{1}{ \nu_{\mathbf x}^q(t) } \left(\hat{u}_{\mathbf x}(t)-\hat{q}_{\mathbf x}(t)\right) + (1- \beta(t) ) \hat{s}_{\mathbf x}(t-1)  , \\
 \label{TeS-damping-4}
{\nu}_{\mathbf x}^{s}(t) &=  \beta(t)  \frac{1}{\nu_{\mathbf x}^{q}(t)}\left(1-\frac{\nu_{\mathbf x}^{u}(t)}{\nu_{\mathbf x}^{q}(t)}\right) + (1- \beta(t) ) {\nu}_{\mathbf x}^{s}(t-1), 
\end{align}
and the following are inserted between step 9 and step 10 in Algorithm~\ref{tab:A2}:
\begin{align}
 \label{TeS-damping-5}
{\bar{\mathbf{a}}^{\ell}_{i,x_i}}(t) &=  \beta(t) {\hat{\mathbf{a}}^{\ell}_{i,x_i}}(t) + (1- \beta(t) ) {\bar{\mathbf{a}}^{\ell}_{i,x_i}}(t-1).
\end{align}
The newly defined state variables ${\bar{\mathbf{a}}^{\ell}_{i,x_i}}(t)$ are then used in place of ${\hat{\mathbf{a}}^{\ell}_{i,x_i}}(t)$ in step 13 in Algorithm~\ref{tab:A2}. 


In addition, the adaptive damping strategy can be easily implemented by following \eqref{KL-min}-\eqref{KL-min-4} in Appendix~\ref{appendixF} with ${\mathsf{Z}}$ and $\mathcal{Z}$ respectively replaced by ${\mathsf{A}}$ and $\mathcal{A}$, where we also approximate the cost \eqref{KL-min-3} by using, in place of $\mathcal{U}$, an independent Gaussian tensor whose component means and variances are matched to those of $\mathcal{U}$. Taking the joint TeS-AMP posterior approximation $b_{\mathsf{A}}(t)$ to be the product of the marginals from \eqref{TeS-condtional-pdf-1}, the resulting component means and variances are
\begin{align} 
\mathbb{E}_{b_{\mathsf{A}}(t)}\left\{u_{\mathbf{x}}\right\} 
&=\sum_{\ell=1}^{r} 
\mathbb{E}_{b_{\mathsf{A}}(t)}\left\{\mathsf{a}_{1, x_{1}}^{\ell}\right\} \cdots \mathbb{E}_{b_{\mathsf{A}}(t)}\left\{\mathsf{a}_{d, x_{d}}^{\ell}\right\} =
\sum_{\ell=1}^{r} 
\hat{\mathbf{a}}_{1, x_{1}}^{\ell}(t) \cdots \hat{\mathbf{a}}_{d,x_{d}}^{\ell}(t) =\bar{q}_{\mathbf{x}}(t),
\nonumber \\ 
\operatorname{Var}_{b_{\mathsf{A}}(t)}\left\{u_{\mathbf{x}}\right\} 
&=\sum_{\ell'=1}^{r} 
\sum_{\substack{A \subset\{1, \ldots, d\} \\ A \neq \emptyset}}
\left(\prod_{i^{\prime} \in A} \nu_{i^{\prime}, \mathbf{x} }^{\mathbf{a}, \ell' }(t) 
\prod_{i^{\prime \prime} \in\{1, \ldots, d\} \backslash A} 
\hat{\mathbf{a}}_{i^{\prime \prime}, \mathbf{x}}^{ \ell' }(t)^{2}\right) .
\end{align}

The approximate $t$-th iteration cost becomes
\begin{align} 
\label{TeS-hat-J-t-1}
\hat{J}(t)=&~\sum_{ \substack{1 \leq i \leq d \\ 1 \leq \ell \leq r \\ 1 \leq x_i \leq n_i }} 
D\left(p_{\mathsf{a}_{i, x_{i}}^{\ell} | r_{\left[\ell, i, x_{i}\right]}}\left(\cdot | \hat{r}_{\left[\ell, i, x_{i}\right]}(t) ; \nu_{\left[\ell, i, x_{i}\right]}^{r}(t)\right) \| p_{\mathsf{a}_{i, x_{i}}^{\ell}}(\cdot)\right) 
\nonumber \\
&- \sum_{\mathbf{x}} \mathbb{E}_{\mathsf{u}_{\mathbf{x}}  \sim \mathcal{N}\left(\mathbb{E}_{b_{\mathsf{A}}(t)}\left\{u_{\mathbf{x}} \right\} ; \mathrm{Var}_{b_{\mathsf{A}}(t)} \left\{u_{\mathbf{x}} \right\}\right)} {\left\{ \log p_{\mathsf{v}_{\mathbf{x}} | \mathsf{u}_{\mathbf{x}} } \left(v_{\mathbf{x}} | u_{\mathbf{x}}\right)  \right\} } 
\nonumber \\
\overset{(a)}{=}&~\sum_{ \substack{1 \leq i \leq d \\ 1 \leq \ell \leq r \\ 1 \leq x_i \leq n_i }} 
\int_{\mathbf{a}_{i}^{\ell}\left(x_{i}\right)} 
\mathcal{N}\left(\mathbf{a}_{i}^{\ell} \left(x_{i}\right) ; 
\hat{\mathbf{a}}_{i, x_{i}}^{\ell}(t),
\nu_{i, x_{i}}^{\mathbf{a}, \ell}(t)\right)
\log
\frac{
\mathcal{N}\left(\mathbf{a}_{i}^{\ell} \left(x_{i}\right) ; 
\hat{\mathbf{a}}_{i, x_{i}}^{\ell}(t),
\nu_{i, x_{i}}^{\mathbf{a}, \ell}(t)\right)
}{
\mathcal{N}\left(
\mathbf{a}_{i}^{\ell} \left(x_{i}\right) ; 
\hat{\mathbf{a}}_{\text{prior}},
\nu_{\text{prior}}^{\mathbf{a}} 
\right)
}
\nonumber \\
&- \sum_{\mathbf{x}} \mathbb{E}_{\mathsf{u}_{\mathbf{x}}  \sim \mathcal{N}\left(\mathbb{E}_{b_{\mathsf{A}}(t)}\left\{u_{\mathbf{x}} \right\} ; \mathrm{Var}_{b_{\mathsf{A}}(t)} \left\{u_{\mathbf{x}} \right\}\right)} {\left\{ \log p_{\mathsf{v}_{\mathbf{x}} | \mathsf{u}_{\mathbf{x}} } \left(v_{\mathbf{x}} | u_{\mathbf{x}}\right)  \right\} } ,
\end{align}
where $(a)$ leverages the Gaussianity of the approximated posterior of $\mathsf{a}_{i, x_{i}}^{\ell}$, $\hat{\mathbf{a}}_{\text{prior}}$ and $\nu_{\text{prior}}^{\mathbf{a}}$ represent the assumed prior mean and variance. Intuitively, the first term in \eqref{TeS-hat-J-t-1} penalizes the deviation between the (TeS-AMP approximated) posterior and the assumed prior on $\mathcal{A}$, and the second term rewards highly likely estimates $\mathcal{U}$.

\section{Additional Experimental Results} \label{appendixG}

\subsection{Experimental Setup}

We consider a particular observation model wherein the elements of $\mathcal{U}$ are potentially AWGN-corrupted at a subset of indices $\Omega$ and unobserved at the remaining indices, which validates the performance of the proposed algorithm applied to tensor completion application in \eqref{V-U}.  For the performance of the proposed algorithm applied to tensor decomposition, we set $\Omega$  to all ones to satisfy the model in \eqref{V-U}.

At the observed entries $\mathbf{x} \in \Omega$, the quantities $\hat{s}_{\mathbf{x}}$ and $\nu^s_{\mathbf{x}}$ are respectively calculated using the AWGN expressions \eqref{hat-s-2} and \eqref{nu-s-2}, while at the missing entries $\mathbf{x} \notin \Omega$, where $v_{\mathbf{x}}$ in invariant to the value of $u_{\mathbf{x}}$, we will have the mean $\mathbb{E}\left\{\mathrm{u}_{\mathbf{x}} | \mathrm{p}_{\mathbf{x}}=\hat{p}_{\mathbf{x}}; \nu_{\mathbf{x}}^{p}\right\}=\hat{p}_{\mathbf{x}}$
 and the variance $\mathrm{Var}\left\{\mathrm{u}_{\mathbf{x}} | \mathrm{p}_{\mathbf{x}}=\hat{p}_{\mathbf{x}} ; \nu_{\mathbf{x}}^{p}\right\}=\nu_{\mathbf{x}}^{p}$,
 so that $\hat{s}_{\mathbf{x}}  = 0$ and $\nu^{s}_{\mathbf{x}} = 0$. This shows that $\nu^s$ can be interpreted as inverse residual variances and $\hat{s}$  as $\nu^s$-scaled residuals. In summary, the possibly incomplete AWGN model yields
\begin{align} 
\hat{s}_{\mathbf{x}} =&~
\left\{
\begin{array}{ll}
{\frac{v_{\mathbf{x}}-\hat{p}_{\mathbf{x}} }{\nu_{\mathbf{x}}^{p}+\nu^{w}}} & 
{{\mathbf{x}} \in \Omega,} 
\\ {0}& 
{{\mathbf{x}} \notin \Omega,}
\end{array}
\right. \\ 
\nu_{\mathbf{x}}^{s} =&~
\left\{
\begin{array}{ll}
{\frac{1}{\nu_{\mathbf{x}}^{p}+\nu^{w}}} & {{\mathbf{x}} \in \Omega,} 
\\
{0} & {{\mathbf{x}} \notin \Omega.}
\end{array}
\right.
\end{align}

Similar to the approaches in~\cite{parker2014bilinear,parker2014bilinear2,vila2013expectation} we select the variance
\begin{align} 
\nu^{w}=\frac{\left\|P_{\Omega}(\mathcal{V})\right\|_{F}^{2}}{(\text{SNR}+1)|\Omega|},
\end{align}
where SNR is an initial estimate of the signal-to-noise ratio that can be set at $100$.

Regarding the termination of the algorithm, we terminate the algorithm when it converges or diverges. The algorithm has been converged if the generated tensor at the end of an iteration is very close to that in the previous iterations (when the Frobenius norm of their difference divided by the Frobenius norm of the previous one is less than $0.001$). And the algorithm has diverged if the Frobenius norm of the generated tensor at the end of an iteration divided by the Frobenius norm of the generated tensor at the end of the first iteration is more than $10^6$.

Assume that we want to generate a tensor of size $n_1\times \cdots \times n_d$ with TR-rank $r_1,\cdots,r_d$ with random components. We consider the TR decomposition and choose each entry of $\mathcal{Z}_{i}^{\ell_i,\ell_{i+1}}$ independently and according to standard Gaussian distribution $\mathcal{N} (0, 1)$, for $1 \leq i \leq d,~1 \leq \ell_i \leq r_i$. Then, we generate the tensor according to \eqref{TR-decomposition-1}. As for the CP tensor, we also use a similar way to generate it.

\subsection{Baseline for Comparison}

We are interested in comparing the performance of TeG-AMP with two other methods, explained below.

\subsubsection{Alternating Minimization}
For low TR-rank tensor completion, we apply the ALS method proposed in~\cite{wang2017efficient} for comparison, where the TR decomposition structure ${\mathcal U} = \sum_{\ell_{1}, \ldots, \ell_{d}=1}^{r_{1}, \ldots, r_{d}}  \mathcal{Z}_1^{\ell} \otimes \mathcal{Z}_2^{\ell} \otimes \cdots \otimes \mathcal{Z}_d^{\ell}$ is utilized. Starting with some initial $\mathcal{Z}_{i_0}^{\ell} \in \mathbb{R}^{n_{i_0}}$ for $1\leq i \leq d$ and $1 \leq \ell_{i_0} \leq r_{i_0}$, at the $k$-th iteration, we update all $\mathcal{Z}_{i,k}^{\ell}$'s one by one by solving the following convex programs
\begin{align} 
&\min_{\mathcal{Z}_{i',k}^{\ell'} \in \mathbb{R}^{n_{i'}}} \Bigg\| {\mathcal U}_{\Omega} - \Bigg( \sum_{\mathbf{l}=1}^{\mathbf r}  
\mathcal{Z}_{1,k_{(1,\ell_1,\ell_2)}}^{\ell}\otimes
\mathcal{Z}_{2,k_{(2,\ell_2,\ell_3)}}^{\ell} \otimes 
\cdots \otimes 
\mathcal{Z}_{d,k_{(d,\ell_d,\ell_1)}}^{\ell} \Bigg)_{\Omega} \Bigg\|_{\mathcal{F}},
\end{align}
where $k_{(i,\ell_i,\ell_{i+1})} = k$ if $i\leq i'$, $\ell_i \leq \ell_i'$ and $\ell_{i+1} \leq \ell_{i+1}'$ and $k_{(i,\ell_i,\ell_{i+1})} = k-1$ otherwise. The iterations continue until the algorithm converges or diverges (either can be determined in the code similar to the way explained earlier for TeG-AMP). Regarding the initialization, we use the tensor ring approximation algorithm in~\cite{wang2017efficient} to obtain an approximated decomposition as the initial values. We refer the readers to the standard argument in~\cite{wang2017efficient} for more details.

For low CP-rank tensor completion, we apply the ALS method that uses the tensor structure $\mathcal{U}=\sum_{\ell=1}^{r} \mathbf{a}_{1}^{\ell} \otimes \mathbf{a}_{2}^{\ell} \otimes \ldots \otimes \mathbf{a}_{d}^{\ell}$, where $\mathbf{a}_{i}^{\ell} \in \mathbb{R}^{N_i}$ for $1\leq i \leq d$ and $1\leq \ell \leq r$. Starting with some initial $\mathbf{a}_{i_0}^{\ell} \in \mathbb{R}^{n_{i_0}}$ for $1\leq i \leq d$ and $1\leq \ell \leq r$ (described later), at the $k$-th iteration, we update all $\mathbf{a}_{i,k}^{\ell}$'s one by one by solving the following convex programs
\begin{align} 
\min_{\mathbf{a}_{i', k}^{\ell'} \in \mathbb{R}^{n_{i'}} }
\left\|
\mathcal{U}_{\Omega} - 
\left(
\sum_{\ell=1}^{r} 
\mathbf{a}_{1, k_{(1, \ell)}}^{\ell} \otimes 
\mathbf{a}_{2, k_{(2, \ell)}}^{\ell} \otimes \ldots \otimes 
\mathbf{a}_{d, k_{(d, \ell)}}^{\ell}
\right)_{\Omega}
\right\|_{\mathcal{F}}, \nonumber
\end{align}
where $k_{(i, \ell)} = k$ if $i \leq i'$ and $\ell \leq \ell'$ and $k_{(i, \ell)} = k-1$ otherwise. The iterations continue until the algorithm converges or diverges (either can be determined in the code similar to the way explained earlier for TeG-AMP). Regarding the initialization for CP alternating minimization (AltMin), we first use the Matlab toolbox `Tensorlab' to calculate the CP decomposition of $\mathcal{U}_{\Omega}$, where $\mathcal{U}_{\Omega}$ is the sampled tensor with entries equal to zero at non-sampled indices. Then, we choose the leading $r$ rank-1 components to obtain a rank-$r$ initialization for AltMin.

\begin{figure}[!htb]
	\centering
		
	\subfloat[][]{\includegraphics[width=2.3in]{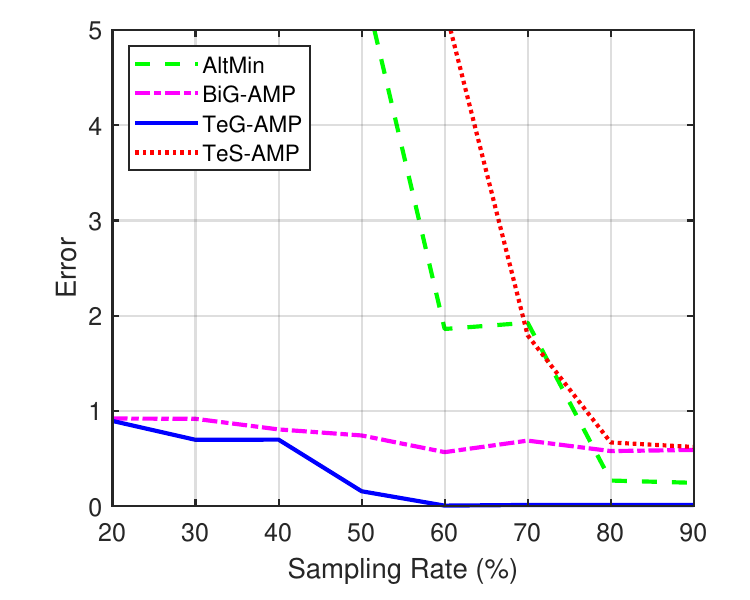}}
        \subfloat[][]{\includegraphics[width=2.3in]{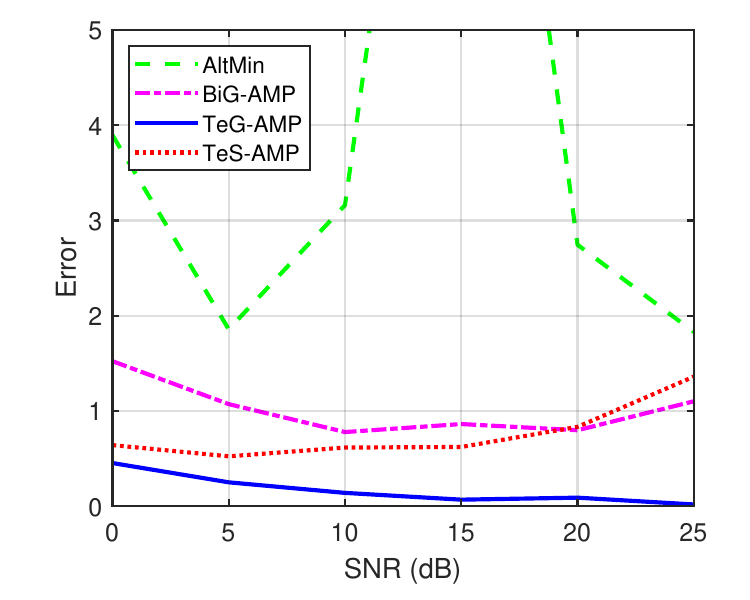}}
	
	\caption{Comparison results for tensor $\mathcal{U} \in \mathbb{R}^{6 \times 7 \times 8}$ of TR-rank $2,2,3$. (a) in noiseless case; (b) in noisy case.}
	\label{figure:TRmodelNoiseless}
\end{figure}

\begin{figure}[!htb]
	\centering
		
	\subfloat[][]{\includegraphics[width=2.3in]{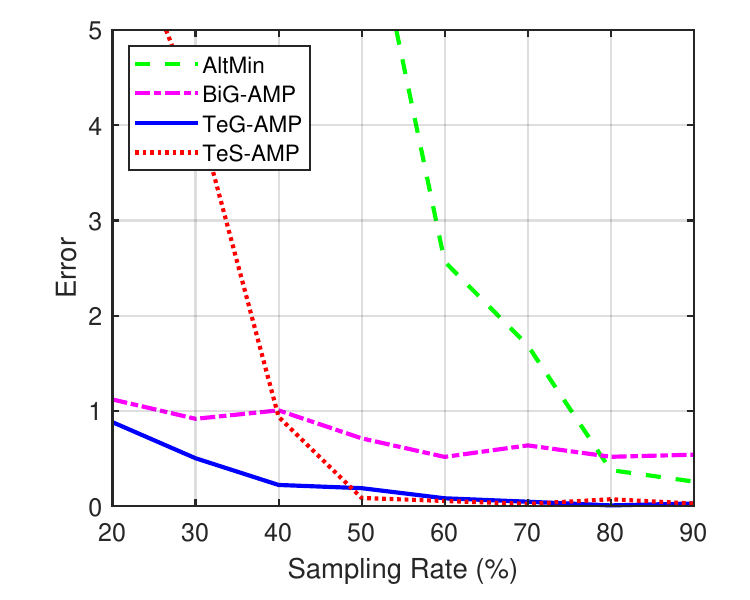}}
        \subfloat[][]{\includegraphics[width=2.3in]{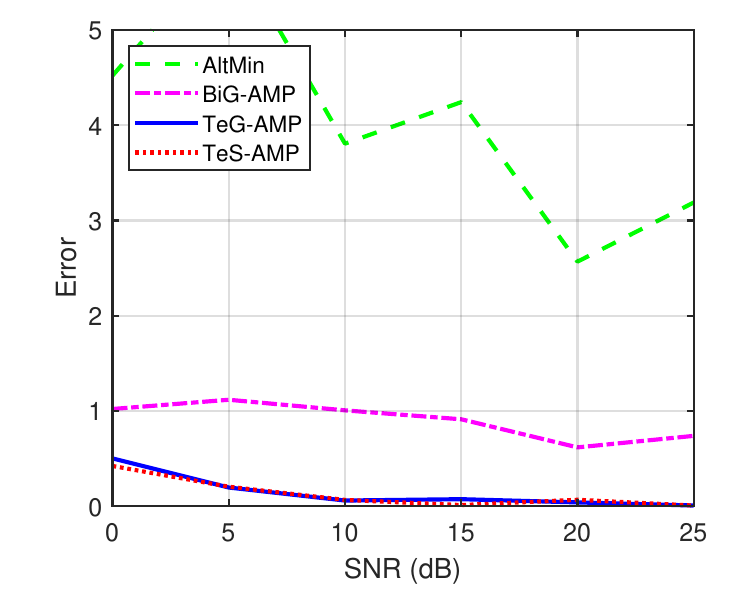}}
 
	\caption{Comparison results for tensor $\mathcal{U} \in \mathbb{R}^{6 \times 7 \times 8}$ of CP-rank $2$. (a) in noiseless case; (b) in noisy case.}
	\label{figure:CPmodelNoiseless}
\end{figure}

\subsubsection{BiG-AMP on Unfoldings of the Tensor}

We also compare our TeG-AMP with the BiG-AMP proposed in~\cite{parker2014bilinear}. The $i$-th unfolding of the $d$-dimensional tensor $\mathcal{U}$ is an $(N_1 N_2 \cdots N_i) \times (N_{i+1} \cdots N_d)$ matrix that is obtained through merging the first $i$ dimensions together and the rest of the dimensions together, $i = 1,\cdots, d-1$. Note that in general, the CP-rank is an upper bound on the rank of the unfolding, however, in our example, it is exactly the unfolding rank due to the random structure. As for low TR-rank tensor, we set the rank of the unfolding as the ceiling of the average of $r_1,\cdots,r_d$.

One simple approach is to apply the two-dimensional analysis BiG-AMP for each of the unfoldings and compare the one that works the best with TeG-AMP and AltMin. However, it is easy to verify that this approach is not efficient due to ignoring the tensor structure and treating it as a matrix structure. Hence, the purpose of comparing this method with other methods is only for verifying that TeG-AMP is well-designed to take advantage of the tensor structure.

\subsection{Comparison of TeG-AMP and TeS-AMP}

\subsubsection{Results of Low TR-rank Tensor}

We first compare the performance of the proposed TeG-AMP and TeS-AMP based on low TR-rank tensor. We generate a tensor $\mathcal{U} \in \mathbb{R}^{6 \times 7 \times 8}$ of rank $2,2,3$, by generating its corresponding TR decomposition components randomly. Then, we sample each entry of $\mathcal{U}$ independently with probability $p_{\Omega}$, where we vary the value of $p_{\Omega}$ from 0.2 to 0.8. Due to the random nature of our tests, and in order to make the error curves more accurate and smooth, for each value of $p_{\Omega}$ we run ``25'' random tests and calculate the average error for each of the four mentioned methods, where, as stated above, the error is defined as
\begin{align}
\epsilon=\frac{\left\|\left(\mathcal{U}^{*} - \mathcal{U}\right)_{\bar{\Omega}}\right\|_{\mathcal{F}}}{\left\| \mathcal{U}_{\bar{\Omega}}\right\|_{\mathcal{F}}}.
\end{align}

		
	

The error curves for this simulation are shown in Figure~\ref{figure:TRmodelNoiseless}(a). We can see that the proposed TeS-AMP cannot be applied in low TR-rank tensor completion since it is builded based on CP decomposition, this also reflects the fact that the TeG-AMP algorithm makes good use of the structure of TR decomposition. In addition, we present the error curves against SNRs in Figure~\ref{figure:TRmodelNoiseless}(b), where the sampling rate is set as 60\%. We can find that the TeG-AMP is much more resistant to noise than TeS-AMP under low TR-rank tensor setting.



\begin{figure}[!tb]
       \centering
        \vspace{-0.48cm} 
        
       \subfloat[][]{\includegraphics[width=2.3in]{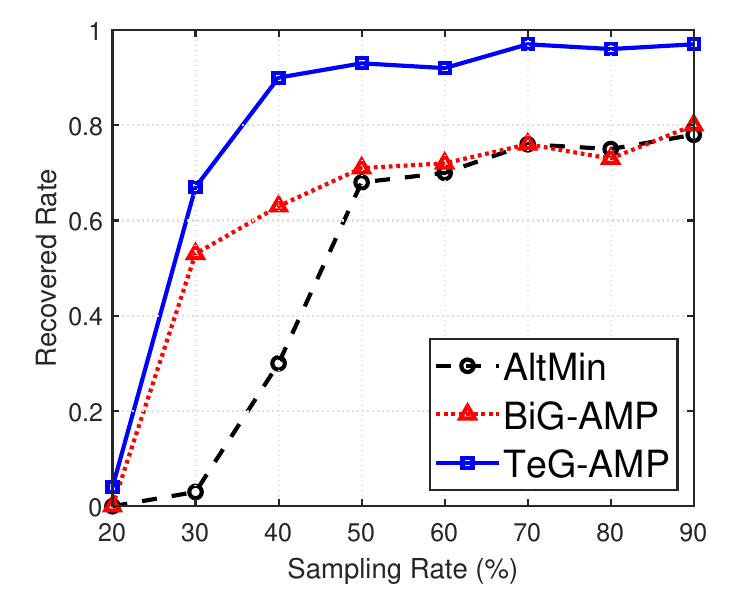}}
       \subfloat[][]{\includegraphics[width=2.3in]{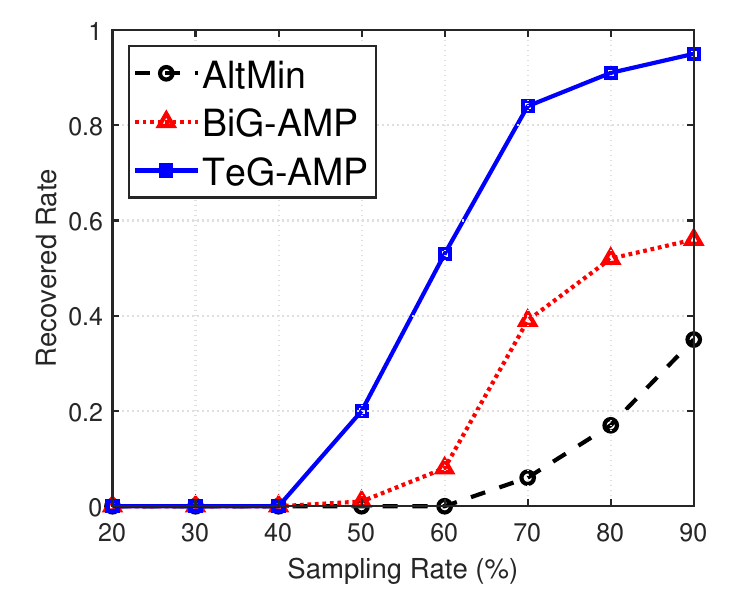}}
       \vspace{-0.3cm} 

        \subfloat[][]{\includegraphics[width=2.3in]{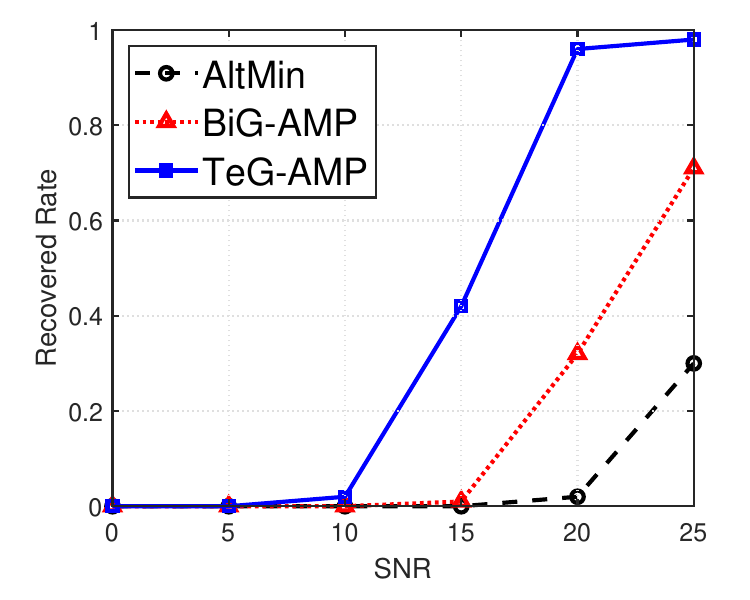}}
	\subfloat[][]{\includegraphics[width=2.3in]{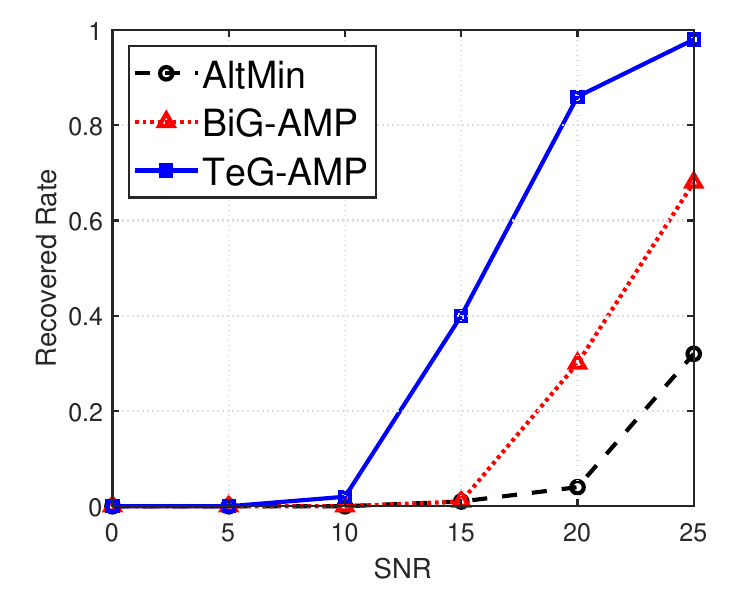}}

    \caption{Comparison results for CP-rank tensor $\mathcal{U} \in \mathbb{R}^{6 \times 7 \times 8}$. The sampling rate in noisy cases is $100\%$. (a) CP-rank $2,3,3$ in noiseless cases; (b) CP-rank $2,2,2$ in noiseless cases; (c) CP-rank $2,3,3$ in noisy cases; (d) CP-rank $2,2,2$ in noisy cases.}
    \label{figure:CPmodelNoise}
\end{figure}

%
%
%
%

		
	

\subsubsection{Results of Low CP-rank Tensor}

Next, we evaluate the performance of the proposed methods for low CP-rank tensor completion. We generate a tensor $\mathcal{U} \in \mathbb{R}^{6 \times 7 \times 8}$ of rank $2$, by similarly generating its corresponding CP decomposition components randomly. 
The error curves are shown in Figure~\ref{figure:CPmodelNoiseless}. We can see that the proposed TeG-AMP and TeS-AMP methods all work well in noiseless case, while the AltMin and BiG-AMP methods suffer from performance loss since they not take advantage of CP decomposition structure. Note that the TeS-AMP algorithm has much lower complexity compared with the TeG-AMP algorithm, makes it more suitable for low CP-rank tensor recovery. The error curves in noisy case are shown in Figure~\ref{figure:CPmodelNoiseless}(b), we can find that both TeG-AMP and TeS-AMP have good anti-noise performance for low CP-rank tensor completion.

Note that among the three mentioned approached, AltMin is the only one that is very dependent on the initialization. BiG-AMP, TeG-AMP and TeS-AMP are very stable by changing the random initialization. However, a relatively slight change in the initialization of AltMin can cause resulting an extremely bad output. Moreover, the initialization of AltMin is based on the TR decomposition of the sampled tensor. Hence, a big reduction in the sampling rate will cause a much worst initialization and therefore, for relatively low sampling rates, AltMin performs very poorly. The sensitivity of alternation minimization on the initialization is one of its weak points in comparison with AMP-based methods.

\subsection{Additional Results}

\subsubsection{Simulation Results of CP-rank Tensors}

In Figure~\ref{figure:CPmodelNoise}, we simulated the error curves for CP-rank under the same enactment as the TR-rank did. It shows that TeG-AMP is directly suitable for CP-rank, and still keeps the superiority compared to the other methodologies.

\subsubsection{Simulation Results of TT-rank Tensors}

\begin{figure}[!htb]
	\centering
	\subfloat[][]{\includegraphics[width=2.3in]{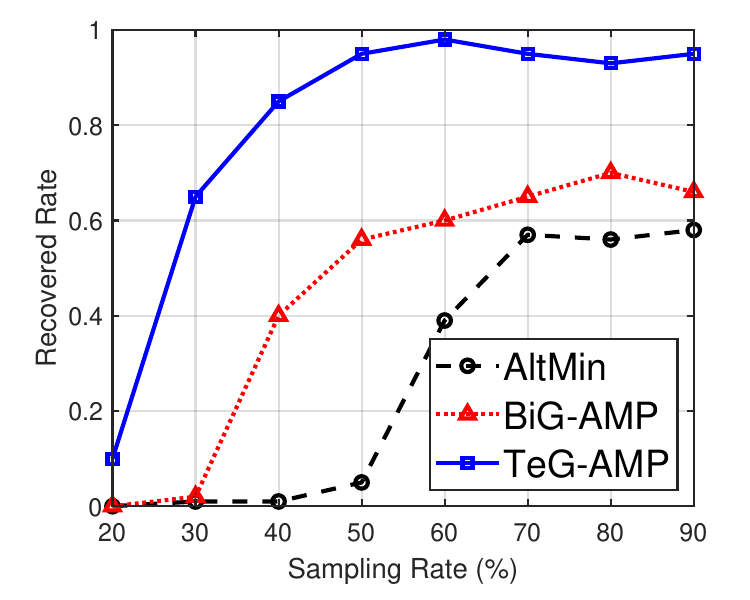}}
	\subfloat[][]{\includegraphics[width=2.3in]{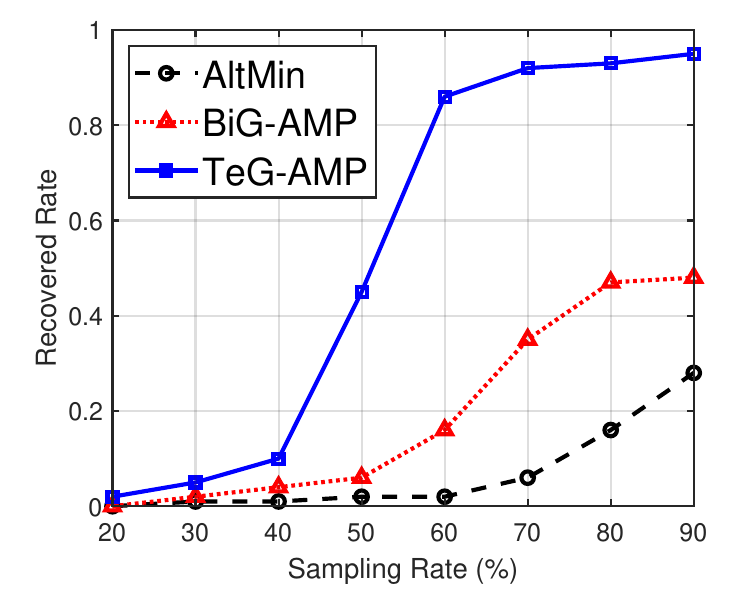}}
    
        \subfloat[][]{\includegraphics[width=2.3in]{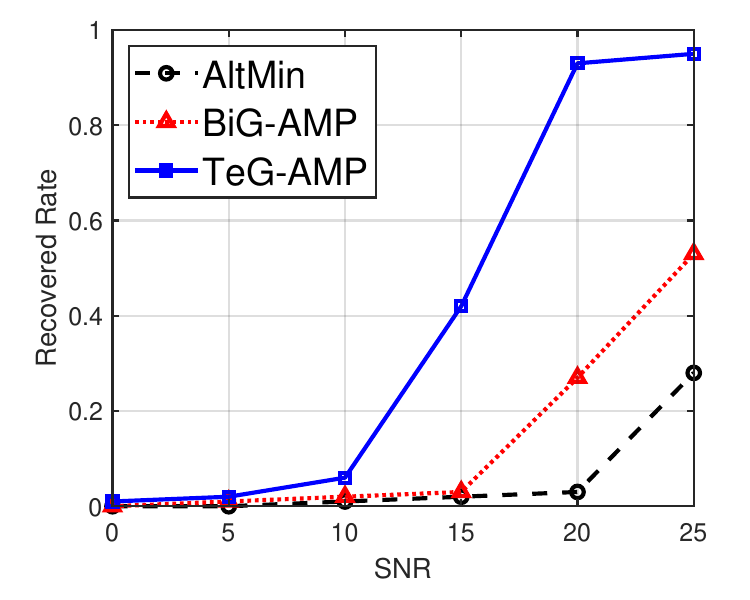}}
	\subfloat[][]{\includegraphics[width=2.3in]{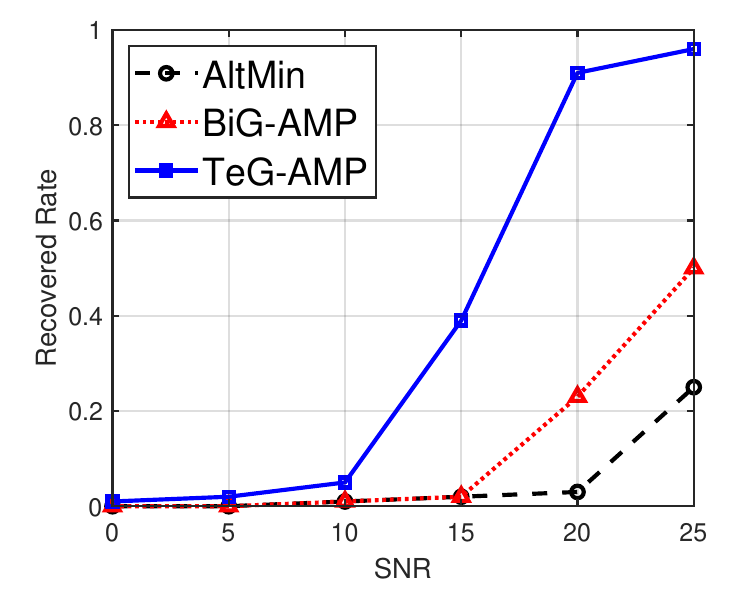}}
	
	\caption{Comparison results for TT-rank tensor $\mathcal{U} \in \mathbb{R}^{6 \times 7 \times 8}$. The sampling rate in noisy cases is $100\%$. (a) TT-rank $1,3,3$ in noiseless cases; (b) TT-rank $1,2,2$ in noiseless cases; (c) TT-rank $1,3,3$ in noisy cases; (d) TT-rank $1,2,2$ in noisy cases.}
        \label{figure:TTmodelNoisy}
\end{figure}

As shown in Section 4.2., TT can be taken as a special form of TR. We then simulated the recovery curves for TT-rank tensors under the same enactment as the TR-rank did. Figure~\ref{figure:TTmodelNoisy} shows that TeG-AMP is directly suitable for TT-rank, and still keeps the superiority compared to the other methodologies.

\subsubsection{Supplemental Simulation Results of MNIST digits}
\label{appendix-Results of MNIST digits}
\begin{figure}[!tb]
	\centering
	\subfloat[][]{\includegraphics[width=3.5in]{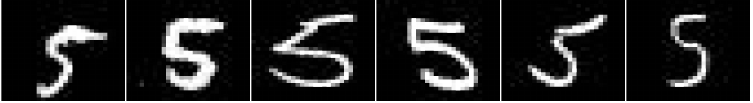}}

	\subfloat[][]{\includegraphics[width=3.5in]{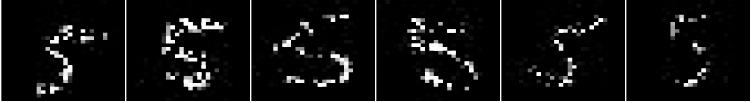}}

	\subfloat[][]{\includegraphics[width=3.5in]{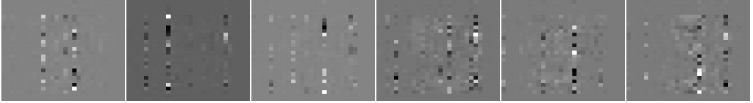}}
	
	\subfloat[][]{\includegraphics[width=3.5in]{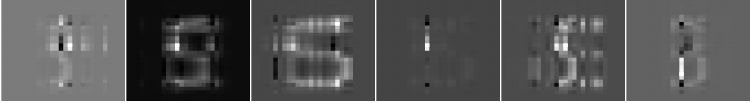}}
		
	\subfloat[][]{\includegraphics[width=3.5in]{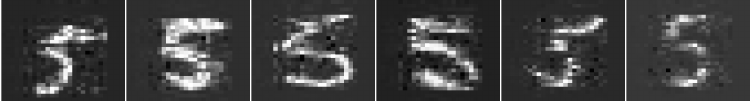}}
	
	\caption{MNIST digit `5'. The sampling rate is $40\%$. (a) Ground truth; (b) Sampling Results; (c) Recovered digits via AltMin, NMSE$ = 9.8921$; (d) Recovered digits via BiG-AMP, NMSE$ = 2.1803$; (e) Recovered digits via TeG-AMP, NMSE$ = 0.4823$.}
	\label{figure_mnist_5_0.4}
\end{figure}

\begin{figure}[!tb]
	\centering
	\subfloat[][]{\includegraphics[width=3.5in]{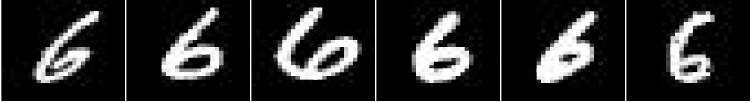}}

	\subfloat[][]{\includegraphics[width=3.5in]{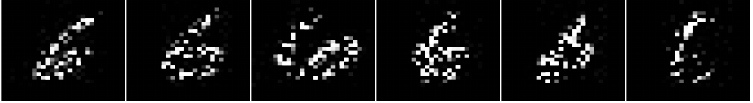}}

	\subfloat[][]{\includegraphics[width=3.5in]{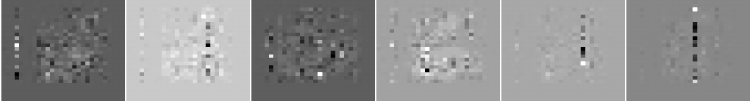}}
	
	\subfloat[][]{\includegraphics[width=3.5in]{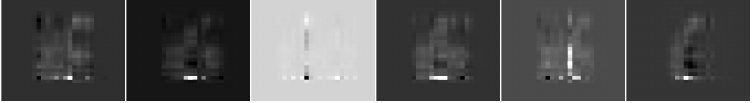}}
		
	\subfloat[][]{\includegraphics[width=3.5in]{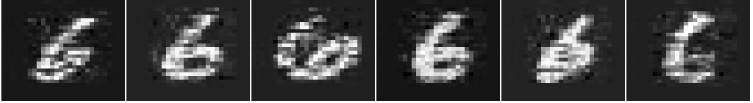}}
	
	\caption{MNIST digit `6'. The sampling rate is $40\%$. (a) Ground truth; (b) Sampling Results; (c) Recovered digits via AltMin, NMSE$ = 7.4701$; (d) Recovered digits via BiG-AMP, NMSE$ = 2.2070$; (e) Recovered digits via TeG-AMP, NMSE$ = 0.4425$.}
	\label{figure_mnist_6_0.4}
\end{figure}

\begin{figure}[!tb]
	\centering
	\subfloat[][]{\includegraphics[width=3.5in]{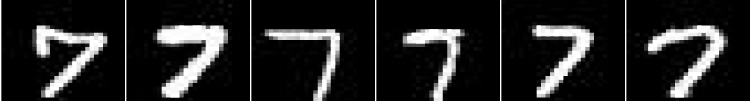}}

	\subfloat[][]{\includegraphics[width=3.5in]{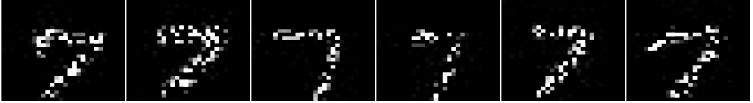}}

	\subfloat[][]{\includegraphics[width=3.5in]{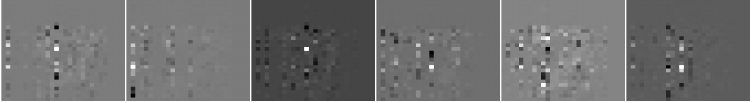}}
	
	\subfloat[][]{\includegraphics[width=3.5in]{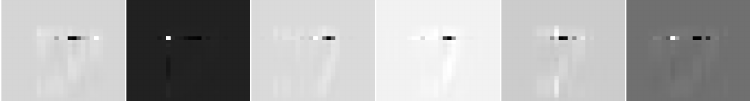}}
		
	\subfloat[][]{\includegraphics[width=3.5in]{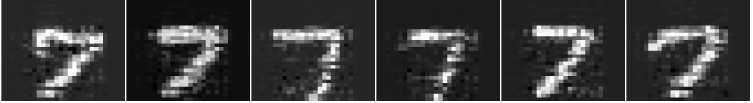}}
	
	\caption{MNIST digit `7'. The sampling rate is $40\%$. (a) Ground truth; (b) Sampling Results; (c) Recovered digits via AltMin, NMSE$ = 7.7645$; (d) Recovered digits via BiG-AMP, NMSE$ = 6.9849$; (e) Recovered digits via TeG-AMP, NMSE$ = 0.4465$.}
	\label{figure_mnist_7_0.4}
\end{figure}

\begin{figure}[!tb]
	\centering
	\subfloat[][]{\includegraphics[width=3.5in]{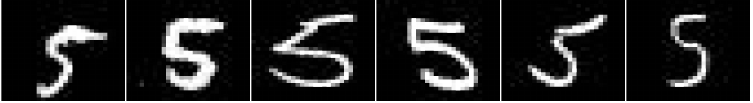}}

	\subfloat[][]{\includegraphics[width=3.5in]{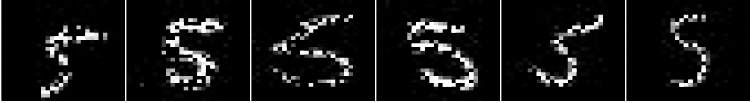}}

	\subfloat[][]{\includegraphics[width=3.5in]{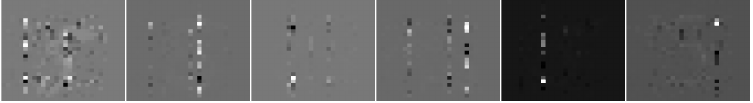}}
	
	\subfloat[][]{\includegraphics[width=3.5in]{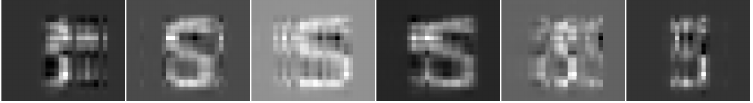}}
		
	\subfloat[][]{\includegraphics[width=3.5in]{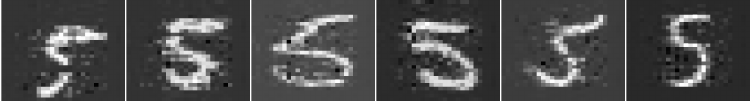}}
	
	\caption{MNIST digit `5'. The sampling rate is $60\%$. (a) Ground truth; (b) Sampling Results; (c) Recovered digits via AltMin, NMSE$ = 7.1096$; (d) Recovered digits via BiG-AMP, NMSE$ = 0.6384$; (e) Recovered digits via TeG-AMP, NMSE$ = 0.4020$.}
	\label{figure_mnist_5_0.6}
\end{figure}

\begin{figure}[!tb]
	\centering
	\subfloat[][]{\includegraphics[width=3.5in]{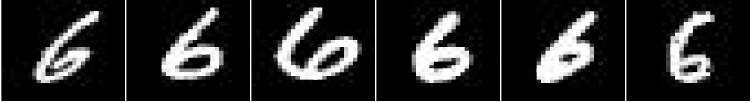}}

	\subfloat[][]{\includegraphics[width=3.5in]{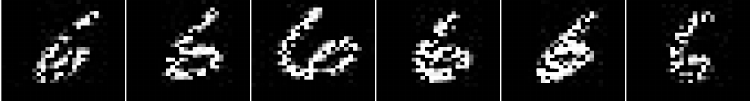}}

	\subfloat[][]{\includegraphics[width=3.5in]{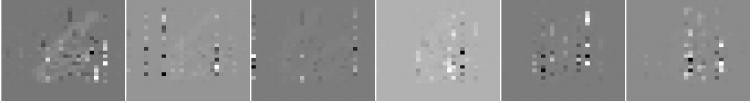}}
	
	\subfloat[][]{\includegraphics[width=3.5in]{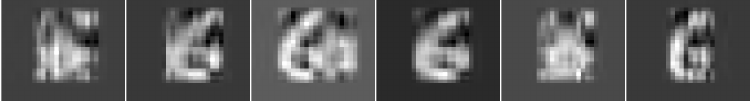}}
		
	\subfloat[][]{\includegraphics[width=3.5in]{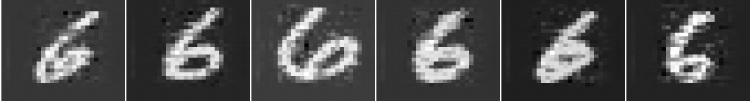}}
	
	\caption{MNIST digit `6'. The sampling rate is $60\%$. (a) Ground truth; (b) Sampling Results; (c) Recovered digits via AltMin, NMSE$ = 3.4961$; (d) Recovered digits via BiG-AMP, NMSE$ = 0.5171$; (e) Recovered digits via TeG-AMP, NMSE$ = 0.2839$.}
	\label{figure_mnist_6_0.6}
\end{figure}

\begin{figure}[!tb]
	\centering
	\subfloat[][]{\includegraphics[width=3.5in]{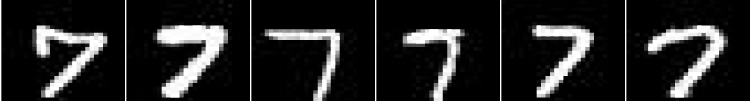}}

	\subfloat[][]{\includegraphics[width=3.5in]{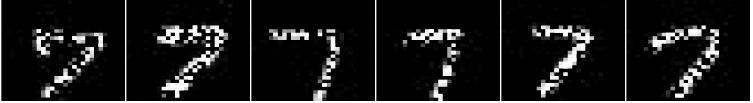}}

	\subfloat[][]{\includegraphics[width=3.5in]{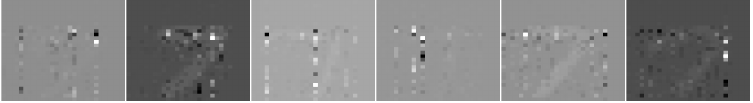}}
	
	\subfloat[][]{\includegraphics[width=3.5in]{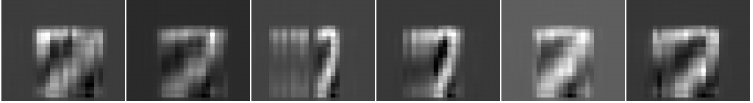}}
		
	\subfloat[][]{\includegraphics[width=3.5in]{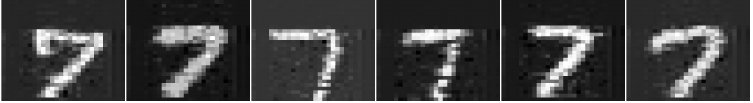}}
	
	\caption{MNIST digit `7'. The sampling rate is $60\%$. (a) Ground truth; (b) Sampling Results; (c) Recovered digits via AltMin, NMSE$ = 5.6153$; (d) Recovered digits via BiG-AMP, NMSE$ = 0.4866$; (e) Recovered digits via TeG-AMP, NMSE$ = 0.3113$.}
	\label{figure_mnist_7_0.6}
\end{figure}

\begin{figure}[!tb]
	\centering
	\subfloat[][]{\includegraphics[width=3.5in]{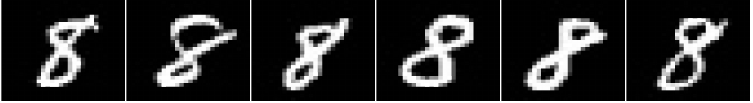}}

	\subfloat[][]{\includegraphics[width=3.5in]{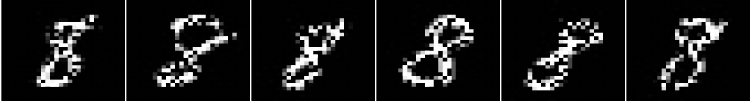}}

	\subfloat[][]{\includegraphics[width=3.5in]{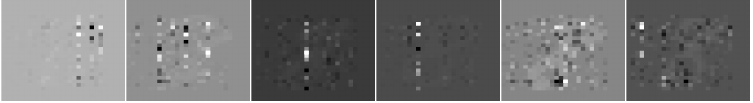}}
	
	\subfloat[][]{\includegraphics[width=3.5in]{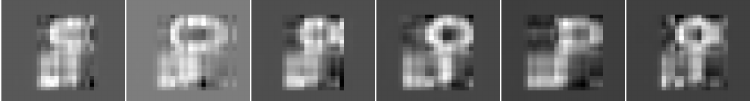}}
		
	\subfloat[][]{\includegraphics[width=3.5in]{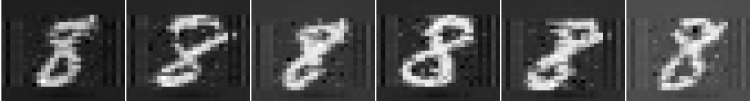}}
	
	\caption{MNIST digit `8'. The sampling rate is $60\%$. (a) Ground truth; (b) Sampling Results; (c) Recovered digits via AltMin, NMSE$ = 5.1195$; (d) Recovered digits via BiG-AMP, NMSE$ = 0.5249$; (e) Recovered digits via TeG-AMP, NMSE$ = 0.3311$.}
	\label{figure_mnist_8_0.6}
\end{figure}

The recovered results of MNIST digits with size $28\times 28 \times 6$ and TR rank $14\times 14 \times 6$ are shown in Figure~\ref{figure:MovingChar} and Figures~\ref{figure_mnist_5_0.4}-\ref{figure_mnist_8_0.6}. 
The recover errors of TeG-AMP, BiG-AMP and AltMin methods are presented in Table~\ref{table2}. 
We can see that the proposed TeG-AMP method performs essentially better than AltMin and BiG-AMP methods, since it takes full advantage of the tensor structure. The recovered digits are clearly visible. In contrast, the digits recovered by BiG-AMP and AltMin methods are drowned in noise.


\begin{figure}[!tb]
	\centering
	\subfloat[][]{\includegraphics[width=3.5in]{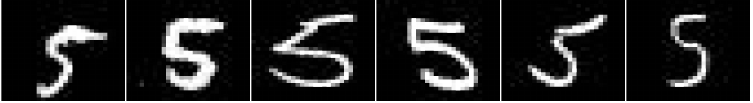}}

	\subfloat[][]{\includegraphics[width=3.5in]{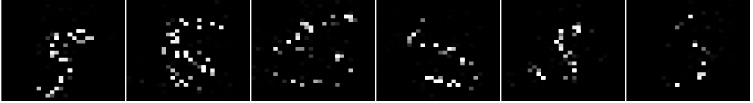}}

	\subfloat[][]{\includegraphics[width=3.5in]{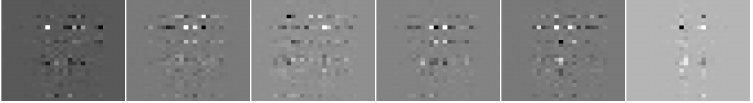}}
	
	\subfloat[][]{\includegraphics[width=3.5in]{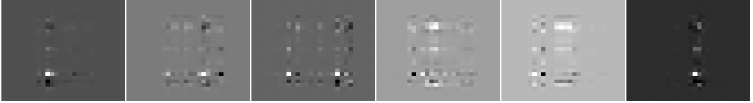}}
		
	\subfloat[][]{\includegraphics[width=3.5in]{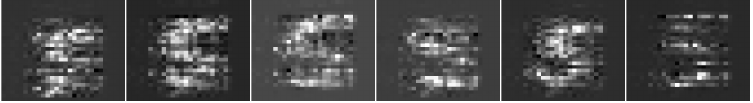}}
	
	\caption{MNIST digit `5'. The sampling rate is $20\%$. (a) Ground truth; (b) Sampling Results; (c) Recovered digits via AltMin, NMSE$ = 9.2409$; (d) Recovered digits via BiG-AMP, NMSE$ = 2.3660$; (e) Recovered digits via TeG-AMP, NMSE$ = 0.6986$.}
	\label{figure_mnist_5_0.2}
\end{figure}

\begin{figure}[!tb]
	\centering
	\subfloat[][]{\includegraphics[width=3.5in]{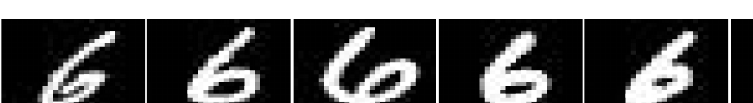}}

	\subfloat[][]{\includegraphics[width=3.5in]{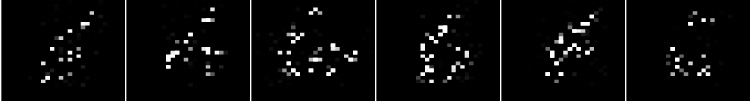}}

	\subfloat[][]{\includegraphics[width=3.5in]{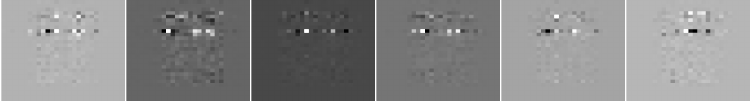}}
	
	\subfloat[][]{\includegraphics[width=3.5in]{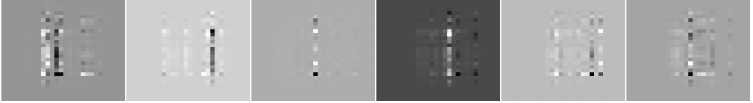}}
		
	\subfloat[][]{\includegraphics[width=3.5in]{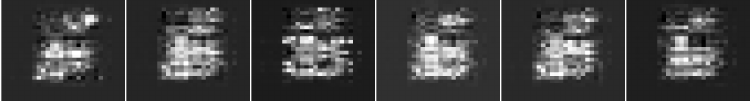}}
	
	\caption{MNIST digit `6'. The sampling rate is $20\%$. (a) Ground truth; (b) Sampling Results; (c) Recovered digits via AltMin, NMSE$ = 8.0302$; (d) Recovered digits via BiG-AMP, NMSE$ = 2.5594$; (e) Recovered digits via TeG-AMP, NMSE$ = 0.6422$.}
	\label{figure_mnist_6_0.2}
\end{figure}

\begin{figure}[!tb]
	\centering
	\subfloat[][]{\includegraphics[width=3.5in]{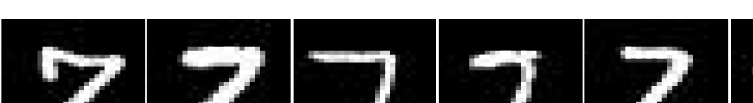}}

	\subfloat[][]{\includegraphics[width=3.5in]{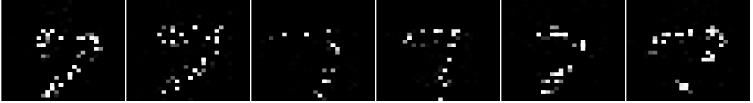}}

	\subfloat[][]{\includegraphics[width=3.5in]{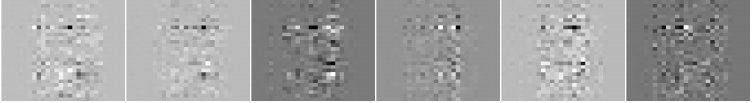}}
	
	\subfloat[][]{\includegraphics[width=3.5in]{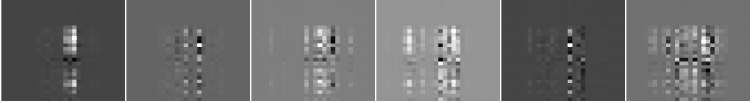}}
		
	\subfloat[][]{\includegraphics[width=3.5in]{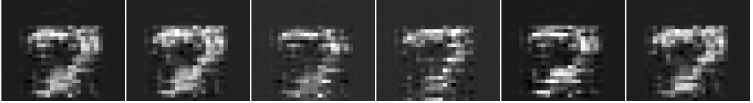}}
	
	\caption{MNIST digit `7'. The sampling rate is $20\%$. (a) Ground truth; (b) Sampling Results; (c) Recovered digits via AltMin, NMSE$ = 8.3517$; (d) Recovered digits via BiG-AMP, NMSE$ = 6.9261$; (e) Recovered digits via TeG-AMP, NMSE$ = 0.6166$.}
	\label{figure_mnist_7_0.2}
\end{figure}

\begin{figure}[!tb]
	\centering
	\subfloat[][]{\includegraphics[width=3.5in]{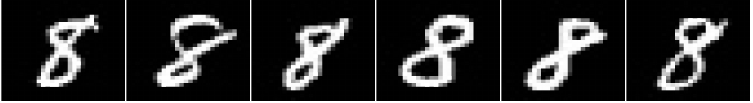}}

	\subfloat[][]{\includegraphics[width=3.5in]{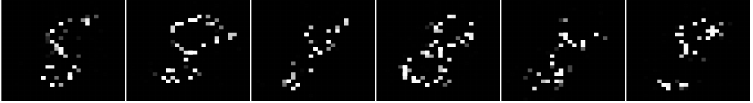}}

	\subfloat[][]{\includegraphics[width=3.5in]{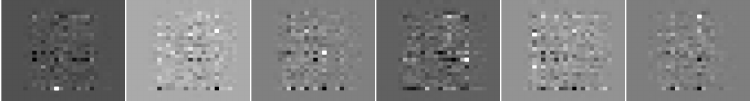}}
	
	\subfloat[][]{\includegraphics[width=3.5in]{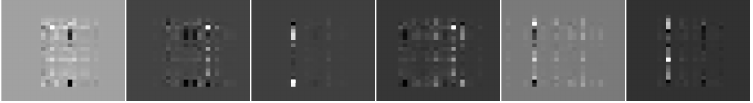}}
		
	\subfloat[][]{\includegraphics[width=3.5in]{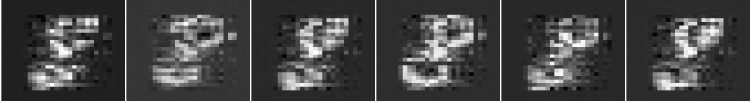}}
	
	\caption{MNIST digit `8'. The sampling rate is $20\%$. (a) Ground truth; (b) Sampling Results; (c) Recovered digits via AltMin, NMSE$ = 5.7151$; (d) Recovered digits via BiG-AMP, NMSE$ = 2.1466$; (e) Recovered digits via TeG-AMP, NMSE$ = 0.6148$.}
	\label{figure_mnist_8_0.2}
\end{figure}

\end{document}